\documentclass[10pt,twocolumn,letterpaper]{article}

\usepackage{iccv}
\usepackage{enumitem}
\usepackage{epsfig}
\usepackage{graphicx}
\usepackage{amsmath}
\usepackage{amssymb}
\usepackage{bm}
\usepackage{mathrsfs}
\usepackage{booktabs}
\usepackage{caption}
\usepackage{subcaption}
\usepackage{tabularx}
\usepackage{float}
\usepackage{microtype}

\usepackage{times}

\usepackage[
backend=biber,
bibstyle=ieee, 
citestyle=numeric,
sorting=ynt,
doi=false,
hyperref=true,
isbn=false,
sortcites,
url=false,
maxbibnames=99,
maxalphanames=3,
minalphanames=3,
maxcitenames=2,
mincitenames=1,
backref=true,
firstinits=false  %
]{biblatex}
\renewbibmacro*{pageref}{. \iflistundef{pageref}{}{{\printlist[pageref][-\value{listtotal}]{pageref}}}}

\usepackage[breaklinks=true,letterpaper=true,colorlinks,bookmarks=false]{hyperref}

\usepackage{nameref}
\usepackage[capitalise,noabbrev,sort]{cleveref}
\crefname{equation}{}{}
\numberwithin{equation}{section} %
\usepackage[toc,page]{appendix}
\usepackage{lipsum}

\usepackage{amsthm}
\usepackage{thmtools}
\usepackage{enumitem}
\usepackage{multirow}
\makeatletter
\let\NAT@parse\undefined
\makeatother
\newtheorem{theorem}{Theorem}[section]
\newtheorem{theoreminformal}{Theorem}
\newtheorem{lemma}{Lemma}[section]

\theoremstyle{remark}
\newtheorem{remark}{Remark}[section]

\theoremstyle{definition}

\Crefname{theoreminformal}{Theorem}{Theorems}

\iccvfinalcopy %

\def\iccvPaperID{10306} %

\input{sam_macros.def}  %

\let\mathbf\relax
\newcommand{\mathbf}{\boldsymbol}

\newcommand{\ours}{TILTED}

\newcommand{\rotsuffix}{\textsuperscript{\text{SO(3)}}}

\newcommand{\Proj}{\texttt{Proj}}
\newcommand{\projparams}{\vtau}
\newcommand{\Reduce}{\texttt{Reduce}}

\newcommand{\FeatureGrid}{\boldsymbol{F}}

\newcommand{\VoxelTrilerp}{\texttt{VoxelTrilerp}}
\newcommand{\voxelparams}{\boldsymbol{\phi}}

\newcommand{\Decoder}{\texttt{MLP}}
\newcommand{\decoderparams}{\boldsymbol{\theta}}

\newcommand{\Interp}{\texttt{Interp}}
\newcommand{\position}{\boldsymbol{p}}
\newcommand{\positionx}{p_x}
\newcommand{\positiony}{p_y}
\newcommand{\positionz}{p_z}
\newcommand{\latent}{\boldsymbol{Z}}
\newcommand{\TallProj}{\vphantom{\mathbb{R}^W}\Proj}

\newcommand{\grtr}{\natural}

\newcommand{\rough}{\mathrm{rough}}

\newcommand{\SG}{}

\begin{document}

\title{Canonical Factors for Hybrid Neural Fields}

\author{
    Brent Yi$^1$ \quad Weijia Zeng$^1$ \quad Sam Buchanan$^2$ \quad Yi Ma$^1$\\
    {\vspace{-0.5em}}\\
    $^1$UC Berkeley \quad \quad $^2$TTI-Chicago
}

\twocolumn[{%
    \renewcommand\twocolumn[1][]{#1}%
    \maketitle
    \centering
    \vspace{-10pt}
    \newcommand{\teaserwidth}{.98\textwidth}
    \includegraphics[width=\teaserwidth]{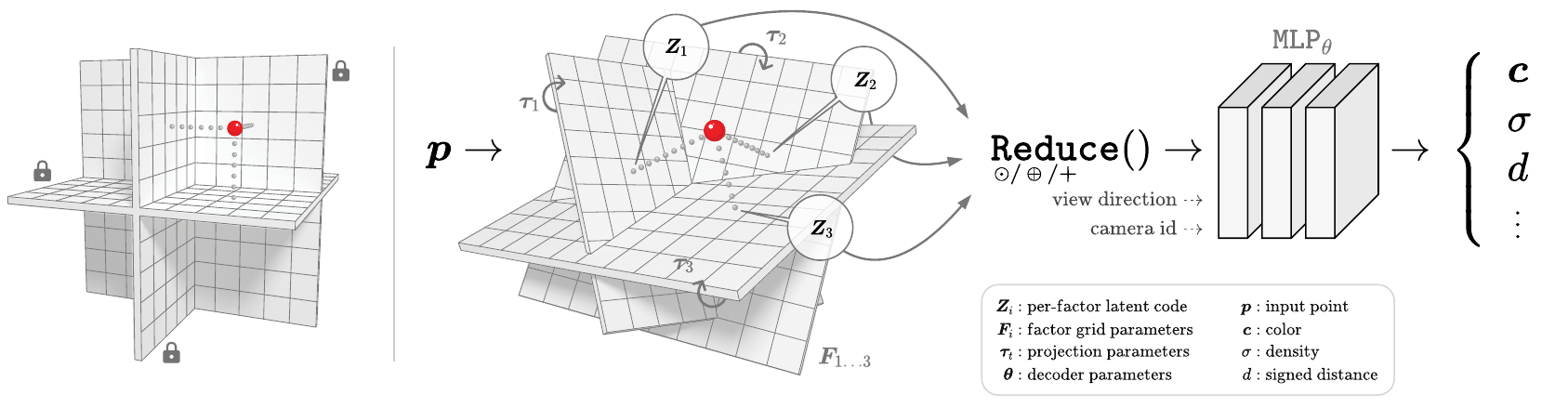}
    \captionof{figure}{
        \label{fig:method}
        \textbf{Learned transforms for factored feature volumes.} 
        Latent decompositions with fixed, axis-aligned projections (left) introduce biases for axis-aligned signals.
        A more robust, transform-invariant latent
        decomposition (TILTED) is obtained by treating
        projections to feature grids as learnable functions, here parameterized
        by $\projparams_t$.
    }%

    \vspace{2em}
}]

\ificcvfinal\thispagestyle{empty}\fi

\begin{abstract}
    Factored feature volumes offer a simple way to build more compact, efficient, and intepretable neural fields, but also introduce biases that are not necessarily beneficial for real-world data.
    In this work, we (1) characterize the undesirable biases that these
    architectures have for axis-aligned signals---they can lead to radiance
    field reconstruction differences of as high as $2$ PSNR---and (2) explore how learning a set of canonicalizing transformations can improve representations by removing these biases.
    We prove in a two-dimensional model problem that simultaneously learning these transformations together with
    scene appearance succeeds with drastically improved efficiency.
    We validate the resulting architectures, which we call TILTED, using image, signed distance, and radiance field reconstruction tasks, where we observe improvements across quality, robustness, compactness, and runtime.
    Results demonstrate that \ours{} can enable capabilities comparable to baselines that are 2x larger, while highlighting weaknesses of neural field evaluation procedures.

\end{abstract}
\vspace{-4mm}

\section{Introduction}
\label{sec:intro}

\vspace{-1mm}

Our physical world layers complexity on top of regularity.
Tucked below the details that imbue our surroundings with character---the
intricate fibers of a fine-grained veneer, the light-catching specularities of
everyday metal, plastic, and glass---we find the simple geometric primitives and symmetries associated with built and natural structure.
3D representations work best when they effectively harness this structure.
Point clouds and voxel grids offer versatility, but their inability to capture structure results in resource usage that can grow too intractably for complex details or expansive scenes.
Meshes capture the uniformity of surfaces for compactness, but can be too restrictive for entities outside of an acceptable regime. %

In this work, we use the theme of structure to study and improve state-of-the-art hybrid neural fields, 
which typically pair neural decoders with factored feature volumes~\cite{chan2021eg3d,chen2022tensorf,fridovich2023kplanes,cao2023hexplane,hu2023Tri-MipRF,gao2023strivec}.
Aided by an ability to exploit sparse and low-rank structure, factorization is simple to implement and offers a host of advantages, such as compactness, efficiency, and interpretability.
However, these advantages hinge on an implicit frame of representation, which is not guaranteed to be aligned with the structure of scenes or signals one aims to represent.
Drawing on insights from both low-rank texture extraction
\cite{Zhang2012-ss} and implicit regularization in optimization
methods for factorization~\cite{Li2017-ic,Stoger2021-an}, we theoretically
characterize the importance of this alignment and then show how it can enable
practical improvements to hybrid neural fields. %
Our contributions are as follows:

  \textbf{(1)} We analyze the fragility of factored
    grids in a two-dimensional model problem, where resource efficiency  on
    simple-to-capture structures can be undermined even by small planar
    rotations (\Cref{sec:theory}). We prove that this fragility can be overcome
    by jointly optimizing over the parameters of a set of \textit{canonical factors}
    and a transformation of domain, when the underlying
    structure is well-aligned in some frame of representation.
    
  \textbf{(2)} We study how this same weakness affects practical neural field architectures, where it can lead to radiance field accuracy differences of as high as 2 PSNR (\Cref{sec:experiments}). We propose optimization of more robust, transform-invariant latent
    decompositions (\ours{}) via the same idea of canonical factors (\Cref{sec:method}).
    \ours{} models are optimized to jointly recover
    factors of a decomposed feature volume with a set of canonicalizing
    transformations, which are simple to incorporate into existing factorization techniques.
    
  \textbf{(3)} We evaluate the \ours{} models on three tasks:
    2D image, signed distance field, and neural radiance field reconstruction (\Cref{sec:experiments}).
    Our experiments highlight biases in existing neural field architecture and evaluation procedures, while
    demonstrating improved quality, robustness, compactness,
    and runtime.  For real scenes, \ours{} can simultaneously improve
    reconstruction detail, halve memory consumption, and accelerate training times by
    25\%.

\section{Related Work}
\label{sec:related-work}

\subsection{Neural Fields}
In its standard form, a neural field is implemented using an MLP that takes coordinates as input and returns a vector of interest.
For example, a basic neural radiance field~\cite{mildenhall2020nerf} with
network parameters $\decoderparams{}$ maps spatial positions
$\position = (\positionx,\positiony,\positionz) \in \mathbb{R}^3$ to RGB
colors $\boldsymbol{c} \in [0,1]^3$ and densities $\sigma \in \mathbb{R}_{\geq 0}$:
\begin{equation}
    \position \xrightarrow[]{\Decoder_{\decoderparams}} (\boldsymbol{c}, \sigma).
    \label{eq:basic-neurfield}
\end{equation}
This framework is highly versatile.
Instead of only position, inputs can include additional conditioning information such as specularity-enabling view directions~\cite{mildenhall2020nerf}, per-camera appearance embeddings~\cite{martin2021nerfw}, or time~\cite{fridovich2023kplanes}.
Instead of radiance, possible outputs also include representations of binary
occupancy~\cite{chen2019learning,mescheder2019occupancy}, signed
distance functions~\cite{park2019deepsdf,saito2019pifu}, joint representations of surfaces and radiance~\cite{yariv2021volsdf,wang2021neus,oechsle2021unisurf}, actions~\cite{florence2022implicit,weng2022neural}, and
semantics~\cite{vora2021nesf,fu2022panoptic,kundu2022panopticneuralfields,lerf2023}.
\ours{} is not tied to specific input or output modalities.

\begin{figure}[t]
    \centering
    \begin{subfigure}{.3\linewidth}
      \centering
      \includegraphics[width=.7\linewidth]{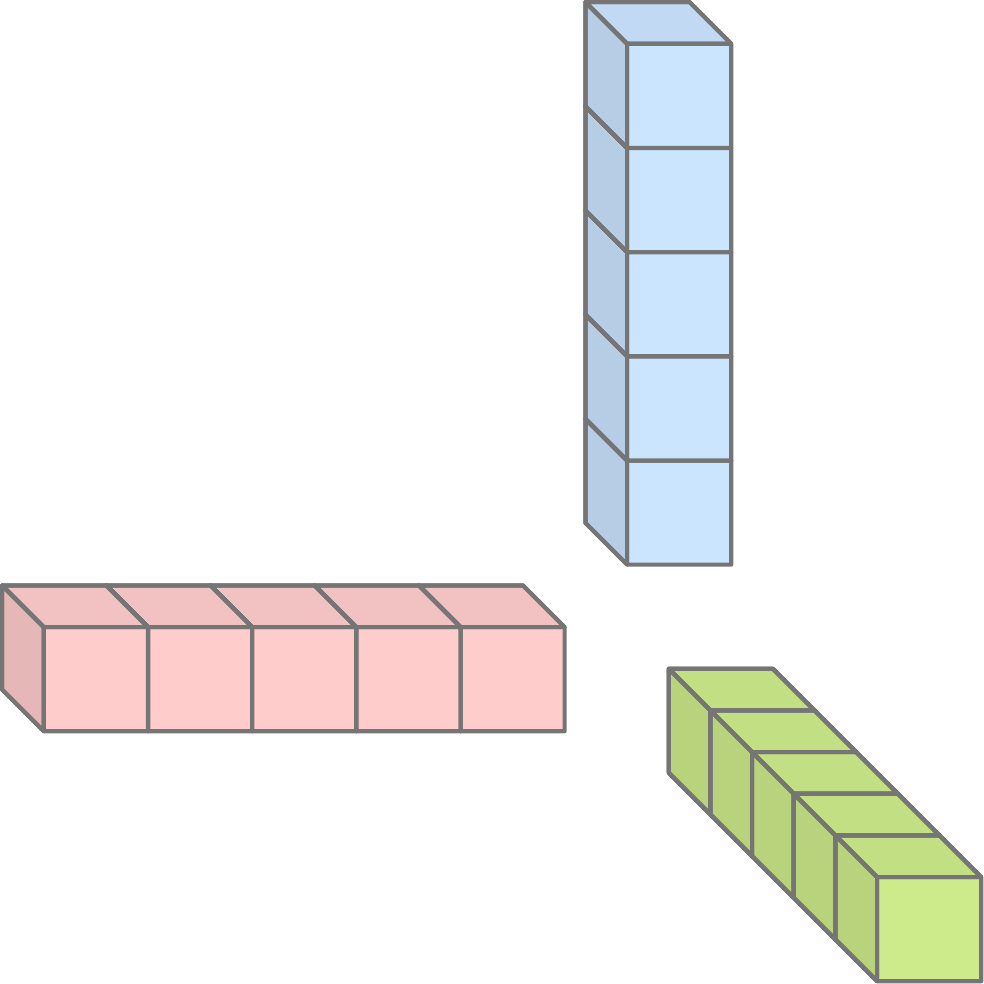}
      \vspace{1em}
      \caption{CP}
      \label{fig:decomp_cp}
    \end{subfigure}%
    \begin{subfigure}{.3\linewidth}
      \centering
      \includegraphics[width=.7\linewidth]{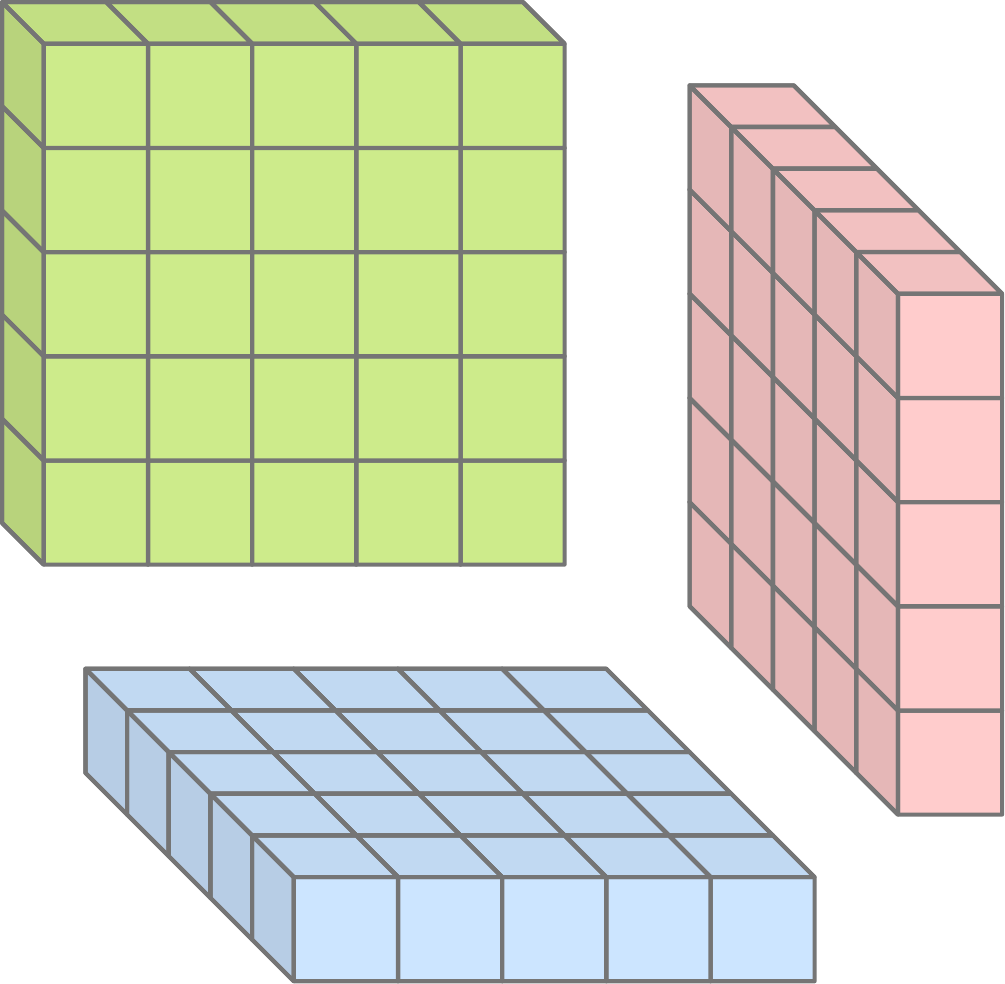}
      \vspace{1em}
      \caption{Tri-plane}
      \label{fig:decomp_triplane}
    \end{subfigure}
    \begin{subfigure}{.3\linewidth}
      \centering
      \includegraphics[width=.7\linewidth]{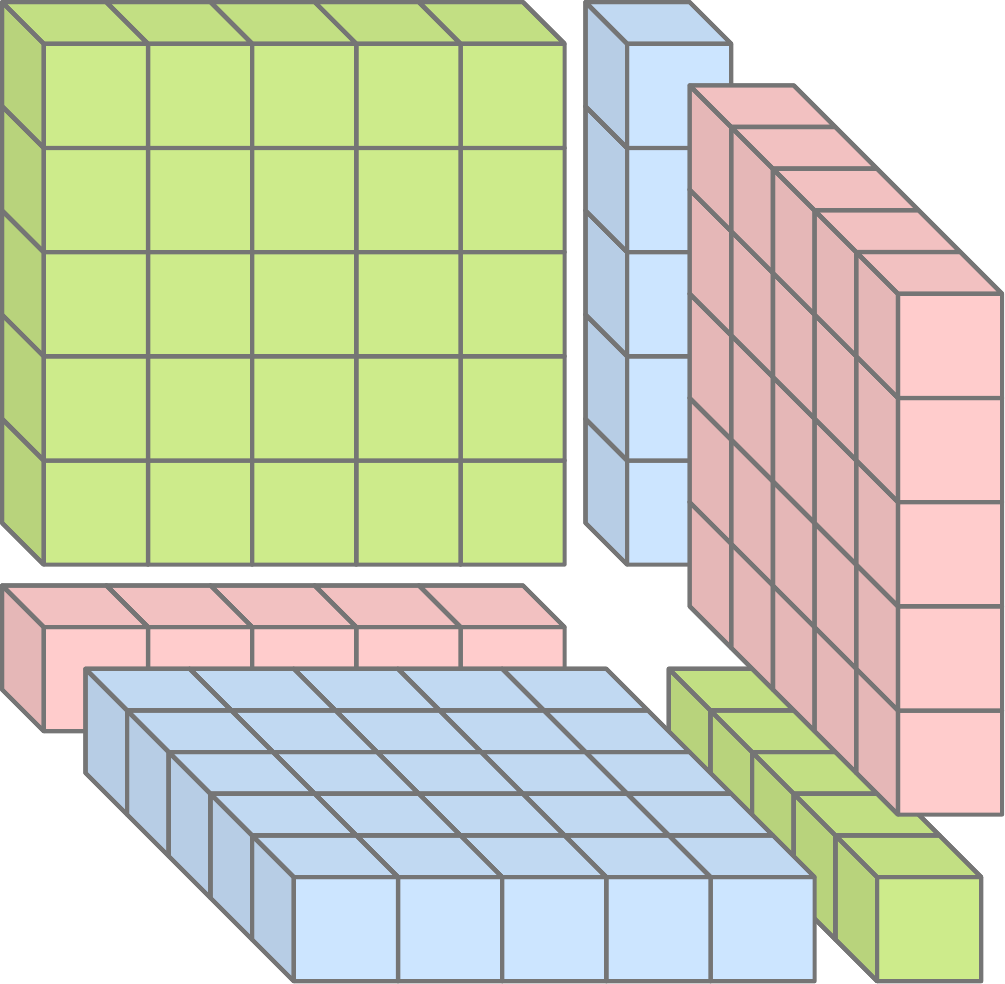}
      \vspace{1em}
      \caption{Vector-matrix}
      \label{fig:decomp_vm}
    \end{subfigure}
   \caption{
       \textbf{Tensor decompositions for 3D features volumes studied by prior work}~\cite{chen2022tensorf,chan2021eg3d,chen2022tensorf,fridovich2023kplanes}.
       Note that all assume a fixed, axis-aligned structure; \ours{} instead proposes to learn transformations of this structure.
   }
   \label{fig:tensor_decompositions}
\end{figure}

\subsection{Hybrid Neural Fields}

When a single MLP is used as a data structure, as in \Cref{eq:basic-neurfield},
all stored information needs to be encoded and entangled in the network weights $\decoderparams{}$.
The result is expensive for both training and inference.  
To address this, several works have proposed forms of \textit{hybrid neural fields}, which have two components:
an explicit geometric data structure from which latent vectors are
interpolated and a neural decoder~\cite{sun2021directvoxgo,ReluField_sigg_22}.
In the case of 3D coordinate inputs and radiance outputs, as in
\Cref{eq:basic-neurfield}, these architectures can be instantiated as
\begin{equation}
    \position \xrightarrow[]{\VoxelTrilerp_{\voxelparams{}}} \latent 
    \xrightarrow[]{\Decoder_{\decoderparams}} (\boldsymbol{c}, \sigma),
    \label{eq:relwork-hybrid}
\end{equation}
where $\texttt{VoxelTrilerp}_{\voxelparams{}}$ interpolates the `latent grid'
parameters $\voxelparams{}$ to produce a latent feature $\latent \in \mathbb{R}^d$,
which is then decoded to standard radiance field outputs by an MLP with
parameters $\decoderparams{}$.

Instead of implementing the latent feature volume $\voxelparams{}$ as a dense
voxel grid, a common pattern is to 
decompose this tensor into
lower-dimensional factors $\voxelparams{} = \{
\FeatureGrid_1 \dots \FeatureGrid_F \}$. In this way, hybrid neural field
approaches that rely on factored feature volumes~\cite{chan2021eg3d,chen2022tensorf,cao2023hexplane,fridovich2023kplanes,chen2023factor,Obukhov2022-zn}
generalize \Cref{eq:relwork-hybrid}
by (i) \textbf{projecting} input coordinates onto each of $F$
lower-dimensional coordinate spaces, (ii) \textbf{interpolating} $F$ feature vectors from
the corresponding factors, and (iii) \textbf{reducing}---for example, by concatenation, multiplication, or addition---the set of latent features into the final latent $\latent$:
\begin{equation}
    \begin{split}
    \latent
    &= \texttt{Reduce}\bigl(
      \bigl[\Interp_{\FeatureGrid_1}(\Proj_{1}(\position))\bigr],
      \dots, \\
      &\hphantom{=\Reduce\bigl(\,\, }
      \bigl[\Interp_{\FeatureGrid_F}(\Proj_{F}(\position))\bigr]
    \bigr).
    \end{split}
    \label{eq:relwork-factored-hybrid}
\end{equation}
Interpolating only on lower-dimensional feature grids $\FeatureGrid_1, \dots,
\FeatureGrid_F$, which may be 1D or 2D when $\position$ is 3D or higher,
provides efficiency advantages. 
We use ~\Cref{eq:relwork-factored-hybrid} to formalize existing factorization techniques in Appendix~\ref{sec:unified_interface_app}.

Hybrid neural fields have many advantages.
In contrast to techniques based on caching and distillation, which require a pretrained neural network~\cite{garbin2021fastnerf,hedman2021baking,yu2021plenoctrees,reiser2021kilonerf,lin2022neurmips,cole2021differentiable_surface_rendering}, hybrid neural field architectures accelerate both training and evaluation.
They also offer unique opportunities in generation~\cite{chan2021eg3d,chen2023singlestage, shue20223d},
real-time rendering~\cite{yariv2023bakedsdf, reiser2023merf},
upsampling~\cite{chen2022tensorf}, incremental growth~\cite{zhu2021niceslam,zhu2023nicerslam,johari2023eslam,meuleman2023progressively},
interpretable regularization~\cite{fridovich2023kplanes}, anti-aliasing~\cite{hu2023Tri-MipRF}, exploiting sparsity~\cite{gao2023strivec}, and dynamic scene reconstruction~\cite{cao2023hexplane, park2023temporal, shao2023tensor4d}.

Existing latent grid factorization methods constrain the \Proj{} operations
to axis-aligned projections (\Cref{fig:tensor_decompositions}). %
Similar to what has been observed in axis-aligned positional encodings~\cite{tancik2020fourfeat} (and pointed out by concurrent work~\cite{gao2023strivec}), this results in a bias for axis-aligned signals.
\ours{} proposes to learn a set of transforms that removes this bias.

\subsection{Learning With Transformations of Domain}

\ours{} improves reconstruction performance
via optimization over transformations of domain, a
mathematical idea dating back to the earliest days of computer vision. A
concrete example is the image registration problem
\cite{Brown1992-vz,Maintz1998-vz,Baker2004-fh,Szeliski2007-ul},
where we seek a transformation $\vtau$ that deforms an observed image $\vY$
to match a target $\vX$ via gradient descent.
\ours{} takes inspiration from
many tried-and-tested techniques for robustly solving problems of this type,
including coarse-to-fine fitting and other regularization schemes (e.g.,
\cite{Lefebure2001-uz,Miller2001-lv,Vural2014-xr}). %
Although this type of `supervised' registration is studied in the context of
neural fields \cite{Goli2022-yw}, it is less relevant to learning neural
implicit models like \Cref{eq:relwork-hybrid,eq:relwork-factored-hybrid}, where
ground-truth is rarely available. Instead, 
we build \ours{} around an
insight of \textcite{Zhang2012-ss}: \textit{for scenes consisting of
    natural or built environments, the transformation that `aligns' the scene
with its intrinsic coordinate frame yields the most compact representation}. In
the case of 2D images, \textcite{Zhang2012-ss} instantiate this principle as a
search for a transformation that minimizes the sum of the singular values of
the image, a relaxation of the rank:
\begin{equation}
    \min_{\vtau}\, \norm*{\vY \circ \vtau}_{*}.
    \label{eq:vanilla-tilt}
\end{equation}
\ours{}
combines this insight with the emerging understanding of
\textit{implicit regularization} in overparameterized matrix
factorization problems \cite{Li2017-ic,Stoger2021-an,Xu2023-ir}, which implies that an implicit bias toward
low-rank structures in factored grid representations learned with gradient
descent obviates the explicit rank regularization of \Cref{eq:vanilla-tilt}.
The 3D variant of $\vtau$ optimized by \ours{} can be interpreted as a special case of a gauge transformation, which prior work has explored for both general neural fields~\cite{zhan2023general} and for texture mapping~\cite{xiang2021neutex}; in context of axis-alignment, it also evokes scene representations that use mixtures of Manhattan frames~\cite{straub2014mixture}.

A parallel line of work seeks to imbue a broader family of neural network
architectures with invariance or `equivariance' to transformations or symmetries that the network should respect. These include
parallel channel networks
\cite{Ciresan2012-ha,10.1093/mnras/stv632,Laptev_2016_CVPR,DBLP:journals/corr/abs-2106-06418},
approaches based on pooling over transformations
\cite{Sohn2012-ml,Kanazawa2014-zf,Worrall_2017_CVPR}, and approaches with
learnable deformation offsets
\cite{Dai2017-lj,10.5555/2969442.2969465,Lin2016-au,Zhu2021-uk}. Other
approaches aim to construct networks that are transformation-invariant by
design \cite{Mallat2012-sl,Wiatowski2018-qr,Buchanan2022-qp}. With \ours{}, we
demonstrate how to combine the benefits of transformation invariance with a
variety of hybrid neural field architectures---as we discuss in Section~\ref{sec:theory}, naive
factorizations are limited in the diversity of structures that they can capture.%

\begin{figure}[!htb]
   \centering
       {\includegraphics[width=0.45\textwidth]{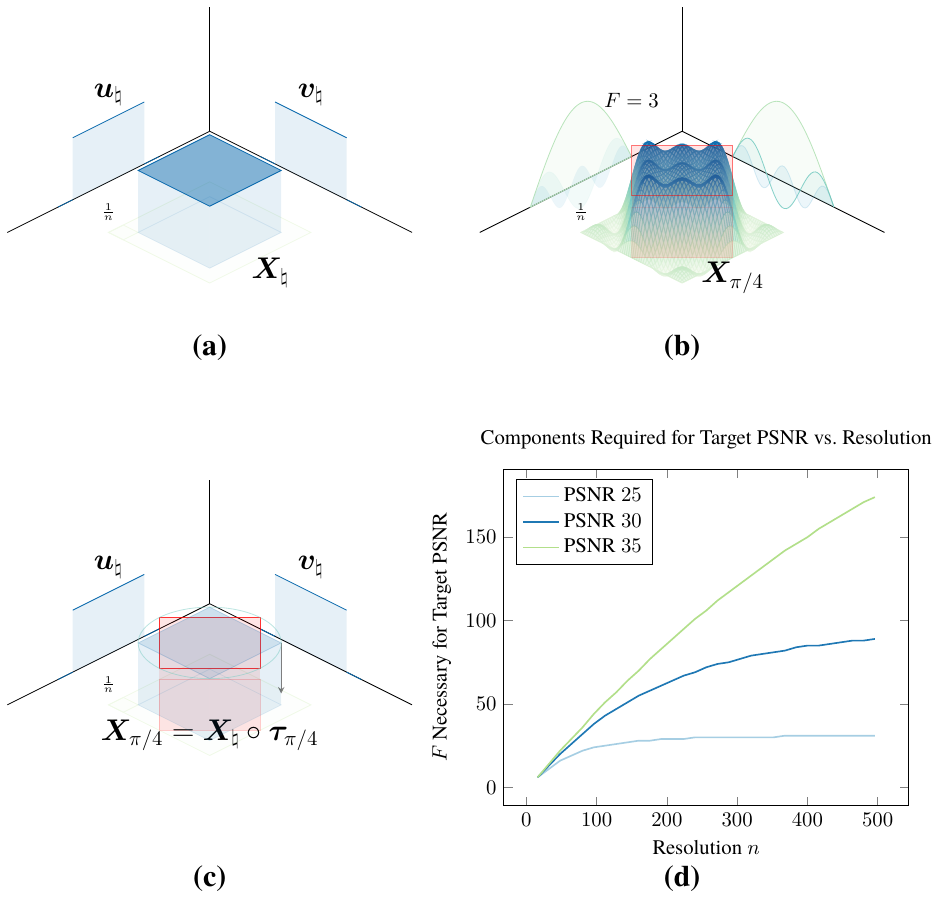}}
   \caption{\textbf{Limitations of low-rank feature grids.} \textbf{(a)}: The
      square template $\vX_{\grtr}$ is axis-aligned, and has
      a maximally-compact (rank one) representation. \textbf{(b)}: After a
      rotation by $\pi/4$ radians, the square template (in red) only changes its
      orientation, but its approximability by a low-rank grid deteriorates
      dramatically. We draw the scaled eigenvectors and approximation for $F=3$.
      \textbf{(c)}: By optimizing over transformations, a rank-one grid can be
      used to represent all rotations of $\vX_{\grtr}$. \textbf{(d)}: We plot the number of
      components needed to achieve varying PSNR levels as a function of image
      resolution for $\nu = \pi/4$. The number of components is always
   significantly larger than is necessary when transform optimization is used. }
   \label{fig:limitations_of_low_rank_grids}
\end{figure}
\section{Low-Rank Grids Are Delicate Creatures}
\label{sec:theory}

In this section, we demonstrate under ideal conditions that it is both desirable and
computationally feasible to recover a minimal set of \textit{canonical factors}
associated with a transformed scene using gradient-based optimization over
appearance and pose.
We omit
the MLP decoder in \Cref{eq:relwork-hybrid} and focus only on the bottleneck imposed
by the factored feature grid of \Cref{eq:relwork-factored-hybrid}.
Note that the capacity of the decoder is tightly constrained by performance considerations; TensoRF~\cite{chen2022tensorf} and K-Planes~\cite{fridovich2023kplanes}, for example, use only a single nonlinearity to decode density and proposal fields respectively.

Concretely, let $\vX_{\grtr} \in
\bbR^{n \times n}$ denote the grayscale image corresponding to the axis-aligned
square pattern in \Cref{fig:limitations_of_low_rank_grids}(a). We can decompose
$\vX_{\grtr}$ as $\vX_{\grtr} = \vu_{\grtr} \vv_{\grtr}\adj$, where
$\vu_{\grtr}$ and $\vv_{\grtr}$ are one-dimensional
square pulses aligned with the support of $\vX_{\grtr}$; $\vX_{\grtr}$ has
rank one, and can be perfectly reconstructed by a maximally-compact low-rank
feature grid.
In contrast, consider exactly the same scene, but with an additional rotation
by an angle of $\nu \in [0, \pi/4]$
applied to yield a transformed scene
$\vX_\nu = \vX_{\grtr} \circ \vtau_{\nu}$
(\Cref{fig:limitations_of_low_rank_grids}(b)).
As $\nu$ approaches its maximum value, %
the rank of the
transformed scene grows to a constant multiple of the resolution $n$,
implying that
\textit{perfect} representation of $\vX_\nu$ by a low-rank feature grid demands
essentially as many components as a generic $n \times n$ matrix. Moreover, 
even
\textit{approximate} representation of the transformed scene by a pure low-rank
grid is inefficient, as we prove for the instance visualized in
\Cref{fig:limitations_of_low_rank_grids}:
\begin{theoreminformal}[informal version of \Cref{thm:inapproximability}] 
   \label{thm:svd-informal}
   There exist absolute constants $c_0, c_1 > 0$ such that for any target channel
   count $F \leq c_0 n^{1/9.5}$, every rank-$F$ approximation 
   $\hat{\vX}$ to $\vX_{\pi/4}$ satisfies 
   \begin{equation*}
      \frac{1}{n^2} \norm*{\hat{\vX} - \vX_{\pi/4}}_{\mathrm{F}}^2
      \geq \frac{c_1}{1 + F}.
   \end{equation*}
\end{theoreminformal}
\Cref{thm:svd-informal} asserts that a broad class of sublinear-rank
approximations to $\vX_{\nu}$ have mean squared error at least as large as the
reciprocal number of components. %
Our proofs suggest this lower bound is tight
up to logarithmic factors---in particular, as we illustrate numerically in
\Cref{fig:limitations_of_low_rank_grids}(d), target PSNR levels that are more
stringent require larger grid ranks $F$ as the image resolution grows.
This situation stands in stark contrast to what can be achieved by capturing
the structure of $\vX_{\grtr}$: regardless of the image resolution, there
exists a single \textit{canonical factor} $\vu_{\grtr}$ which can represent any
observation $\vX_{\nu}$ via composition with a rotation $\vtau_{\nu}$
(\Cref{fig:limitations_of_low_rank_grids}(c)).
We prove that the $F=1$ instantiation of this problem %
successfully represents
$\vX_{\nu}$ in the hard instance visualized in
\Cref{fig:limitations_of_low_rank_grids} 
by jointly optimizing over grid factors and transformations:
\begin{theoreminformal}[informal version of \Cref{thm:tilt-infinite}]
   \label{thm:ntilt-informal}
   The infinite-resolution limit of the optimization procedure
   \begin{equation}
      \min_{\phi, \vu}\, \norm*{
         \vX_{\pi/4} - \left( \vu \vu\adj \right) \circ \vtau_{\phi}
      }_\mathrm{F}^2
      \label{eq:body-tilt-opti}
   \end{equation}
   solved with randomly-initialized constant-stepping gradient descent 
   converges to the true parameters $(\pi/4, \vu_{\grtr})$, up to
   symmetry, at a linear rate.
\end{theoreminformal}
\Cref{thm:ntilt-informal} provides theoretical grounding for \ours{}'s
transformation optimization approach in an idealized setting. Importantly,
\textit{there exist conditions under which the joint learning of the visual
representation and pose parameters provably succeeds}. 
The proof %
reveals a key 
principle underlying the success of this disentangled representation
learning: there is a symbiotic relationship between the model's
representation accuracy and its alignment accuracy, due to its constrained
capacity (i.e., $F = 1$ feature channels). More precisely, incremental
improvements to representation quality under inaccurate alignment
help the model localize the scene content and create texture gradients that
promote improvements to alignment; meanwhile, improvements to alignment allow
the model to leverage its constrained capacity to more accurately represent the
scene. %

\section{\ours{}}
\label{sec:method}

To instantiate the optimization procedure \Cref{eq:body-tilt-opti} in practice, we study a family of architectures that we call \ours{}, implemented based on two goals:
  \textbf{(1) Robustness.} \ours{} aims to be able to capture a broader set of structures than methods based on existing
  factorization techniques.
    Reconstruction ability should be invariant to simple transformations like rotations; as discussed theoretically in Section~\ref{sec:theory}
    and later empirically in Section~\ref{sec:experiments}, this does not hold for naively decomposed feature volumes.
  \textbf{(2) Generality.} \ours{} does not attempt to re-invent the
    wheel; instead, it is designed to be compatible with and build directly
    upon existing
    approaches~\cite{chen2022tensorf,fridovich2023kplanes} for
    factoring feature volumes.

Rather than assuming that the projection functions $\Proj_i$ in
\Cref{eq:relwork-factored-hybrid} are static and axis-aligned, \ours{} aims at recovery of \textit{canonical factors} by replacing the fixed and axis-aligned $\Proj_i$ with learnable functions $\Proj_{i,\projparams}$,  where $\projparams$ is a set of learnable transformation parameters.
By substituting into \Cref{eq:relwork-factored-hybrid},
the feature volume interpolation function then becomes:
\begin{equation}\label{eq:factored_hybrid_field_learnable_proj}
    \begin{split}
    \latent
    &= \texttt{Reduce}\bigl(
      \bigl[\Interp_{\FeatureGrid_1}(\Proj_{1,\projparams}(\position))\bigr],
      \dots, \\
      &\hphantom{=\Reduce\bigl(\,\, }
      \bigl[\Interp_{\FeatureGrid_F}(\Proj_{F,\projparams}(\position))\bigr]
    \bigr).
    \end{split}
\end{equation}
The transformations $\projparams$ enable mapping from arbitrary scene
coordinates to canonicalized domains for each factor $\FeatureGrid_i$.
As illustrated in \Cref{fig:method}, this can be interpreted as a
reconfiguration of factors to best align with and capture the 
structure of target signals.

\subsection{Applying Transformations}

The design space for the parameterization of $\projparams$ and how it is applied to input coordinates $\position$ is large.
We develop \ours{} for the case where $\projparams$ is a set of $T$ randomly
initialized rotations $\projparams = \{\projparams_1 \dots \projparams_T\}$,
parameterized by the unit circle $\Sph^1$ in 2D and $\Sph^3$ (the universal
cover of the set of rotation matrices $\mathrm{SO}(3)$; i.e., unit quaternions) in 3D.
We suffix variants with the value of $T$; \ours{}-4, for example, refers to \ours{} with $4$ learned rotations.

All experiments build atop feature volumes studied in prior work: for 3D, the CP~\cite{carroll1970cp_decomp}, vector-matrix~\cite{chen2022tensorf}, and K-Planes~\cite{fridovich2023kplanes} decompositions, which are each detailed in Appendix~\ref{sec:unified_interface_app}.
K-Planes in 3D is equivalent to a tri-plane~\cite{chan2021eg3d}, but uses a
multiplicative reduction.
We characterize each decomposition architecture using the channel dimension $d$ of its reduced latent vector $\latent \in \mathbb{R}^d$.
We constrain $T$ such that it evenly divides $d$, and apply rotations to the input coordinates such that each rotation $\projparams_t$ is used to compute $d / T$ of the final output channels.
This can be interpreted as a vectorized alternative to instantiating $T$ instances of a given decomposition, each with channel count $d / T$, applying a different learned rotation to the input of each decomposition, and concatenating outputs.
The resulting formulation has several desirable qualities:

\textbf{Robustness.}
When $\projparams$ is defined by a family of transforms and optimized from a random initialization, we see two related advantages.
First, as established in Section~\ref{sec:theory}, the latent feature volume
becomes able to represent signals that are not axis-aligned with
vastly improved parameter efficiency. %
Second, reconstruction becomes invariant to the transformation group encompassed by $\projparams$.
When $\projparams$ is constrained to rotations, a rotation applied to the scene
becomes equivalent to a rotation applied to the random initialization of
$\projparams$. 

\textbf{Convergence.}
Transformation optimization problems like camera registration are typically challenging and prone to local minima, but optimization in TILTED is better positioned to succeed.
We initialize many transforms: for any given structure in a scene, only one of these many transforms needs to fall into the basin of attraction for success.
Optimization of individual transforms is also highly symmetric.
Consider rotation optimization over a 2D grid: each increment of 90 degrees results in a representation with equivalent structure.
Our theoretical analysis, namely \Cref{thm:tilt-infinite}, verifies that these
properties are sufficient for optimization to succeed under idealized
conditions.

\textbf{Overhead.}
Rotations in this form are inexpensive both to store and apply.
Standard hybrid neural fields can have on the order of millions of parameters; a library of geometric transformations requires only dozens.
Because coordinate transformations reduce to simple matrix multiplications, the runtime penalty is also small.

\subsection{Coarse-to-Fine Optimization}\label{sec:coarse_to_fine}

When optimizing over transformations, high frequency signals produce undesirable local minima.
We improve convergence via two coarse-to-fine optimization strategies. %

\begin{figure}[t]
  \centering
  \includegraphics[width=\linewidth]{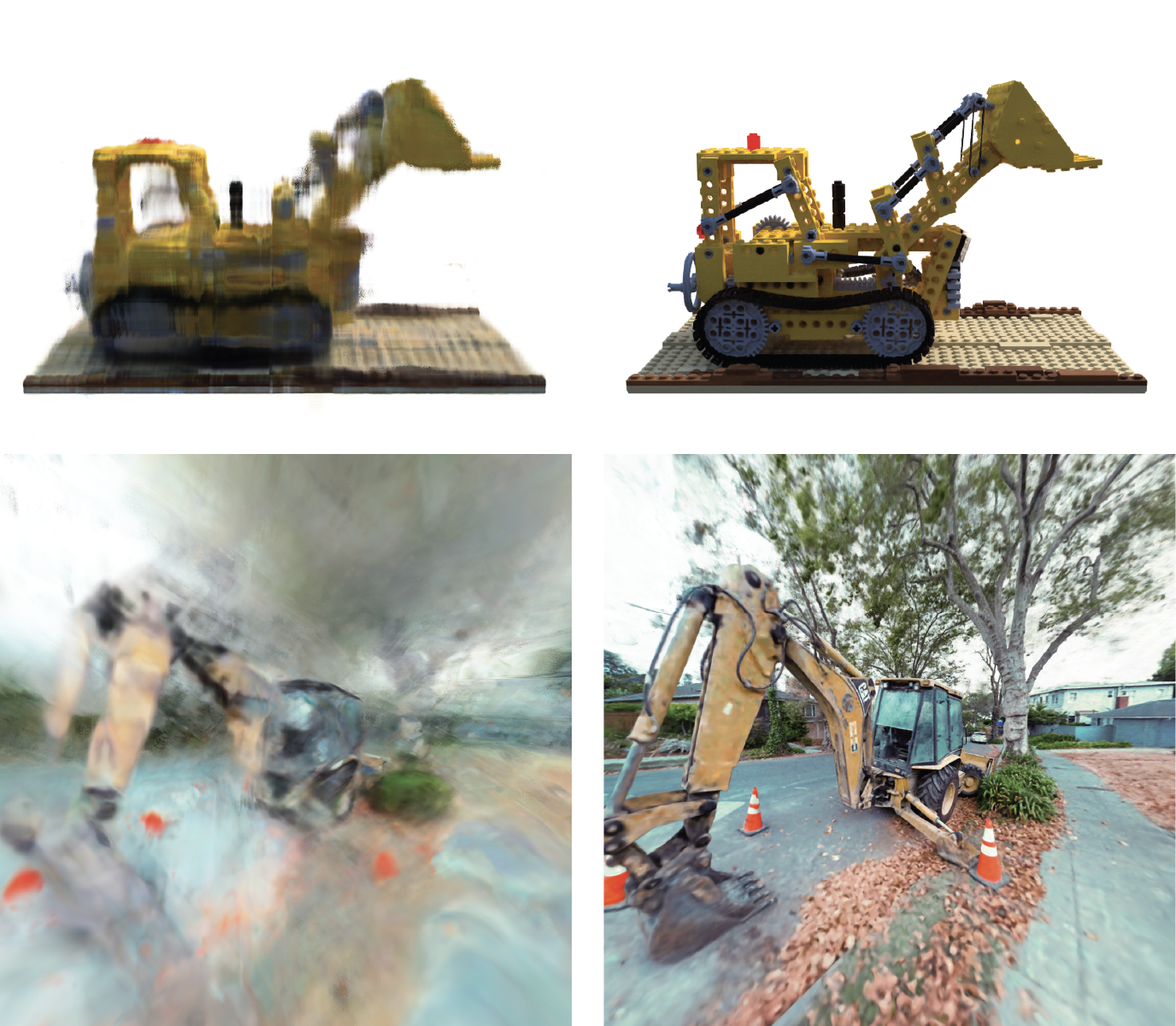}

   \caption{
       \textbf{
           Two-phase optimization.
       }
       Two \ours{} neural fields are trained: the first using a rank-constrained
       \textit{bottleneck} representation (left); all parameters are discarded except for the projection parameters
       $\projparams_\text{bneck}$, which are used for initialization of the final
       representation (right). %
    }
   \label{fig:two_phase}
\end{figure}

\textbf{Dynamic low-pass filtering.}
Similar to prior work~\cite{wang2022rodin,chen2022tensorf}, we encode features interpolated in \ours{}'s RGB and SDF experiments with a Fourier embedding~\cite{tancik2020fourfeat}.
When these features are used, we adopt the coarse-to-fine strategy proposed for learning deformation in Nerfies~\cite{park2021nerfies} and for camera registration in BARF~\cite{lin2021barf}.
Given $J$ frequencies, we weight the $j$-th frequency band via:
\begin{equation*}
    w^j_k(\eta_k) = \frac{1 - \cos(\pi\text{clamp}(\eta_k - j, 0, 1))}{2}
\end{equation*}
where $k$ is the training step count and $\eta_k$ is interpolated from a linear schedule $\in [0, J]$.

\textbf{Two-phase optimization.}
Effective recovery of $\projparams$ is coupled with the rank of latent decompositions.
As rank is increased, high-frequency signals become easier to express and overfit to;
as a target signal becomes more explainable without a well-aligned latent structure, optimizers have less incentive to push $\projparams$ toward improved solutions.

To disentangle the $\projparams$ recovery from the capacity of latent feature volumes in radiance field experiments, we apply a two-phase strategy inspired by structure from motion, where techniques like the 8-point algorithm can be used to initialize Newton-based bundle adjustment.
In the first phase, we train a hybrid field using a channel-limited CP decomposition, which has limited representational capacity.
This produces ``bottlenecked'' MLP decoder parameters $\decoderparams_\text{bneck}$, feature grid parameters $\voxelparams_\text{bneck}$, and projection parameters $\projparams_\text{bneck}$.
We discard all parameters but $\projparams_\text{bneck}$, and then simply set:
\begin{equation*}
    \projparams_\text{init} = \projparams_\text{bneck}
\end{equation*}
to initialize the final, more expressive neural field. %
Example reconstructions after each phase are displayed in Figure~\ref{fig:two_phase}.

\section{Experiments}
\label{sec:experiments}

\newcommand{\SO}{\text{SO}}
\newcommand{\so}{\mathfrak{so}}

\subsection{2D Image Reconstruction}

\begin{figure}[t]
  \centering
  \includegraphics[width=0.8\linewidth]{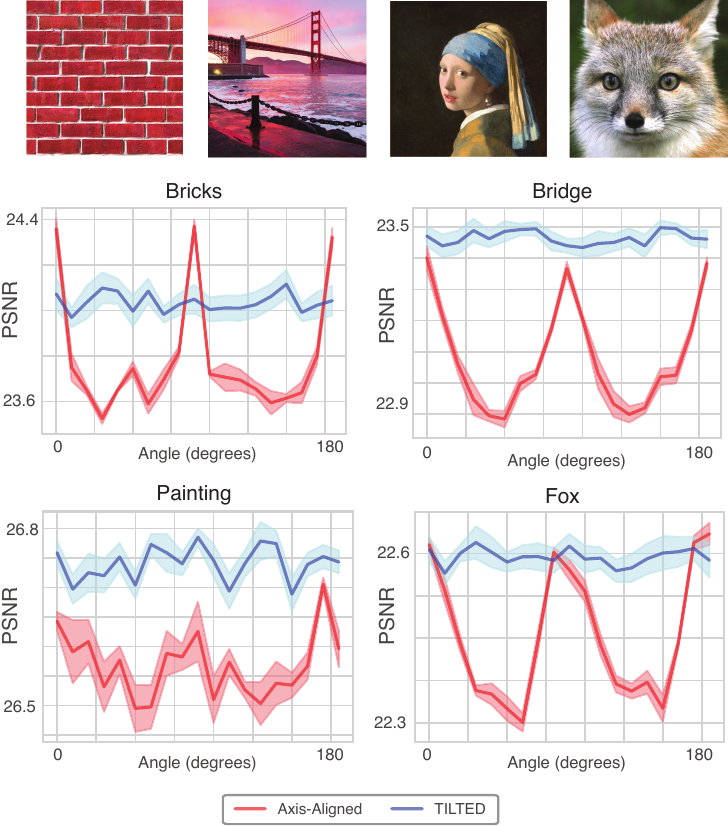}

   \caption{
        \textbf{Evaluation images and results for 2D image reconstruction.}
        We apply rotations to each input image, and plot holdout PSNR for a model trained at each angle.
        Axis-aligned feature decompositions are sensitive to transformations of the input, while \ours{} retains a constant PSNR across angles.
    }
   \label{fig:exp-2d-quadplot}
\end{figure}

To build intuition in a simple setting, we begin by studying \ours{} for 2D
image reconstruction with low-rank feature grids, analogous to our theoretical
studies in \Cref{sec:theory}.
To evaluate sensitivity to image orientation, we evaluate two model variants---with an axis-aligned decomposition and with a \ours{} decomposition---on four images
rotated by angles sampled uniformly between 0 and 180, at 10 degree intervals.
The setup of models can be interpreted as the 2D version of a CP decomposition-based neural field~\cite{chen2022tensorf,gao2023strivec}.
In the axis-aligned case,
latent grids are decomposed into $d=64$ vector pairs, where each vector $\in \mathbb{R}^{128}$.
The full latent grid can be computed by concatenating the outer products of each pair.
In the \ours{} variant, we introduce a set of $T=8$ 2D rotations, each of which are applied to $d / T$ vector pairs.
We run experiments that evoke the partially-observable nature of most reconstruction tasks: we fit a hybrid field with a 2-layer, 32-unit decoder
to a randomly subsampled half of the pixels in an image for training, and use the other half for evaluation. 
Results over 5 seeds are reported in Figure~\ref{fig:exp-2d-quadplot};
differences are mild compared to more complex tasks, but we nonetheless observe:

  \textbf{(1) \ours{} improves robustness.}
    When an axis-aligned decomposition is used, recovered PSNRs are more
    volatile, with a difference of as high as 1 PSNR for the \textit{Bricks}
    test image.  With the introduction of learned transforms, reconstruction
    quality becomes stable to input rotations.
    
\begin{figure}[t]
  \centering
  \includegraphics[width=.7\linewidth]{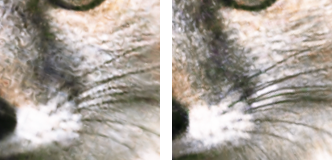}

   \caption{
       \textbf{Fine details without (left) and with (right) \ours{}.}
        The \ours{} reconstruction of the whiskers mitigates artifacts from axis-aligned factors.
    }
   \label{fig:whisker_zoom}
\end{figure}

  \textbf{(2) \ours{} improves detail recovery.}
  We qualitatively evaluate results by zooming into reconstructed images in Figure~\ref{fig:whisker_zoom}.
  \ours{} improves reconstruction particularly in fine features like the
  whiskers, which are jagged and bottlenecked by the factorization in the
  axis-aligned case, but rendered with fewer artifacts when we apply \ours{}.

\subsection{Signed Distance Field Reconstruction}

\begin{table}[t]
	\begin{center}
            \small
            \begin{tabular}{lllllll}
	       \toprule
	        IoU $\uparrow$ & 30 & 60 & 90\\
            \cmidrule(r){1-1} \cmidrule(l){2-4}
	       K-Planes & 0.949{\scriptsize$\pm$0.015} & 0.952{\scriptsize$\pm$0.015} & 0.952{\scriptsize$\pm$0.016}\\
              w/ \ours{} & \textbf{0.989{\scriptsize$\pm$0.002}} & \textbf{0.990{\scriptsize$\pm$0.002}} & \textbf{0.991{\scriptsize$\pm$0.002}}\\
              \midrule
	        IoU $\uparrow$ & 45 & 90 & 135\\
            \cmidrule(r){1-1} \cmidrule(l){2-4}
	       Vector-Matrix & 0.970{\scriptsize$\pm$0.007} & 0.979{\scriptsize$\pm$0.005} & 0.982{\scriptsize$\pm$0.003}\\
              w/ \ours{} & \textbf{0.982{\scriptsize$\pm$0.003}} & \textbf{0.989{\scriptsize$\pm$0.002}} & \textbf{0.988{\scriptsize$\pm$0.003}}\\
	       \bottomrule
            \end{tabular}
	\end{center}
	\caption{
             \textbf{Aggregated metrics across models used for SDF experiments.}
             Three channel count variations are used for each latent decomposition structure. \ours{} improves reconstructions consistently.
         }
	\label{table:sdf_summary_table}
\end{table}

Next, we study the impact of \ours{} on reconstruction of signed distance fields.
We follow the mesh sampling strategy used for studying signed distance fields in prior work~\cite{muller2022instant,chen2023factor} to produce a set of 8M training points and 16M evaluation points, and then train hybrid fields based on both VM and K-Plane decompositions.
Evaluation metrics are reported using intersection-over-union (IoU).

We sweep reconstructions based on both K-Planes and VM, with three channel
counts for each architecture, on 8 different object meshes. Each representation uses 3
resolutions---32, 128, and 256. For K-Planes, we use channel counts of 30, 60, and 90; for VM, we use channel counts of 45, 90, 135. %
All experiments use a 3-layer, 64-unit decoder and 5 transforms. We observe:

  \textbf{(1) Improved reconstructions across architectures and models.}
  We report the average IoU for eight objects in Table~\ref{table:sdf_summary_table}.
  \ours{} improves results for all decomposition and channel count variants.
  When we disaggregate results by object (Appendix~\ref{app:sdf_disaggregated}), \ours{} outperforms its axis-aligned
  counterpart in all but one (of 48) examples. %

\begin{figure}[t]
  \centering
  \includegraphics[width=.8\linewidth]{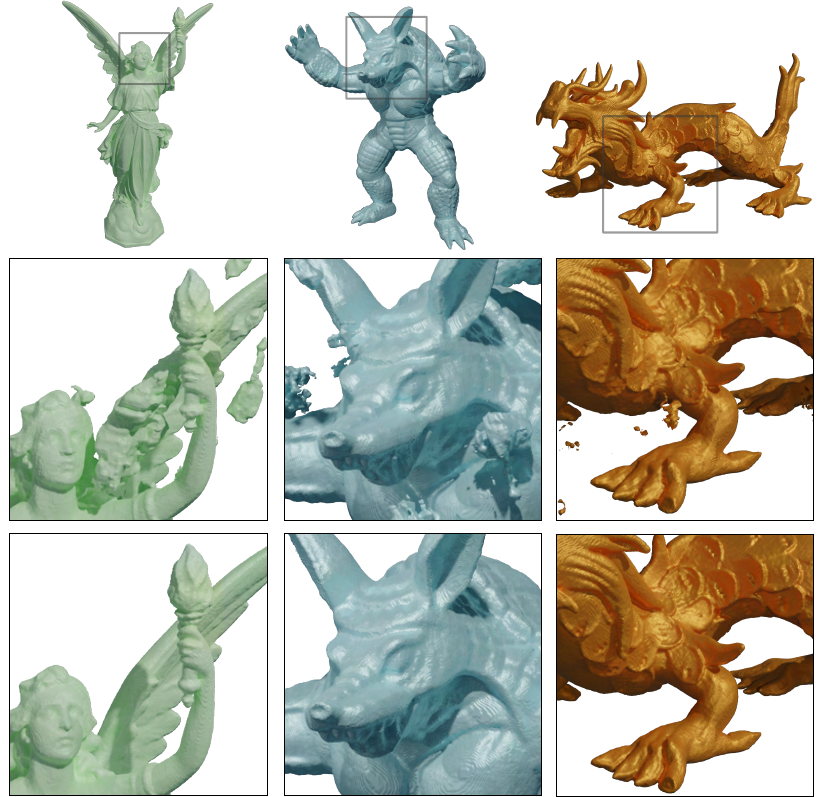}

   \caption{
        \textbf{Signed-distance field reconstruction before (above) and after (below) \ours{}.}
        \ours{} reduces floating artifacts without expressiveness-limiting
        regularization.
    }
   \label{fig:sdf_qualitative}
\end{figure}

  \textbf{(2) Implicit 3D regularization.}
  To better understand how \ours{} impacts SDF reconstruction, we apply marching cubes~\cite{lorensen1987marching} to learned fields after training.
  Qualitative examples are shown in Figure~\ref{fig:sdf_qualitative}.
  Renders reveal that the hybrid field architectures we use, which were proposed for and have not been extensively studied beyond the context of radiance fields, are prone to floating artifacts in recovered meshes.
  The typical solution for artifacts like these is to adjust the model size or regularization, for example to increase channel count or encourage spatial smoothness with total variation.
  We find that the canonical factors of \ours{} achieves a similar effect without expanding the
  factorization size or changing the optimized cost function.

\subsection{Neural Radiance Fields}

Our final set of experiments evaluate \ours{} for neural radiance fields applied to both synthetic and real data.

\subsubsection{Synthetic Study}

We begin with a quantitative study using the NeRF-Synthetic~\cite{mildenhall2020nerf} dataset.
While this dataset is commonly used for evaluation of neural field architectures~\cite{mildenhall2020nerf,chen2022tensorf,yu2021plenoxels,fridovich2023kplanes,chen2023factor,sun2021directvoxgo}, it is unrealistic because objects are rendered in Blender and perfectly axis-aligned.
The bricks of the Lego scene, for example, are aligned with the coordinate system that camera poses are defined in.
To understand how this impacts representations, we compare NeRF-Synthetic against the rotated variant used by~\cite{tancik2020fourfeat}.
We refer to this dataset as NeRF-Synthetic\rotsuffix{}, because we apply a uniformly sampled $\text{SO}(3)$ rotation to all camera poses.
Robustness against this operation is critical for real-world data, where
axis-alignment is rarely well-defined, let alone provided. %

For each of the NeRF-Synthetic and NeRF-Synthetic\rotsuffix{} datasets, we train every combination of:
  \textbf{(1)} two decompositions: VM and K-Planes, each 32 channels per factor,
  \textbf{(2)} three parameterizations of $\projparams$: axis-aligned (baseline), 4 transforms, and 8 transforms,
  \textbf{(3)} eight scenes: chair, drums, ficus, hotdog, lego, materials, mic, and ship, and
  \textbf{(4)} three random seeds: we use 0, 1, and 2. %
To eliminate the possibility of bounding box clipping artifacts interfering
with results, we use enlarged scene bounding boxes of $[-1.6, 1.6]$; this exerts a noticeable but uniform penalty on PSNR metrics relative to results with smaller standard bounding boxes.
We additionally incorporate the proposal fields, histogram loss, and distortion loss proposed by MipNeRF-360~\cite{barron2021mipnerf360}.
Our core conclusions are:

  \textbf{(1) Naive hybrid representations have strong axis-alignment biases.}
Results from the axis-aligned factorizations mirror our theoretical results in Section~\ref{sec:theory}.
When an axis-aligned decomposition is used, the quality of reconstructions becomes highly sensitive to the orientation of the target input.
In Table~\ref{table:naive_nerf_rot}, we observe as high as a 2 PSNR drop from scene rotation on the Lego dataset.
In contrast, \ours{} is designed with invariance in mind, and is thus robust to these transformations.  %

\textbf{(2) \ours{} improves reconstructions.}
On the NeRF-Synthetic\rotsuffix{} dataset, we observe performance increases from learned transforms, increasing the number of optimized transformations, and adopting two-phase optimization.
Table~\ref{table:tilted_blender} highlights how \ours{} improves PSNRs for the NeRF-Synthetic\rotsuffix{} dataset, while Table~\ref{table:nerf_ablation} demonstrates how components of our method (multiple transforms and two-phase optimization) improve results.

\begin{table}[t]
    \begin{center}
        \resizebox{\linewidth}{!}{%
            \begin{tabular}{lll}
                \toprule
                & K-Planes & VM\\
                \cmidrule(lr){2-2} \cmidrule(l){3-3}
                Lego & \textbf{35.31{\scriptsize$\pm$0.02}} $\to$ 33.29{\scriptsize$\pm$0.11} & \textbf{34.24{\scriptsize$\pm$0.04}} $\to$ 32.63{\scriptsize$\pm$0.01}\\
                Avg. & \textbf{32.12{\scriptsize$\pm$0.02}} $\to$ 31.62{\scriptsize$\pm$0.04} & \textbf{31.30{\scriptsize$\pm$0.03}} $\to$ 30.76{\scriptsize$\pm$0.03}\\
                \bottomrule
            \end{tabular}
        }
    \end{center}
    \vspace{-1em}
    \caption{
        \textbf{PSNR decrease of prior methods, before and after random scene rotation.}
        Metrics are reported from NeRF-Synthetic (standard, axis-aligned) $\to\ $ NeRF-Synthetic\rotsuffix (randomly rotated).
        Without \ours{}, a simple rotation of the scene coordinate frame can lead to as high as a 2 PSNR drop in performance.
     }
    \label{table:naive_nerf_rot}
\end{table}

\begin{table}[t]
    \begin{center}
        \resizebox{\linewidth}{!}{%
        \begin{tabular}{lllll}
            \toprule
            & K-Planes & w/ TILTED & VM & w/ TILTED \\
            \cmidrule(lr){2-3} \cmidrule(l){4-5}
            Lego & 33.29{\scriptsize$\pm$0.11} & \textbf{34.35{\scriptsize$\pm$0.07}} & 32.63{\scriptsize$\pm$0.01} & \textbf{33.90{\scriptsize$\pm$0.06}}\\
            Avg. & 31.62{\scriptsize$\pm$0.04} & \textbf{31.91{\scriptsize$\pm$0.04}} & 30.76{\scriptsize$\pm$0.03} & \textbf{31.08{\scriptsize$\pm$0.02}}\\
            \bottomrule
        \end{tabular}
        }
    \end{center}
    \vspace{-1em}
    \caption{
        \textbf{PSNR improvement after incorporating \ours{}, on the NeRF-Synthetic\rotsuffix{} dataset.}
        \ours{} offers transform-invariant reconstruction quality and moderate PSNR improvements.
     }
    \label{table:tilted_blender}
\end{table}

\begin{table}[t]
    \begin{center}
        \begin{tabular}{rll}
            \toprule
            & 8 transforms & 4 transforms\\
            \cmidrule(lr){2-2} \cmidrule(l){3-3}
            Two-Phase & \textbf{34.35{\scriptsize$\pm$0.07}} & 34.19{\scriptsize$\pm$0.22}  \\
            Without & 33.95{\scriptsize$\pm$0.15} & 33.83{\scriptsize$\pm$0.08}  \\
            \bottomrule
        \end{tabular}
    \end{center}
    \vspace{-1em}
    \caption{
        \textbf{Ablations on the Lego synthetic dataset.}
        Two-phase optimization and an increased number of transforms synergistically improve reconstruction quality.
        Similar but weaker trends can be found in less structured scenes.
        Reported metrics use the K-Planes model.
     }
    \label{table:nerf_ablation}
\end{table}

\subsubsection{Real-World Study}

\begin{figure*}[hbtp]
  \vspace{2em} %
  \centering
  \begin{subfigure}{.475\linewidth}
      \centering
      \includegraphics[width=\linewidth]{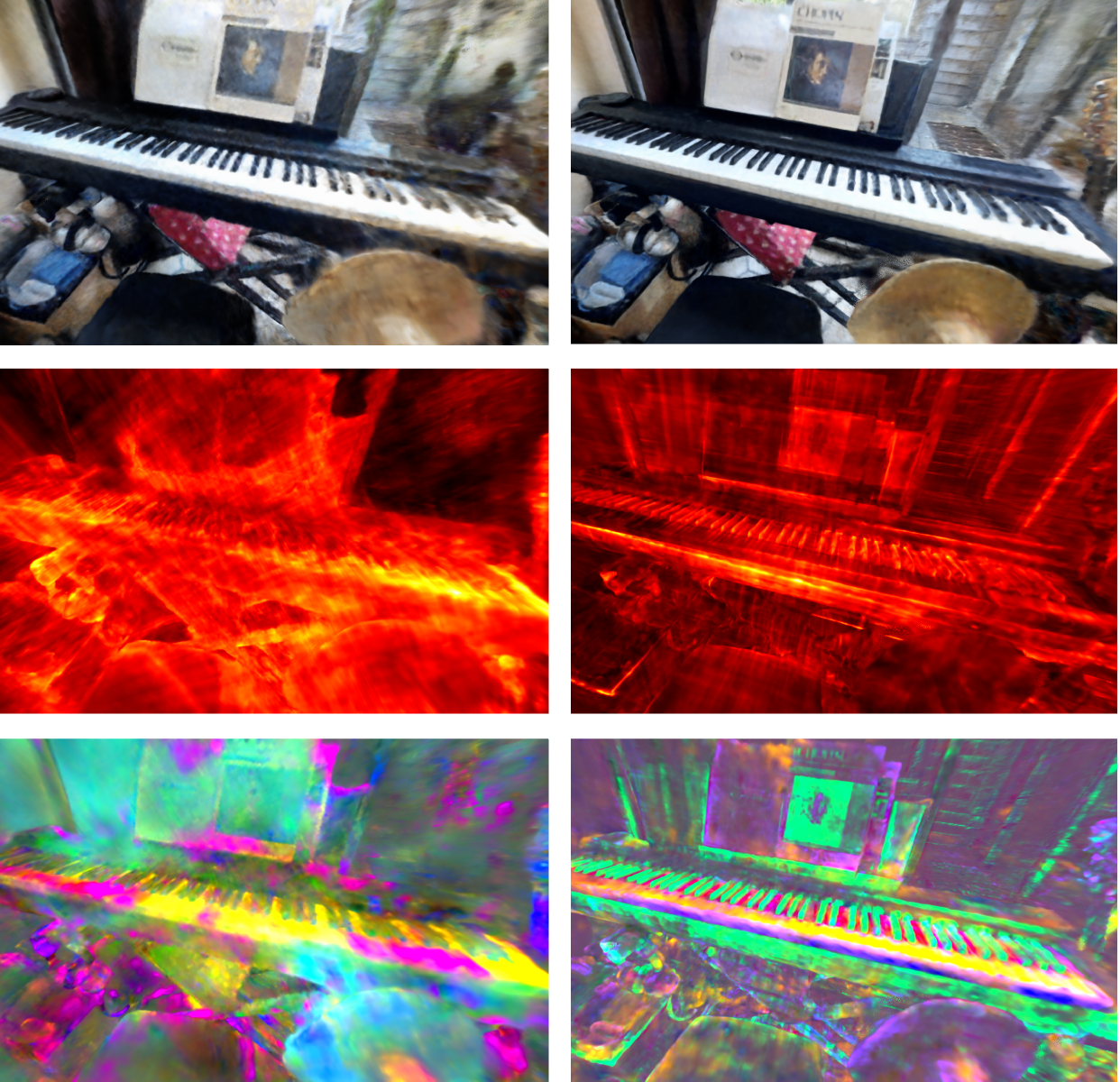}
      \caption{Kitchen}
  \end{subfigure}
  \hfill
  \begin{subfigure}{.475\linewidth}
      \vspace{1em}
      \centering
      \includegraphics[width=\linewidth]{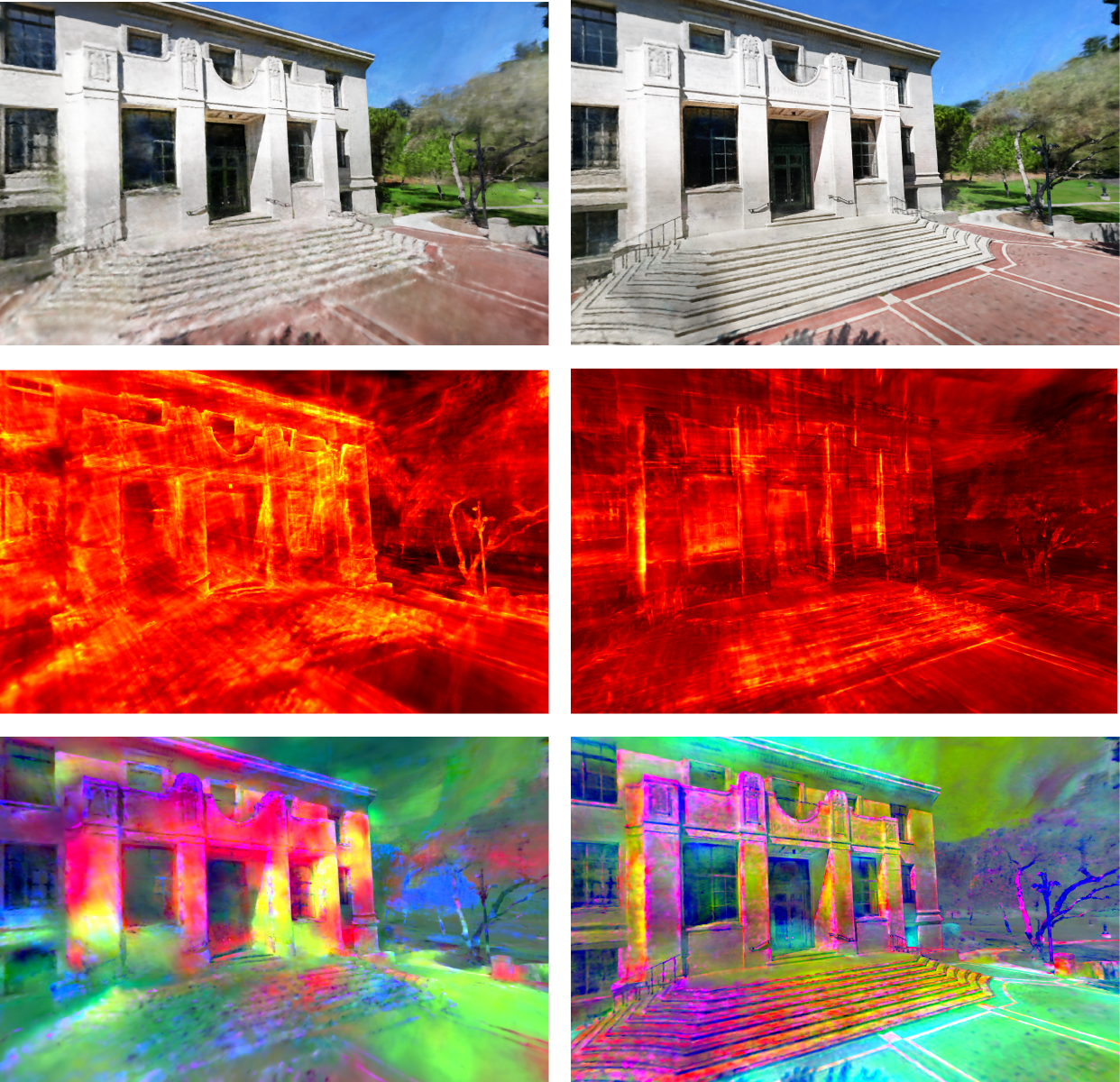}
      \caption{Giannini}
  \end{subfigure}
  \begin{subfigure}{.475\linewidth}
      \vspace{1em}
      \centering
      \includegraphics[width=\linewidth]{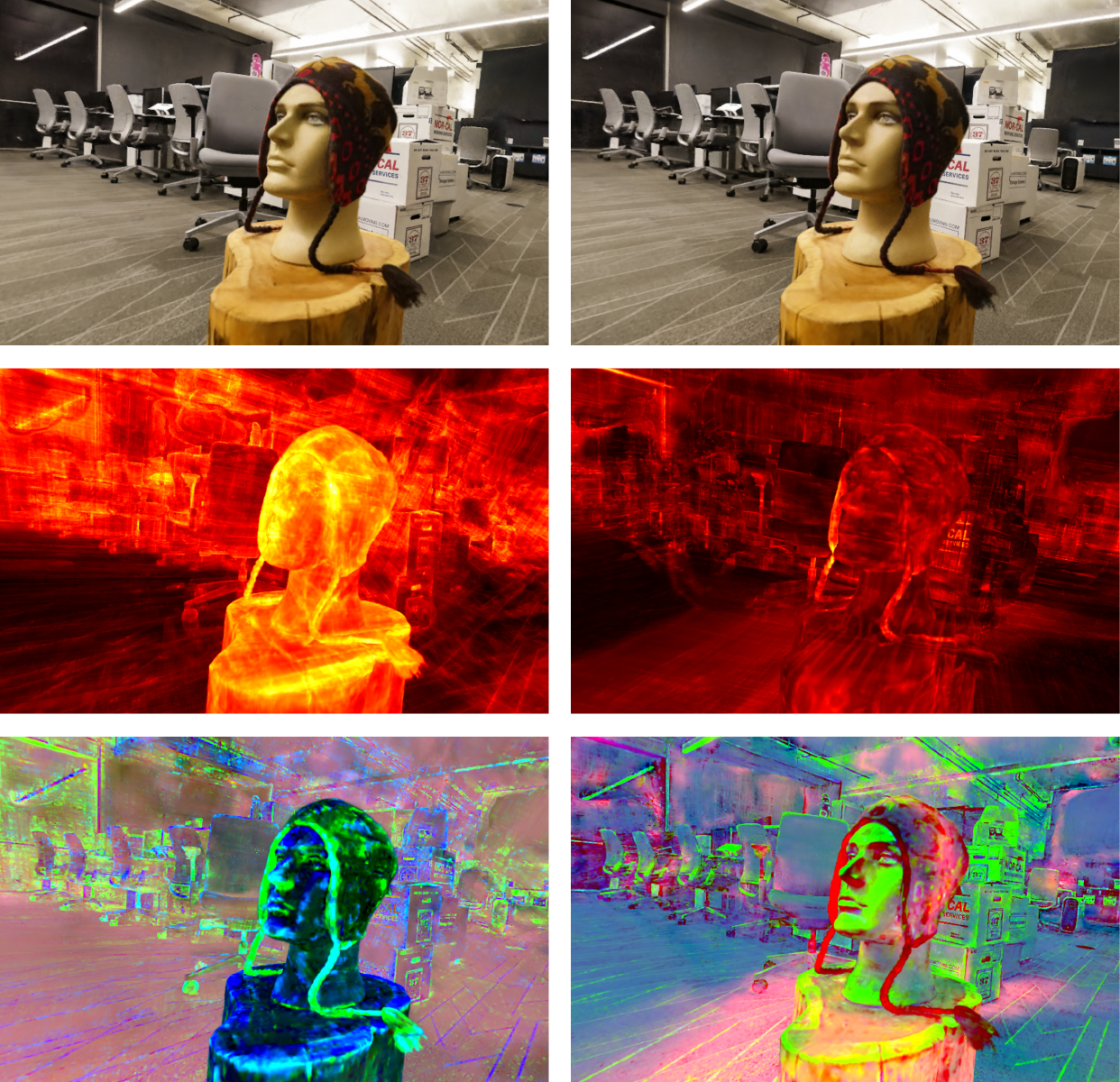}
      \caption{Stump}
  \end{subfigure}
  \hfill
  \begin{subfigure}{.475\linewidth}
      \vspace{1em}
      \centering
      \includegraphics[width=\linewidth]{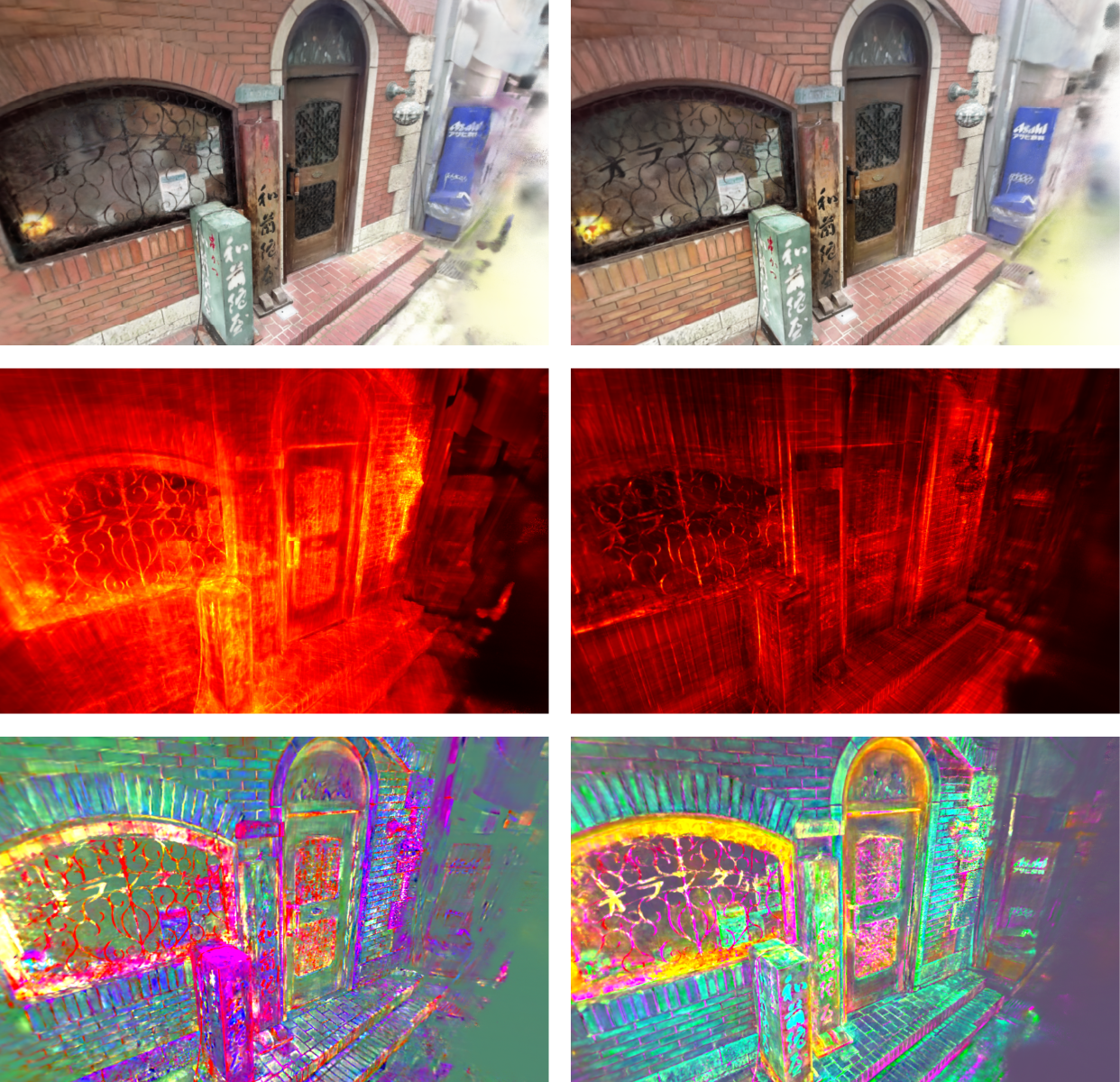}
      \caption{Storefront}
  \end{subfigure}
  
   \caption{
        \textbf{Real-world radiance field comparisons, before (left) and after (right) \ours{}.}
        For each scene, we arrange in three rows the outputs of \textbf{(i)} rendering RGB images, \textbf{(ii)} visualizing the structure-revealing $\ell^2$-norm of interpolated features, and \textbf{(iii)} mapping the top three principal components of interpolated features to RGB.
        TILTED feature volumes result in better reconstruction quality, with more structured, interpretable, and expressive features.
        Results in this figure are from K-Planes.
    }
   \label{fig:real_world_nerf}
   \vspace{2em} %
\end{figure*}

In our final set of experiments, we apply \ours{} to 18 real-world scenes made available via Nerfstudio~\cite{tancik2023nerfstudio}.
We modify architectures with (a) an $\ell^\infty$ norm-based scene contraction (Equation~\ref{eq:scene_contraction}) to handle the unbounded nature of real-world data, (b) camera pose optimization to account for noisy camera poses, and (c) NeRF-W-style appearance embeddings~\cite{martin2021nerfw}. %
Once camera pose optimization and per-camera appearance embeddings are enabled, we lose the ability to reliably compute evaluation metrics~\cite{tancik2023nerfstudio}.
Instead, we examine how incorporating \ours{} impacts training PSNRs and qualitative results.

  \textbf{(1) On real-world data, \ours{} can simultaneously halve the parameter count of a model, accelerate training by 25\%, and improve reconstructions.}
In Table~\ref{table:nerfstudio_train_losses}, we compare standard factored neural field representations with two techniques for improving reconstructions: doubling the feature volume channel count and \ours{}.
Compared to an axis-aligned model of the same size, \ours{} improves reconstruction performance on all scenes.
It also outperforms axis-aligned models with 2x higher channel counts in most cases (72\% of the time for VM, 56\% for K-Planes), thus cutting parameter count by almost half while training 25\% faster (11:04 vs 14:46 for 30k steps).

\textbf{(2) Recovered transforms align factors to underlying scene geometry.}
In Figure~\ref{fig:real_world_nerf}, we visualize both renders and underlying latent features.
We display a norm-based latent visualization, which involves volume rendering a map of feature norms using standard NeRF densities, and a PCA-based approach, which maps latent vectors to RGB.
Evoking the model problem result of \Cref{thm:ntilt-informal}, \ours{} factors interpretably align themself to the geometry of scenes while enabling more detailed and expressive feature volumes.

\textbf{(3) Standard evaluations incentivize axis-alignment biases.}
Despite significantly outperforming axis-aligned baselines on both real-world data and NeRF-Synthetic, we note that \ours{} underperforms against baselines on the axis-aligned NeRF-Synthetic dataset.
This hints at room for further performance optimizations of our method, while highlighting flaws in the way that radiance fields architectures are often evaluated. %
Aiming to improve standard evaluation metrics (like PSNR on the NeRF-Synthetic dataset) can end up undermining real-world capabilities.

\begin{table}[t]
	\begin{center}
        \resizebox{\linewidth}{!}{%
\begin{tabular}{lcc}
\toprule
Dataset & K-Plane / 2x / TILTED & VM / 2x / TILTED\\
\cmidrule(r){1-1} \cmidrule(lr){2-2} \cmidrule(l){3-3}
Kitchen         & 25.95 / 26.91 / \textbf{27.12} & 25.63 / 26.54 / \textbf{26.90} \\
Floating   & 24.58 / \textbf{25.17} / 25.06 & 24.03 / 24.70 / \textbf{25.04} \\
Poster          & 33.14 / 33.71 / \textbf{33.79} & 32.84 / 33.49 / \textbf{33.61} \\
Redwoods       & 23.55 / 24.08 / \textbf{24.12} & 23.22 / 23.81 / \textbf{23.85} \\
Stump           & 26.82 / \textbf{27.29} / 27.28 & 26.33 / 26.83 / \textbf{26.97} \\
Vegetation      & 21.62 / 22.10 / \textbf{22.10} & 21.11 / 21.55 / \textbf{21.73} \\
BWW    & 24.64 / \textbf{25.06} / 24.95 & 24.22 / 24.75 / \textbf{24.80} \\
Library         & 25.24 / 25.68 / \textbf{25.78} & 25.50 / 25.78 / \textbf{25.84} \\
Storefront      & 29.71 / \textbf{30.12} / 29.87 & 29.15 / 29.77 / \textbf{29.87} \\
Dozer           & 22.37 / \textbf{22.88} / 22.69 & 21.91 / \textbf{22.46} / 22.40 \\
Egypt           & 20.69 / 21.10 / \textbf{21.12} & 20.84 / \textbf{21.17} / 21.09 \\
Person          & 24.83 / 24.93 / \textbf{25.36} & 25.28 / 25.38 / \textbf{25.39} \\
Giannini   & 20.51 / \textbf{20.90} / 20.82 & 20.27 / \textbf{20.64} / 20.60 \\
Sculpture       & 23.20 / 23.40 / \textbf{23.40} & 22.86 / 23.07 / \textbf{23.28} \\
Plane           & 22.75 / 23.00 / \textbf{23.01} & 22.53 / \textbf{22.84} / 22.74 \\
Aspen           & 15.99 / 16.20 / \textbf{16.21} & 15.98 / 16.15 / \textbf{16.20} \\
Desolation      & 22.14 / \textbf{22.40} / 22.25 & 21.88 / 22.11 / \textbf{22.12} \\
Campanile       & 24.27 / \textbf{24.64} / 24.37 & 23.97 / \textbf{24.35} / 24.19 \\
\bottomrule
\end{tabular}
        }
	\end{center}
        \vspace{-1em}
	\caption{
             \textbf{For real-world data, TILTED improves PSNRs on all evaluated scenes, typically outperforming even much larger axis-aligned models.}
             We compare: standard hybrid neural fields (K-Plane, VM), axis-aligned fields with channel counts doubled (2x), and the fields with the original channel count but addition of TILTED (TILTED).
         }
	\label{table:nerfstudio_train_losses}
\end{table}

\section{Conclusion}
\label{sec:discussion}
We demonstrate the importance of alignment for factored feature volumes via \ours{}, an extension to existing hybrid neural field architectures based on the idea of \textit{canonical factors}.
By aligning to and thus capturing the structure of a scene, \ours{} enables improvements across reconstruction detail, compactness, and runtime.
We also developed the theoretical foundations for this methodology; 
our analysis can be viewed as providing the first provable guarantee for
explicit disentangled representation learning with visual data beyond
spatial deconvolution (e.g., \cite{Kuo2020-pr}), here disentangling appearance and pose.

Many directions exist for extending our work, both practically and
theoretically.
On the practical side, these include further explorations of
convergence characteristics, more diverse families of
transformations, and ``scaling laws'' --- how methods like \ours{} interact with larger representations or scenes.
On the theoretical side, future work includes extending our results to overparameterized models,
MLPs, and scenes with visual clutter.
We also note that our work compares \ours{} neural fields only to their axis-aligned equivalents: while these representations have unique advantages over alternatives, many applications will still benefit from alternative techniques~\cite{muller2022instant,kerbl20233d}.

\textbf{Acknowledgements.}
This material is based upon work supported by the National Science Foundation
Graduate Research Fellowship Program under Grant DGE 2146752. YM
acknowledges partial support from the ONR grant
N00014-22-1-2102, the joint Simons Foundation-NSF DMS grant 2031899, and a
research grant from TBSI.
The authors thank Justin Kerr, Chung
Min Kim, Sara Fridovich-Keil, Druv Pai, and members of the Nerfstudio team for implementation references, technical discussions, and suggestions.

{\small
   \printbibliography
}

\newpage
\onecolumn
\appendices

\section{Implementation Details}

\subsection{Tangent-space optimization}
Due to manifold constraints, rotations cannot be naively optimized using standard first-order optimizers.
In \ours, we address this via a Riemannian ADAM~\cite{becigneul2018riemannian} approach.
Each $\projparams_t$ is stored as a unit-complex vector ($\in \Sph^1$) for 2D
experiments and as a unit quaternion ($\in \Sph^3$) for 3D experiments, but gradients are computed
with respect to tangent spaces corresponding to the standard $\mathfrak{so}(2)$
and $\mathfrak{so}(3)$ Lie algebras.
ADAM~\cite{kingma2014adam} is applied to scale tangent-space gradients $\xi_t^k$ at each training step $k$, and an exponential map is used in place of addition to apply updates:
\begin{equation*}
    \projparams_{t,k+1} = \projparams{}_{t,k} \text{Exp}(\alpha_k \xi_{t,k})
\end{equation*}
where $\alpha_k$ is the learning rate for $\projparams{}$ at step $k$.
For experiments with real world data, we refine camera poses using this same mechanism.

\subsection{Handling boundaries}
One benefit of axis-aligned latent decompositions is that they make bounding boxes intuitive: all coordinates used for interpolation can be constrained to lie within a well-defined input domain.
When we apply geometric transformations to the domain of factors, however, the regions of the input space that each factor covers stop fully overlapping.
To resolve this for bounded scenes, we apply simple coordinate clipping.
Toroidal boundary conditions, similar to what is used in Factor Fields~\cite{chen2023factor}, can also be used.
For unbounded scenes, we adopt an $\ell^\infty$ norm-based scene contraction function~\cite{barron2021mipnerf360,tancik2023nerfstudio}:

\newcommand{\positionnorm}{\| \position \|_\infty}
\begin{equation}
    \begin{split}
        \text{contract}(\position) &= \begin{cases}\
            \position & \positionnorm \leq 1\\
            (2 - \frac{1}{\positionnorm})
            (\frac{\position}{\positionnorm}) & \positionnorm > 1
        \end{cases}
    \end{split}
    \label{eq:scene_contraction}
\end{equation}
When applied \textit{after} $\projparams$, note that scene contraction places all points in the range $[-2, 2]$, which mitigates boundary concerns entirely.

\subsection{Regularization}
\label{sec:app-regularization}
We implement two standard regularization terms: spatial total variation (TV) on feature grids and the distortion loss proposed by MipNeRF 360~\cite{barron2021mipnerf360}.
NeRF experiments additionaly rely on a pair of proposal fields, which require an additional interlevel loss~\cite{barron2021mipnerf360}.
We also found benefit in including a sparsity-encouraging regularization term based on the $\ell_{2,1}$ matrix norm.
This can be interpreted as forming a matrix $\mathbf{A}$ with columns $(\mathbf{a}_{1}, \dots, \mathbf{a}_{F * T})$, where each column vector $\mathbf{a}_i$ contains parameters that correspond to a unique transformation and factor pair $\projparams_t$, $\FeatureGrid_f$.
The final regularization term is computed by summing the $\ell_2$ norms of each column vector.

All coefficients and additional implementation details can be found in our code release.

\section{Concretizing Factored Feature Volumes}
\label{sec:unified_interface_app}
In this section, we concretize how feature volume decompositions used by prior work can be instantiated using the common notation that we present:
\begin{equation}\label{eq:factored_hybrid_field}
    \begin{split}
    \latent
    &= \texttt{Reduce}\bigl(
      \bigl[\Interp_{\FeatureGrid_1}(\Proj_{1}(\position))\bigr],
      \dots, \\
      &\hphantom{=\Reduce\bigl(\,\, }
      \bigl[\Interp_{\FeatureGrid_F}(\Proj_{F}(\position))\bigr]
    \bigr),
    \end{split}
\end{equation}
where, as earlier, $\position$ is an input coordinate and $\latent$ is an output that can be used to regress quantities like radiance or signed distance.
This unified formulation, which closely mirrors the structure of our implementation, enables integration of the latent registration mechanism proposed by \ours{} in a general-purpose way.

\subsection{Vector outer products}
\label{sec:cp_decomp}
Among the best-known approaches for factoring tensors is the classic CANDECOMP/PARAFAC (CP) decomposition, which has been studied as a baseline for factoring latent grids in prior work~\cite{chen2022tensorf}.
In 3D, the CP decomposition is equivalent to a single vector-matrix decomposition when the matrix rank is constrained to rank-$1$ and can thus be represented with a vector outer product.

To build CP-decomposed latent structures, a channel dimension is included to instantiate three paired 1D feature grids and projection functions:
\begin{equation*}
    \begin{split}
        \FeatureGrid_1 &\in \mathbb{R}^{w \times c}\\
        \FeatureGrid_2 &\in \mathbb{R}^{h \times c}\\
        \FeatureGrid_3 &\in \mathbb{R}^{d \times c}\quad
    \end{split}
    \begin{split}
        \TallProj_1(\position) &= \positionx \in \mathbb{R}\\
        \TallProj_2(\position) &= \positiony \in \mathbb{R}\\
        \TallProj_3(\position) &= \positionz \in \mathbb{R}
    \end{split}
\end{equation*}
Where $w$, $h$, and $d$ are the spatial dimensions of the voxel grid we aim to represent, and $c$ is a channel count.
After interpolation, an element-wise (Hadamard) product $\odot$ is used to reduce interpolated latents $\latent_1$, $\latent_2$, and $\latent_3$ into the final latent $\latent$:
\begin{equation}\label{eq:hadamard_reduce}
    \Reduce(\latent_1, \latent_2, \latent_3) = \latent_1 \odot \latent_2 \odot \latent_3
\end{equation}

\subsection{Tri-plane architectures}
\label{sec:triplane_decomp}
Beginning in generative 3D~\cite{chan2021eg3d,wang2022rodin}, several works have evaluated \textit{tri-plane} architectures for decomposing latent 3D grids.
The key idea of a tri-plane is to build feature grids along the XY, YZ, and XZ planes (Figure~\ref{fig:decomp_triplane}), which are dramatically more compact than a full 3D tensor and conducive to generative architectures developed for 2D image synthesis.
Using the notation described above, this can be concretized by setting $F=3$ and defining three axis-aligned factors and projection functions:
\begin{equation*}
    \begin{split}
        \FeatureGrid_1 &\in \mathbb{R}^{w \times h \times c}\\
        \FeatureGrid_2 &\in \mathbb{R}^{h \times d \times c}\\
        \FeatureGrid_3 &\in \mathbb{R}^{w \times d \times c}\quad
    \end{split}
    \begin{split}
        \TallProj_1(\position) &= (\positionx, \positiony) \in \mathbb{R}^2\\
        \TallProj_2(\position) &= (\positiony, \positionz) \in \mathbb{R}^2\\
        \TallProj_3(\position) &= (\positionx, \positionz) \in \mathbb{R}^2
    \end{split}
\end{equation*}
As described in the general case above, projected coordinates are used to interpolate per-projection latent vectors $\latent_1$, $\latent_2$, and $\latent_3$ from the corresponding set of feature grids $\FeatureGrid_1$, $\FeatureGrid_2$, and $\FeatureGrid_3$, which are passed through a $\Reduce$ operation to produce the final latent vector $\latent$.

Several choices exist for $\Reduce$.
EG3D~\cite{chan2021eg3d}, which adapts a StyleGAN2~\cite{karras2020stylegan2} architecture for 3D generation of faces and cats, uses element-wise summation:
\begin{equation*}
    \Reduce(\latent_1, \latent_2, \latent_3) = \latent_1 + \latent_2 + \latent_3
\end{equation*}
Rodin~\cite{wang2022rodin}, which adapts latent diffusion~\cite{rombach2021latentdiffusion} for 3D generation of avatars, adopts concatenation:
\begin{equation*}
    \Reduce(\latent_1, \latent_2, \latent_3) = \latent_1 \oplus \latent_2 \oplus \latent_3
\end{equation*}
Outside of generative models, K-Planes~\cite{fridovich2023kplanes} demonstrates that a Hadamard product for reduction is advantageous when applied with both linear and MLP decoders.
In \ours{}, we adopt the K-Planes naming for tri-plane architectures due to the use of product-based reduction.

\subsection{Vector-matrix pairs}
\label{sec:vm_decomp}
Rather than building a representation using only matrix components, TensoRF~\cite{chen2022tensorf} proposes a factorization of voxel grids using three vector-matrix (VM) pairs (Figure~\ref{fig:decomp_vm}).
The corresponding factors and projection functions can be formalized as:
\begin{equation*}
    \begin{split}
        \FeatureGrid_1 &\in \mathbb{R}^{w \times c}\\
        \FeatureGrid_2 &\in \mathbb{R}^{h \times d \times c}\\
        \FeatureGrid_3 &\in \mathbb{R}^{h \times c}\\
        \FeatureGrid_4 &\in \mathbb{R}^{w \times d \times c}\\
        \FeatureGrid_5 &\in \mathbb{R}^{h \times c}\\
        \FeatureGrid_6 &\in \mathbb{R}^{w\times h \times c}\quad
    \end{split}
    \begin{split}
        \TallProj_1(\position) &= \positionx\\
        \TallProj_2(\position) &= (\positiony, \positionz)\\
        \TallProj_3(\position) &= \positiony\\
        \TallProj_4(\position) &= (\positionx, \positionz)\\
        \TallProj_5(\position) &= \positionz\\
        \TallProj_6(\position) &= (\positionx, \positiony)
    \end{split}
\end{equation*}
The result is 6 interpolated latent vectors $\latent_{1 \dots 6}$.
Components from each vector-matrix pair are multiplied to produce 3 vectors, which are then concatenated:
\begin{equation*}
    \Reduce(\latent_{1 \dots 6}) = \bigoplus_{i=1,3,5} \bigl[ \latent_i \odot \latent_{i+1} \bigr]
\end{equation*}
After reduction, the latent $\latent$ is passed to an MLP decoder to regress quantities like radiance or signed distance.

\subsection{Multi-resolution factors}\label{sec:multiresolution_factors}

The decomposition architectures presented in Sections \ref{sec:cp_decomp}, \ref{sec:triplane_decomp}, and \ref{sec:vm_decomp} all assume that decompositions exist at only one resolution per scene.
In practice, it can be advantageous to aggregate features at varying spatial resolutions~\cite{muller2022instant,fridovich2023kplanes}.

Adapting the notation above to the multi-resolution setting is straightforward.
K-Planes, for example, runs all experiments at four resolutions: $64 \times 64$, $128 \times 128$, $256 \times 256$, and $512 \times 512$.
Generalizing to $R$ resolutions and per-resolution scale factor $s_r$, the process for interpolating multi-resolution K-Planes features can be written with our abstractions as:
\begin{equation*}
    \begin{split}
        \FeatureGrid_{r,1} &\in \mathbb{R}^{w_r \times h_r \times c}\\
        \FeatureGrid_{r,2} &\in \mathbb{R}^{h_r \times d_r \times c}\\
        \FeatureGrid_{r,3} &\in \mathbb{R}^{w_r \times d_r \times c}\quad
    \end{split}
    \begin{split}
        \TallProj_{r,1}(\position) &= (s_r\positionx, s_r\positiony)\\
        \TallProj_{r,2}(\position) &= (s_r\positiony, s_r\positionz)\\
        \TallProj_{r,3}(\position) &= (s_r\positionx, s_r\positionz)
    \end{split}
\end{equation*}
for $r={1 \dots R}$.
For the $\Reduce$ operator, the Hadamard product is applied within each resolution, and concatenation is applied across resolutions:
\begin{equation*}
    \begin{split}
        \Reduce(\{\latent_{r,i}\}_{r,i}) \quad= \bigoplus_{r=1\dots R}\bigl[ \latent_{r,1} \odot  \latent_{r,2} \odot  \latent_{r,3}  \bigr]
    \end{split}
\end{equation*}
\ours{} applies this pattern to all 3D experiments.

\section{Additional Results}

\subsection{Disaggregated SDF results}
\label{app:sdf_disaggregated}
\subsubsection{SDF results, with random scene rotation}

In this section, we report disaggregated results from our SDF reconstruction experiments, with and without \ours{}.
We apply a random global rotation to each mesh in these results.

\begin{table}[H]
	\begin{center}
        \begin{tabular}{cccccccccc}
	   \toprule
	       Methods & Avg & Bunny & Lucy & Chair & Armadillo & Dragon & Cheburashka & Beast & Happy \\
          \cmidrule(r){1-1} \cmidrule(l){2-2} \cmidrule(l){3-10}

	   K-Planes-30 & 0.949 & 0.969 & 0.933 & 0.937 & 0.952 & 0.935 & 0.980 & 0.922 & 0.967 \\
	   \ \ w/ \ours{} & 0.989 & 0.996 & \textbf{0.987} & 0.987 & 0.993 & 0.977 & 0.995 & 0.988 & 0.990 \\
	   K-Planes-60 & 0.949 & 0.982 & 0.939 & 0.922 & 0.955 & 0.922 & 0.979 & 0.918 & 0.978 \\
	   \ \ w/ \ours{} & 0.990 & 0.996 & 0.982 & \textbf{0.993} & 0.989 & 0.983 & \textbf{0.997} & 0.984 & \textbf{0.993} \\
	   K-Planes-90 & 0.946 & 0.967 & 0.951 & 0.898 & 0.946 & 0.929 & 0.989 & 0.913 & 0.974 \\
	   \ \ w/ \ours{} & \textbf{0.991} & \textbf{0.996} & 0.981 & 0.991 & \textbf{0.994} & \textbf{0.986} & 0.995 & \textbf{0.990} & 0.992 \\
	   \bottomrule
        \end{tabular}
	\end{center}
	\caption{\textbf{K-Plane results for SDF reconstruction \textit{with} random scene rotation.} We report IoUs with 30, 60, and 90 channels.}
\end{table}

\vspace{1.5em}
\begin{table}[H]
	\begin{center}
            \begin{tabular}{lccccccccc}
	       \toprule
	       Methods & Avg & Bunny & Lucy & Chair & Armadillo & Dragon & Cheburashka & Beast & Happy \\
	       \cmidrule(r){1-1} \cmidrule(l){2-2} \cmidrule(l){3-10}

	       VM-45 & 0.866 & 0.974 & 0.802 & 0.950 & 0.913 & 0.821 & 0.969 & 0.977 & 0.519 \\
	       \ \ w/ \ours{} & 0.974 & 0.994 & 0.973 & 0.936 & 0.988 & 0.952 & 0.979 & 0.981 & 0.991 \\
	       VM-90 & 0.946 & 0.982 & 0.956 & 0.948 & 0.984 & 0.762 & 0.981 & 0.972 & 0.985 \\
	       \ \ w/ \ours{} & 0.977 & 0.995 & \textbf{0.984} & 0.897 & \textbf{0.994} & 0.978 & \textbf{0.995} & 0.987 & 0.989 \\
	       VM-135 & 0.982 & 0.988 & 0.969 & 0.974 & 0.987 & 0.971 & 0.986 & \textbf{0.988} & 0.991 \\
	       \ \ w/ \ours{} & \textbf{0.988} & \textbf{0.996} & 0.982 & \textbf{0.976} & 0.992 & \textbf{0.981} & 0.994 & 0.987 & \textbf{0.994} \\
	       \bottomrule
            \end{tabular}
	\end{center}
	\caption{\textbf{Vector-matrix results for SDF reconstruction \textit{with} random scene rotation.} We report IoUs with 45, 90, and 135 channels.}
\end{table}

\subsubsection{SDF results, without random scene rotation}

In this section, we report SDF reconstruction metrics when we turn off random scene rotation.
Metrics here are similar to those when we include random scene rotation.
In the main paper body, we report metrics with random rotation included.

\begin{table}[H]
	\begin{center}
            \begin{tabular}{lccccccccc}
	       \toprule
	       Methods & Avg & Bunny & Lucy & Chair & Armadillo & Dragon & Cheburashka & Beast & Happy \\
	       \cmidrule(r){1-1} \cmidrule(l){2-2} \cmidrule(l){3-10}

	       K-Planes-30 & 0.949 & 0.970 & 0.945 & 0.965 & 0.945 & 0.843 & 0.989 & 0.970 & 0.966 \\
	       \ \ w/ \ours{} & 0.989 & 0.996 & 0.983 & 0.988 & 0.992 & 0.979 & 0.995 & 0.988 & 0.990 \\
	       K-Planes-60 & 0.952 & 0.972 & 0.954 & 0.964 & 0.951 & 0.842 & 0.993 & 0.972 & 0.969 \\
	       \ \ w/ \ours{} & 0.990 & \textbf{0.997} & 0.982 & \textbf{0.991} & 0.991 & \textbf{0.981} & \textbf{0.996} & 0.989 & \textbf{0.993} \\
   	     K-Planes-90 & 0.952 & 0.977 & 0.945 & 0.959 & 0.961 & 0.838 & 0.994 & 0.971 & 0.971 \\
              \ \ w/ \ours{} & \textbf{0.991} & 0.996 & \textbf{0.985} & 0.990 & \textbf{0.996} & 0.979 & 0.994 & \textbf{0.995} & 0.992 \\
	       \bottomrule
            \end{tabular}
	\end{center}
	\caption{\textbf{K-Plane results for SDF reconstruction \textit{without} random scene rotation.} We report IoUs with 30, 60, and 90 channels.}
\end{table}

\begin{table}[H]
	\begin{center}
              \begin{tabular}{lccccccccc}
	       \toprule

	       Methods & Avg & Bunny & Lucy & Chair & Armadillo & Dragon & Cheburashka & Beast & Happy \\
	       \cmidrule(r){1-1} \cmidrule(l){2-2} \cmidrule(l){3-10}

	       VM-45 & 0.970 & 0.990 & 0.927 & 0.975 & 0.970 & 0.952 & 0.988 & 0.981 & 0.980 \\
	       \ \ w/ \ours{} & 0.982 & 0.995 & 0.980 & 0.980 & 0.970 & 0.975 & 0.988 & 0.982 & 0.989 \\
	       VM-90 & 0.979 & 0.993 & 0.971 & \textbf{0.991} & 0.955 & 0.960 & 0.992 & 0.983 & 0.988 \\
	       \ \ w/ \ours{} & \textbf{0.989} & 0.995 & 0.985 & 0.989 & 0.993 & 0.976 & 0.993 & \textbf{0.987} & 0.991 \\
	       VM-135 & 0.982 & 0.993 & 0.973 & 0.987 & 0.991 & 0.964 & 0.977 & 0.981 & 0.989 \\
              \ \ w/ \ours{} & 0.988 & \textbf{0.996} & \textbf{0.985} & 0.989 & \textbf{0.994} & \textbf{0.983} & \textbf{0.997} & 0.966 & \textbf{0.993} \\
              \bottomrule
            \end{tabular}
	\end{center}
	\caption{\textbf{Vector-matrix results for SDF reconstruction \textit{without} random scene rotation.} We report IoUs with 45, 90, and 135 channels.}
\end{table}

\subsubsection{Ablations on coarse-to-fine optimization}

We report an ablation for the low pass-based coarse-to-fine optimization in Table~\ref{table:barf_ablation}.

\begin{table}[H]
	\begin{center}
		\begin{tabular}{lllllll}
			\toprule
			Methods & Lucy & Dragon \\
			\cmidrule(r){1-1} \cmidrule(l){2-3}
                K-Planes-90 \ours{} & \textbf{0.985} & \textbf{0.979}\\
                K-Planes-90 \ours{} w/o coarse-to-fine & 0.974 & 0.977 \\
			\cmidrule(r){1-1} \cmidrule(l){2-3}
                VM-135 \ours{} & \textbf{0.985} & \textbf{0.983}\\
                VM-135 \ours{}, w/o coarse-to-fine & 0.975 & 0.976\\
			\bottomrule
		\end{tabular}
	\end{center}
	\caption{
		\textbf{Ablation for coarse-to-fine optimization inspired by Nerfies~\cite{park2021nerfies} and BARF~\cite{lin2021barf}.}
		Coarse-to-fine optimization improves IoUs for \ours{} SDF reconstructions.
	}\label{table:barf_ablation}
\end{table}

\subsection{2D Results}
\subsubsection{Experiments on various latent grid resolutions}
In this section, we vary latent grid resolution for the 2D image reconstruction task.
\ours{} improves results across resolutions.
\begin{table*}[h!]
	\begin{center}
            \begin{tabular}{lllllll}
	       \toprule
              Grid Resolution & 32 & 64 & 128 & 256 & 512 & 1024 \\
              \cmidrule(r){1-1} \cmidrule(l){2-7}
               Fox (axis-aligned) & 21.26 & 21.98 & 22.31 & 21.63 & 17.23 & 10.34\\
               Fox (\ours{}) & \textbf{21.33} & \textbf{22.19} & \textbf{22.52} & \textbf{22.23} & \textbf{19.21} & \textbf{17.00}\\
              \cmidrule(r){1-1} \cmidrule(l){2-7}
               Bridge (axis-aligned) & 20.95 & 21.96 & 22.99 & 23.63 & 23.46 & 20.90\\
               Bridge (\ours{}) & \textbf{21.43} & \textbf{22.28} & \textbf{23.34} & \textbf{24.16} & \textbf{24.08} & \textbf{22.23}\\
              \cmidrule(r){1-1} \cmidrule(l){2-7}
               Painting (axis-aligned) & 25.59 & 26.16 & 26.51 & 26.76 & 26.40 & 18.22\\
               Painting (\ours{}) & \textbf{25.83} & \textbf{26.5} & \textbf{26.81} & \textbf{26.94} & \textbf{26.81} & \textbf{22.15}\\
	       \bottomrule
            \end{tabular}
	\end{center}
	\caption{
            \textbf{Varying latent grid resolutions for 2D image reconstruction.} 
        }
	\label{table:2D Various Grid Size}
\end{table*}

\section{Proofs for \Cref{sec:theory}}
\label{sec:proofs_base}
We assume throughout these appendices that $n \geq 2$.

\paragraph{Notation.} 

We write $\bbR$ for the reals, $\bbZ$ for the integers, and $\bbN$ for the positive integers.
For positive integers $m$ and $n$, we let $\bbR^{m}$ and $\bbR^{m \times n}$
denote the spaces of real-valued $m$-dimensional vectors and $m$-by-$n$
matrices %
(resp.). We write $\ve_{i}$, $\ve_{ij}$, etc.\ to denote the
elements of the canonical basis of these spaces, and $\One_m$ and $\Zero_{m,n}$ (etc.) to denote
their all-ones and all-zeros elements (resp.). 
We write $\ip{}{}$ and $\norm{}_F$ to denote the euclidean inner product
and associated norm of these spaces. We will write the $\ell^p$ norms
$\norm{\vx}_p = (\sum_i \abs{x_i}^p)^{1/p}$, with $\norm{\vx}_\infty= \max_i\,
\abs{x_i}$, of these spaces as either
$\norm{}_p$ or $\norm{}_{\ell^p}$ depending on context.
We will use the notation $\norm{}$ to denote the operator norm (the largest
singular value) on $m \times n$ matrices.
If $\vA \in \bbR^{m \times n}$, we write $\vA\adj \in \bbR^{n \times m}$ for
its (conjugate) transpose.
For matrices $\vA$ and $\vB$, we write $\vA \kron \vB$ to denote their tensor
product---if indices $(i, j)$ index $\vA$ and $(k, l)$ index $\vB$, we have
$(\vA \kron \vB)_{ijkl} = (\vA)_{ij} (\vB)_{kl}$.

As a technical tool (in \Cref{sec:svd_proofs}), and as a mathematical
abstraction (in \Cref{sec:ntilt_proofs}), we will frequently work with
``continuum'' images defined on the square $[-1, 1]^2 \subset \bbR^2$. 
By default, we will use ``image coordinates'' for $\vx \in \bbR^2$ (in order to
match the usual matrix-type indexing of discrete images), which corresponds in
the canonical basis to the positively-oriented frame $[-\ve_2, \ve_1]$. We will
formally write these coordinates as $\vx = (s, t)$.
For an image $X : \bbR^2 \to [0, 1]$ we will write $\norm{X}_{L^p} =
(\int_{\bbR^2} \abs{X(\vx)}^p \diff \vx)^{1/p}$ for the $L^p$ norms, and
$\norm{X}_{L^\infty} = \sup_{\vx \in \bbR^2}\, \abs{X(\vx)}$ when $X$ is
bounded. The space $L^2(\bbR^d)$ is a Hilbert space; as for
finite-dimensional vector spaces, we will write $\ip{}{}_{L^2}$ for its
associated inner product (which we take to be linear in its second argument),
and if $\sT : L^2 \to L^2$ is a bounded operator we will write $\norm{\sT}$ for
its (operator) norm and $\sT\adj$ for
its adjoint. Similarly, we will use notation defined above for matrix
operations for its analogous application to $L^2$ functions (e.g., tensor
products).
If $\vtau : \bbR^2 \to \bbR^2$ is a continuous function (e.g., a rotation of
the domain) and $X: \bbR^2 \to \bbR$ is an image, we write $X \circ \vtau$ for
their composition (the ``deformed image''). For sufficiently regular functions
$f, g : \bbR^2 \to \bbR$, we define their convolution $(f \ast g)(\vx) =
\int_{\bbR^2} f(\vx') g(\vx - \vx') \diff \vx'$; this operation is symmetric
and defines an element of $L^p$ when (say) $f$ is in $L^1$ and $g$ is in $L^p$.
We will use $\Ind{A}$ to denote the indicator function associated to an event
$A$ in a probability space; typically $A$ will be a subset of $\bbR^2$ (e.g.,
describing a continuous image) or a discrete set (e.g., describing the
Kronecker delta $\Ind{i = j}$ in summations).

Just as with discrete images, which can either be thought of as a function on
the discrete grid $\set{0, \dots, m-1} \times \set{0, \dots, n-1}$,
representing sampled intensity values, or a matrix (i.e., a finite-dimensional
operator) that aggregates those values, ``continuum images'' can also be
thought of as either functions or operators; if $f \in L^2(\bbR^2)$, we will
write $\sT_{f} : L^2(\bbR) \to L^2(\bbR)$ for the ``Fredholm operator''
associated to an $L^2$ function $f$, defined by $\sT_{f}[g] = \int_{\bbR}
f(\spcdot, x) g(x) \diff x$. %
If $\sT : L^2(\bbR) \to L^2(\bbR)$ is
bounded, we denote its Hilbert-Schmidt norm by $\norm{\sT}_{\HS} = (\sum_{n \in
\bbN} \norm{\sT u_n}_{L^2(\bbR)}^2)^{1/2}$, where $(u_n)_{n \in \bbN}$ is any
orthonormal basis of $L^2(\bbR)$; when $\sT_f$ is a Fredholm operator, we have
$\norm{\sT_f}_{\HS} = \norm{f}_{L^2(\bbR^2)}$, analogous to the Frobenius norm
of a matrix. We will exploit this correspondence in the sequel, often without
mentioning it, to identify a function $f \in L^2(\bbR^2)$ with its Fredholm
operator $\sT_f$ when convenient (c.f.\ \cite[\S B]{Heil2011-zk}); for example,
for $f \in L^2(\bbR)$ we will write $ff\adj : L^2(\bbR) \to L^2(\bbR)$ to
denote its induced Fredholm operator, which satisfies $ff\adj[g] = f
\ip{f}{g}_{L^2(\bbR)}$, and we will identify it with its $L^2(\bbR^2)$
representative satisfying $ff\adj(s, t) = f(s) f(t)$.
Consult the first few paragraphs
in \Cref{sec:svd_proofs} for specialized notation used in low-rank
approximation proofs, and the proof of \Cref{lem:l2-sampling} for notation used
in proofs that require harmonic analysis.

\paragraph{Problem setup.} 

We analyze a simple model problem that captures the improved efficiency of
\ours{} compared to competing approaches for compactly representing
non-axis-aligned scenes. Consider the following class of two-dimensional
greyscale images: let $m, n \in \bbN$ denote the image height and width, write
$\vc = [\tfrac{m-1}{2}, \tfrac{n-1}{2}]\adj$ for the image center (we use
zero-indexing), and define a centered square template by
\begin{equation}
   (\vX_{\grtr})_{ij} = \begin{cases}
      1 & \norm*{[i, j]\adj - \vc}_\infty \leq \alpha \min\set{c_0, c_1}    \\
      0 & \ow,
   \end{cases} 
   \label{eq:square_definition}
\end{equation}
where $0 < \alpha < 1$ controls the size of the square; we are interested in
$\alpha < 1/\sqrt{2}$, for a square that takes up a constant fraction of the
image pixels. We consider a rotational motion model for the square template
$\vX_{\grtr}$: for a parameter $\nu\in [0, 2\pi)$ corresponding to the rotation
about the image center $\vc$, let $\vtau_\nu : \bbR^2 \to \bbR^2$ denote the
(continuum) transformation corresponding to
\begin{equation}
   \begin{bmatrix}
      s \\
      t
   \end{bmatrix}
   \mapsto
   \begin{bmatrix}
      \cos \nu & -\sin \nu \\
      \sin \nu & \cos \nu
   \end{bmatrix}
   \left(
      \begin{bmatrix}
         s \\
         t
      \end{bmatrix}
      - \vc
   \right)
   + \vc,
   \label{eq:2d_rotation}
\end{equation}
and consider the class of observations 
\begin{equation}
   \mathfrak{S} = \set*{\vX \in \bbR^{m \times n} \given 
      X_{ij} = 
      \begin{cases}
         1 & \norm*{\vtau_{-\nu}(i, j) - \vc}_\infty \leq \alpha \min\set{c_0, c_1}    \\
         0 & \ow
      \end{cases} 
   }.
   \label{eq:template_class}
\end{equation}
In our lower bounds on low-rank compression in \Cref{sec:svd_proofs}, we will
work with a ``directly-sampled'' observation following the model
\Cref{eq:template_class}. In \Cref{sec:ntilt_proofs}, we will work in a
continuum idealization where it is more convenient to describe the observations
in a shifted coordinate system, which we now describe.

In our proofs, we will work in a shifted coordinate system so that the center
of the square \Cref{eq:square_definition} lies at the origin of the coordinate
system.
In particular, in these appendices we consider the image grid $\set{0,
1, \dots, m-1} \times \set{0, 1, \dots, n-1} - \vc$, corresponding to the grid
\begin{equation*}
   G_{\vc} = \set*{(i, j) \given i \in \set{-(m-1)/2, \dots, (m-1)/2}, 
   j \in \set{-(n-1)/2, \dots, (n-1)/2}}.
\end{equation*}
We will often %
index vectors and matrices by their coordinates in $G_{\vc}$ and
its derived grids, rather than in the standard image grid, due to the
straightforward one-to-one correspondence between grids.
Without loss of generality, we will assume that $m \leq n$. 
Let us then note that in $G_{\vc}$ coordinates, \Cref{eq:square_definition}
admits the equivalent rank-one expression
\begin{equation}
   \vX_{\grtr} = \vu_{\grtr} \vv_{\grtr} \adj,
   \quad
   (\vu_{\grtr})_i =
   \begin{cases}
      1 & \abs*{i} \leq \frac{\alpha}{2}(m-1) \\
      0 & \ow,
   \end{cases},
   \enspace
   (\vv_{\grtr})_j =
   \begin{cases}
      1 & \abs*{j} \leq \frac{\alpha}{2}(m-1) \\
      0 & \ow,
   \end{cases}.
   \label{eq:odd_size_lr_square}
\end{equation}
We will require, roughly, that $0 < \alpha < \tfrac{1}{\sqrt{2}}$, so that
there are %
no boundary issues with rotated versions of the template
\Cref{eq:odd_size_lr_square}. 

The template definition \Cref{eq:odd_size_lr_square} implies that as the image
size $m, n$ become large, $\vX_{\grtr}$ samples the same fixed continuum template
$X_{\grtr} : [-1, 1] \to \set{0, 1}$ defined by
\begin{equation}
   X_{\grtr}(s, t) = \Ind{\abs{s} \leq \alpha, \abs{t}\leq \alpha}.
   \label{eq:template-continuum}
\end{equation}
To make this correspondence, it is necessary to scale the grid $G_{\vc}$ by the
factor $2/(m-1)$: this corresponds to the grid
\begin{equation}
   G = \set*{(i, j) \given i \in \set*{-1, -1 + \frac{2}{m-1}, \dots, 1 -
      \frac{2}{m-1}, 1}, 
   j \in \set*{-\frac{n-1}{m-1}, \dots, \frac{n-1}{m-1}}}.
   \label{eq:image_ndc}
\end{equation}
It is then evident that if $(i, j) \in G$, one has $(\vX_{\grtr})_{(m-1)i/2,
(m-1)j/2} =
X_{\grtr}(i, j)$.

The possible complication that one may have rectangular images with $n > m$ is
actually not essential---to see this, note that we always have the block
structure
\begin{equation*}
   \vX_{\grtr} = \begin{bmatrix}
      \Zero & \bar{\vX}_{\grtr} & \Zero',
   \end{bmatrix}
\end{equation*}
where $\bar{\vX}_{\grtr}$ follows the definition \Cref{eq:odd_size_lr_square}, but
with $m = n$, and $\Zero$ and $\Zero'$ are zero matrices of appropriate sizes.
This shows that $\vX_{\grtr}$ and $\bar{\vX}_{\grtr}$ have the same nonzero singular
values, the same left singular vectors, and right singular vectors that are in
one-to-one correspondence (simply prepend and append the appropriate number of
zeros to the singular vectors of $\bar{\vX}_{\grtr}$). This implies that in our
proofs for the SVD approach in \Cref{sec:svd_proofs}, we may assume that $m =
n$ without any loss of generality.

\subsection{Proofs for \Cref{thm:svd-informal}}
\label{sec:svd_proofs}

As mentioned previously, without loss of generality we assume $m = n$ in this
section. We will therefore write $\vX \in \bbR^{n \times n}$ for the
observation, and use $m$ as a free parameter.

\paragraph{Problem setting.} 

We study the special case of $\nu_{\grtr} = \pi/4$, so that the observation
\begin{equation*}
   (\vX)_{ij} = \Ind{\norm{(\vtau_{\pi/4})_{ij}}_{\infty} \leq \alpha}
\end{equation*}
corresponds to a ``diamond''.
This case makes the rank of the transformed
image as large as possible.

\paragraph{Continuum surrogate.} Our analysis will proceed by relating the
singular value decomposition of $\vX$ %
to the spectrum of
an `infinite resolution' surrogate $X$, defined as
\begin{equation*}
   X(s, t) = X_{\grtr}(s\cos \nu_{\grtr} + t \sin \nu_{\grtr}, -s \sin
   \nu_{\grtr} + t \cos \nu_{\grtr}).
\end{equation*} 
Whereas $\supp X_{\grtr} = [-\alpha, \alpha]^2$, we have $\supp X = [-\sqrt{2}\alpha,
\sqrt{2}\alpha]^2$. The `infinite resolution' analogue of taking the singular
value decomposition of an image is the Schmidt decomposition (c.f.\
\cite{Chan2000-sc}) of the image's associated Fredholm operator: %
define $\sT_{X} : L^2([-1,+1]) \to L^2([-1,+1])$ by
\begin{equation*}
   \sT_{X}[f](s) =
   \int_{[-1, +1]} X(s, t) f(t) \diff t,
\end{equation*}
and note by the geometry of the diamond $X$ that
\begin{equation}
   \sT_{X}[f](s) =
   \int_{-(\sqrt{2}\alpha - \abs{s})}^{\sqrt{2}\alpha - \abs{s}} f(t) \diff t,
   \label{eq:diamond_iop}
\end{equation}
so that in particular $\sT_{X}$ is self-adjoint and Hilbert-Schmidt (hence
compact). The spectral theorem for compact operators on a Hilbert space
\cite{Heil2011-zk} then implies that $\sT_{X}$ diagonalizes in an orthonormal
basis of eigenfunctions $(e_k)_{k \in \bbN} \subset L^2([-1, +1])$ with
corresponding eigenvalues $(\lambda_k)_{k \in \bbN} \subset \bbR$:
\begin{equation}
   \sT_{X} = \sum_{k \in \bbN} \lambda_k e_k e_k\adj,
   \label{eq:integral_op_diagonalized}
\end{equation}
where the equality must be interpreted in the sense of $L^2 \to L^2$. We will 
derive a closed-form expression for \Cref{eq:integral_op_diagonalized} for the
diamond (\Cref{lem:diamond_diagonalized}), and use a truncation and
discretization of it as an approximate diagonalization of the discrete diamond
$\vX$.

\paragraph{Approximation guarantees with the SVD.} The use of an
infinite-dimensional surrogate to analyze $\vX$ requires the instantiation of
some approximation machinery. We quantify reconstruction performance in terms
of squared error. For any matrix $\vM \in \bbR^{n \times n}$, we
write $\sigma_1(\vM) \geq \sigma_2(\vM) \geq \dots \geq \sigma_n(\vM) \geq 0$
for its singular values. The singular value decomposition asserts that for any
$\vM$, there exist orthogonal matrices $\vU(\vM)$ and $\vV(\vM)$ such that
\begin{equation*}
   \vM =
   \vU 
   \underbrace{\begin{bmatrix}
         \sigma_1(\vM) & & \\
         & \ddots & \\
         & & \sigma_n(\vM)
   \end{bmatrix}}_{\vSigma}
   \vV\adj.
\end{equation*}
We recall that $\norm{\vM}_{\frob}^2 = \sum_{i=1}^n \sigma_i^2(\vM)$. The
``rank-$k$'' SVD approximation to $\vM$ is defined as\footnote{The ``scare
   quotes'' are to draw attention to the fact that if $\vM$ has rank strictly
   less than $k$, this approximation is not actually rank $k$---its rank is no
larger than $\rank(\vM)$.}
\begin{equation*}
   \mathrm{SVD}_k(\vM) = 
   \vU 
   \begin{bmatrix}
      \sigma_1(\vM) & & & \\
      & \ddots & & \\
      & & \sigma_k(\vM) & \\
      & & & \Zero_{n-k, n-k}
   \end{bmatrix}
   \vV\adj.
\end{equation*}
Following \cite{Bhatia1997-ly}, for $p \geq 1$ we write $\norm{\vM}_{(k)}^{(p)} =
\bigl(\sum_{i=1}^k \sigma_i^p(\vM)\bigr)^{1/p}$ for the Ky Fan $p$-norms of a
matrix $\vM$. These are indeed norms in the mathematical sense (e.g., 
\cite[\S IV.2, eqn.\ IV.47]{Bhatia1997-ly}). From the celebrated
Eckart-Young-Mirsky theorem \cite{Eckart1936-ep,Mirsky1960-ek}, we have
\begin{equation*}
   \inf_{\rank(\vM) \leq k}\,
   \norm*{
      \vM - \vX
   }_{\frob}^2
   =
   \sum_{i=k+1}^n \sigma_i^2(\vX)
   = \norm{\vX}_{\frob}^2 - \left(\norm{\vX}_{(k)}^{(2)}\right)^2,
\end{equation*}
and it is evident that $\vM = \mathrm{SVD}_k(\vX)$ achieves the infimum in this
formula: that is,
\begin{equation*}
   \norm*{\mathrm{SVD}_k(\vX) - \vX}_{\frob}^2 
   = \norm{\vX}_{\frob}^2 - \left(\norm*{\vX}_{(k)}^{(2)}\right)^2.
\end{equation*}
It follows that we can obtain lower bounds on the approximation error of
SVD-based compression of $\vX$ via upper bounds on the Ky Fan $2$-norms of
$\vX$.

For any $\vXi \in \bbR^{n \times n}$, we have from the triangle inequality
\begin{align*}
   \norm*{\vX}_{(k)}^{(2)}
   &\leq
   \norm{\vXi}_{(k)}^{(2)}
   +
   \norm*{\vXi - \vX}_{(k)}^{(2)} \\
   &\leq
   \norm{\vXi}_{(k)}^{(2)}
   +
   \norm*{\vXi - \vX}_{\frob},
   \labelthis \label{eq:kyfan-approx-ub}
\end{align*}
where the second inequality simply worst-cases over all $n$ singular values of
the residual. \Cref{eq:kyfan-approx-ub} is the basis of our approximation
argument: we will choose $\vXi$ as a matrix whose spectral decay is known, and
which gives a good approximation to the actual diamond matrix $\vX$. In
particular, we will consider a family of approximations $\vXi_m$, with $m \in
\bbN$, defined as
\begin{equation}
   (\vXi_m)_{ij} =
   \sum_{l=1}^m \lambda_l g_l(i) g_l(j),
   \label{eq:discretized-truncated-ests}
\end{equation}
with coordinates $(i, j) \in G$ and with notation as defined in
\Cref{lem:diamond_diagonalized}. 
We discuss the sources of error in these approximations momentarily; let us
first introduce additional notation to write these matrices more compactly.
Define $\vU_m \in \bbR^{n \times m}$ by 
\begin{equation*}
   (\vU_m)_{ij} = g_j(i); \quad i \in \set{k \given \exists l : (k, l) \in G},
   \enspace j \in [n]
\end{equation*}
(the slightly abstruse indexing notation simply defines the projection of $G$
onto either of its coordinate factors),
and let $\vLambda_m \in \bbR^{m \times m}$ be a diagonal matrix with
$\lambda_l$ on its $l$-th diagonal entry. Then 
\begin{equation}
   \vXi_m = \vU_m \vLambda_m \vU_m\adj.
   \label{eq:discretized-truncated-ests-mtx}
\end{equation}
For technical reasons, we will need to consider a further level of
approximation induced by smoothing the nonsmooth square pattern $X_{\grtr}$. For $\sigma^2 > 0$, we write
$\varphi_{\sigma^2}(t) = 1/\sqrt{2\pi\sigma^2} \exp(-\tfrac{1}{2 \sigma^2}t^2)$
for the one-dimensional standard gaussian, and $m_{\sigma^2} =
\varphi_{\sigma^2} \varphi_{\sigma^2}\adj$
for its two-dimensional analogue. Let $f \conv
g$ denote the convolution of $L^2(\bbR^d)$ signals $f$ and $g$.
Then define a smoothed family of approximations
\begin{equation}
   (\tilde{\vXi}_m)_{ij} =
   \sum_{l=1}^m \lambda_l (\varphi_{\sigma^2} \conv g_l)(i) 
   (\varphi_{\sigma^2} \conv g_l)(j).
   \label{eq:discretized-truncated-ests-moll}
\end{equation}
As above, let $\tilde{\vXi}_m$ denote the matrix representation of this
construction:
\begin{equation}
   \tilde{\vXi}_m = \tilde{\vU}_m \vLambda_m \tilde{\vU}_m\adj.
   \label{eq:discretized-truncated-ests-moll-mtx}
\end{equation}

Relative to the continuum diamond $X$, there
are three main sources of error in the approximations
\Cref{eq:discretized-truncated-ests-moll}. The parameter $m$ controls a truncation
of the infinite series of eigenfunctions that defines $\sT_{X}$, and the grid
resolution (proportional to $n$) controls a discretization error relative to the continuum image
$X$. In addition, the smoothing scale $\sigma^2$ controls a further error,
since the smoothed eigenfunctions do not coincide with eigenfunctions of the
`smoothed operator'. These three parameters are in tension---choosing $m$ larger recovers more terms
in the series defining $\sT_{X}$, but when the grid resolution is fixed at
$2/(n-1)$, the fact that the eigenfunctions $g_l$ become more and more
oscillatory at larger values of $l$ suggests a larger and larger discretization
error, and a need for a smaller and smaller smoothing scale $\sigma^2$ to avoid
destroying the spectral structure of the eigenfunctions $g_l$.
We will choose these parameters in tandem with the SVD rank $k$ in
\Cref{eq:kyfan-approx-ub} in order to guarantee as strong of a lower bound on
the approximation error as possible.

\paragraph{Main result.} Our main result is an inapproximability result for
sublinear low-rank approximations to $\vX$, up to a threshold. 

\begin{theorem}
   \label{thm:inapproximability}

   There are absolute constants $c, C, C' > 0$ such that the following holds. 
   Let $\nu_{\grtr} = \pi/4$ and $\alpha = 1/\sqrt{2}$, and consider the observation
   \begin{equation*}
      (\vX)_{ij} = \Ind{\norm{(\vtau_{\nu_{\grtr}})_{ij}}_{\infty} \leq \alpha}.
   \end{equation*}
   For every $n \geq \max \set{C, C'k^{1/9.5}}$, one has for every $\hat{\vX}
   \in \bbR^{n \times n}$ with rank no larger than $k$
   \begin{equation*}
      \frac{1}{n^2}\norm*{\hat{\vX} -
         \vX
      }_{\frob}^2
      \geq
      \frac{c}{1+k}.
   \end{equation*}

   \begin{proof}
      We instantiate the argument discussed in the previous paragraph,
      culminating in \Cref{eq:kyfan-approx-ub}. 
      Below, we will occasionally not calculate precise constants for simplicity, and
      similarly we will fix $\alpha = 1/\sqrt{2}$, allowing us to treat it as
      an absolute constant.
      Put
      \begin{equation*}
         \bar{X} = \varphi_{\sigma^2}^{\otimes 2} \conv X
      \end{equation*}
      for the smoothed observation (recall that $(\vX)_{ij} =
      X(i, j)$
      for $(i, j) \in G$), let $\bar{G} = (-1, -1) + \tfrac{2}{n-1} \bbZ^2$
      denote the infinitely-extended grid $G$ defined in \Cref{eq:image_ndc},
      and let $(\bar{\vX})_{ij} = \bar{X}(i, j)$ for $(i, j) \in \bar{G}$.
      The inclusion $G \subset \bar{G}$ means that we can naturally think of
      $\bar{\vX}$ as a matrix indexed by $G$ as well (via restriction), and we
      will write $\norm{}_{\ell^2(G)}=\norm{}_{\frob}$ and $\norm{}_{\ell^2(\bar{G})}$ to
      denote the respective norms.
      Observe that, by linearity of the convolution operation and
      \Cref{lem:diamond_diagonalized}, we have for $(i, j) \in \bar{G}$
      \begin{align*}
         (\bar{\vX})_{ij}
         &= (\varphi_{\sigma^2}^{\otimes 2} \conv X)(i, j) \\
         &=
         \sum_{l=1}^\infty \lambda_l (\varphi_{\sigma^2} \conv g_l)(i) 
         (\varphi_{\sigma^2} \conv g_l)(j) \\
         &=
         (\tilde{\vXi}_m)_{ij} + 
         \left(
            \underbrace{
               \sum_{l=m+1}^\infty \lambda_l
               (\varphi_{\sigma^2} \conv g_l)
               (\varphi_{\sigma^2} \conv g_l)\adj
            }_{\Delta_{\mathrm{tail}}}
         \right)(i, j).
         \labelthis \label{eq:barX_splitting}
      \end{align*}
      Let $\hat{\vX}$ be any approximation to $\vX$ with rank at most $k$.
      For technical convenience, we want to compare $\ell^2(G)$ norms to
      $\ell^2(\bar{G})$ norms---note that these are distinct when we consider
      our smoothed approximation $\bar{\vX}$, because convolution with the
      mollifier enlarges the support to be outside of $[-1, 1]^2$.
      We extend $\hat{\vX}$ and $\vX$ to all of $\bar{G}$ by zero-padding,
      and note that
      \begin{equation*}
         \norm*{\hat{\vX} - \vX}_{\frob}
         =
         \norm*{\hat{\vX} - \vX}_{\ell^2(G)}
         =
         \norm*{\hat{\vX} - \vX}_{\ell^2(\bar{G})}.
      \end{equation*}
      By the triangle inequality, we have
      \begin{equation*}
         \norm*{\hat{\vX} - \vX}_{\ell^2({G})}
         \geq
         \norm*{\hat{\vX} - \bar{\vX}}_{\ell^2({G})}
         -
         \norm*{\bar{\vX} - \vX}_{\ell^2({G})},
      \end{equation*}
      We can apply the EYM theorem to obtain
      \begin{equation}
         \norm*{\hat{\vX} - \bar{\vX}}_{\ell^2({G})}
         \geq
         \sqrt{
            \norm*{\bar{\vX}}_{\ell^2({G})}^2
            -
            \left(
               \norm*{
                  \bar{\vX}
               }_{(k)}^{(2)}
            \right)^2
         }.
         \label{eq:eym-to-plug}
      \end{equation}
      Notice that, by \Cref{eq:barX_splitting} and the fact that the Ky Fan
      $2$-norms are mathematically norms, we have 
      \begin{align*}
         \left(
            \norm*{
               \bar{\vX}
            }_{(k)}^{(2)}
         \right)^2
         &\leq
         \left(
            \norm*{
               \tilde{\vXi}_m
            }_{(k)}^{(2)}
            +
            \norm*{
               \vDelta_{\mathrm{tail}} 
            }_{(k)}^{(2)}
         \right)^2 \\
         &=
         \left(
            \norm*{
               \tilde{\vXi}_m
            }_{(k)}^{(2)}
         \right)^2
         +
         \left(
            \norm*{
               \vDelta_{\mathrm{tail}} 
            }_{(k)}^{(2)}
         \right)^2
         +2
         \norm*{
            \tilde{\vXi}_m
         }_{(k)}^{(2)}
         \norm*{
            \vDelta_{\mathrm{tail}} 
         }_{(k)}^{(2)} \\
         &\leq
         \left(
            \norm*{
               \tilde{\vXi}_m
            }_{(k)}^{(2)}
         \right)^2
         +
         \norm*{
            \vDelta_{\mathrm{tail}} 
         }_{\ell^2(G)}^2
         +2
         \norm*{
            \tilde{\vXi}_m
         }_{(k)}^{(2)}
         \norm*{
            \vDelta_{\mathrm{tail}} 
         }_{\ell^2(G)}
         \\
         &\leq
         \left(
            \norm*{
               \tilde{\vXi}_m
            }_{(k)}^{(2)}
         \right)^2
         +
         \norm*{
            \vDelta_{\mathrm{tail}} 
         }_{\ell^2(\bar{G})}^2
         +2
         \norm*{
            \tilde{\vXi}_m
         }_{(k)}^{(2)}
         \norm*{
            \vDelta_{\mathrm{tail}} 
         }_{\ell^2(\bar{G})}.
      \end{align*}
      Moreover, by
      \Cref{lem:diamond_approx_mollified_decay,lem:xim-tail-sum-decay}, we have
      \begin{align*}
         \left(
            \norm*{\tilde{\vXi}_m}_{(k)}^{(2)}
         \right)^2
         \leq
         \frac{n^2}{4}
         \left(
         4\alpha^2 - \frac{16\alpha^2}{\pi^2} \frac{1}{2 \min\set{m,k} + 1}
         \right)
         &+
         Cn(m(1 + \log m)^{1/2} + n\sigma^2 m^2)
         \\
         &\quad+
         C'(m^2 (1 + \log m) + n^2 \sigma^4 m^4).
      \end{align*}
      Meanwhile, by \Cref{eq:barX_splitting}, we have that
      $\Delta_{\mathrm{tail}}$ is in $L^1(\bbR^2)$, and its $L^1$ norm is no
      larger than that of $\bar{X}$.
      Applying \Cref{lem:l2-sampling} thus implies
      \begin{align*}
         \norm*{
            \vDelta_{\mathrm{tail}} 
         }_{\ell^2(\bar{G})}^2
         &\leq
         \frac{n^2}{4} \norm{\Delta_{\mathrm{tail}}}_{L^2}^2
         +
         \frac{C}{\sigma^4}(1 + n\sigma).
      \end{align*}
      By Young's inequality, we have that $\norm{\Delta_{\mathrm{tail}}}_{L^2}$
      is less than the corresponding tail sum without smoothing.
      Now notice that, by orthogonality,
      \begin{align*}
         \norm*{
            \sum_{l = m+1}^\infty \lambda_l g_l g_l \adj
         }_{L^2}^2
         &=
         \sum_{l = m+1}^\infty \lambda_l^2 \\
         &\leq
         \frac{32\alpha^2}{\pi^2} \frac{1}{2m + 1},
      \end{align*}
      following the arguments in the proof of \Cref{lem:xim-tail-sum-decay}
      (the last estimate assumes $m \geq 1$). Thus
      \begin{align*}
         \norm*{
            \vDelta_{\mathrm{tail}} 
         }_{\ell^2(\bar{G})}^2
         &\leq
         \frac{32\alpha^2n^2}{4\pi^2} \frac{1}{2m + 1}
         +
         \frac{1}{\sigma^4}(1 + n\sigma).
      \end{align*}
      Combining these estimates, we have
      \begin{align*}
         \left(
            \norm*{
               \bar{\vX}
            }_{(k)}^{(2)}
         \right)^2
         &\leq
         n^2 \alpha^2
         +
         \frac{4n^2}{\pi^2} \frac{1}{2m + 1}
         -
         \frac{2n^2/\pi^2}{2\min\set{m, k} + 1}
         \\
         &\quad+
         Cn(m(1 + \log m)^{1/2} + n\sigma^2 m^2)
         + C'm^2 (1 + \log m)
         C' n^2 \sigma^4 m^4
         +
         C''(1 + n\sigma)/\sigma^4
         \\
         &\quad+
         C'''\left(
            n + \sqrt{nm \log^{1/2} m} + n\sigma m 
            + m \sqrt{\log m} + n\sigma^2 m^2
         \right)
         \left(
            \frac{n}{\sqrt{m}} + \sqrt{\frac{1 + n\sigma}{\sigma^4}}
         \right).
      \end{align*}
      Inspecting these residuals with some foresight, it is clear that the
      smoothing-induced terms will typically dominate: these require that $m
      \sigma \ltsim 1$, but for the best lower bound we want $m$ to be large,
      whereas the residuals of size $n / \sigma^3$ penalize us for choosing
      $\sigma$ to be too small.
      We will make the choices $m = n^{4/19}$ and $\sigma = m^{-5/4}$.
      Evaluating the residual in the
      previous expression shows that for $n$ sufficiently large, there is an
      absolute constant $C>0$ such that
      \begin{equation*}
         \left(
            \norm*{
               \bar{\vX}
            }_{(k)}^{(2)}
         \right)^2
         \leq
         n^2 \alpha^2
         +
         \frac{4n^2}{\pi^2} \frac{1}{2m + 1}
         -
         \frac{2n^2/\pi^2}{2\min\set{m, k} + 1}
         + 
         C n^{36/19}.
      \end{equation*}
      Similarly, when $m \geq C k$ for a sufficiently large constant $C$, this
      bound is upper bounded by
      \begin{equation*}
         \left(
            \norm*{
               \bar{\vX}
            }_{(k)}^{(2)}
         \right)^2
         \leq
         n^2 \alpha^2
         -
         \frac{cn^2}{k + 1}
         + 
         C' n^{36/19}.
      \end{equation*}
      Now, plugging this estimate into \Cref{eq:eym-to-plug} after requiring
      that $k \leq c\sqrt{m} = c n^{2/19}$ for an absolute constant $c>0$, we have
      \begin{equation*}
         \norm*{\hat{\vX} - \vX_{\nu_{\grtr}}}_{\ell^2({G})}
         \geq
         \sqrt{
            \norm*{\bar{\vX}}_{\ell^2({G})}^2
            -
            n^2 \alpha^2
            + \frac{cn^2}{1 + k}
            - C' n^{36/19}
         }
         -
         \norm*{\bar{\vX} - \vX_{\nu_{\grtr}}}_{\ell^2({G})}.
      \end{equation*}
      We just need to estimate the remaining error terms.
      By \Cref{lem:smoothing-error}, we have for $m \geq 2^{2/3}$
      \begin{equation*}
         \norm*{\bar{\vX} - \vX}_{\ell^2(G)}^2 \leq
         \frac{n^2\sigma^8}{\pi^2}
         +
         \frac{2n}{\pi} + \frac{n^2\sqrt{48\sigma^2 \log(1/\sigma)}}{\pi}.
      \end{equation*}
      We have chosen $\sigma = n^{-5/19} \leq n^{-1/4}$, which makes the
      residuals in this expression of order smaller than $n^{3/2}$: for
      sufficiently large $n$, 
      \begin{equation*}
         \norm*{\bar{\vX} - \vX}_{\ell^2(G)}^2 \leq
         Cn^{3/2}
      \end{equation*}
      for an absolute constant $C>0$.
      Similarly, by this last bound and \Cref{lem:directly-sampled-norm}
      together with the triangle inequality, we have
      for $n$ sufficiently large
      \begin{align*}
         \norm*{\bar{\vX}}_{\ell^2({G})}^2
         &\geq
         \left(
            \norm*{{\vX}_{\nu_{\grtr}}}_{\ell^2({G})}
            -
            \norm*{\bar{\vX} - \vX}_{\ell^2(G)}
         \right)^2
         \\
         &=
         \norm*{{\vX}}_{\ell^2({G})}^2
         -2\norm*{{\vX}}_{\ell^2({G})}
         \norm*{\bar{\vX} - \vX}_{\ell^2(G)}
         \\
         &\geq
         n^2 \alpha^2
         -5n
         -Cn^{7/4}
      \end{align*}
      where we use the trivial upper bound $\norm{\vX}_{\ell^2(G)} \leq n$.
      Plugging into our previous EYM estimate and noticing that the previous
      residual dominates, we have for $n$ sufficiently large
      \begin{align*}
         \norm*{\hat{\vX} - \vX}_{\ell^2({G})}
         &\geq
         \sqrt{
            \frac{cn^2}{1 + k} - C n^{36/19}
         }
         - C' n^{3/4}.
      \end{align*}
      For the RMSE, given that $k \leq c\sqrt{m} = c n^{2/19}$ for an absolute
      constant $c>0$,
      we can use the inequality $\sqrt{1 - x} \geq 1-x$ for $0 \leq x \leq 1$
      (following from concavity) to get
      \begin{equation*}
         \frac{1}{n}\norm*{\hat{\vX} - \vX}_{\ell^2({G})}
         \geq
         \frac{c}{\sqrt{1+k}} - C n^{-1/4}.
      \end{equation*}
      This residual is redundant when $k \ltsim n^{2/19}$, and we can therefore
      conclude
      \begin{equation*}
         \frac{1}{n^2}\norm*{\hat{\vX} - \vX}_{\ell^2({G})}^2
         \geq
         \frac{c}{1+k}
      \end{equation*}
      for a sufficiently small absolute constant $c$.

   \end{proof}

\end{theorem}

\begin{remark}
   \Cref{thm:inapproximability} asserts lower bounds up to a threshold $k
   \ltsim n^{2/19}$. Based on empirical evidence and certain key residuals in
   the proof of \Cref{lem:diamond_approx_mollified_decay}, we believe it should be possible to assert the same lower bound
   up to scalings $k \ltsim n/ \log^c(n)$, for some $c > 0$, although our
   arguments are insufficient to this task. The main technical issue we contend
   with in the proof of \Cref{thm:inapproximability} is the nonsmoothness of
   the underlying image $X_{\grtr}$, which in our case necessitates the use of somewhat
   technical smoothing arguments. Some lemmas that we develop to this end,
   especially \Cref{lem:l2-sampling,lem:mollified-orthogonality}, are 
   suboptimal, and improvements would improve the rates. 
   At the same time, the perturbation framework we have developed in the proof
   of \Cref{thm:inapproximability} relies as little as possible on specific
   analytical properties of the spectral decomposition of the
   infinite-dimensional surrogate $X = X_{\grtr} \circ \vtau_{\nu_{\grtr}}$ for
   the observation
   $\vX$, encapsulated in \Cref{lem:diamond_diagonalized}; instead, we use
   only relatively coarse properties of the spectral decomposition of $X$,
   including bounds on norms of the eigenfunctions and their derivatives, the rate
   of decay of the eigenvalues, and regularity of the boundary of the support
   of $X_{\grtr}$. It is likely that more precise estimates tailored at the
   specific properties of $X$ would lead to straightforward improvements of the
   rates, but the resulting loss of generality would be undesirable for
   modeling templates and scenes beyond the model $X_{\grtr}$.
   Accordingly, we believe the crux of our argument should be applicable to
   templates $X_{\grtr}$ that have better regularity without having to go
   through smoothing arguments, which should yield improved rates in a more
   general setting.
\end{remark}

\begin{remark}

   The fact that the observation $\vX$ corresponds to the ``directly sampled''
   observation $(i, j) \mapsto \Ind{\norm{\vtau_{\nu_{\grtr}}(i, j)}_{\infty}
   \leq \alpha}$
   is not essential to our arguments in \Cref{thm:inapproximability}---indeed,
   the same perturbation framework would work with minor adaptations for any
   nonuniformly-sampled grid that is sufficiently close to the uniform sampling
   grid $G$. For example, defining $\vX$ in terms of a resampled version of the
   square template $\vX_{\grtr}$ using any compactly-supported interpolation
   kernel, such as the bilinear interpolation kernel (c.f.\
   \Cref{sec:app_resampling}), would require only minor adaptations, analogous
   to our treatment of the smoothing error in \Cref{lem:smoothing-error}.
   Although nonrealistic as a model for image acquisition, it is an interesting
   mathematical problem to extend our framework to the case of nonuniform grids
   that are far from the uniform grid $G$, such as grids induced by random
   sampling locations. Such an extension would require novel ideas,
   particularly in the core perturbation result, \Cref{lem:l2-sampling}.
\end{remark}

\subsubsection{Supporting Results}

\begin{lemma}
   \label{lem:diamond_diagonalized}

   Define a sequence
   \begin{equation}
      \lambda_k = (-1)^{k-1} \frac{4\sqrt{2}\alpha}{\pi( 2k - 1)}, \quad k = 1,
      2, \dots,
      \label{eq:diamond_eigenvals}
   \end{equation}
   and functions $g_k : [-1, 1] \to \bbR$ by
   \begin{equation}
      g_k(s) = \begin{cases}
         \frac{1}{\sqrt{\alpha\sqrt{2}}}\cos\left(  \frac{\pi}{2\sqrt{2}{\alpha}} (2k-1) s \right) &
         \abs{s} \leq \sqrt{2}\alpha \\
         0 & \ow,
      \end{cases} 
      \quad k =  1, 2, \dots
      \label{eq:diamond_eigenfuncs}
   \end{equation}
   Then the functions $g_k$ form an orthonormal basis for the range of the
   (compact, self-adjoint) operator $\sT_{X}$, and we have the decomposition
   \begin{equation*}
      \sT_{X} =
      \sum_{k \in \bbN} \lambda_k g_k g_k\adj.
   \end{equation*}

   \begin{proof}
      We take the formula \Cref{eq:diamond_iop} as our starting point. Because
      of the spectral theorem for self-adjoint compact operators on a Hilbert
      space, we have the decomposition \Cref{eq:integral_op_diagonalized} for
      $\sT_X$. Our approach will be to study the eigenvalue equation
      \begin{equation}
         \sT_X[g] = \lambda g, \enspace \lambda \neq 0, g \neq 0,
         \label{eq:diamond_eigen}
      \end{equation}
      and to produce a large enough family of solutions $(\lambda, g)$ to this
      equation that we can assert that we have produced the eigenvalues and
      eigenfunctions asserted by the spectral theorem in
      \Cref{eq:integral_op_diagonalized}. To begin, we make several preliminary
      observations about solutions to the eigenvalue equation
      \Cref{eq:diamond_eigen}. First, we note
      from \Cref{eq:diamond_iop} and the change of variables formula that
      \begin{equation*}
         \sT_X[f](\sqrt{2}\alpha s) = 
         \sqrt{2}{\alpha} \int_{-(1 - \abs{s})}^{1 - \abs{s}}
         f(\sqrt{2}\alpha t) \diff t,
      \end{equation*}
      so that, if for $\veps > 0$ we write $\sS_{\veps}[g](u) = g(\veps u)$ as
      the dilation operator (which satisfies $\sS_{\veps}\inv =
      \sS_{\veps\inv}$), we have
      \begin{equation}
         \sT_{X} = \sS_{\sqrt{2}\alpha}\bar{\sT}_{X}\sS_{\sqrt{2}\alpha}\inv,
         \label{eq:TX_bar_TX}
      \end{equation}
      where $ \bar{\sT}_{X} : L^2([-1, 1]) \to L^2([-1, 1])$ is defined as
      \begin{equation*}
         \bar{\sT}_{X}[f](s) = \sqrt{2}\alpha \int_{-(1 - \abs{s})}^{1 - \abs{s}}f(t) \diff t.
      \end{equation*}
      In particular, $\sT_{X}$ is similar to the operator $\bar{\sT}_{X}$. We
      therefore focus our analysis on $\bar{\sT}_{X}$ below. Next, note that by
      the Schwarz inequality, we have
      \begin{align*}
         \abs*{
            \bar{\sT}_{X}[f](s)
         }
         &\leq
         \sqrt{2}\alpha \norm{f}_{L^2} \norm{\Ind{[-(1-\abs{s}),
         1-\abs{s}]}}_{L^2} \\
         &=
         4\alpha \norm{f}_{L^2} \sqrt{1 - \abs{s}}.
      \end{align*}
      In particular, we have $\bar{\sT}_{X}[f](\pm 1) = 0$ for any $f \in L^2$.
      Thus, if $f$ is moreover a solution to
      \Cref{eq:diamond_eigen}, it is necessary that $f(\pm 1) = 0$,
      giving us boundary conditions for the eigenvalue equation.
      Similarly, the formula \Cref{eq:diamond_iop} shows that
      $\bar{\sT}_{X}[f](s) = \bar{\sT}_{X}[f](-s)$ for any $f \in L^2$, so any
      $f$ solving \Cref{eq:diamond_eigen} also satisfies even symmetry.

      We proceed with a standard bootstrapping argument---we start by seeking
      only solutions to \Cref{eq:diamond_eigen} that are infinitely
      differentiable. 
      For any $\abs{s} > 0$, differentiating
      \Cref{eq:diamond_iop} gives the equivalent boundary value problem 
      \begin{equation*}
         \lambda g'(s) = -\sqrt{2}\alpha \sign(s) \left(
            g(1 - \abs{s}) - g(-(1 - \abs{s}))
         \right), \quad
         g(\pm 1)= 0
      \end{equation*}
      for the eigenvalue equation \Cref{eq:diamond_eigen}. By even symmetry of
      $g$, this is equivalent to the problem
      \begin{equation*}
         \lambda g'(s) = -2\sqrt{2}\alpha\, g(1 - s), \quad
         g(1)= 0,\enspace g'(0) = 0
      \end{equation*}
      with $g \in C^\infty([0, 1])$. Differentiating once more to eliminate the
      `space reversal' on the RHS, we obtain the (necessary) system
      \begin{equation*}
         g'' + \frac{8\alpha^2}{\lambda^2} g = 0, \quad
         g(1)= 0,\enspace g'(0) = 0.
      \end{equation*}
      This is a second-order linear ODE. It has as its solutions
      \begin{equation*}
         g(s) = A \cos\left(\frac{2\sqrt{2}\alpha}{\abs{\lambda}} s\right)
         + B\sin\left(\frac{2\sqrt{2}\alpha}{\abs{\lambda}} s\right)
      \end{equation*}
      for constants $A, B$ to be determined with the boundary conditions. The
      condition $g'(0) = 0$ implies that $B=0$. The condition $g'(1) = 0$
      implies either that $A=0$ or that
      \begin{equation*}
         \frac{2\sqrt{2}\alpha}{\abs{\lambda}} 
         \in \frac{\pi}{2} \left( 2\bbZ + 1 \right),
      \end{equation*}
      i.e., that the frequency is an odd multiple of $\pi/2$. This implies
      \begin{equation*}
         \abs{\lambda_k} = \frac{4\sqrt{2}\alpha}{\pi( 2k + 1)}, \quad k = 0, 1,
         \dots,
      \end{equation*}
      and in particular
      \begin{equation*}
         g_k(s) = A_k \cos \left( \frac{\pi}{2} (2k + 1) s \right), \quad k = 0, 1,
         \dots,
      \end{equation*}
      where the constants $A_k$ can be determined such that $g$ has unit $L^2$
      norm. We have
      \begin{align*}
         \int_{-1}^1 g_k(s) g_{k'}(s) \diff s
         &=
         \frac{1}{2}\int_{-1}^1 \left( \cos( \pi(k - k') s) + \cos(\pi(k + k' + 1) s)
         \right) \diff s \\
         &= \left( \Ind{k = k'} + \Ind{k + k' + 1 = 0} \right) \\
         &= \Ind{k = k'}.
      \end{align*}
      In particular, $A_k = 1$. It remains to determine the signs of the
      eigenvalues $\lambda_k$. We calculate
      \begin{align*}
         \bar{\sT}_X[g_k](s)
         &= 2\sqrt{2}\alpha \int_{0}^{1-\abs{s}}\cos \left( \frac{\pi}{2} (2k +
         1) s \right) \\
         &= \abs{\lambda_k} \sin\left( \frac{\pi}{2} (2k + 1) (1 - \abs{s})
         \right) \\
         &=\abs{\lambda_k} \sin\left( \frac{\pi}{2} (2k + 1) \right)
         \cos \left( \frac{\pi}{2} (2k + 1) s \right) \\
         &= (-1)^k \abs{\lambda_k} g_k(s).
      \end{align*}
      In particular, the functions $g_k$ form a mutually orthogonal set of
      eigenfunctions of $\bar{\sT}_X$ with corresponding eigenvalues
      \begin{equation*}
         \lambda_k = (-1)^k \frac{4\sqrt{2}\alpha}{\pi( 2k + 1)}, \quad k = 0,
         1, \dots
      \end{equation*}
      To conclude, we note that from \Cref{eq:TX_bar_TX} that the functions
      $f_k : [-1, +1] \to \bbR $ defined by
      \begin{equation*}
         f_k(s) = \begin{cases}
            \frac{1}{\sqrt{\alpha\sqrt{2}}}\cos\left(  \frac{\pi}{2\sqrt{2}{\alpha}} (2k+1) s \right) &
            \abs{s} \leq \sqrt{2}\alpha \\
            0 & \ow,
         \end{cases} 
         \quad k = 0, 1, \dots
      \end{equation*}
      form an orthonormal basis for the image of $\sT_{X}$, and together with
      the eigenvalues $\lambda_k$ defined above
      provide a Schmidt decomposition of the operator $\sT_{X}$:
      \begin{equation*}
         \sT_{X} =
         \sum_{k \in \bbN_0} \lambda_k f_k f_k\adj.
      \end{equation*}
      This completes the proof.
   \end{proof}
\end{lemma}

\begin{lemma}
   \label{lem:diamond_approx_mollified_decay}

   For all $m \in \bbN$, any $k \in [n]$, and any $\sigma^2 > 0$, one has for
   the operator defined in \Cref{eq:discretized-truncated-ests-mtx}
   \begin{equation*}
      \norm*{\tilde{\vXi}_m}_{(k)}^{(2)}
      \leq
      \frac{n}{2}\norm*{\vLambda_m}_{(k)}^{(2)}
      +
      \frac{4m(1 + \log m)^{1/2}}{\alpha}
      +
      \frac{\pi n \sigma^2 m^2}{32\sqrt{2}\alpha}.
   \end{equation*}

   \begin{proof}
      We build from the matrix representation
      \Cref{eq:discretized-truncated-ests-mtx} of $\tilde{\vXi}_m$. The idea of the
      proof is straightforward: if $\tilde{\vU}_m$ had orthonormal columns, we would
      have immediately
      \begin{equation}
         \norm*{\tilde{\vXi}_m}_{(k)}^{(2)} 
         = \norm*{\vLambda_m}_{(k)}^{(2)},
         \label{eq:diamond_approx_decay_fake}
      \end{equation}
      by unitary invariance. Because of discretization and smoothing errors,
      $\tilde{\vU}_m$ is not an orthonormal $m$-frame, so
      \Cref{eq:diamond_approx_decay_fake} does not hold. However, when $n$ is
      large and $m$ is not too large relative to $n$, we can guarantee that
      $\tilde{\vU}_m$ is close to orthonormal, which we will combine with a
      perturbation result (\Cref{lem:matrix-am-gm}) to obtain the claim.

      By \Cref{lem:matrix-am-gm} and the triangle inequality, we have
      \begin{align*}
         \norm*{\tilde{\vXi}_m}_{(k)}^{(2)} 
         &\leq
         \norm*{
            \abs{\vLambda_m}^{1/2} \tilde{\vU}_m\adj \tilde{\vU}_m\abs{\vLambda_m}^{1/2}
         }_{(k)}^{(2)}  \\
         &\leq
         \tfrac{n}{2} \norm*{\vLambda_m}_{(k)}^{(2)}
         + \norm*{
            \abs{\vLambda_m}^{1/2} \left(\tilde{\vU}_m\adj \tilde{\vU}_m - \tfrac{n}{2} \vI \right)
            \abs{\vLambda_m}^{1/2}
         }_{(k)}^{(2)} \\
         &\leq
         \tfrac{n}{2}\norm*{\vLambda_m}_{(k)}^{(2)}
         + \norm*{
            \abs{\vLambda_m}^{1/2} \left(\tilde{\vU}_m\adj \tilde{\vU}_m - \tfrac{n}{2}\vI \right)
            \abs{\vLambda_m}^{1/2}
         }_{\frob},
         \labelthis \label{eq:xim-norm-bound-start}
      \end{align*}
      where in the final inequality we worst-case the residual by summing over
      all singular values. We will bound the residual term in
      \Cref{eq:xim-norm-bound-start} by bounding the magnitude of each of its
      elements.
      For $j =0, 1,
      \dots, m-1$, let $\tilde{\vu}_{m, j}$ denote
      the $j$-th column of $\tilde{\vU}_m$, and let $\pi_1(G)$ denote the projection of
      the rectangle $G$ onto its first coordinate. 
      Then $(2/n)\ip{\tilde{\vu}_{m, j}}{\tilde{\vu}_{m, j'}}$ is a Riemann sum
      corresponding to the integral of the function $(\varphi_{\sigma^2} \conv
      g_j) (\varphi_{\sigma^2} \conv g_{j'})$ over $[-1, 1]$.  We have from the
      Leibniz rule
      \begin{align*}
         \norm*{ (\varphi_{\sigma^2} \conv g_j)(\varphi_{\sigma^2}\conv g_{j'}) }_{\Lip} 
            &= \norm{\varphi_{\sigma^2} \conv g_j}_{L^\infty} 
            \norm{\varphi_{\sigma^2} \conv g_{j'}}_{\Lip} 
            + \norm{\varphi_{\sigma^2} \conv g_j}_{\Lip}
            \norm{\varphi_{\sigma^2} \conv g_{j'}}_{L^\infty}  \\
         &= \frac{1}{\sqrt{2}\alpha} \left(
            \norm{g_j}_{\Lip} + \norm{g_{j'}}_{\Lip}
         \right) \\
         &=
         \frac{\pi(j + j' + 1)}{2\alpha^2},
      \end{align*}
      where we use the fact that convolution with a gaussian does not increase
      $L^\infty$ norms (a special case of Young's inequality \cite[Ch.\ I, Thm
      1.3]{Stein1971-ed}) nor Lipschitz seminorms (the functions $g_j$ are all
      in $C^\infty$, so we can differentiate under the integral and then apply
      Jensen's inequality, since the gaussian has unit $L^1$ norm).
      Thus, by \Cref{lem:lipschitz_grid_1d} and
      \Cref{lem:diamond_diagonalized} and the triangle inequality, we have
      \begin{align*}
         \abs*{
            \ip{\tilde{\vu}_{m, j}}{\tilde{\vu}_{m, j'}}
            -
            \frac{n}{2} \Ind{j = j'}
         }
         &\leq
         \abs*{
            \ip{\tilde{\vu}_{m, j}}{\tilde{\vu}_{m, j'}}
            -
            \frac{n}{2} 
            \ip{\varphi_{\sigma^2} \conv g_j}{\varphi_{\sigma^2} \conv g_{j'}}_{L^2}
         }
         +
         \abs*{
            \frac{n}{2} 
            \ip{\varphi_{\sigma^2} \conv g_j}{\varphi_{\sigma^2} \conv g_{j'}}_{L^2}
            -
            \frac{n}{2} \Ind{j = j'}
         } \\
         &\leq
         \frac{n}{2}
         \abs*{
            \ip{\varphi_{\sigma^2} \conv g_j}{\varphi_{\sigma^2} \conv g_{j'}}_{L^2}
            -
            \Ind{j = j'}
         }
         +
         \frac{\pi(j + j' + 1)}{2\alpha^2}.
         \labelthis \label{eq:U_almost_orthogonal-pre}
      \end{align*}
      To handle the remaining residual, we will apply
      \Cref{lem:mollified-orthogonality}. This gives
      \begin{equation*}
         \frac{n}{2}
         \abs*{
            \ip{\varphi_{\sigma^2} \conv g_j}{\varphi_{\sigma^2} \conv g_{j'}}_{L^2}
            -
            \Ind{j = j'}
         }
         \leq
         \frac{n\sigma^2}{2} \norm{g_j'}_{L^2} \norm{g_{j'}'}_{L^2},
      \end{equation*}
      and from \Cref{lem:diamond_diagonalized} and a $L^1$-$L^\infty$ estimate,
      we have
      \begin{equation*}
         \norm{g_j'}_{L^2}
         \leq
         \frac{\pi(2j + 1)}{2\sqrt{2} \alpha},
      \end{equation*}
      whence
      \begin{equation*}
         \frac{n}{2}
         \abs*{
            \ip{\varphi_{\sigma^2} \conv g_j}{\varphi_{\sigma^2} \conv g_{j'}}_{L^2}
            -
            \Ind{j = j'}
         }
         \leq
         \frac{n\sigma^2 \pi^2(2j + 1)(2j' + 1)}{16\alpha^2}.
      \end{equation*}
      In particular, substituting this estimate into
      \Cref{eq:U_almost_orthogonal-pre} gives
      \begin{align*}
         \abs*{
            \ip{\tilde{\vu}_{m, j}}{\tilde{\vu}_{m, j'}}
            -
            \frac{n}{2} \Ind{j = j'}
         }
         \leq
         \frac{n\sigma^2 \pi^2(2j + 1)(2j' + 1)}{16\alpha^2}
         +
         \frac{\pi(j + j' + 1)}{2\alpha^2}.
         \labelthis \label{eq:U_almost_orthogonal}
      \end{align*}
      From the definition of $\vLambda_m$, it follows
      \begin{align*}
         \norm*{
            \abs{\vLambda_m}^{1/2} \left(\vU_m\adj \vU_m - \tfrac{n}{2}\vI \right)
            \abs{\vLambda_m}^{1/2}
         }_{\frob}^2
         &\leq
         \frac{16}{\alpha^2}
         \sum_{1 \leq i, j \leq m} \frac{(i + j - 1)^2}{(2i - 1)(2j - 1)}
         +
         \frac{\pi^2 n^2 \sigma^4}{2^{11} \alpha^2}
         \sum_{1 \leq i, j \leq m} 
         (2i - 1)(2j - 1).
      \end{align*}
      The second sum evaluates to $m^4$. For the first sum, note that
      $i + j - 1 = \half( (2i - 1) + (2j - 1) )$, so
      \begin{align*}
         \frac{(i + j - 1)^2}{(2i - 1)(2j - 1)} 
         &= \frac{1}{4} \left(
            \sqrt{\frac{2i - 1}{2j - 1}} + \sqrt{\frac{2j - 1}{2i - 1}}
         \right)^2 \\
         &\leq
         \frac{1}{2} \left(
            \frac{2i - 1}{2j - 1} + \frac{2j - 1}{2i - 1}
         \right),
      \end{align*}
      by the inequality $a + b \leq 2(a^2 + b^2)$. When summed over the grid
      $[m]^2$, the two functions of $i, j$ in the last inequality must be equal
      by symmetry. Thus
      \begin{align*}
         \sum_{1 \leq i, j \leq m} \frac{(i + j - 1)^2}{(2i - 1)(2j - 1)}
         &\leq
         \sum_{1 \leq i,j \leq m} \frac{2i - 1}{2j - 1} \\
         &= m^2 \sum_{j=1}^m \frac{1}{2j - 1}.
      \end{align*}
      By the usual estimates $\log m \leq \sum_{j=1}^m \tfrac{1}{j} \leq 1 +
      \log m$ for the harmonic numbers, we have $\sum_{j=1}^m \tfrac{1}{2j -
      1} \leq 1 + \log(2m) - \half \log m$, which in turn is less than $1 +
      \log m$ when $m \geq 4$. In addition, one can check numerically that the
      same estimate holds for $m \in \set{1, 2, 3, 4}$. We have thus shown
      \begin{equation*}
         \norm*{
            \abs{\vLambda_m}^{1/2} \left(\vU_m\adj \vU_m - \tfrac{n}{2}\vI \right)
            \abs{\vLambda_m}^{1/2}
         }_{\frob}^2
         \leq
         \frac{8m^2(1 + \log m)}{\alpha^2}
         + 
         \frac{\pi^2 n^2 \sigma^4 m^4}{2^{11} \alpha^2},
      \end{equation*}
      which establishes the claim when combined with our previous estimates.

   \end{proof}
\end{lemma}

\begin{lemma}
   \label{lem:xim-tail-sum-decay}

   For all $m \in \bbN$ and any $k \in [n]$, one has for the operator defined
   in \Cref{eq:discretized-truncated-ests-mtx}
   \begin{equation*}
      \norm*{\vLambda_m}_{(k)}^{(2)}
      \leq
      \sqrt{
         4\alpha^2 - \frac{16\alpha^2}{\pi^2} \frac{1}{2 \min\set{m,k} + 1}
      }.
   \end{equation*}
   \begin{proof}
      We have by definition
      \begin{align*}
         \left(
            \norm*{\vLambda_m}_{(k)}^{(2)}
         \right)^2
         &=
         \frac{32 \alpha^2}{\pi^2}
         \sum_{i=1}^{\min \set{k, m}}
         \frac{1}{(2k -1)^2} \\
         &=
         \frac{32 \alpha^2}{\pi^2}
         \left(
            \sum_{i=1}^{\infty} \frac{1}{(2k -1)^2}
            - \sum_{i=1 + \min\set{k,m}}^{\infty} \frac{1}{(2k -1)^2}
         \right).
         \labelthis \label{eq:xim-tail-sum-decay-est}
      \end{align*}
      For the first term, we have
      \begin{align*}
         \sum_{i=1}^{\infty} \frac{1}{(2k -1)^2}
         &=
         \sum_{i=1}^\infty \frac{1}{k^2}
         - \sum_{i=1}^\infty \frac{1}{(2k)^2} \\
         &=
         \frac{3}{4} \sum_{i=1}^\infty \frac{1}{k^2}\\
         &= \frac{\pi^2}{8}.
      \end{align*}
      For the second term, we have from the integral test estimate
      \begin{align*}
         \sum_{i=1 + \min\set{k,m}}^{\infty} \frac{1}{(2k -1)^2}
         &\geq
         \int_{1 + \min\set{k,m}}^\infty \frac{1}{(2t - 1)^2}\diff t \\
         &=
         \frac{\half}{2 \min\set{k, m} + 1}.
      \end{align*}
      Plugging into \Cref{eq:xim-tail-sum-decay-est} and taking square roots
      gives the claim.
   \end{proof}
\end{lemma}

\begin{lemma}
   \label{lem:smoothing-error}
   Consider the smoothed template $\bar{X}$, as in
   \Cref{thm:inapproximability}, with sampling $\bar{\vX}$ on the grid $G$.
   Let $\vX$ denote the ``directly sampled'' template
   \begin{equation*}
      ({\vX})_{ij} 
      = \Ind{\norm{(\vtau_{\pi/4})_{ij}}_{\infty} \leq \alpha},
   \end{equation*}
   where we recall \Cref{eq:motion_field_param_0,eq:motion_field_param_1}.
   Then if $\sigma \leq \half$ and $\alpha = \tfrac{1}{\sqrt{2}}$, one has
   \begin{equation*}
      \norm*{\bar{\vX} - \vX}_{\ell^2(G)}^2 \leq
      \frac{n^2\sigma^8}{\pi^2}
      +
      \frac{2n}{\pi} + \frac{n^2\sqrt{48\sigma^2 \log(1/\sigma)}}{\pi}.
   \end{equation*}
   \begin{proof}

      Note that by definition
      \begin{equation*}
         \bar{X}(i, j)
         =
         \int_{\bbR^2}
         \varphi_{\sigma^2}((i,j) - \vx') X(\vx') \diff \vx'.
      \end{equation*}
      Because
      \begin{equation*}
         \varphi_{\sigma^2}(\vx)
         =
         \frac{1}{2\pi\sigma^2} e^{-\frac{1}{2\sigma^2} \norm{\vx}_2^2},
      \end{equation*}
      for $\norm{\vx}_2^2 \geq 12 \sigma^2 \log(1 / \sigma)$, one has
      \begin{equation}
         \varphi_{\sigma^2}(\vx)
         \leq \frac{\sigma^4}{2\pi}.
         \label{eq:gaussian-decay}
      \end{equation}
      Since $\norm{\vx}_2 \geq \norm{\vx}_{\infty}$ and $\norm{\vR_{\nu} \vx}_2
      = \norm{\vx}_2$, 
      if $\norm{\vR_{\pi/4} \vx}_\infty^2 \geq 12 \sigma^2 \log(1/\sigma)$, 
      then \Cref{eq:gaussian-decay} also holds.

      Consider the set
      \begin{equation}
         S = \set*{(i, j) \in G \given 
            \abs*{ \norm*{(\vtau_{\pi/4})_{ij}}_\infty - \alpha} >
            \sqrt{ 12 \sigma^2 \log(1/\sigma) }
         },
         \label{eq:farsquare}
      \end{equation}
      and write $S^\compl$ for the complement of $S$ relative to $G$. 
      First, suppose that $(i, j) \in S$ is not in the support of $X$.
      We have $X(i, j) = 0$, and
      \begin{align*}
         \bar{X}(i, j)
         &=
         \int_{\bbR^2}
         \varphi_{\sigma^2}((i,j) - \vx') X(\vx') \diff \vx'
         \\
         &=
         \int_{\norm{\vR_{\pi/4} \vx}_\infty \leq \alpha}
         \varphi_{\sigma^2}((i,j) - \vx') X(\vx') \diff \vx'
         + \int_{\norm{\vR_{\pi/4} \vx}_\infty \geq \alpha}
         \varphi_{\sigma^2}((i,j) - \vx') X(\vx') \diff \vx'
         \\
         &\leq
         \frac{\sigma^4}{2\pi}
         \int_{\norm{\vR_{\pi/4} \vx}_\infty \leq \alpha}
         X(\vx') \diff \vx'
         \\
         &= \frac{\sigma^4}{\pi}
      \end{align*}
      by the triangle inequality. Evidently $\bar{X}(i, j) \geq 0$ as well.
      A symmetric argument applies when $(i, j) \in
      S$ is in the support of $X$, except that we obtain
      \begin{align*}
         \bar{X}(i, j)
         &=
         \int_{\bbR^2}
         \varphi_{\sigma^2}(\vx') X((i,j) - \vx') \diff \vx'
         \\
         &=
         \int_{\norm{\vR_{\pi/4}( (i,j) - \vx')}_\infty \leq \alpha}
         \varphi_{\sigma^2}(\vx') \diff \vx'
         \\
         &=
         1-
         \int_{\norm{\vR_{\pi/4}( (i,j) - \vx')}_\infty \geq \alpha}
         \varphi_{\sigma^2}(\vx') \diff \vx',
      \end{align*}
      whence
      \begin{align*}
         \abs*{ \bar{X}(i, j) - 1 } 
         &\leq
         \int_{\norm{\vR_{\pi/4}( (i,j) - \vx')}_\infty \geq \alpha}
         \varphi_{\sigma^2}(\vx') \diff \vx' \\
         &\leq
         \int_{\norm{\vR_{\pi/4}\vx'}_\infty \geq \sqrt{12\sigma^2
         \log(1/\sigma)}}
         \varphi_{\sigma^2}(\vx') \diff \vx'
         \\
         &\leq
         \int_{\norm{\vx'}_\infty \geq \sqrt{6\sigma^2
         \log(1/\sigma)}}
         \varphi_{\sigma^2}(\vx') \diff \vx'.
      \end{align*}
      The last integral can be estimated with the fact that the gaussian tail
      integral is bounded by the density---in particular, \cite[Proposition
      2.1.2]{Vershynin2018-br} gives
      \begin{equation*}
         \int_{\norm{\vx'}_\infty \geq \sqrt{6\sigma^2
         \log(1/\sigma)}}
         \varphi_{\sigma^2}(\vx') \diff \vx'
         \leq
         \left(
            \frac{1}{\sqrt{12\pi \log(1/\sigma)}}e^{-3\log(1/\sigma)}
         \right)^2
         =
         \frac{\sigma^6}{12 \pi \log(1/\sigma)},
      \end{equation*}
      so we have
      \begin{equation*}
         \abs*{ \bar{X}(i, j) - 1 } 
         \leq
         \frac{\sigma^4}{\pi},
      \end{equation*}
      when $\sigma \leq \half$.
      Thus, we have shown that for any $(i, j) \in S$, we have
      \begin{equation}
         \abs{X(i, j) - \bar{X}(i, j)} \leq \sigma^4 / \pi.
         \label{eq:real-smoothed-far-est}
      \end{equation}

      Next, we argue that the cardinality $\abs{S}$
      is sufficiently large. We will do this by bounding the size of
      $S^\compl$. We have by inequalities for $\ell^p$ norms and
      \Cref{eq:motion_field_param_0,eq:motion_field_param_1}
      \begin{equation*}
         \frac{1}{\sqrt{2}}
         \norm*{\begin{bmatrix} i \\ j \end{bmatrix}}_2
         \leq\norm*{\vtau_{\nu ij}}_{\infty}
         \leq
         \norm*{\begin{bmatrix} i \\ j \end{bmatrix}}_2,
      \end{equation*}
      so if we define
      \begin{align*}
         S' &= \set*{(i, j) \in G \given 
            \abs*{ \norm*{\begin{bmatrix} i \\ j \end{bmatrix}}_2 - \alpha} \leq
            \sqrt{24\sigma^2 \log(1/\sigma)}
         }, %
      \end{align*}
      we have $S^\compl \subset S'$.
      The square $[-1, 1]$ is covered by the union of balls of radius $\sqrt{2}/(n-1)$
      centered at each point of the grid $G$. Consider the subset
      \begin{equation*}
         U = \set*{ (u, v) \in [-1, 1] \given \abs*{ \sqrt{u^2 + v^2} - \alpha }
         \leq \frac{\sqrt{2}}{n-1} + \sqrt{24\sigma^2 \log(1/\sigma)}}.
      \end{equation*}
      Then by the triangle inequality and the above covering reasoning, $S' +
      \set{(u, v) \in \bbR^2 \given \sqrt{u^2 + v^2} \leq \sqrt{2}/(n-1)}
      \subset U$, from which it follows by a volume bound
      \begin{equation*}
         \abs{S^\compl} \pi \left( \frac{\sqrt{2}}{n-1} \right)^2
         \leq
         \Vol( U ).
      \end{equation*}
      Because $U$ is an annulus, we calculate readily 
      \begin{equation*}
         \Vol(U) 
         = 4\alpha \left(
            \frac{\sqrt{2}}{n-1} + \sqrt{24\sigma^2 \log(1/\sigma)}
         \right),
      \end{equation*}
      whence
      \begin{equation*}
         \abs{S^\compl}
         \leq 
         \frac{2n}{\pi} + \frac{n^2\sqrt{48\sigma^2 \log(1/\sigma)}}{\pi},
      \end{equation*}
      where the last inequality worst-cases over our condition on $\alpha$.

      Now, to conclude, we have by the above
      \begin{align*}
         \norm*{ \bar{\vX} - \vX}_{\frob}^2
         &=
         \sum_{(i, j) \in S} \left(
            (\bar{\vX})_{ij} - (\vX \circ \vtau_{\nu})_{ij}
         \right)^2
         +
         \sum_{(i, j) \in S^\compl} \left(
            (\bar{\vX})_{ij} - (\vX \circ \vtau_{\nu})_{ij}
         \right)^2 \\
         &\leq
         \frac{n^2\sigma^8}{\pi^2}
         +
         \abs{S^\compl} \sup_{(i, j) \in G}
         \left( (\bar{\vX})_{ij} - (\vX \circ \vtau_{\nu})_{ij}\right)^2 \\
         &\leq
         \frac{n^2\sigma^8}{\pi^2}
         +
         \frac{2n}{\pi} + \frac{n^2\sqrt{48\sigma^2 \log(1/\sigma)}}{\pi},
      \end{align*}
      because both matrices have entries in $[0, 1]$.

   \end{proof}
\end{lemma}

\begin{lemma}
   \label{lem:directly-sampled-norm}
   For $\nu = \pi/4$, %
   consider the ``directly sampled'' infinite-resolution template
   \begin{equation}
      (\bar{\vX})_{ij} = \Ind{\norm{(\vtau_{\nu})_{ij}}_{\infty} \leq \alpha},
      \label{eq:bar_X_defn}
   \end{equation}
   where we recall \Cref{eq:motion_field_param_0,eq:motion_field_param_1}. Then
   one has
   \begin{equation*}
      \norm*{\bar{\vX}}_{\frob} \geq n^2 \alpha^2 - 5n.
   \end{equation*}
   \begin{proof}

      Consider the case $\nu = \pi/4$. By rotational symmetry of $\bar{\vX}$ by
      multiples of $\pi/2$, and discarding the sum over the central axes when
      $n$ is odd, we have
      \begin{align*}
         \norm*{\bar{\vX}}_{\frob}^2
         &=\sum_{(i, j) \in G} 
         \Ind{ \max \set{\abs{i+j}, \abs{i-j}} \leq \sqrt{2}\alpha} \\
         &= \sum_{i=0}^{n-1} \sum_{j=0}^{n-1}
         \Ind{ \max \set{\abs{i-(n-1 - j)}, \abs{i-j}} \leq
         (n-1)\alpha/\sqrt{2}} \\
         &\geq
         4\sum_{i=0}^{\floor{\tfrac{n-1}{2}}}
         \sum_{j=0}^{\floor{\tfrac{n-1}{2}}}
         \Ind{\abs{i-(n - 1 - j)} \leq (n-1)\alpha/\sqrt{2}}.
      \end{align*}
      So, by the integral test estimate (because the summand is monotone
      increasing as a function of both $i$ and $j$ when the other is fixed) and
      nonnegativity,
      \begin{align*}
         \norm*{\bar{\vX}}_{\frob}^2
         &\geq
         4\int_{0}^{\floor{\tfrac{n-1}{2}}}
         \int_{0}^{\floor{\tfrac{n-1}{2}}}
         \Ind{\abs{i-(n - 1 - j)} \leq (n-1)\alpha/\sqrt{2}}
         \diff i \diff j \\
         &=
         4(n-1)^2
         \int_{0}^{\tfrac{1}{n-1}\floor{\tfrac{n-1}{2}}}
         \int_{0}^{\tfrac{1}{n-1}\floor{\tfrac{n-1}{2}}}
         \Ind{\abs{i-(1 - j)} \leq \alpha/\sqrt{2}}
         \diff i \diff j \\
         &\geq
         4(n-1)^2
         \int_{0}^{\half - \tfrac{1}{n-1}}
         \int_{0}^{\half - \tfrac{1}{n-1}}
         \Ind{\abs{i-(1 - j)} \leq \alpha/\sqrt{2}}
         \diff i \diff j \\
         &\geq
         4(n-1)^2
         \left(
            \int_{0}^{\half}
            \int_{0}^{\half}
            \Ind{\abs{i-(1 - j)} \leq \alpha/\sqrt{2}}
            \diff i \diff j
            -
            2
            \int_{0}^{\half} \int_{\half - \tfrac{1}{n-1}}^{\half}
            \Ind{\abs{i-(1 - j)} \leq \alpha/\sqrt{2}}
            \diff i \diff j
         \right),
      \end{align*}
      where in the final inequality we used permutation symmetry of the
      integral as a function of $(i, j)$ to simplify the residual.
      Now, the region of integration in the first term in the last line of the
      previous expression is equivalent to $\set{(i, j) \given (\half - i) +
      (\half - j) \leq \alpha/\sqrt{2}}$, which defines a right triangle
      with two side lengths equal to $\alpha / \sqrt{2}$. Because $\alpha <
      1/\sqrt{2}$, the integral evaluates to $\alpha^2 / 4$. Meanwhile, the
      integral in the second term is no larger than $1/2(n-1)$, because the
      integrand is bounded by $1$. Thus
      \begin{equation*}
         \norm*{\bar{\vX}}_{\frob}^2
         \geq
         (n-1)^2 \alpha^2 - 4(n-1).
      \end{equation*}
      Distributing in this expression and worst-casing slightly gives the
      claim.

   \end{proof}
\end{lemma}

\subsection{Proofs for \Cref{thm:ntilt-informal}}
\label{sec:ntilt_proofs}

\paragraph{Problem setup (and continuum idealization).} 
Let $k \in \bbN$, and consider an observation $\vX \in \bbR^{m \times n}$ drawn
from the class \Cref{eq:template_class}, with rotation parameter $\nu_{\grtr}$ (so that
$(\vX)_{ij} = X_{\grtr} \circ \vtau_{-\nu_{\grtr}}(i, j)$ if $(i, j) \in G$, following
\Cref{eq:image_ndc}).
For $\vU \in \bbR^{m \times k}$, $\vV \in \bbR^{n \times k}$, we study the
optimization objective
\begin{equation}
   \sL_{\mathrm{discrete}}(\nu, \vU, \vV)
   =
   \frac{1}{2}\norm*{
      \vX - \left(
         \vU \vV\adj
      \right) \circ \vtau_{-\nu}
   }_{\frob}^2,
   \label{eq:tilt-objective-discrete}
\end{equation}
where here in the context of discrete images, the transformations
$\vtau_{-\nu}$ must be implemented with resampling (we give a brief overview of
this idea in \Cref{sec:app_resampling}). The resampling operation can be chosen
to be continuously differentiable, making it amenable to gradient-based
optimization on the objective $\sL_{\mathrm{discrete}}$, but it introduces a host of
discretization-based artifacts to the optimization process that are challenging
to treat.\footnote{In particular, note that for any $\nu$, $\vM \mapsto \vM
   \circ \vtau_{\nu}$, as defined in \Cref{sec:app_resampling}, is a linear
   operator. If we call this operator $\sA_\nu$, it can be seen from
   \Cref{sec:app_resampling} that $\sA_\nu \adj \sA_\nu$ is a banded matrix, but
   it is \textit{not} incoherent---this means that the analysis of the problem
   \Cref{eq:tilt-objective-discrete} requires tools other those developed to
   analyze matrix sensing under the RIP (c.f.\ \cite{Stoger2021-an,Chi2018-au}).
   The situation is further complicated by the fact that the objective
   \Cref{eq:tilt-objective-discrete} simultaneously learns the sensing matrix (in
   matrix sensing parlance) and the low-rank factorization, a setting that has not
been considered in prior work.}
We will simplify the situation by considering a continuum limit of the
objective \Cref{eq:tilt-objective-discrete}, and a corresponding continuum
gradient-like iteration for its solution. 

Consider operators $\vU : \bbR^{k} \to L^2(\bbR)$, $\vV : \bbR^k \to
L^2(\bbR)$. These operators can be thought of as `matrices', whose columns
are $L^2(\bbR)$ functions---note that in the continuum, following
\Cref{eq:template-continuum} we have
\begin{equation*}
   X_{\grtr}(s, t) = \underbrace{\Ind{\abs{s}\leq \alpha}}_{u_{\grtr}(s)} 
   \underbrace{\Ind{\abs{t}\leq \alpha}}_{v_{\grtr}(t)},
\end{equation*}
i.e., as an operator, $X_{\grtr} = u_{\grtr} v_{\grtr}\adj$. The corresponding continuum analogue
of the observation $\vX$ is the deformed template $X = X_{\grtr} \circ
\vtau_{-\nu_{\grtr}}$. To mirror the smoothing effect of a continuous interpolation
kernel imposed in \Cref{eq:resampling}, we introduce an extra gaussian
smoothing filter $\varphi_{\sigma^2}(s, t) = (2\pi\sigma^2)^{-1}
e^{-(s^2 + t^2)/2\sigma^2}$ to the objective function, yielding the objective
\begin{equation}
   \sL^\sigma(\nu, \vU, \vV)
   =
   \frac{1}{2}
   \norm*{
      \varphi_{\sigma^2} \ast \left(
         X - \left( \vU \vV\adj \right) \circ \vtau_{-\nu}
      \right)
   }_{L^2}^2.
   \label{eq:tilt-objective}
\end{equation}
Notice that, by its definition \Cref{eq:2d_rotation}, the map $f \mapsto f
\circ \vtau_{-\nu}$ is a unitary transformation (apply the change of variables
formula in the integral defining the $L^2$ inner product). Using in addition
the Lie group structure of the rotation matrices \Cref{eq:2d_rotation}, 
we have that for any $f \in L^2(\bbR^2)$, any $\nu$ and any $\sigma^2$,
\begin{align*}
   (\varphi_{\sigma^2} \ast (f \circ \vtau_{\nu}))(\vx)
   &=
   \int_{\bbR^2} \varphi_{\sigma^2}(\vx') f(\vR_{\nu}(\vx - \vx')) \diff \vx'
   \\
   &=
   \int_{\bbR^2} \varphi_{\sigma^2}(\vR_{-\nu}(\vx')) f(\vR_{\nu}(\vx) -
   \vx') \diff \vx' \\
   &=
   \int_{\bbR^2} \varphi_{\sigma^2}(\vx') f(\vR_{\nu}(\vx) -
   \vx') \diff \vx' \\
   &= 
   (\varphi_{\sigma^2} \ast f)\circ \vtau_{\nu}(\vx)
   \labelthis \label{eq:gaussian-rigid-commute}
\end{align*}
by the change of variables formula and rotational invariance of the gaussian
function. In words, rigid motions commute with gaussian smoothing.
Applying this result together with the unitary transformation property, we can
write our objective as
\begin{align*}
   \sL^\sigma(\nu, \vU, \vV)
   &=
   \frac{1}{2}
   \norm*{
      \varphi_{\sigma^2} \ast X
      - (\varphi_{\sigma^2} \ast (\vU \vV\adj)) \circ \vtau_{-\nu}
   }_{L^2}^2 \\
   &=
   \frac{1}{2}
   \norm*{
      (\varphi_{\sigma^2} \ast X) \circ \vtau_{\nu}
      - \varphi_{\sigma^2} \ast (\vU \vV\adj)
   }_{L^2}^2 \\
   &=
   \frac{1}{2}
   \norm*{
      \varphi_{\sigma^2} \ast \left(
         X_{\grtr} \circ \vtau_{\nu - \nu_{\grtr}}
         - \vU \vV\adj
      \right)
   }_{L^2}^2.
   \labelthis \label{eq:tilt-objective-ez}
\end{align*}
We emphasize that \Cref{eq:tilt-objective-ez,eq:tilt-objective} are equal, but
\Cref{eq:tilt-objective-ez} is more straightforward to differentiate.

\paragraph{Simplifications to \Cref{eq:tilt-objective-ez}.} 
Our analysis will apply to a simplified version of the general objective
\Cref{eq:tilt-objective-ez}. We discuss the simplifications we make here.

\begin{enumerate}
   \item \textbf{Single-channel factorization ($k=1$).} We analyze a
      critically-parameterized version of the problem
      \Cref{eq:tilt-objective-ez}, where $k=1$. This leads to the objective
      \begin{equation}
         \sL^\sigma(\nu, u, v)
         =
         \frac{1}{2}
         \norm*{
            \varphi_{\sigma^2} \ast \left(
               X_{\grtr} \circ \vtau_{\nu - \nu_{\grtr}}
               - uv\adj 
            \right)
         }_{L^2}^2,
         \label{eq:tilt-objective-ezz}
      \end{equation}
      where $u, v \in L^2(\bbR)$. When the transformation component of
      \Cref{eq:tilt-objective-ezz} is omitted, this simplification is analogous
      to consideration of the rank-one matrix factorization problem
      \cite[\S3]{Chi2018-au}; because the untransformed square template
      $X_{\grtr}$
      has ``rank one'' (in a suitable, generalized sense), perfect
      reconstruction is still possible in our setting. Although the rank-one
      case is a vast simplification over the problem
      \Cref{eq:tilt-objective-ez} with general $k$, we begin our analysis here
      because the introduction of the simultaneous transformation optimization
      component to \Cref{eq:tilt-objective-ezz} represents a nontrivial
      complication with respect to existing analyses (c.f.\
      \cite{Zhang2020-on,Ge2017-tj}). We anticipate that the emerging
      understanding of overparameterized matrix sensing will be useful in
      generalizing our results to the setting of general $k$
      \cite{Stoger2021-an,Li2017-ic,Xu2023-ir,Ward2023-rw}.
   \item \textbf{Symmetric factorization.} Because the square template
      $X_{\grtr}$ is
      self-adjoint as a Fredholm operator (in other words, the template
      satisfies $X_{\grtr}(s, t) = X_{\grtr}(t, s)$), it is reasonable to reduce the search
      space in \Cref{eq:tilt-objective-ezz} to factorizations where $u = v$.
      This gives the problem
      \begin{equation}
         \sL^\sigma(\nu, u)
         =
         \frac{1}{2}
         \norm*{
            \varphi_{\sigma^2} \ast \left(
               X_{\grtr} \circ \vtau_{\nu - \nu_{\grtr}}
               - uu\adj 
            \right)
         }_{L^2}^2.
         \label{eq:tilt-objective-ezzz}
      \end{equation}
      All of our experiments with \ours{} make use of general, asymmetric grid
      factors, so a theoretical understanding of the general asymmetric case
      (when the target template $X_{\grtr}$ is asymmetric) remains crucial for future
      work. In this connection, we note that theoretical analyses of asymmetric
      matrix factorization typically add an additional ``balancing''
      regularizer to the objective \Cref{eq:tilt-objective-ez} (c.f.\
      \cite{Chi2018-au,Ge2017-tj,Zhang2020-on,Haeffele2015-ka})---in our
      experiments, the $2, 1$ regularizer described in
      \Cref{sec:app-regularization} plays this role.
\end{enumerate}

\paragraph{Gradient-like iterations for alignment and factorization.}

Obtaining a gradient iteration for the objective \Cref{eq:tilt-objective-ez}
can be done straightforwardly with respect to the finite-dimensional $\nu$
variable:
making essential use of the duality on $L^2(\bbR)$ and 
the fact that the convolution of two gaussian functions is another gaussian
function, with variance equal to the sum of the factors' variances, 
we calculate in \Cref{lem:tilt_grad_hess} 
\begin{align*}
   \nabla_{\nu} \sL^\sigma(\nu, u) &= 
   -\ip*{\varphi_{\sigma^2}\ast
      \left( 
         u u\adj\circ \vtau_{\nu_{\grtr} - \nu}
      \right)
   }{
      \ip*{
         \nabla_{\vx}[\varphi_{\sigma^2} \ast X_{\grtr}]
      }{
         \begin{bmatrix}
            0 & -1 \\
            1 & 0
         \end{bmatrix}
         (\spcdot)
      }_{\ell^2}
   }_{L^2(\bbR^2)}.
   \labelthis \label{eq:tilt-grad-nu}
\end{align*}
Differentiation with respect to the $u$ factor in
\Cref{eq:tilt-objective-ezzz} requires a slightly more technical notion of
gradient. To limit technicality in the analysis, we study instead an
infinite-dimensional analogue of a projected gradient descent method, where
after each update to the $u$ variable we project it onto the ``unit sphere''
in $L^2(\bbR)$ as $u \mapsto u / \norm{u}_{L^2(\bbR)}$. 
Moreover, when performing factorization, we optimize only the unsmoothed loss
$\sL^{0}(\nu, \spcdot)$.  We recall in \Cref{lem:r1approx-to-power} that in
this setting, whenever the factorization target $X_{\grtr} \circ \vtau_{\nu -
\nu_{\grtr}}$ is not negative (as an operator), it is equivalent in this
setting to seek the largest positive eigenvalue of the (symmetrized) operator
corresponding to the factorization target. Moreover, as long as the
factorization target has no negative eigenvalues of significant
magnitude,\footnote{This is typically the case; see
   \Cref{lem:diamond_diagonalized}. Our proofs show that this structure persists
throughout iterations of our algorithm, as well, in order to guarantee that our
algorithm succeeds.} this process is achieved by the power method, which in our
setting produces iterates
\begin{equation}
   u_{k+1} = \frac{
      \left(
         \sT_{ X_{\grtr} \circ \vtau_{\nu - \nu_{\grtr}}}
         +
         \sT_{ X_{\grtr} \circ \vtau_{\nu - \nu_{\grtr}}}\adj
      \right)u_{k}
   }{
      \norm*{
         \left(
            \sT_{ X_{\grtr} \circ \vtau_{\nu - \nu_{\grtr}}}
            +
            \sT_{ X_{\grtr} \circ \vtau_{\nu - \nu_{\grtr}}}\adj
         \right)u_{k}
      }_{L^2(\bbR)}
   },
   \label{eq:power_method_structure}
\end{equation}
before outputting an approximate factor for the target, which we will write as 
$\mathsf{P}(k, u_0, \nu) \in L^2(\bbR)$ (specifying the dependence on the power
method's initialization $u_0$ and the rotation $\nu$ applied to the template):
\begin{equation}
   \mathsf{P}(k, u_0, \nu) 
   =
   \sqrt{\half u_k \adj (
      \sT_{ X_{\grtr} \circ \vtau_{\nu - \nu_{\grtr}}}
      +
      \sT_{ X_{\grtr} \circ \vtau_{\nu - \nu_{\grtr}}}\adj
   ) u_k}
   u_k.
   \label{eq:power_method_output}
\end{equation}
\textit{A priori}, this may return a complex-valued function; we will take care
in our analysis to show that this never occurs when the iteration count is set
appropriately.

\paragraph{Our algorithm.}
The key mathematical property underlying the success of \ours{} in practical
experiments is the fact that \textit{incremental improvements to representation
   (factorization) help promote incremental improvements to alignment, and vice
versa}. 
The algorithm we study theoretically is a simplified version of \ours{}, but
nonetheless captures this complex interplay and sheds light on why it succeeds
in practice. The major simplification we impose is that rather than jointly
updating the $\nu$ (alignment) iterates and the $u$ (factorization) iterates,
we will update them individually in consecutive blocks, as in an alternating
minimization procedure. Our algorithm separates into five distinct stages,
described below.

\subparagraph{Stage one: rough representation.}
From a ``flat'' initialization for the scene
\begin{equation}
   u_0 = \Ind{[-1, +1]},
   \label{eq:u-init}
\end{equation}
we perform $T_{\rough}$ iterations of power method
\Cref{eq:power_method_output}, to generate a
roughly-localized representation of the template $X_{\grtr}$:
\begin{equation}
   u_{\rough}
   =
   \mathsf{P}(T_{{\rough}}, u_0, 0).
   \label{eq:u-rough-defn}
\end{equation}
This procedure corresponds to the initial iterations of \ours{} in practical
experiments, where the uninformative initialization $u_0$ does not produce
sufficient gradients (in texture or geometry) for alignment to occur.
The roughly-localized output $u_{\rough}$ usefully ends up with both
\textit{texture} and a rough \textit{shape} profile that promotes subsequent
alignment. 

\subparagraph{Stage two: rough alignment.}
Given a step size $\beta$, we perform $T_{\nu}$ iterations of
gradient descent on the alignment objective with smoothing level
$\sigma_{\SG}$, initialized randomly:
\begin{equation}
   \nu_0 \sim \Unif{[0, 2\pi]},
   \label{eq:nu-init}
\end{equation}
and with the factorization iterate at the output of the previous rough
representation step:
\begin{align*}
   \nu_{k+1} &=
   \nu_{k} - \beta \nabla_{\nu} \sL^{\sigma_{\SG}}(\nu_{k}, u_{\rough}),
   \qquad k = 0, 1, \dots, T_{\nu}-1.
   \labelthis \label{eq:tilt-grad-nu-iter}
\end{align*}
We write $\hat{\nu} = \nu_{T_{\nu}}$.
The alignment problem in \Cref{eq:tilt-objective-ezzz} has multiple optimal
solutions, due to the symmetries of the alignment target $X_{\grtr}$---this
means the optimization landscape is \textit{not} globally convex. At these
initial iterations of the alignment procedure, with a non-informative
initialization, we rely on the presence of strong gradient in the objective landscape to bring our initial iterate close to one of
the several equivalent optimal solutions. At a technical level, this style of
analysis mirrors those used in other global analyses of nonconvex optimization
landscapes \cite{Zhang2020-on}.

\subparagraph{Stage three: refined representation.}

This final stage  of the algorithm takes advantage of the roughly-localized alignment
output from the previous stage to improve the representation quality
further---the initial roughly-localized template $u_{\rough}$ from the first
stage is better localized, and its edges sharpened to match those of the target
$X_{\grtr}$. Accordingly, we run $T_{u}$ iterations of power method
\Cref{eq:power_method_output},
started with the outputs of the previous stages:
\begin{equation}
   \hat{u} = \mathsf{P}(T_{u}, u_{\rough}, \hat{\nu}).
   \label{eq:u-hat-defn}
\end{equation}
The algorithm's output is the pair $(\hat{\nu}, \hat{u})$.

\paragraph{Main result.}

Our main result establishes convergence of our alternating minimization
version of \ours{} to the true parameters $(\nu_{\grtr}, u_{\grtr})$, up to
symmetry, in a ``hard'' instance of the problem: where $\nu_{\grtr} = \pi/4$
(as we studied in \Cref{sec:svd_proofs}) and $\alpha = \tfrac{1}{\sqrt{2}}$
(corresponding to an `in focus' target).

\begin{theorem}
   \label{thm:tilt-infinite}

   Consider the iterations encompassed by
   \Cref{eq:u-rough-defn,eq:tilt-grad-nu-iter,eq:u-hat-defn}, with
   initializations
   \Cref{eq:u-init,eq:nu-init}. Suppose $\alpha = \tfrac{1}{\sqrt{2}}$
   and $\nu_{\grtr} = \pi/4$.
   There are absolute constants $c_1$, $C_1$, $C_2 > 0$
   such that for any parameters $\sigma$, $\beta$ satisfying
   \begin{align*}
      \sigma{\SG}^2 &\leq \tfrac{1}{10^4}, \\
      \beta &\leq c_1,
   \end{align*}
   for any $0 < \veps \leq \tfrac{1}{768}$, if the iteration counts satisfy
   \begin{align*}
      T_{\rough} &\geq -C_1 \log(\sigma_{\SG}^2 \veps),
      \\
      T_{\nu} &\geq -\frac{C_2  \log(3\veps)}{\beta}, 
      \\
      T_u &\geq 16,
   \end{align*}
   then with probability over the random initialization of $\nu_0$ at least
   $4/7$, one has
   \begin{align*}
      \min \set*{
         \abs{\hat{\nu} - \nu_{\grtr}} \mod \frac{\pi}{2} ,
         \frac{\pi}{2} - \left( \abs{\hat{\nu} - \nu_{\grtr}} \mod
         \frac{\pi}{2} \right)
      }
      &\leq 3\veps, \\
      \norm{\hat{u} - u_{\grtr}}_{L^2(\bbR)} &\leq 31 \sqrt{\veps}.
   \end{align*}
   In particular, the template parameters are recovered up to symmetry.

   \begin{proof}
      Following the alternating structure of the algorithm we study, the proof
      separates into a distinct stage for each phase of the algorithm. For
      concision, we will not carefully track the value of absolute constants in
      some parts of the proof; expressions such as $c, c_1, \dots$ and $C, C_1, \dots$ will
      denote small (respectively, large) absolute constants whose value may
      change from line to line unless otherwise noted. We will also use the expression $f \ltsim g$ to
      denote the statement ``there exists an absolute constant $C > 0$ such that $f
      \leq Cg$'' for functions $f$, $g$, and analogously for $f \gtsim g$.

      \paragraph{Rough factorization stage.}

      We will apply \Cref{lem:power-method-convergence} to the final iterate
      \Cref{eq:u-rough-defn}, obtained via the power method
      \Cref{eq:power_method_output}; to this end, we need to check properties of the
      operator
      \begin{equation*}
         \frac{1}{2}
         \left(
            \sT_{X_{\grtr} \circ \vtau_{\pi/4}}
            + \sT_{X_{\grtr} \circ \vtau_{\pi/4}}\adj
         \right)
         =
         \sT_{X_{\grtr} \circ \vtau_{\pi/4}}
      \end{equation*}
      (the simplification uses symmetry properties of the $\pi/4$-rotated
      template), and of the initialization \Cref{eq:u-init}.
      Notice that by \Cref{lem:diamond_diagonalized},
      we have
      \begin{equation*}
         \norm*{
            \sT_{X_{\grtr} \circ \vtau_{\pi/4}}
         }
         =
         \lambda_{\max}\left(
            \sT_{X_{\grtr} \circ \vtau_{\pi/4}}
         \right)
         =
         \frac{4}{\pi},
      \end{equation*}
      which we will denote as $\lambda_1$ (in the notation of
      \Cref{lem:power-method-convergence}),
      and its corresponding unit eigenvector is $v_1(s) = \cos(\pi s /2)
      \Ind{\abs{s} \leq 1}$.
      Moreover, \Cref{lem:diamond_diagonalized} shows that the sequence of
      eigenvalue magnitudes of this operator are a decreasing function of index
      $k$, and therefore
      \begin{equation*}
         \abs*{
            \frac{\lambda_{k}}{\lambda_1}
         }
         \leq
         \frac{4}{3\pi}
         \cdot
         \frac{\pi}{4}
         = 1 - \frac{2}{3}.
      \end{equation*}
      In addition, we calculate
      \begin{equation*}
         \ip{u_0}{v_1}_{L^2(\bbR)}
         =
         \ip{\Ind{[-1, 1]}}{\cos(\pi s /2)}_{L^2(\bbR)}
         =
         \frac{4}{\pi},
      \end{equation*}
      and evidently $\norm{u_0}_{L^2(\bbR)} = \sqrt{2}$.
      Applying the second conclusion of \Cref{lem:power-method-convergence}, we
      thus get
      \begin{equation}
         \norm*{ 
            u_{\rough}
            - 
            \frac{2}{\sqrt{\pi}}\cos (\tfrac{\pi}{2} (\spcdot)) \Ind{[-1, 1]}
         }_{L^2(\bbR)}
         \leq
         \frac{\sqrt{\pi}}{3^{T_{{\rough}} - 1}}
         \label{eq:tilt-proof-roughfact-result}
      \end{equation}
      as long as $T_{{\rough}} \geq 2$.

      \paragraph{Alignment stage.}
      To establish progress by the iteration \Cref{eq:tilt-grad-nu-iter},
      we combine a standard optimization analysis under a lower bound on the
      magnitude of the gradient with
      \Cref{lem:gradient-lower-bounds-nu}, which gives a lower bound on the
      `nominal' value of the gradient, and a basic perturbation analysis that
      uses the control we have established in the previous step between
      $u_{\rough}$ and its nominal value. 

      First, by the fact that $T_{{\rough}} \geq 2$ and $\sigma_{\SG} \leq
      \tfrac{1}{100}$, we can apply \Cref{lem:alignment-objective-sg}.
      We perform a landscape analysis 
      of the loss $\sL^{\sigma_{\SG}}(\spcdot, u_{\rough})$, where we relate it
      to properties of the `nominal loss' $\sL^{\sigma_{\SG}}(\spcdot,
      \bar{u}_{\rough})$, to guarantee progress of the gradient iteration
      \Cref{eq:tilt-grad-nu-iter}.
      The initialization $\nu_{0} \sim \Unif{[0, 2\pi]}$,
      and by \Cref{lem:gradient-lower-bounds-nu}, we see
      that the objective $\nu \mapsto \sL^{\sigma}(\nu, u)$ (for any $\sigma$
      and any $u$) is $\pi/2$-periodic and has reflection symmetry about
      $\nu_{\grtr}$ on the interval $[\nu_{\grtr} - \pi/4, \nu_{\grtr} + \pi/4]$.
      This implies that the landscape (and hence the behavior of the gradient
      descent iterates) is determined for all $\nu$ by its behavior on the
      domain $[\nu_{\grtr} - \pi/4, \nu_{\grtr} + \pi/4]$, and we can therefore
      assume that $\nu_{0} \in [\nu_{\grtr} - \pi/4, \nu_{\grtr} +
      \pi/4]$; it then follows by the uniform initialization that 
      with probability at least $(\pi/7) / (\pi/4) = 4/7$, we have
      \begin{equation*}
         \abs{\nu_{0} - \nu_{\grtr} } \leq \pi/7.
      \end{equation*}
      This means we can invoke the lower bound in
      \Cref{lem:gradient-lower-bounds-nu} to obtain that
      \begin{align*}
         \sign(\nu_{0} - \nu_{\grtr})
         \cdot
         \nabla_{\nu}\sL^{\sigma_{\SG}}(\nu_{0}, \bar{u}_{\rough})
         &\gtsim
         \sin( \abs{\nu_{0} - \nu_{\grtr}} )
         \\
         &\gtsim
         \abs{\nu_{0} - \nu_{\grtr}},
      \end{align*}
      where the last inequality uses that $\sin x \geq (2/\pi) x $ when $0 \leq
      x \leq \pi/2$.
      Meanwhile, by the first estimate of the second assertion in
      \Cref{lem:alignment-objective-sg} and
      \Cref{eq:tilt-proof-roughfact-result}, this implies
      \begin{align*}
         \sign(\nu_{ 0} - \nu_{\grtr})
         \cdot
         \nabla_{\nu}\sL^{\sigma_{\SG}}(\nu_{0}, {u}_{\rough})
         &\gtsim
         \abs{\nu_{0} - \nu_{\grtr}}
         -
         \frac{3^{-T_{{\rough}}}}{\sigma_{\SG}^2}.
         \labelthis \label{eq:tilt-proof-rough-alignment-glb}
      \end{align*}
      Next, using the upper bound in \Cref{lem:gradient-lower-bounds-nu}, we
      have that for any $\nu$,
      \begin{equation*}
         \abs*{
            \nabla_{\nu}\sL^{\sigma_{\SG}}(\nu, \bar{u}_{\rough})
         }
         \ltsim
         \abs*{\nu - \nu_{\grtr}}.
      \end{equation*}
      Combining this with 
      the first estimate of the second assertion in
      \Cref{lem:alignment-objective-sg} and
      \Cref{eq:tilt-proof-roughfact-result}, as above, we obtain
      \begin{equation*}
         \abs*{
            \nabla_{\nu}\sL^{\sigma_{\SG}}(\nu, u_{\rough})
         }
         \ltsim
         \abs*{\nu - \nu_{\grtr}}
         +
         \frac{3^{-T_{{\rough}}}}{\sigma_{\SG}^2}.
      \end{equation*}
      In particular, choosing $\beta \leq c$ for an absolute
      constant $c \leq 1$ and $T_{{\rough}} \gtsim -\log(\sigma_{\SG}^2
      \veps)$ for any $\veps >0$, we have
      \begin{equation}
         \beta
         \abs*{
            \nabla_{\nu}\sL^{\sigma_{\SG}}(\nu, u_{\rough})
         }
         \leq
         \abs*{\nu - \nu_{\grtr}}
         +
         \veps.
         \label{eq:tilt-proof-gradient-bound}
      \end{equation}
      Now, for $\gamma > 0$, define the domains
      \begin{equation*}
         S_{\gamma} = \set{\nu \in \bbR \given \abs{\nu - \nu_{\grtr}} \geq \gamma}.
      \end{equation*}
      Fix $0 < \veps < \pi/14$.
      We are going to argue that after $T_{\nu}$ iterations of
      \Cref{eq:tilt-grad-nu-iter}, the last iterate $\hat{\nu}$ satisfies
      $\hat{\nu} \in S_{3\veps}^\compl$.
      We start by proving two invariants of the sequence of iterates
      \Cref{eq:tilt-grad-nu-iter}.
      First, note that
      \begin{align*}
         \abs{\nu_{k+1} - \nu_{\grtr}}
         &=
         \abs{\nu_{k} - \beta \nabla_{\nu}
            \sL^{\sigma_{\SG}}(\nu_{k},
         u_{\rough}) - \nu_{\grtr}}
         \\
         &\leq
         2 \abs{\nu_{ k} - \nu_{\grtr}} + \veps,
         \labelthis \label{eq:tilt-proof-alignment-iterates-weak-blowup}
      \end{align*}
      by the triangle inequality and \Cref{eq:tilt-proof-gradient-bound}.
      Next, suppose that for some $k$, we have $\nu_{ k} \in S_{\veps}
      \cap S_{\pi/7}^\compl$.
      By \Cref{eq:tilt-proof-rough-alignment-glb},
      if $T_{{\rough}} \gtrsim -\log(\sigma_{\SG}^2 \veps)$, we have
      from \Cref{eq:tilt-proof-rough-alignment-glb} (via
      \Cref{lem:gradient-lower-bounds-nu})
      \begin{equation}
         \sign(\nu_{k} - \nu_{\grtr})
         \cdot
         \nabla_{\nu}\sL^{\sigma_{\SG}}(\nu_{k}, {u}_{\rough})
         \gtsim
         \abs{\nu_{k} - \nu_{\grtr}}.
         \label{eq:tilt-proof-rough-alignment-glb-farset}
      \end{equation}
      Suppose first that $\nu_{k} - \nu_{\grtr} \geq 0$.
      Then \Cref{eq:tilt-proof-rough-alignment-glb-farset} becomes
      \begin{equation*}
         \nabla_{\nu}\sL^{\sigma_{\SG}}(\nu_{k}, {u}_{\rough})
         \gtsim
         \nu_{k} - \nu_{\grtr}.
      \end{equation*}
      In particular, the gradient at $\nu_{k}$ is nonnegative. 
      This implies
      \begin{align*}
         \nu_{k+1} - \nu_{\grtr}
         &=
         \nu_{k} - \beta
         \nabla_{\nu}\sL^{\sigma_{\SG}}(\nu_{k}, u_{\rough})
         - \nu_{\grtr}
         \\
         &\leq
         \left(
            1 - c_0 \beta
         \right)
         \left(
            \nu_{k} - \nu_{\grtr}
         \right)
         \\
         &<
         \nu_{k} - \nu_{\grtr},
      \end{align*}
      since $\beta \leq 1$, where $c_0 > 0$ is an absolute
      constant that we may assume is no larger than $\half$.
      We also have, by \Cref{eq:tilt-proof-gradient-bound}, that
      \begin{equation*}
         \nu_{k+1} - \nu_{\grtr}
         =
         \nu_{k} - \beta
         \nabla_{\nu}\sL^{\sigma_{\SG}}(\nu_{k}, u_{\rough})
         - \nu_{\grtr}
         \geq -\veps.
      \end{equation*}
      But $\nu_{k} \in S_{\veps}$, so $\veps \leq \abs{\nu_{k} -
      \nu_{\grtr}}$.
      We conclude
      \begin{equation}
         \abs*{
            \nu_{k+1} - \nu_{\grtr}
         }
         \leq
         \max \set*{
            \veps,
            \left(
               1 - c_0 \beta
            \right)
            \abs*{\nu_{k} - \nu_{\grtr}}
         }
         \leq
         \abs*{\nu_{k} - \nu_{\grtr}},
         \label{eq:tilt-proof-alignment-iterates-far-contraction}
      \end{equation}
      and a completely analogous argument implies the same conclusion in the
      case where $\nu_{k} - \nu_{\grtr} \leq 0$.
      As a consequence, suppose now that for some $k$, we have $\nu_{k} \in
      S_{3\veps}^\compl$. If in fact $\nu_{k} \in
      S_{\veps}^\compl$, we know immediately from
      \Cref{eq:tilt-proof-alignment-iterates-weak-blowup} that $\nu_{ k+1} \in
      S_{3\veps}^\compl$. %
      On the other hand, if instead $\nu_{k} \in S_{\veps} \cap
      S_{3\veps}^\compl$, we have immediately from
      \Cref{eq:tilt-proof-alignment-iterates-far-contraction} that
      $\nu_{k+1} \in S_{3\veps}^\compl$.
      We conclude the full non-escape invariant:
      \begin{equation}
         \nu_{k} \in S_{3\veps}^\compl
         \implies
         \nu_{k+1} \in S_{3\veps}^\compl.
         \label{eq:tilt-proof-alignment-iterates-noescape}
      \end{equation}
      We can now give an inductive argument to obtain the desired convergence,
      namely that
      \begin{equation*}
         \hat{\nu} \in S_{3\veps}^\compl.
      \end{equation*}
      First, if $\nu_{0} \in S_{3\veps}^\compl$, we are done
      immediately, by \Cref{eq:tilt-proof-alignment-iterates-noescape}.
      If not, then by the preceding parameter choices and assumption on the
      initialization we have $\nu_{0} \in S_{3\veps} \cap
      S_{\pi/7}^\compl$ and therefore
      $\nu_{0} \in S_{\veps} \cap S_{\pi/7}^\compl$, 
      so that by \Cref{eq:tilt-proof-alignment-iterates-far-contraction},
      we obtain
      \begin{equation*}
         \nu_{1} \in %
         S_{
            \max \set{\veps,
               (1 - c_0\beta) \abs{\nu_{0} - \nu_{\grtr}}
            }
         }^\compl.
      \end{equation*}
      At this point, notice that by assumption $(1 - c_0 \beta)
      \abs{\nu_{0} - \nu_{\grtr}} \geq \tfrac{3}{2} \veps$,
      so in fact
      \begin{equation*}
         \nu_{1} \in %
         S_{
            (1 - c_0\beta) \abs{\nu_{0} - \nu_{\grtr}}
         }^\compl.
      \end{equation*}
      Proceeding inductively in this way, it follows that at iteration $k \in
      \bbN$ we either have $\nu_{k} \in S_{3\veps}^\compl$ or that
      $\nu_{k-1} \in S_{3\veps} \cap S^\compl_{\pi/7}$ and 
      \begin{equation*}
         \nu_{k} \in
         S_{
            (1 - c_0\beta) \abs{\nu_{k-1} - \nu_{\grtr}}
         }^\compl.
      \end{equation*}
      Unraveling this recurrence gives
      \begin{equation*}
         \nu_{k} \in
         S_{
            (1 - c_0\beta)^k \abs{\nu_{0} - \nu_{\grtr}}
         }^\compl.
      \end{equation*}
      Thus, as soon as
      \begin{equation*}
         T_{\nu} \gtrsim \frac{\log(3\veps)}{
            \log(1 - c_0 \beta)
         },
      \end{equation*}
      we have $\hat{\nu} \in S_{3\veps}^\compl$, i.e.\ that
      \begin{equation}
         \abs{\hat{\nu} - \nu_{\grtr}} \leq 3\veps.
         \label{eq:nu-hat-result}
      \end{equation}

      \paragraph{Refined factorization stage.} In this stage, we run power
      method to refine the factorization $u_{\rough}$, using the fact that
      $\hat{\nu} \approx \nu_{\grtr}$ to argue that the relevant operator
      for the refinement power method \Cref{eq:u-hat-defn} (c.f.\
      \Cref{eq:power_method_structure,eq:power_method_output})
      is sufficiently close to $\sT_{X_{\grtr}}$ (a self-adjoint operator,
      hence equal to its Hermitian part) that we can guarantee the progress
      of the power method \Cref{eq:u-hat-defn} by applying spectral properties
      of $\sT_{X_{\grtr}} = u_{\grtr} u_{\grtr}\adj$ in
      \Cref{lem:power-method-convergence}.
      More precisely, if we define the kernel
      \begin{equation*}
         \hat{X} = 
         \frac{1}{2}\left(
            X_{\grtr} \circ \vtau_{\hat{\nu} - \nu_{\grtr}} 
            + X_{\grtr} \circ \vtau_{-(\hat{\nu} - \nu_{\grtr})}
         \right),
      \end{equation*}
      the refinement power method \Cref{eq:u-hat-defn} involves the operator
      $\sT_{\hat{X}}$, since for any $\nu \in \bbR$ and any $f, g \in
      L^2(\bbR)$
      \begin{align*}
         \ip*{\sT_{X_{\grtr} \circ \vtau_{\nu}}[f]}{g}_{L^2(\bbR)}
         &=
         \iint_{\bbR^2} X_{\grtr} \circ \vtau_{\nu}(s,t) g(s) f(t) \diff s
         \diff t,
      \end{align*}
      and by \Cref{eq:so2-reflection-commuting} in the proof of
      \Cref{lem:curl-field-symmetries} and the fact that $X_{\grtr}(s, t) =
      X_{\grtr}(t, s)$, we have that
      \begin{equation*}
         X_{\grtr} \circ \vtau_{\nu}(s,t) 
         =
         X_{\grtr} \circ \vtau_{-\nu}(t,s) ,
      \end{equation*}
      whence
      \begin{align*}
         \ip*{\sT_{X_{\grtr} \circ \vtau_{\nu}}[f]}{g}_{L^2(\bbR)}
         &=
         \ip*{\sT_{X_{\grtr} \circ \vtau_{-\nu}}[g]}{f}_{L^2(\bbR)},
      \end{align*}
      so that 
      \begin{equation*}
         \frac{1}{2} \left(
            \sT_{
               X_{\grtr} \circ \vtau_{\hat{\nu} - \nu_{\grtr}} 
            }
            +
            \sT_{
               X_{\grtr} \circ \vtau_{\hat{\nu} - \nu_{\grtr}} 
            }\adj
         \right)
         =
         \sT_{\hat{X}}.
      \end{equation*}
      Thus, to accomplish our goal, we
      need to prove that the spectrum of $\sT_{\hat{X}}$ has a gap.
      In the sequel, we will repeatedly use the fact that $\sT_{\hat{X}}$ and
      $\sT_{X}$ are self-adjoint Fredholm operators, and that for any Fredholm
      operator with kernel $f \in L^2(\bbR^2)$, we have
      \begin{equation}
         \norm{\sT_{f}}_{\HS} = \norm{f}_{L^2(\bbR^2)}.
         \label{eq:fredholm-isomorphism}
      \end{equation}
      Conversely, if $\sT : L^2(\bbR) \to L^2(\bbR)$ is a self-adjoint
      Hilbert-Schmidt operator, it is in particular a compact operator,
      which can be written as
      $\sT = \sum_{i=1}^\infty \lambda_i v_i v_i\adj$ for
      eigenvalues $(\lambda_i)_{i \in \bbN} \subset \bbR$ and an orthonormal basis of
      eigenfunctions $(v_i)_{i \in \bbN}$, with $\norm{\sT}_{\HS} = \sum_{i\in
      \bbN}\lambda_i^2$ (c.f.\ \cite[\S B]{Heil2011-zk}); if
      we put $f(s, t) = \sum_{i \in \bbN} \lambda_i v_i(s) v_i(t)$, 
      then $f \in L^2(\bbR^2)$ and we 
      have $\sT = \sT_{f}$. This together with \Cref{eq:fredholm-isomorphism}
      shows that $f \mapsto \sT_{f}$ is an isometry of Banach spaces, and we
      will exploit this in the sequel to simplify our notation, writing (for
      example) $\norm{f}_{\HS}$ to denote $\norm{\sT_{f}}_{\HS}$ if $f \in
      L^2(\bbR^2)$, and so on.

      First, we note that $\sT_{\hat{X}}$ is Hilbert-Schmidt, by the triangle
      inequality and the fact that it is a Fredholm operator (together with the
      fact that $f \mapsto f \circ \vtau_{\nu}$ is a unitary transformation of
      $L^2(\bbR^2)$):
      \begin{align*}
         \norm{\sT_{\hat{X}}}_{\HS}
         = \norm{\hat{X}}_{L^2(\bbR^2)}
         &\leq
         \frac{1}{2}
         \left(
            \norm*{X_{\grtr} \circ \vtau_{\hat{\nu} - \nu_{\grtr}}
            }_{L^2(\bbR^2)}
            + 
            \norm{X_{\grtr} \circ \vtau_{-(\hat{\nu} - \nu_{\grtr})}}_{L^2(\bbR^2)}
         \right)
         \\
         &=
         \norm*{X_{\grtr}}_{L^2(\bbR^2)}
         =
         \norm*{\sT_{X_{\grtr}}}_{\HS}.
         \labelthis \label{eq:hatX-HS-bounded-Xgrtr}
      \end{align*}
      This means that $\sT_{\hat{X}}$ is a compact operator; it is also
      self-adjoint, by construction.
      We have therefore $\sT_{\hat{X}} = \sum_{i=1}^\infty \lambda_i v_i v_i\adj$ for
      eigenvalues $(\lambda_i)_{i \in \bbN} \subset \bbR$ and an orthonormal basis of
      eigenfunctions $(v_i)_{i \in \bbN}$ (c.f.\ \cite[\S B]{Heil2011-zk}).
      Without loss of generality, we assume that the sequence is ordered such
      that $\abs{\lambda_1} = \norm{\hat{X}}$.
      To show that the spectrum has a gap in the way that is needed to apply
      \Cref{lem:power-method-convergence}, we need to show that $\lambda_1 > 0$
      and that the rest of the spectrum is bounded in magnitude away from
      $\lambda_1$. We will establish the latter first. Note that
      \begin{align*}
         \sup_{i \geq 2}\, \abs{\lambda_i}^2
         \leq
         \sum_{i = 2}^\infty \abs{\lambda_i}^2
         &=
         -
         \abs{\lambda_1}^2
         +
         \sum_{i=1}^\infty \abs{\lambda_i}^2
         \\
         &= \norm{\hat{X}}_{\HS}^2 - \norm{\hat{X}}^2.
      \end{align*}
      By the triangle inequality and (as above) the fact that $\norm{} \leq
      \norm{}_{\HS}$, we have
      \begin{align*}
         \norm{\hat{X}}
         &\geq
         \norm{X_{\grtr}}
         -
         \norm{X_{\grtr} - \hat{X}}
         \\
         &\geq
         \norm{X_{\grtr}}
         -
         \norm{X_{\grtr} - \hat{X}}_{\HS}
         \\
         &=
         \norm{X_{\grtr}}_{\HS}
         -
         \norm{X_{\grtr} - \hat{X}}_{\HS}
      \end{align*}
      where we used the fact that $X_{\grtr} = u_{\grtr} u_{\grtr}\adj$, so
      that $\norm{X_{\grtr}}_{\HS} = u_{\grtr}\adj u_{\grtr}$ coincides with
      $\norm{X_{\grtr}} = \sup_{\norm{f}_{L^2}\leq 1}\, \abs{\ip{u_{\grtr}}{f}}
      \norm{u_{\grtr}}_{L^2} = \norm{u_{\grtr}}_{L^2}^2$ (apply the Schwarz
      inequality).
      Thus, if $\norm{X_{\grtr} - \hat{X}}_{\HS} \leq \norm{X_{\grtr}}_{\HS}$,
      we have
      \begin{align*}
         \sup_{i \geq 2}\, \abs{\lambda_i}^2
         &\leq
         \norm{\hat{X}}_{\HS}^2
         -\left(
            \norm{X_{\grtr}}_{\HS}
            -
            \norm{X_{\grtr} - \hat{X}}_{\HS}
         \right)^2
         \\
         &\leq
         2 \norm{X_{\grtr}}_{\HS}
         \norm{X_{\grtr} - \hat{X}}_{\HS},
      \end{align*}
      where the second inequality applies \Cref{eq:hatX-HS-bounded-Xgrtr}
      and discards the (negative) second-order term.
      Since $u_{\grtr} = \Ind{[-1/\sqrt{2}, 1/\sqrt{2}]}$,
      we have $\norm{X_{\grtr}}_{\HS} = \sqrt{2}$,
      and to proceed using the above it suffices to control
      $ \norm{X_{\grtr} - \hat{X}}_{\HS}$ and show that it is no larger than
      $\sqrt{2}$.
      To this end, we have
      \begin{align*}
         \norm*{\hat{X} - X_{\grtr}}
         &\leq
         \norm*{\hat{X} - X_{\grtr}}_{\mathrm{HS}}
         \\
         &\leq
         \frac{1}{2}
         \norm*{
            X_{\grtr} \circ \vtau_{\hat{\nu} - \nu_{\grtr}} 
         - X_{\grtr}}_{\mathrm{HS}}
         +
         \frac{1}{2}
         \norm*{
            X_{\grtr} \circ \vtau_{-(\hat{\nu} - \nu_{\grtr})} 
         - X_{\grtr}}_{\mathrm{HS}},
      \end{align*}
      where the second line uses the triangle inequality.
      Below, write $\alpha = 1/\sqrt{2}$.
      One has for any $\veps \in \bbR$
      \begin{align*}
         \norm*{ X_{\grtr} \circ \vtau_{\veps} - X_{\grtr}}_{\mathrm{HS}}^2
         &=
         2\norm*{ X_{\grtr}}_{L^2(\bbR^2)}^2
         - 2\ip{X_{\grtr} \circ \vtau_{\veps}}{X_{\grtr}}_{L^2(\bbR^2)} \\
         &=
         8\alpha^2 - 
         2\ip{X_{\grtr} \circ \vtau_{\veps}}{X_{\grtr}}_{L^2(\bbR^2)},
      \end{align*}
      because the operators are Fredholm operators and $\vtau_{\veps}$ is a
      unitary transformation. To estimate 
      the cross term, we argue geometrically.
      For $\abs{\veps} \leq 1$, we have the estimate
      $\abs{\cos \veps} + \abs{\sin \veps} \leq 1 + \abs{\veps}$. %
      This implies that 
      \begin{align*}
         \norm{\vR_{\veps} \vx}_{\infty}
         &\leq
         \norm*{\vR_{\veps}}_{\infty \to \infty} \norm{\vx}_{\infty}
         \\
         &\leq
         \norm{\vx}_{\infty} \left( \abs{\cos\veps} + \abs{\sin\veps} \right)
         \\
         &\leq
         \norm{\vx}_{\infty}(1 + \abs{\veps}).
      \end{align*}
      Points $\vx \in \bbR^2$ where the inner product
      $ \ip{X_{\grtr} \circ \vtau_{\veps}}{X_{\grtr}}_{L^2(\bbR^2)}$ is
      positive (it is nonnegative, because both of the integrands are
      nonnegative) are those where $\norm{\vx}_\infty \leq \alpha$ and
      $\norm{\vR_{\veps} \vx}_{\infty} \leq \alpha$. By the previous estimate,
      both conditions occur when $\norm{\vx}_{\infty} \leq \alpha / (1 +
      \abs{\veps})$.
      Since $1/(1 + \abs{\veps})^2 \geq (1 - \abs{\veps})^2$ if $\abs{\veps}
      \leq 1$, this implies
      \begin{align*}
         \ip{X_{\grtr} \circ \vtau_{\veps}}{X_{\grtr}}_{L^2(\bbR^2)}
         &\geq
         \int_{\norm{\vx}_\infty \leq \alpha/(1 + \abs{\veps})}
         \diff \vx \\
         &= \frac{4\alpha^2}{(1 + \abs{\veps})^2} \\
         &\geq 4\alpha^2 - 8\alpha^2 \abs{\veps},
      \end{align*}
      so
      \begin{equation*}
         \norm*{ X_{\grtr} \circ \vtau_{\veps} - X_{\grtr}}_{\mathrm{HS}}^2
         \leq
         16\alpha^2\abs{\veps},
      \end{equation*}
      and thus, since $\abs{\nu_{\grtr} - \hat{\nu}} \leq 1$ by the
      reduction-by-symmetry to the domain $[\nu_\grtr - \pi/4, \nu_\grtr +
      \pi/4]$ given in the previous phase of the argument,
      \begin{equation}
         \norm{\hat{X} - X_{\grtr}}
         \leq \norm{\hat{X} - X_{\grtr}}_{\HS}
         \leq 2\sqrt{2} \sqrt{\abs{\hat{\nu} - \nu_{\grtr}}}.
         \label{eq:angle-perturbation}
      \end{equation}
      In particular, if $\abs{\hat{\nu} - \nu_{\grtr}} \leq \fourth$, 
      we have 
      $\norm{\hat{X} - X_{\grtr}}_{\HS} \leq \norm{X_{\grtr}}_{\HS}$,
      and by the above
      \begin{equation*}
         \sup_{i \geq 2}\, \abs{\lambda_i}
         \leq
         2\sqrt{2} \abs*{\nu_{\grtr} - \hat{\nu}}^{1/4}.
      \end{equation*}
      Meanwhile, under this condition the above estimates yield
      \begin{align*}
         \abs{\lambda_1} = \norm{\hat{X}} 
         &\leq \norm{X_{\grtr}} + \norm{X_{\grtr} - \hat{X}} \\
         &\leq \sqrt{2} + 2\sqrt{2} \abs{\nu_{\grtr} - \hat{\nu}}^{1/2},
         \labelthis \label{eq:tilt-refinedfact-opnorm-ub}
      \end{align*}
      and moreover, by the Schwarz inequality,
      \begin{align*}
         \ip{u_{\grtr}}{\hat{X}[u_{\grtr}]}_{L^2(\bbR)}
         &\geq
         \ip{u_{\grtr}}{X_{\grtr}[u_{\grtr}]}
         -
         \norm{u_{\grtr}}_{L^2(\bbR)}^2 \norm{\hat{X} - X_{\grtr}}
         \\
         &\geq
         \norm{u_{\grtr}}_{L^2(\bbR)}^4
         -
         2 \sqrt{2} \norm{u_{\grtr}}_{L^2(\bbR)}^2 \abs{\nu_{\grtr} -
         \hat{\nu}}^{1/2}
      \end{align*}
      Since $\norm{u_{\grtr}}_{L^2(\bbR)} = 2^{1/4}$, 
      this means that if $\abs{\hat{\nu} -
      \nu_{\grtr}} \leq \tfrac{1}{64}$, we have
      \begin{equation*}
         \ip*{\frac{u_{\grtr}}{\norm{u_{\grtr}}_{L^2}}}{\hat{X}\left[\frac{u_{\grtr}}{\norm{u_{\grtr}}_{L^2}}\right]}_{L^2(\bbR)}
         \geq
         \frac{3\sqrt{2}}{4},
      \end{equation*}
      and since, under this condition, we have
      \begin{equation*}
         \sup_{i \geq 2}\, \abs{\lambda_i}
         \leq
         1,
      \end{equation*}
      we conclude that
      \begin{equation*}
         \max_{i \in \bbN} \lambda_i
         =
         \max_{\norm{u}_{L^2(\bbR)} \leq 1}\, \ip{u}{\hat{X}[u]}_{L^2(\bbR)}
         \geq
         \ip*{\frac{u_{\grtr}}{\norm{u_{\grtr}}_{L^2}}}{\hat{X}\left[\frac{u_{\grtr}}{\norm{u_{\grtr}}_{L^2}}\right]}_{L^2(\bbR)}
         >
         \sup_{i\geq2}\, \abs{\lambda_i}.
      \end{equation*}
      (c.f.\ the proof of \Cref{lem:r1approx-to-power}).
      In particular, we have $\lambda_1 > 0$.
      Following as well the above arguments, we have
      if $\abs{\hat{\nu} - \nu_{\grtr}} \leq \tfrac{1}{64}$
      \begin{equation}
         \abs{\lambda_1}
         \geq \sqrt{2} - 2\sqrt{2} \abs{\nu_{\grtr} - \hat{\nu}}^{1/2}
         \geq
         \frac{3\sqrt{2}}{4},
         \label{eq:tilt-refinedfact-opnorm-lb}
      \end{equation}
      which implies the gap condition
      \begin{equation}
         \frac{
            \sup_{i\geq2}\, \abs{\lambda_i}
         }
         {
            \lambda_1
         }
         \leq
         \frac{2 \abs{\nu_{\grtr} - \hat{\nu}}^{1/4}}{1 - 2\abs{\nu_{\grtr} -
         \hat{\nu}}^{1/2}}
         \leq
         \frac{8}{3}
         \abs{\nu_{\grtr} - \hat{\nu}}^{1/4}
         <1
         \label{eq:refined-fact-gap}
      \end{equation}
      under the preceding assumptions.
      Using these characterizations of the spectral gap, we can conveniently
      also apply \cite[Proposition 6.1]{Davis1970-po}
      to obtain
      \begin{equation*}
         \norm*{\frac{u_{\grtr}}{\norm{u_{\grtr}}_{L^2}} \frac{u_{\grtr}\adj}{\norm{u_{\grtr}}_{L^2}} - v_1 v_1 \adj}_{\HS} \leq
         \frac{2}{\sqrt{2}-1} \norm{X_{\grtr} - \hat{X}}_{\HS}
         \leq
         14 \sqrt{\abs{\nu_{\grtr} - \hat{\nu}}}.
      \end{equation*}
      Meanwhile, we have
      \begin{equation*}
         \norm*{\frac{u_{\grtr}}{\norm{u_{\grtr}}_{L^2}}
         \frac{u_{\grtr}\adj}{\norm{u_{\grtr}}_{L^2}} - v_1 v_1 \adj}_{\HS}^2
         =
         2 \left(
            1 - 
            \ip*{\frac{u_{\grtr}}{\norm{u_{\grtr}}_{L^2}}}{v_1}^2
         \right).
      \end{equation*}
      Since $v_1$ is only defined up to
      sign, let us suppose without loss of generality that $v_1$ is such that
      $\ip{v_1}{u_{\grtr}}_{L^2} \geq 0$. Proceeding, we then obtain
      \begin{align*}
         \norm*{\frac{u_{\grtr}}{\norm{u_{\grtr}}_{L^2}}
         \frac{u_{\grtr}\adj}{\norm{u_{\grtr}}_{L^2}} - v_1 v_1 \adj}_{\HS}^2
         &=
         2 \left(
            1 - 
            \ip*{\frac{u_{\grtr}}{\norm{u_{\grtr}}_{L^2}}}{v_1}
         \right)
         \left(
            1 + 
            \ip*{\frac{u_{\grtr}}{\norm{u_{\grtr}}_{L^2}}}{v_1}
         \right)
         \\
         &\geq
         2 \left(
            1 - 
            \ip*{\frac{u_{\grtr}}{\norm{u_{\grtr}}_{L^2}}}{v_1}
         \right)
         \\
         &=
         \norm*{
            \frac{u_{\grtr}}{\norm{u_{\grtr}}_{L^2}}
            -
            v_1
         }_{L^2(\bbR)}^2.
      \end{align*}
      Combining, this gives
      \begin{equation}
         \norm*{
            2^{-1/4} u_{\grtr}
            -
            v_1
         }_{L^2(\bbR)}
         \leq
         14 \sqrt{\abs{\nu_{\grtr} - \hat{\nu}}}.
         \label{eq:davis-kahan}
      \end{equation}
      This allows us to lower bound the correlation between the power method
      initialization $u_{\rough}$ and the target eigenvector: by the triangle
      inequality and the Schwarz inequality,
      \begin{align*}
         \ip{v_1}{u_{\rough}}
         &\geq
         \ip{v_1}{\bar{u}_{\rough}} - \norm{u_{\rough} -
         \bar{u}_{\rough}}_{L^2(\bbR)}
         \\
         &\geq
         2^{-1/4} \ip{u_{\grtr}}{\bar{u}_{\rough}} 
         - \norm{v_1 - 2^{-1/4} u_{\grtr}}_{L^2(\bbR)}
         - \norm{u_{\rough} - \bar{u}_{\rough}}_{L^2(\bbR)}.
      \end{align*}
      We calculate
      \begin{equation*}
         2^{-1/4} 
         \ip{u_{\grtr}}{\bar{u}_{\rough}} 
         =
         \frac{2}{2^{1/4} \sqrt{\pi}}
         \int_{-1}^1 \cos(\pi x / 2) \diff x
         =
         \frac{8}{2^{1/4} \pi^{3/2}},
      \end{equation*}
      so that, by \Cref{eq:tilt-proof-roughfact-result,eq:davis-kahan},
      we have for $T_{\rough} \geq 4$ and $\abs{\nu_{\grtr} - \hat{\nu}} \leq
      \tfrac{1}{256}$ that
      \begin{equation}
         \ip{v_1}{u_{\rough}}
         \geq
         \frac{1}{4}.
         \label{eq:refined-fact-initcorr}
      \end{equation}
      Moreover, it allows us to control the unnormalized distance between the
      power method output and the target in terms of the normalized distance:
      the triangle inequality gives
      \begin{align*}
         \norm*{u_{\grtr} - \sqrt{\lambda_1} v_1}_{L^2(\bbR)}
         &\leq
         \max \set*{ \norm{u_{\grtr}}_{L^2(\bbR)}, \sqrt{\lambda_1}}
         \norm*{\frac{u_{\grtr}}{\norm{u_{\grtr}}_{L^2}} - v_1}_{L^2(\bbR)}
         + \abs{\norm{u_{\grtr}}_{L^2(\bbR)}- \sqrt{\lambda_1}}
         \\
         &\leq
         14 \left( 2^{1/4} + 2^{3/8} \abs{\nu_{\grtr} - \hat{\nu}}^{1/4}
         \right)
         \sqrt{\abs{\nu_{\grtr} - \hat{\nu}}}
         + \abs{\norm{u_{\grtr}}_{L^2(\bbR)}- \sqrt{\lambda_1}},
      \end{align*}
      where the second line applies
      \Cref{eq:tilt-refinedfact-opnorm-ub}.
      Meanwhile, 
      \Cref{eq:tilt-refinedfact-opnorm-ub,eq:tilt-refinedfact-opnorm-lb} imply
      \begin{equation*}
         2^{1/4} \left(
            \sqrt{ 1 - 2 \abs{\nu_{\grtr} - \hat{\nu}}^{1/2} } - 1
         \right)
         \leq 
         \sqrt{\lambda_1} - 2^{1/4}
         \leq
         2^{1/4} \left(
            \sqrt{ 1 + 2 \abs{\nu_{\grtr} - \hat{\nu}}^{1/2} } - 1
         \right),
      \end{equation*}
      and given that $2 \abs{\nu_{\grtr} - \hat{\nu}}^{1/2} \leq \tfrac{1}{8}$
      by our assumptions,
      the inequalities $\sqrt{1 + x} \leq 1 + x/2$ and $\sqrt{1 - x} \geq 1 -
      x$ (the latter valid if $0 \leq x \leq 1$)
      lead to the bounds
      \begin{equation*}
         -2^{5/4} \abs{\nu_{\grtr} - \hat{\nu}}^{1/2}
         \leq 
         \sqrt{\lambda_1} - 2^{1/4}
         \leq
         2^{5/4} \abs{\nu_{\grtr} - \hat{\nu}}^{1/2},
      \end{equation*}
      so, plugging into the previous estimate, we obtain
      \begin{align*}
         \norm*{u_{\grtr} - \sqrt{\lambda_1} v_1}_{L^2(\bbR)}
         \leq
         30\sqrt{\abs{\nu_{\grtr} - \hat{\nu}}}
         \labelthis \label{eq:refined-fact-unnormalized-diff}
      \end{align*}
      after worst-casing constants.
      We can finally apply \Cref{lem:power-method-convergence} with the
      properties \Cref{eq:refined-fact-gap,eq:refined-fact-initcorr} together
      with the triangle inequality and \Cref{eq:refined-fact-unnormalized-diff}
      to obtain
      that the iteration \Cref{eq:u-hat-defn} satisfies
      \begin{equation*}
         \norm{\hat{u} - u_{\grtr}}_{L^2(\bbR)}
         \leq
         30\sqrt{\abs{\nu_{\grtr} - \hat{\nu}}}
         +
         28\left(
            \frac{8}{3} \abs{\nu_{\grtr} - \hat{\nu}}^{1/4}
         \right)^{T_u}
      \end{equation*}
      after worst-casing constants slightly.

      \paragraph{Concluding the result.} Finally, we instantiate our results
      above with appropriate parameter choices to obtain the desired
      conclusion.
      We have shown the following: there are absolute constants $c_1$, $C_1$,
      $C_2 > 0$
      such that for any parameters $\sigma$, $\beta$ satisfying
      \begin{align*}
         \sigma{\SG}^2 &\leq \tfrac{1}{10^4}, \\
         \beta &\leq c_1,
      \end{align*}
      for any $0 < \veps \leq \tfrac{1}{768}$, if the iteration counts satisfy
      \begin{align*}
         T_{\rough} &\geq -C_1 \log(\sigma_{\SG}^2 \veps),
         \\
         T_{\nu} &\geq -\frac{C_2  \log(3\veps)}{\beta}, 
         \\
         T_u &\geq 16,
      \end{align*}
      then with probability over the random initialization of $\nu_0$ at least
      $4/7$, one has
      \begin{align*}
         \abs{\hat{\nu} - \nu_{\grtr}} &\leq 3\veps, \\
         \norm{\hat{u} - u_{\grtr}}_{L^2(\bbR)} &\leq 31 \sqrt{\veps}.
      \end{align*}
      The condition on $T_{u}$ is quite mild because the rate of convergence
      improves with the quality of the output of the alignment stage of the
      algorithm, since the target $u_{\grtr} u_{\grtr}\adj$ is rank one.
      The condition on $T_{\nu}$ above takes advantage of the fact that when $0
      \leq x \leq \half$, we have by concavity $\log(1 - x) \geq (-2\log 2) x$ to
      simplify the stated bound in our previous work.

   \end{proof}

\end{theorem}

\begin{remark}
   \Cref{thm:tilt-infinite} establishes a linear rate of convergence of both
   the alignment iterates $\nu_{k}$ and the representation iterations that
   generate $u_{\rough}$ and $\hat{u}$ to the true parameters of the template,
   up to symmetry. The dependence on the other problem parameters in these
   rates, namely the smoothing $\sigma^2$ and the step size $\beta$, is about
   as mild as one would hope for: the smoothing level $\sigma^2$ only enters
   the rates logarithmically, and the step size is only required to be smaller
   than an absolute constant, which is reflected as a linear dependence in the
   rate of convergence of the alignment step of the algorithm.
   The issue of smoothing represents an interesting conceptual takeaway from
   our analysis, with regards to modern 3D representation approaches like
   \ours{} which do not explicitly incorporate a classical coarse-to-fine
   smoothing schedule, as in, for example, image registration
   \cite{Lefebure2001-uz}. Our proofs demonstrate that the reason smoothing is
   not necessary for precise local convergence is that \textit{computational
      constraints on the representation capacity of the method (here, a
      rank-one matrix) and the $L^2$ loss create texture gradients (i.e.,
   blurry images) when optimizing the representation}, and these texture
   gradients cause the subsequent alignment landscape to be smoother than
   one might otherwise expect. For example, when $\sigma_{\SG}^2$ is small,
   the smoothed template $\varphi_{\sigma_{\SG}^2} \conv X_{\grtr}$ has a
   Lipschitz constant on the order of $1/\sigma_{\SG}$, due to sharp edges
   in $X_{\grtr}$; nevertheless, this sharpness does not reflect in our
   rates as a consequence of the blessings of capacity-constrained inexact
   representation.

   \Cref{thm:tilt-infinite} contains a hypothesis on the probability of
   success, which is asserted to be at least $\tfrac{4}{7}$; since this value
   is larger than $\half$, it is theoretically possible to boost the success
   probability to an arbitrarily-high level by running multiple independent
   trials of the algorithm and aggregating the outputs appropriately. In
   practice, of course, the algorithm succeeds with probability one on the
   template $X_{\grtr}$: the discrepancy is due to the technical need to prove
   that the nonconvex alignment landscape $\nu \mapsto \sL^{\sigma_{\SG}}(\nu,
   u_{\rough})$ has suitable negative curvature in neighborhoods of the
   maximizers $(\nu_{\grtr} + \tfrac{\pi}{4}) + \tfrac{\pi}{2} \bbZ$.
   This ``benign global geometry'', in the sense of \textcite{Zhang2020-on},
   has been studied in other contexts
   \cite{Chi2018-au,Gilboa2019-px,Stoger2021-an}, although the mixture
   of discrete and continuous symmetries in the $(\nu, u)$ landscape of
   \Cref{eq:tilt-objective-ezzz} is somewhat distinguished.
   In general, we have endeavored to keep the optimization analysis in
   \Cref{thm:tilt-infinite} as elementary as possible, and we have not made
   attempts to optimize absolute constants. Simple and standard modifications
   to the proofs can be made to yield slightly better constants and rates, at
   the cost of additional technicality
   \cite{Nesterov2004-oz,Chi2018-au,Zhang2020-on}.

\end{remark}

\begin{remark}

   We discuss three directions of extension for \Cref{thm:tilt-infinite} below.
   \begin{enumerate}
      \item \textbf{Joint factorization and alignment.} 
         The algorithm we study, in its use of the power method as a standin
         for matrix factorization as in \Cref{eq:tilt-objective-ezzz} as well
         as its use of ``block'' alternating minimization iterations rather
         than alternating gradient steps on $\nu$ and $u$, differs from our
         implementation of \ours{} in ways that present important directions
         for further technical improvement.
         It seems to us that extending our analysis to the case of alternating
         gradient steps would require some additional conceptual insight (e.g.,
         the identification of a conserved quantity):
         various technical components of the current alignment argument are
         delicate and break when the factorization target is not constant.
         A similar issue is associated with the extension of the result to
         $\nu_{\grtr} \neq \pi/4$, although this case seems ``easier''; e.g.,
         perturbative analogues of \Cref{lem:diamond_diagonalized} for other
         values of $\nu_{\grtr}$ would suffice here.
         In the alternating $(\nu, u)$ setting, there is also a challenge
         associated with the fact that texture gradients induced by inexact
         factorization become smaller as the alignment becomes more accurate,
         making the landscape nonsmooth. Rather than introducing auxiliary
         smoothing in this setting, it may be most relevant to practice to
         study the nonsmooth landscape directly, \`{a} la \cite{Bai2019-xk};
         the analysis can be less technical in our setting since there is no
         statistical component to the problem.
         Extending the loss \Cref{eq:tilt-objective-ezzz} to the setting of
         overparameterized matrix factorization is also interesting; developing
         this extension in the context of observations $X_{\grtr}$ with
         background clutter, as discussed below, may be the most natural
         setting.
      \item \textbf{Extensions to multi-object scenes and three dimensions.}
         Establishing a 
         direct 3D analogue of \Cref{thm:tilt-infinite} seems to be mostly technical.
         Extending the result to apply to scenes $X_{\grtr}$ with other objects
         and background clutter present seems more challenging: the proof of
         \Cref{thm:tilt-infinite} presents a perturbative framework for
         analyzing \ours{} that should
         not be hard to extend to `perturbed' observations $X_{\grtr}$ when the
         magnitude of the perturbation is small, but getting insights into how
         the algorithm can be changed to cope with the kinds of
         structured perturbations that arise in real-world scenes (e.g., a
         scene with objects with shape content that cannot be axis-aligned as
         well as the square, like people, but that nonetheless contains enough
         `prominent' axis-alignable components, such as buildings and roads in a
         built environment, for success to be possible) seems to require novel ideas.
         One path forward here could be to introduce additional appearance
         components to the square $X_{\grtr}$, such as a texture, and study
         conditions under which these are sufficiently decorrelated with the
         shape components of the template for success to remain possible.
         Another possibility is to study the overparameterized case and
         separate distinct factorization components into ``groups'', as in our
         implementation of \ours{}, which have their own transformations and
         can thus represent distinct parts of the scene; the necessary
         symmetry-breaking aspects of such a result feel reminiscent of
         analyses of dictionary learning \cite{Zhang2020-on}, but feel
         significantly more challenging due to the need to localize different
         objects via factorization in the setting of \ours{}.
      \item \textbf{Computing with a MLP.} An extremely important avenue for
         extension of \Cref{thm:tilt-infinite} is to go beyond the linear
         representation studied there and introduce a neural network for
         representing the scene. This presents an additional challenge with
         respect to disentangling appearance and shape versus our current
         analysis: there, the rank-one capacity constraint on the factorization
         leads to inexact intermediate factorizations that create texture
         gradients to help alignment, and alignment improvements help to
         further improve the representation. With a MLP, there seems to be some
         natural capacity to represent coordinate rotations (c.f.\
         \Cref{sec:app_resampling})---understanding how initialization and
         implicit bias of gradient descent training preserves the disentangled
         learning of appearance and alignment that we prove occurs in the
         linear model is a fascinating direction for future work.
   \end{enumerate}
\end{remark}

\subsubsection{Supporting Results}

\begin{lemma}
   \label{lem:power-method-convergence}
   Let $\sT : L^2(\bbR) \to L^2(\bbR)$ be a nonzero self-adjoint Hilbert-Schmidt
   operator with corresponding eigenvalues $(\lambda_k)_{k\in \bbN}$ and
   orthonormal basis $(v_k)_{k \in \bbN} \subset L^2(\bbR)$. Without loss of
   generality, suppose that $\norm{\sT} = \abs{\lambda_1} > 0$. Suppose
   moreover that $\lambda_1 > 0$,\footnote{If $\lambda_1 < 0$, it is necessary
      to take absolute values in order to obtain the result asserted here:
      consider for example the case where $\sT = -\Id$, so that $u_k = (-1)^k u_0$ if
   $\norm{u_0}_{L^2(\bbR)} = 1$.} and that the spectrum has a gap, i.e., that for
   some $0 < \gamma \leq 1$ we have $\lambda_1 - \abs{\lambda_k} \geq \gamma
   \lambda_1$ for all $k > 1$.  Consider power method on $\sT$, starting from
   initialization $u_0 \in L^2(\bbR)$:
   \begin{equation*}
      u_{k+1} = \frac{\sT u_k}{\norm{\sT u_k}_{L^2(\bbR)}}.
   \end{equation*}
   Then if $\abs{\ip{v_1}{u_0}_{L^2(\bbR)}} \geq \eta > 0$,
   it holds
   \begin{equation*}
      \norm*{ u_k - \sign(\ip{v_1}{u_0}_{L^2(\bbR)}) v_1 }_{L^2(\bbR)}^2
      \leq
      2(1 - \gamma)^{2k} \frac{\norm{u_0}_{L^2(\bbR)}^2}{\eta^2}.
   \end{equation*}
   In addition, if the iteration count satisfies
   \begin{equation*}
      k \geq 
      \frac{
         \log( \eta / 4\norm{u_0}_{L^2(\bbR)} )
      }{
         \log(1 - \gamma)
      },
   \end{equation*}
   then the `rank-one approximating factor' error satisfies
   \begin{equation*}
      \norm*{ 
         \sqrt{u_k \adj \sT u_k} u_k 
         - \sqrt{\lambda_1} \sign(\ip{v_1}{u_0}_{L^2(\bbR)}) v_1 
      }_{L^2(\bbR)}^2
      \leq
      18{\lambda_1} (1 - \gamma)^{2k} \frac{\norm{u_0}_{L^2(\bbR)}^2}{\eta^2}.
   \end{equation*}

   \begin{proof}
      We apply the standard argument; under the assumption of a Hilbert-Schmidt
      operator with a gapped spectrum, as we have made here,
      the standard argument's convergence is actually dimension-free, in
      contrast to the general case (c.f.\ \cite{Kuczynski1992-fu}).
      We use basic notions from the analysis of self-adjoint Hilbert-Schmidt
      operators in the proof (see \cite[\S B]{Brezis2011-wx}).

      The main observation to make is that the update equation for $u_{k+1}$ in
      the definition of the power method is a $0$-absolutely-homogeneous
      function of $u_k$. This implies
      \begin{equation*}
         u_{k} = \frac{\sT^k u_0}{\norm*{\sT^k u_0}_{L^2(\bbR)}}.
      \end{equation*}
      It is clear that this iteration is well-defined, i.e., that for every $k$
      one has $\sT^k u_0 \neq 0$, by the assumption that $\abs{\ip{v_1}{u_0}} >
      0$ (and the fact that $\abs{\lambda_1} > 0$), since, as we will use
      below,
      \begin{align*}
         \sT^k u_0
         &=
         \sum_{l=1}^\infty
         \lambda_l^k \ip{v_l}{u_0}_{L^2(\bbR)} v_l,
         \\
         \norm*{\sT^k u_0}_{L^2(\bbR)}^2
         &=
         \sum_{l=1}^\infty
         \lambda_l^{2k} \abs{\ip{v_l}{u_0}_{L^2(\bbR)}}^2,
      \end{align*}
      since the sequence $(v_l)$ is an orthonormal basis.
      Moreover, notice that if we initialize the power method with $-u_0$
      instead of $u_0$, we end up only changing the sign of the output $u_k$;
      hence we can assume below without loss of generality that
      $\ip{v_1}{u_0} > 0$.
      Now, expanding the square shows that
      \begin{align*}
         \norm*{u_k - v_1}_{L^2}^2
         &=
         \norm*{
            \frac{\sT^k u_0}{\norm*{\sT^k u_0}_{L^2(\bbR)}}
            -
            v_1
         }_{L^2}^2
         \\
         &=
         2\left(
            1 - \frac{\lambda_1^k \ip{v_1}{u_0}_{L^2(\bbR)}}{
               \norm{\sT^k u_0}_{L^2(\bbR)}
            }
         \right).
      \end{align*}
      Since $\lambda_1 > 0$, we have
      \begin{align*}
         \frac{
            \norm{\sT^k u_0}_{L^2(\bbR)}
         }
         {
            \lambda_1^k \ip{v_1}{u_0}_{L^2(\bbR)}
         }
         &=
         \left(
            \frac{1}{\lambda_1^{2k} \ip{v_1}{u_0}_{L^2(\bbR)}^2}
            \sum_{l=1}^\infty
            \lambda_l^{2k} \abs{\ip{v_l}{u_0}_{L^2(\bbR)}}^2
         \right)^{1/2}
         \\
         &=
         \left(
            1 + 
            \sum_{l=2}^\infty
            \left(
               \frac{ \abs{\lambda_l} } { \abs{\lambda_1} }
            \right)^{2k}
            \frac{
               \ip{v_l}{u_0}_{L^2(\bbR)}^2
            }{
               \ip{v_1}{u_0}_{L^2(\bbR)}^2
            }
         \right)^{1/2}.
      \end{align*}
      The gapped assumption implies that
      \begin{equation*}
         \frac{ \abs{\lambda_l} } { \abs{\lambda_1} }
         \leq
         1-\gamma,
      \end{equation*}
      and the lower bound $ \ip{v_1}{u_0}_{L^2(\bbR)}^2 \geq \eta^2$ then
      implies
      \begin{equation*}
         \frac{
            \norm{\sT^k u_0}_{L^2(\bbR)}
         }
         {
            \lambda_1^k \ip{v_1}{u_0}_{L^2(\bbR)}
         }
         \leq
         \left(
            1 + (1 - \gamma)^{2k} \frac{\norm{u_0}_{L^2(\bbR)}^2}{\eta^2}
         \right)^{1/2}.
      \end{equation*}
      For $x \geq 0$, the function $x \mapsto 1 - (1 + x)^{-1/2}$ is increasing
      and concave, and therefore satisfies $1 - (1+x)^{-1/2} \leq \half x$. As
      a result, we have
      \begin{equation*}
         \norm*{u_k - v_1}_{L^2}^2
         \leq
         2(1 - \gamma)^{2k} \frac{\norm{u_0}_{L^2(\bbR)}^2}{\eta^2},
      \end{equation*}
      as claimed.

      To obtain the claimed estimate for the rank-one approximating factor
      error, we use the triangle inequality and a bound for the square root
      when its argument is sufficiently far from $0$.
      Notice that $\sqrt{\lambda_1} = \sqrt{v_1 \adj \sT v_1}$,
      and by the triangle inequality
      \begin{align*}
         \abs*{u_k\adj \sT u_k - \lambda_1}
         &\leq
         \abs*{
            u_k\adj \sT u_k - u_k\adj \sT v_1
         }
         +
         \abs*{
            u_k\adj \sT v_1 - v_1\adj \sT v_1
         }
         \\
         &\leq
         2 \norm{\sT}
         \norm*{
            u_k - v_1
         }_{L^2(\bbR)}
         \\
         &=
         2\lambda_1 
         \norm*{
            u_k - v_1
         }_{L^2(\bbR)}.
         \labelthis \label{eq:approx-eigenvalue-distance}
      \end{align*}
      by the Schwarz inequality and the fact that $u_k$ and $v_1$ are unit
      norm.
      This implies
      \begin{equation*}
         u_k\adj \sT u_k
         \geq
         \lambda_1\left(
            1 - 2 \norm{u_k - v_1}_{L^2(\bbR)}
         \right).
      \end{equation*}
      As a result, if
      \begin{equation*}
         k \geq 
         \frac{
            \log( \eta / 4\norm{u_0}_{L^2(\bbR)} )
         }{
            \log(1 - \gamma)
         },
      \end{equation*}
      we have $u_k\adj \sT u_k > 0$, and square roots can be taken without
      worry.
      We have by the triangle inequality
      \begin{align*}
         \norm*{
            \sqrt{u_k \adj \sT u_k}
            u_k
            -
            \sqrt{\lambda_1} v_1
         }_{L^2(\bbR)}
         &\leq
         \norm*{
            \sqrt{u_k \adj \sT u_k}
            u_k
            -
            \sqrt{v_1 \adj \sT v_1}
            u_k
         }_{L^2(\bbR)}
         +
         \norm*{
            \sqrt{v_1 \adj \sT v_1}
            u_k
            -
            \sqrt{v_1 \adj \sT v_1}
            v_1
         }_{L^2(\bbR)}
         \\
         &\leq
         \abs*{
            \sqrt{u_k \adj \sT u_k}
            -
            \sqrt{v_1 \adj \sT v_1}
         }
         +
         \sqrt{\lambda_1}
         \norm*{
            u_k - v_1
         }_{L^2(\bbR)},
      \end{align*}
      since $u_k$ and $v_1$ are unit norm.
      Now, by the fundamental theorem of calculus, we have for any $x, y \geq
      0$
      \begin{equation*}
         \abs*{ \sqrt{x} - \sqrt{y} }
         =
         \frac{1}{2} \abs*{
            \int_y^x z^{-1/2} \diff z
         }
         \leq
         \frac{\abs{x-y}}{2 \sqrt{\min \set{x, y}}},
      \end{equation*}
      and in our setting, we have shown above
      \begin{equation*}
         u_k \adj \sT u_k
         \geq
         \left(
            1 - \frac{1}{\sqrt{2}}
         \right)
         \lambda_1
      \end{equation*}
      by our choice of $k$.
      Since $2(1 - 1/\sqrt{2})^{1/2} \geq 1$, 
      it follows
      \begin{align*}
         \abs*{
            \sqrt{u_k \adj \sT u_k}
            -
            \sqrt{v_1 \adj \sT v_1}
         }
         &\leq
         \frac{1}{\sqrt{\lambda_1}}
         \abs*{
            u_k \adj \sT u_k
            -
            v_1 \adj \sT v_1
         }
         \\
         &\leq
         2 \sqrt{\lambda_1}
         \norm*{u_k - v_1}_{L^2(\bbR)}.
      \end{align*}
      where we used \Cref{eq:approx-eigenvalue-distance} in the final line.
      Consequently, we have shown
      \begin{equation*}
         \norm*{
            \sqrt{u_k \adj \sT u_k}
            u_k
            -
            \sqrt{\lambda_1} v_1
         }_{L^2(\bbR)}
         \leq
         3 \sqrt{\lambda_1}
         \norm*{u_k - v_1}_{L^2(\bbR)},
      \end{equation*}
      as claimed.

   \end{proof}
\end{lemma}

\begin{lemma}
   \label{lem:tilt_grad_hess}
   For the objective $\sL^\sigma$ defined in \Cref{eq:tilt-objective-ez}, one has
   \begin{align*}
      \nabla_{\nu} \sL^\sigma(\nu, \vU, \vV) 
      &= 
      -\ip*{\varphi_{\sigma^2}\ast
         \left( 
            \vU \vV\adj\circ \vtau_{\nu_{\grtr} - \nu}
         \right)
      }{
         \ip*{
            \nabla_{\vx}[\varphi_{\sigma^2} \ast X_{\grtr}]
         }{
            \begin{bmatrix}
               0 & -1 \\
               1 & 0
            \end{bmatrix}
            (\spcdot)
         }_{\ell^2}
      }_{L^2(\bbR^2)},
      \\
      \nabla_{\vU} \sL^\sigma(\nu, \vU, \vV) &=
      -\left[ \varphi_{2\sigma^2} \ast
         \left( 
            X_{\grtr} \circ \vtau_{\nu - \nu_{\grtr}} - \vU \vV\adj
         \right)
      \right]\vV
      \\
      \nabla_{\vV} \sL^\sigma(\nu, \vU, \vV) &=
      -\left[\varphi_{2\sigma^2} \ast
         \left( 
            X_{\grtr} \circ \vtau_{\nu_{\grtr} - \nu} - \vV \vU\adj
         \right)
      \right]\vU.
   \end{align*}

   \begin{proof}
      For the gradients of $\sL^\sigma$ with respect to $(\vU, \vV)$, direct
      calculation using the chain rule for the Fr\'{e}chet derivative and
      the duality of $L^2(\bbR)$ gives
      \begin{align*}
         \nabla_{\vU} \sL^\sigma(\nu, \vU, \vV) &=
         -\left[ \varphi_{2\sigma^2} \ast
            \left( 
               X_{\grtr} \circ \vtau_{\nu - \nu_{\grtr}} - \vU \vV\adj
            \right)
         \right]\vV
         \\
         \nabla_{\vV} \sL^\sigma(\nu, \vU, \vV) &=
         -\left[\varphi_{2\sigma^2} \ast
            \left( 
               X_{\grtr} \circ \vtau_{\nu - \nu_{\grtr}} - \vU \vV\adj
            \right)
         \right]\adj \vU.
      \end{align*}
      It is convenient to simplify the adjoint operation in the second
      expression. First, note that $f \mapsto [\varphi_{2\sigma^2} \ast (
      X_{\grtr}
      \circ \vtau_{\nu - \nu_{\grtr}} - \vU \vV\adj) ][f]$ is a bounded operator on
      $L^2(\bbR)$, because its $L^2 \to L^2$ operator norm is bounded by its
      Hilbert-Schmidt norm, which is finite:
      \begin{equation*}
         \norm*{
            \varphi_{2\sigma^2} \ast ( X_{\grtr} \circ \vtau_{\nu - \nu_{\grtr}} - \vU
            \vV\adj)
         }_{\mathrm{HS}}^2
         =
         \int_{\bbR^2}
         [
         \varphi_{2\sigma^2} \ast ( X_{\grtr} \circ \vtau_{\nu - \nu_{\grtr}} - \vU
         \vV\adj)
         ](s, t)^2 \diff s \diff t
         < +\infty,
      \end{equation*}
      by Young's inequality for convolutions. This allows us to use Fubini's
      theorem freely in the sequel.
      For any $f, g \in L^2(\bbR)$, we have
      \begin{align*}
         &\ip*{
            \left[\varphi_{2\sigma^2} \ast
               \left( 
                  X_{\grtr} \circ \vtau_{\nu - \nu_{\grtr}} - \vU \vV\adj
               \right)
            \right][f]
         }
         {
            g
         }_{L^2(\bbR)}
         \\
         &\qquad=
         \int_{\bbR}  g(s) \left(
            \int_{\bbR} \left[
               \varphi_{2\sigma^2} \ast (X_{\grtr} \circ \vtau_{\nu -
               \nu_{\grtr}} -
               \vU\vV\adj)
            \right](s, t) f(t) \diff t
         \right) \diff s \\
         &\qquad=
         \int_{\bbR}
         \int_{\bbR} 
         \int_{\bbR^2}
         f(t) g(s)
         \varphi_{2\sigma^2}((s, t) - \vx) 
         (X_{\grtr} \circ \vtau_{\nu - \nu_{\grtr}} - \vU\vV\adj)(\vx)
         \diff \vx
         \diff t
         \diff s \\
         &\qquad=
         \int_{\bbR}
         \int_{\bbR} 
         \int_{\bbR^2}
         \iiint
         f(t) g(s)
         \varphi_{2\sigma^2}((t, s) - \vx) 
         (X_{\grtr} \circ \vtau_{\nu - \nu_{\grtr}} - \vU\vV\adj)\left(
            \begin{bmatrix} 1 & 0 \\ 0 & 1 \end{bmatrix} \vx
         \right)
         \diff \vx
         \diff t
         \diff s,
         \labelthis \label{eq:smoothed-residual-adjoint-tmp}
      \end{align*}
      where the third equality uses a unitary change of variables $\vx \mapsto
      \begin{bmatrix} 0 & 1 \\ 1 & 0 \end{bmatrix} \vx$ in the convolution
      integral. Notice that
      \begin{align*}
         (X_{\grtr} \circ \vtau_{\nu - \nu_{\grtr}} - \vU\vV\adj)(s, t)
         &=
         X_{\grtr} \circ \vtau_{\nu - \nu_{\grtr}}(s, t)
         -
         \sum_{i=1}^k u_i(s) v_i(t) \\
         &=
         X_{\grtr}\left(
            \vR_{\nu - \nu_{\grtr}} \begin{bmatrix} 0 & 1 \\ 1 & 0 \end{bmatrix} (t,
            s)
         \right)
         -
         \sum_{i=1}^k  v_i(t) u_i(s).
      \end{align*}
      Moreover, if $\vQ$ is any orthogonal matrix with determinant
      $-1$, then for any $\nu$ one has $\det(\vR_{\nu} \vQ) = -1$; because the
      orthogonal matrices form a Lie group and every $2\times 2$ orthogonal
      matrix with determinant $-1$ is symmetric, it follows 
      \begin{equation}
         \vR_\nu \vQ = \vQ \vR_{\nu}\adj.
         \label{eq:so2-reflection-commuting}
      \end{equation}
      In particular,
      \begin{align*}
         \vR_{\nu - \nu_{\grtr}} \begin{bmatrix} 0 & 1 \\ 1 & 0 \end{bmatrix}
         &=
         \begin{bmatrix} 0 & 1 \\ 1 & 0 \end{bmatrix}\vR_{\nu - \nu_{\grtr}}\adj,
      \end{align*}
      which, together with the fact that $X_{\grtr}(s, t) = X_{\grtr}(t, s)$, implies
      that 
      \begin{equation*}
         (X_{\grtr} \circ \vtau_{\nu - \nu_{\grtr}} - \vU\vV\adj)(s, t)
         = (X_{\grtr} \circ \vtau_{\nu_{\grtr} - \nu} - \vV\vU\adj)(t, s).
      \end{equation*}
      Applying this to \Cref{eq:smoothed-residual-adjoint-tmp} and unwinding
      the preceding steps implies immediately
      \begin{equation*}
         \ip*{
            \left[\varphi_{2\sigma^2} \ast
               \left( 
                  X_{\grtr} \circ \vtau_{\nu - \nu_{\grtr}} - \vU \vV\adj
               \right)
            \right][f]
         }
         {
            g
         }_{L^2(\bbR)}
         =
         \ip*{
            \left[\varphi_{2\sigma^2} \ast
               \left( 
                  X_{\grtr} \circ \vtau_{\nu_{\grtr} - \nu} - \vV \vU\adj
               \right)
            \right][g]
         }
         {
            f
         }_{L^2(\bbR)},
      \end{equation*}
      which implies the claimed expression for the gradients with respect to
      $\vV$.
      The gradient with respect to $\nu$ is a
      similar calculation with the chain rule, but involves some
      simplifications so we reproduce it here. From the chain rule, for any
      $\Delta \nu \in \bbR$ we have for the differential
      \begin{align*}
         \diff_\nu[\sL^\sigma(\spcdot, \vU, \vV)](\Delta \nu)
         &=
         \ip*{
            \varphi_{\sigma^2} \ast \left(
               X_{\grtr} \circ \vtau_{\nu - \nu_{\grtr}} - \vU \vV\adj
            \right)
         }{
            \dac[ X_{\grtr} \circ \vtau_{\nu + t\Delta \nu - \nu_{\grtr}}]
         }_{L^2(\bbR^2)}
         \\
         &=
         \ip*{
            \varphi_{\sigma^2} \ast \left(
               X_{\grtr} \circ \vtau_{\nu - \nu_{\grtr}} - \vU \vV\adj
            \right)
         }{
            \ip*{
               \nabla_{\vx}[\varphi_{\sigma^2} \ast X_{\grtr}] \circ \vtau_{\nu -
               \nu_{\grtr}}
            }{
               \dot{\vR}_{\nu - \nu_{\grtr}} (\spcdot)
            }_{\ell^2}
         }_{L^2(\bbR^2)} \Delta \nu,
         \labelthis \label{eq:tilt-nu-grad-tmp}
      \end{align*}
      where $\dot{\vR}_\nu$ is the elementwise derivative of the expression in
      \Cref{eq:2d_rotation} for $\vR_\nu$, which evaluates as
      \begin{align*}
         \dot{\vR}_\nu
         &=
         \begin{bmatrix}
            -\sin \nu & -\cos \nu \\
            \cos \nu & -\sin \nu
         \end{bmatrix} \\
         &=
         \begin{bmatrix}
            \cos \nu & -\sin \nu \\
            \sin \nu & \cos \nu
         \end{bmatrix}
         \begin{bmatrix}
            0 & -1 \\
            1 & 0
         \end{bmatrix} \\
         &=
         \vR_\nu
         \begin{bmatrix}
            0 & -1 \\
            1 & 0
         \end{bmatrix}.
      \end{align*}
      Note that the expression in the $\ell^2$ inner product in
      \Cref{eq:tilt-nu-grad-tmp} is a function of $\vx \in \bbR^2$.
      In particular, the function
      \begin{equation*}
         \vx \mapsto
         \ip*{
            \vR_{\nu - \nu_{\grtr}}\adj
            \nabla_{\vx}[\varphi_{\sigma^2} \ast X_{\grtr}] \circ \vtau_{\nu -
            \nu_{\grtr}}(\vx)
         }{
            \begin{bmatrix}
               0 & -1 \\
               1 & 0
            \end{bmatrix}
            \vx
         }_{\ell^2}
      \end{equation*}
      gives the rotational component (tangential to the co-incident circle
      centered at the origin) of the rotated gradient vector field of
      $\varphi_{\sigma^2} \ast X_{\grtr}$ at the point $\vx$.
      This gives the expression
      \begin{equation*}
         \nabla_{\nu} \sL^\sigma(\nu, \vU, \vV) = 
         \ip*{\varphi_{\sigma^2}\ast
            \left( 
               X_{\grtr} \circ \vtau_{\nu - \nu_{\grtr}} - \vU \vV\adj
            \right)
         }{
            \ip*{
               \vR_{\nu - \nu_{\grtr}}\adj
               \nabla_{\vx}[\varphi_{\sigma^2} \ast X_{\grtr}] \circ \vtau_{\nu -
               \nu_{\grtr}}
            }{
               \begin{bmatrix}
                  0 & -1 \\
                  1 & 0
               \end{bmatrix}
               (\spcdot)
            }_{\ell^2}
         }_{L^2(\bbR^2)}.
      \end{equation*}
      Using the commutation relationship \Cref{eq:gaussian-rigid-commute} and a
      unitary change of variables $\vx \mapsto \vtau_{\nu_{\grtr} - \nu}(\vx)$, the
      previous expression implies 
      \begin{equation*}
         \nabla_{\nu} \sL^\sigma(\nu, \vU, \vV) = 
         \ip*{\varphi_{\sigma^2}\ast
            \left( 
               X_{\grtr}  - \vU \vV\adj\circ \vtau_{\nu_{\grtr} - \nu}
            \right)
         }{
            \ip*{
               \nabla_{\vx}[\varphi_{\sigma^2} \ast X_{\grtr}]
            }{
               \begin{bmatrix}
                  0 & -1 \\
                  1 & 0
               \end{bmatrix}
               (\spcdot)
            }_{\ell^2}
         }_{L^2(\bbR^2)}.
      \end{equation*}
      As in \Cref{lem:curl-field-symmetries}, let $\sC(\vx)$ denote the
      function of $\vx$ encompassed by the $\ell^2$ inner product. By
      \Cref{lem:curl-field-symmetries}, we have that $\sC(s, t) = -\sC(t, s)$,
      whereas by \Cref{eq:gaussian-rigid-commute} we have that
      $(\varphi_{\sigma^2} \ast X_{\grtr})(s, t) = (\varphi_{\sigma^2} \ast
      X_{\grtr})(t,
      s)$. It follows that these two functions are orthogonal over
      $L^2(\bbR^2)$, so that in particular
      \begin{equation*}
         \nabla_{\nu} \sL^\sigma(\nu, \vU, \vV) = 
         -\ip*{\varphi_{\sigma^2}\ast
            \left( 
               \vU \vV\adj\circ \vtau_{\nu_{\grtr} - \nu}
            \right)
         }{
            \ip*{
               \nabla_{\vx}[\varphi_{\sigma^2} \ast X_{\grtr}]
            }{
               \begin{bmatrix}
                  0 & -1 \\
                  1 & 0
               \end{bmatrix}
               (\spcdot)
            }_{\ell^2}
         }_{L^2(\bbR^2)},
      \end{equation*}
      as claimed.
   \end{proof}
\end{lemma}

\begin{lemma}
   \label{lem:curl-field-symmetries}

   Let
   \begin{equation*}
      \sC(\vx)
      =
      \ip*{
         \nabla_{\vx}[\varphi_{\sigma^2} \ast X_{\grtr}](\vx)
      }{
         \begin{bmatrix}
            0 & -1 \\
            1 & 0
         \end{bmatrix}
         \vx
      }_{\ell^2}
   \end{equation*}
   denote the rotational component of the gradient vector field of the smoothed
   template. Let $\vQ \in \Orth(2)$ satisfy $\det(\vQ) = -1$, and suppose that
   $\vQ$ is a symmetry of the square template $X_{\grtr}$: in particular
   \begin{equation}
      \vQ \in \set*{
         \begin{bmatrix}
            1 & 0 \\
            0 & -1
         \end{bmatrix},
         \begin{bmatrix}
            -1 & 0 \\
            0 & 1
         \end{bmatrix},
         \begin{bmatrix}
            0 & 1 \\
            1 & 0
         \end{bmatrix},
         \begin{bmatrix}
            0 & -1 \\
            -1 & 0
         \end{bmatrix}
      }
      \label{eq:square-reflection-symmetries}
   \end{equation}
   (this is the subgroup of $\mathsf{D}_4$ consisting of symmetries of
   determinant $-1$). Then one has
   \begin{equation*}
      \sC(\vQ \vx) = -\sC(\vx).
   \end{equation*}

   \begin{proof}
      With \Cref{eq:2d_rotation}, we can write
      \begin{equation*}
         \sC(\vx)
         =
         \ip*{
            \nabla_{\vx}[\varphi_{\sigma^2} \ast X_{\grtr}](\vx)
         }{
            \vR_{\pi/2}
            \vx
         }_{\ell^2}.
      \end{equation*}
      For $\vQ$ as in \Cref{eq:square-reflection-symmetries} and using
      \Cref{eq:so2-reflection-commuting}, we have
      \begin{align*}
         \sC(\vQ \vx)
         &=
         \ip*{
            \nabla_{\vx}[\varphi_{\sigma^2} \ast X_{\grtr}](\vQ\vx)
         }{
            \vQ
            \vR_{-\pi/2}
            \vx
         }_{\ell^2} \\
         &=
         -\ip*{
            \nabla_{\vx}[\varphi_{\sigma^2} \ast X_{\grtr}](\vQ\vx)
         }{
            \vQ
            \vR_{\pi/2}
            \vx
         }_{\ell^2} \\
         &=
         -\ip*{
            \vQ
            \nabla_{\vx}[\varphi_{\sigma^2} \ast X_{\grtr}](\vQ\vx)
         }{
            \vR_{\pi/2}
            \vx
         }_{\ell^2}.
         \labelthis \label{eq:curl-field-rotated}
      \end{align*}
      The second line uses \Cref{eq:2d_rotation}, and the third uses that every
      member of \Cref{eq:square-reflection-symmetries} is symmetric. By Young's
      inequality, we have
      \begin{equation*}
         \nabla_{\vx}[\varphi_{\sigma^2} \ast X_{\grtr}]
         =
         \nabla_{\vx}[\varphi_{\sigma^2}] \ast X_{\grtr},
      \end{equation*}
      and because $\varphi_{\sigma^2}$ is invariant to all orthogonal matrices,
      its gradient is equivariant with respect to $\Orth(2)$, so in particular
      \begin{align*}
         (\vQ \nabla_{\vx}[\varphi_{\sigma^2}] \ast X_{\grtr})(\vQ\vx)
         &= 
         \int_{\bbR^2} X_{\grtr}(\vx') \vQ \nabla_{\vx}[\varphi_{\sigma^2}](\vQ\vx - \vx')
         \diff \vx' \\
         &=
         \int_{\bbR^2} X_{\grtr}(\vx') \nabla_{\vx}[\varphi_{\sigma^2}](\vx - \vQ \vx')
         \diff \vx' \\
         &=
         \int_{\bbR^2} X_{\grtr}(\vQ \vx') \nabla_{\vx}[\varphi_{\sigma^2}](\vx - \vx')
         \diff \vx' \\
         &= \int_{\bbR^2} X_{\grtr}(\vx') \nabla_{\vx}[\varphi_{\sigma^2}](\vx - \vx')
         \diff \vx' \\
         &= (\nabla_{\vx}[\varphi_{\sigma^2}] \ast X_{\grtr})(\vx).
      \end{align*}
      Above, the third line uses an orthogonal change of variables in the
      convolution integral, and the fourth uses that $\vQ$ is a symmetry of
      $X_{\grtr}$. By \Cref{eq:curl-field-rotated}, we have that $\sC(\vQ\vx) =
      -\sC(\vx)$.

   \end{proof}
\end{lemma}

\begin{lemma}
   \label{lem:gradient-lower-bounds-nu}

   The following symmetry properties hold:
   \begin{enumerate}
      \item For any $u \in L^2(\bbR)$, $\sigma > 0$,
         the objective $\nu \mapsto \sL^{\sigma}(\nu, u)$ is $\pi/2$-periodic;
      \item For any $u \in L^2(\bbR)$, $\sigma > 0$, and $-\pi/4 \leq \nu -
         \nu_{\grtr} \leq \pi/4$,
         one has $\sL^{\sigma}(\nu - \nu_{\grtr}, u) = \sL^{\sigma}(\nu_{\grtr}
         - \nu, u)$.
   \end{enumerate}
   Moreover, consider the alignment gradient at the nominal rough initial representation:
   \begin{equation*}
      \nu \mapsto \nabla_\nu \sL^{\sigma_{\SG}}(\nu, \bar{u}_{\rough}).
   \end{equation*}
   where the nominal rough initial representation is defined as
   \begin{equation*}
      \bar{u}_{\rough}(s) = 
      \frac{2}{\sqrt{\pi}} \Ind{\abs{s} \leq 1} \cos(\pi s / 2).
   \end{equation*}
   Suppose that $\alpha = \tfrac{1}{\sqrt{2}}$. Then for any $\nu$,
   \begin{equation*}
      \abs*{
         \nabla_{\nu} \sL^{\sigma_{\SG}}(\nu, \bar{u}_{\rough})
      }
      \leq
      256 \abs{\sin(\nu - \nu_{\grtr})}.
   \end{equation*}
   If, in addition, $\sigma_{\SG} \leq 10^{-3}$, then if $\nu$ is sufficiently
   far from maximizers, i.e., if
   \begin{equation}
      \abs*{
         \left(
            \nu - \nu_{\grtr} + \frac{\pi}{4} \mod \frac{\pi}{2}
         \right)
         - \frac{\pi}{4}
      }
      \leq
      \frac{\pi}{7},
      \label{eq:alignment-gradient-lb-region}
   \end{equation}
   one has
   \begin{equation*}
      \sign\left(
         \left(
            \nu - \nu_{\grtr} + \frac{\pi}{4} \mod \frac{\pi}{2}
         \right)
         - \frac{\pi}{4}
      \right)
      \cdot
      \nabla_{\nu} \sL^{\sigma_{\SG}}(\nu, \bar{u}_{\rough})
      \geq
      c_0 \sin \left(
         \abs*{
            \left(
               \nu - \nu_{\grtr} + \frac{\pi}{4} \mod \frac{\pi}{2}
            \right)
            - \frac{\pi}{4}
         }
      \right)
   \end{equation*}
   for an absolute constant $c_0 > 0$.
   In particular, the gradient is nonnegative when $\nu - \nu_{\grtr} \mod
   \pi/2 \leq \pi/7$, and nonpositive when $\nu - \nu_{\grtr} \mod
   \pi/2 \geq \pi/2 - \pi/7$.

   \begin{proof}
      The proof exploits heavily the $\mathsf{D}_4$ symmetries of the square
      template $X_{\grtr}$ (c.f.\ \Cref{lem:curl-field-symmetries}) and of the
      initialization $\bar{u}_{\rough} \bar{u}_{\rough}\adj$.
      Before proceeding with the analysis of the gradient, we go through some
      simplifying reductions based on symmetry. 
      First, by the definition of the loss in
      \Cref{eq:tilt-objective-ezzz}, it suffices to analyze the case where
      $\nu_{\grtr} = 0$, and perform the substitution $\nu \mapsto \nu -
      \nu_{\grtr}$ in
      all results obtained.
      Next, notice that because
      $\vR_{\nu + \pi/2} = \vR_{\pi/2} \vR_{\nu}$ for any $\nu$ (following the
      notation of \Cref{eq:2d_rotation}),
      one has $X_{\grtr}(\vR_{\nu +
      \pi/2}\vx) = X_{\grtr}(\vR_{\nu}\vx)$ for any $\vx$ by symmetry, which implies (c.f.\
      \Cref{eq:tilt-objective-ezzz}) that $\sL^{\sigma_{\SG}}(\nu + \pi/2, u) =
      \sL^{\sigma_{\SG}}(\nu, u)$ for any $\nu, u$. 
      Moreover, by \Cref{eq:so2-reflection-commuting}, one has $\vR_{-\nu} =
      \vQ \vR_\nu \vQ$, where 
      \begin{equation*}
         \vQ = \begin{bmatrix} 0 & 1 \\ 1 & 0 \end{bmatrix}.
      \end{equation*}
      We have that $\vQ$ is an orthogonal matrix with determinant $-1$;
      writing $\vtau_{\vQ} : \bbR^2 \to \bbR^2$ for its induced transformation,
      we have again by symmetry that $X_{\grtr} \circ \vtau_{\pi/2 -
      \nu}(\vx) = X_{\grtr} \circ \vtau_{\nu} \circ \vtau_{\vQ}$.
      Applying then \Cref{eq:gaussian-rigid-commute} (notice that the
      calculation does not use the fact that $\det(\vR_{\nu}) = 1$, and in fact
      any orthogonal matrix yields the same conclusion) together with a unitary
      change of coordinates, we obtain that
      \begin{equation*}
         \sL^{\sigma_{\SG}}(\pi/2 - \nu, \bar{u}_{\rough})
         = \frac{1}{2} \norm*{
            \varphi_{\sigma_{\SG}^2} \ast 
            \left( 
               X_{\grtr} \circ \vtau_{\nu}
               - 
               \bar{u}_{\rough} \bar{u}_{\rough}\adj \circ \vtau_{\vQ}
            \right)
         }_{L^2}^2,
      \end{equation*}
      since $\vQ\adj = \vQ$. But since $\vtau_{\vQ}(s, t) = (t, s)$, it follows
      from symmetry that $\sL^{\sigma_{\SG}}(\pi/2 - \nu, \bar{u}_{\rough}) =
      \sL^{\sigma_{\SG}}(\nu, \bar{u}_{\rough})$.
      We have thus shown that $\nu \mapsto \sL^{\sigma_{\SG}}(\nu, \bar{u}_{\rough})$ is
      \begin{enumerate}
         \item $\pi/2$-periodic;
         \item on $[0, \pi/2]$, symmetric about $\pi/4$.
      \end{enumerate}
      It therefore suffices to assume that $0 \leq \nu \leq \pi/4$ in the
      sequel, since conclusions on this interval can be translated to all $\nu
      \in \bbR$ as stated in the statement of the result by these symmetry
      properties.

      We proceed to estimate the gradient
      \begin{equation*}
         \nabla_{\nu} \sL^{\sigma_{\SG}}(\nu, \bar{u}_{\rough})
         = 
         -\ip*{
            \underbrace{
               \varphi_{\sigma_{\SG}^2} \conv \left( 
                  \bar{u}_{\rough} \bar{u}_{\rough}\adj \circ \vtau_{-\nu}
               \right)
            }_{\sR}
         }{
            \sC
         }_{L^2(\bbR^2)}
      \end{equation*}
      under the preceding assumptions, where $\sC$ is defined as in
      \Cref{lem:curl-field-symmetries}. 
      First, we reduce the $L^2$ integral in the expression for a
      gradient into a difference of integrals over a `fundamental domain'
      depending on the $\sD_4$ symmetries of $X_{\grtr}$ and the
      initialization; this expression will be useful for upper and lower bounds.
      Then, we will establish the lower bound, which is more technical, before
      concluding with the upper bound.
      For $(\vkappa, \pi) \in \set{-1, 1}^2 \times \mathrm{P}(2)$, where
      $\mathrm{P}(2)$ is the set of permutations on $2$ elements,  we consider
      the ``wedge'' domains
      \begin{equation*}
         C_{\vkappa, \pi} = \set*{
            \vx = (s, t) \in \bbR^2 \given
            \kappa_1 s \geq 0,
            \kappa_2 t \geq 0,
            \pi_1(s, \kappa_1 \kappa_2 t) \geq \pi_2(s, \kappa_1 \kappa_2 t)
         }.
      \end{equation*}
      Intuitively, in the third constraint, $\pi$ governs the
      ``direction'' of the inequality, and $\kappa_1 \kappa_2$ selects the
      proper subspace to reflect about.
      One notes that $\cup_{(\vkappa,\pi) \in \set{-1, 1}^2 \times
      \mathrm{P}(2)} C_{\vkappa, \pi} = \bbR^2$,
      and if $(\vkappa, \pi) \neq (\vkappa', \pi')$ then $C_{\vkappa, \pi} \cap
      C_{\vkappa', \pi'}$ has zero Lebesgue measure.
      Because $\sC$ and $\sR$ are smooth functions, it follows
      \begin{align*}
         \nabla_{\nu} \sL^{\sigma_{\SG}}(\nu, \bar{u}_{\rough})
         &= 
         -\ip*{
            \sR
            \left(
               \sum_{(\vkappa, \pi) \in \set{-1, 1}^2 \times \mathrm{P}(2)}
               \Ind{C_{\vkappa, \pi}}
            \right)
         }{
            \sC
         }_{L^2} \\
         &=
         \sum_{(\vkappa, \pi) \in \set{-1, 1}^2 \times \mathrm{P}(2)}
         -\ip*{
            \sR\Ind{C_{\vkappa, \pi}}
         }{
            \sC
         }_{L^2}.
      \end{align*}
      Recall, following \Cref{eq:gaussian-rigid-commute}, that we can freely
      interchange the order of gaussian smoothing and rotation in the
      expression for $\sL^{\sigma_{\SG}}(\nu, \bar{u}_{\rough})$. 
      Since $\bar{u}_{\rough}(s) = \bar{u}_{\rough}(-s)$ for any $s \in \bbR$, by
      the argument above applied to $X_{\grtr}$ we have that for any $\nu$
      \begin{align*}
         \sR \circ \vtau_{\pm \pi/2}
         &=
         \left(
            \varphi_{\sigma_{\SG}^2} \conv \left( 
               \bar{u}_{\rough} \bar{u}_{\rough}\adj \circ \vtau_{-\nu} 
            \right)
         \right)
         \circ \vtau_{\pm \pi/2}  \\
         &=
         \varphi_{\sigma_{\SG}^2} \conv \left( 
            \bar{u}_{\rough} \bar{u}_{\rough}\adj \circ \vtau_{\pm \pi/2} \circ  \vtau_{-\nu}
         \right) \\
         &=
         \varphi_{\sigma_{\SG}^2} \conv \left( 
            \bar{u}_{\rough} \bar{u}_{\rough}\adj \circ \vtau_{-\nu} 
         \right)
         \\
         &= \sR.
      \end{align*}
      Meanwhile, we note that
      \begin{equation*}
         \vR_{\pi/2}
         =
         \begin{bmatrix}
            0 & -1 \\
            1 & 0
         \end{bmatrix}
         =
         \begin{bmatrix}
            0 & 1 \\
            1 & 0
         \end{bmatrix}
         \begin{bmatrix}
            1 & 0 \\
            0 & -1
         \end{bmatrix} ;
         \quad
         \vR_{-\pi/2}
         =
         \begin{bmatrix}
            0 & 1 \\
            -1 & 0
         \end{bmatrix}
         =
         \begin{bmatrix}
            0 & 1 \\
            1 & 0
         \end{bmatrix}
         \begin{bmatrix}
            -1 & 0 \\
            0 & 1
         \end{bmatrix} ,
      \end{equation*}
      so by \Cref{lem:curl-field-symmetries}, $\sC \circ \vtau_{\pm \pi/2} =
      \sC$. Thus, changing coordinates in the $L^2$ integral, we get
      \begin{equation*}
         \nabla_{\nu} \sL^{\sigma_{\SG}}(\nu, \bar{u}_{\rough})
         =
         -4\ip*{
            \sR\left(
               \Ind{C_{\set{1, -1}, \Id}}
               + \Ind{C_{\set{1, 1}, \Id}}
            \right)
         }{
            \sC
         }_{L^2},
      \end{equation*}
      where we recall
      \begin{equation}
         C_{\set{1, -1}, \Id}
         = \set*{
            \vx = (s, t) \in \bbR^2 \given
            s \geq 0,
            t \leq 0,
            s \geq -t
         },
         \quad
         C_{\set{1, 1}, \Id}
         = \set*{
            \vx = (s, t) \in \bbR^2 \given
            s \geq 0,
            t \geq 0,
            s \geq t
         }.
      \end{equation}
      By another change of coordinates and \Cref{lem:curl-field-symmetries}, we
      then have in addition
      \begin{equation}
         \nabla_{\nu} \sL^{\sigma_{\SG}}(\nu, \bar{u}_{\rough})
         \geq
         -4
         \int_{\set{0 \leq t \leq s}}
         \left(
            \sR(s, t) 
            - \sR(s, -t)
         \right)
         \sC(s, t)
         \diff s \diff t
         \label{eq:angle-gradient-localized}
      \end{equation}
      Next, we control $\sR(s, t) - \sR(s, -t)$ using the geometry of the
      rough factorization $\bar{u}_{\rough} \bar{u}_{\rough}\adj$;
      we will then conclude the bound from
      \Cref{eq:angle-gradient-localized}. First, by linearity and
      \Cref{eq:gaussian-rigid-commute}, we have
      \begin{align*}
         \sR(s, t) - \sR(s, -t)
         &=
         \varphi_{\sigma_{\SG}^2} \conv \left( 
            \bar{u}_{\rough} \bar{u}_{\rough}\adj \circ \vtau_{-\nu}
         \right)(s, t)
         -  
         \varphi_{\sigma_{\SG}^2} \conv \left( 
            \bar{u}_{\rough} \bar{u}_{\rough}\adj \circ \vtau_{-\nu}
         \right)(s, -t)
         \\
         &=
         \varphi_{\sigma_{\SG}^2} \conv \left(
            \bar{u}_{\rough} \bar{u}_{\rough}\adj \circ \vtau_{-\nu}
            -
            \bar{u}_{\rough} \bar{u}_{\rough}\adj \circ \vtau_{-\nu} \circ \vtau_{\vQ}
         \right)(s, t),
         \labelthis \label{eq:residual-as-targetdiff}
      \end{align*}
      where $\vQ$ is the matrix representation of the orthogonal transformation
      $(s, t) \mapsto (s, -t)$. 
      Gaussian smoothing is nonnegativity-preserving, so developing a lower
      bound on the difference $\sR(s, t) - \sR(s, -t)$ can be done by
      developing a lower bound on the parenthesized term above.
      By \Cref{eq:so2-reflection-commuting} and
      symmetry, we have 
      \begin{equation}
         \bar{u}_{\rough} \bar{u}_{\rough}\adj \circ
         \vtau_{-\nu} \circ \vtau_{\vQ} = \bar{u}_{\rough} \bar{u}_{\rough}\adj
         \circ \vtau_{\nu}.
         \label{eq:residual-as-targetdiff-flippedpart}
      \end{equation}
      Applying the first conclusion in \Cref{lem:residual-field-estimates}
      together with \Cref{lem:curl-field-nonnegative}, it follows that
      \begin{equation*}
         -(\sR(s, t) - \sR(s, -t)) \sC(s, t) \geq 0
      \end{equation*}
      for every $0 \leq t \leq s$.%
      This means that we can obtain a
      lower bound for the RHS of \Cref{eq:angle-gradient-localized} by
      integrating over a subset of the domain $\set{0 \leq t \leq s}$. 
      When $\sigma_{\SG}$ is small, the field $\sC$ concentrates
      around the boundary of the square template $X_{\grtr}$; we will therefore
      obtain a lower bound for the gradient by integrating in a small strip
      around this region. 
      To this end, the second conclusion in \Cref{lem:residual-field-estimates}
      gives the following quantitative bound, valid for $0 \leq t \leq s \leq
      1$ and all $0 \leq \nu \leq \pi/7$:
      \begin{equation*}
         \bar{u}_{\rough} \bar{u}_{\rough}\adj \circ \vtau_{\nu}(\vx)
         -
         \bar{u}_{\rough} \bar{u}_{\rough}\adj \circ \vtau_{-\nu}(\vx)
         \geq
         \frac{7 \sin \nu}{1000}
         \Ind{-0.137 \leq t - \tfrac{1}{\sqrt{2}} \leq -0.127}
         \Ind{-0.001 \leq s - \tfrac{1}{\sqrt{2}} \leq 0.001}.
      \end{equation*}
      Since we are considering a regime with $\sigma_{\SG}$ small, it is now
      reasonable to simplify this estimate further by worst-casing the
      smoothing that connects it to $\sR$. 
      Because this indicator is a box in the $(s, t)$ plane, its smoothed
      version is a product of smoothed indicators for compact connected
      intervals in $\bbR$. 
      If $I = [-a, a]$ is such an interval (because convolution commutes with
      translations, it will be sufficient to consider such a centered
      interval), we have (see the derivative calculations at the start of the
      proof of \Cref{lem:curl-field-at-opt-estimates}) that $\varphi_{\sigma^2}
      \conv \Ind{I}(x)$ is decreasing (resp.\ increasing) for $x \geq 0$
      (resp.\ $x \leq 0$). Hence the minimum value taken by $\varphi_{\sigma^2}
      \conv \Ind{I}$ among those $x \in I$ is attained at $x \in \set{\pm a}$, 
      where
      \begin{align*}
         \Ind{I}\conv \varphi_{\sigma^2}(a)
         =
         \int_{-a}^a \varphi_{\sigma^2}(x - a) \diff x
         &=
         \int_{0}^{2a} \varphi_{\sigma^2}(x) \diff x
         \\
         &=
         \frac{1}{2} - \int_{2a}^{\infty} \varphi_{\sigma^2}(x) \diff x
         \\
         &\geq
         \frac{1}{2} - \frac{\sigma}{2a\sqrt{2\pi}} e^{-2a^2/\sigma^2},
      \end{align*}
      by the standard estimate for the gaussian tail integral.
      Thus, as soon as $\sigma \leq 2a$, one has
      \begin{equation}
         \Ind{I}\conv \varphi_{\sigma^2}(a)
         \geq
         \frac{1}{4},
         \label{eq:smoothed-indicator-lb-indicator}
      \end{equation}
      which shows that $\Ind{I} \conv \varphi_{\sigma^2} \geq \fourth \Ind{I}$
      if $\sigma \leq 2a$.
      Applying this to our lower bound, it follows that if $\sigma_{\SG} \leq
      \tfrac{1}{500}$, we have
      \begin{equation*}
         -\left(
            \sR(s, t)
            - \sR(s, -t)
         \right)
         \geq
         \frac{7 \sin \nu}{16000}
         \Ind{-0.137 \leq t - \tfrac{1}{\sqrt{2}} \leq -0.127}
         \Ind{-0.001 \leq s - \tfrac{1}{\sqrt{2}} \leq 0.001}.
      \end{equation*}
      Plugging this bound into \Cref{eq:angle-gradient-localized}
      gives
      \begin{equation}
         \nabla_{\nu} \sL^{\sigma_{\SG}}(\nu, \bar{u}_{\rough})
         \geq
         \frac{7 \sin \nu}{4000}
         \iint_{\substack{
               -0.001 \leq s - 1/\sqrt{2} \leq 0.001,\\
               -0.137 \leq t -1/\sqrt{2} \leq -0.127
            }
         }
         \sC(s, t)
         \diff s \diff t.
         \label{eq:angle-gradient-lb}
      \end{equation}
      The remainder of the proof is a relatively tedious calculation over this
      domain of integration.
      We make use of the expressions for $\sC$ derived in
      \Cref{lem:curl-field-nonnegative}:
      \begin{align*}
         \sC(s, t) &= s f(s) f'(t) - t f(t) f'(s), \quad \text{where} \\
         f(x) &= \frac{1}{\sqrt{2\pi \sigma_{\SG}^2}} \int_{-1/\sqrt{2}}^{1/\sqrt{2}}
         e^{-\tfrac{(x - x')^2}{2\sigma_{\SG}^2}} \diff x', \quad x \in \bbR; \\
         f'(x) &= 
         \varphi_{\sigma_{\SG}^2}(x + \alpha) - \varphi_{\sigma_{\SG}^2}(x - \alpha).
      \end{align*}
      Moreover, we recall that $f'(x) \geq 0$ if $x \geq 0$ and $f'(x) \leq 0$
      if $x \leq 0$, as shown in the proof of
      \Cref{lem:curl-field-nonnegative}.
      We have $\abs{s f(s)} \leq 1$ when $s \leq 1$ by Young's convolution
      inequality, and 
      \begin{align*}
         f'(t) &\geq -\varphi_{\sigma_{\SG}^2}(t - 1/\sqrt{2}) \\
         &\geq -\varphi_{\sigma_{\SG}^2}(-0.127) \\
         &\geq -\frac{1}{\sigma_{\SG} \sqrt{2\pi}} e^{-\tfrac{1}{128 \sigma_{\SG}^2}}
      \end{align*}
      for $t$ in the region of integration.
      Similarly, 
      for $t$ in the region of integration
      \begin{align*}
         t f(t) &\geq
         \left(
            \frac{1}{\sqrt{2}} - 0.137
         \right)
         f(1/\sqrt{2} - 0.127)
         \\
         &\geq
         \frac{1}{4}
         \left(
            \frac{1}{\sqrt{2}} - 0.137
         \right),
      \end{align*}
      where the last line uses \Cref{eq:smoothed-indicator-lb-indicator}.
      Finally, we have for $s$ in the domain of integration
      \begin{align*}
         f'(s)
         &=
         \varphi_{\sigma_{\SG}^2}(s - 1/\sqrt{2})
         -
         \varphi_{\sigma_{\SG}^2}(s + 1/\sqrt{2})
         \\
         &\geq
         \varphi_{\sigma_{\SG}^2}(s - 1/\sqrt{2})
         -
         \varphi_{\sigma_{\SG}^2}(\sqrt{2} - 0.001)
         \\
         &\geq
         \varphi_{\sigma_{\SG}^2}(s - 1/\sqrt{2})
         -\frac{1}{\sigma_{\SG} \sqrt{2\pi}} e^{-\tfrac{1}{2\sigma_{\SG}^2}}.
      \end{align*}
      Combining these, we have the lower bound (valid on the domain of
      integration of our gradient lower bound)
      \begin{align*}
         \sC(s, t) &\geq
         \frac{1}{4}
         \left(
            \frac{1}{\sqrt{2}} - 0.137
         \right)
         \left(
            \varphi_{\sigma_{\SG}^2}(s - 1/\sqrt{2})
            -\frac{1}{\sigma_{\SG} \sqrt{2\pi}} e^{-\tfrac{1}{2\sigma_{\SG}^2}}
         \right)
         -\frac{1}{\sigma_{\SG} \sqrt{2\pi}} e^{-\tfrac{1}{128 \sigma_{\SG}^2}}
         \\
         &\geq
         \frac{1}{8}\varphi_{\sigma_{\SG}^2}(s - 1/\sqrt{2})
         -\frac{1}{\sigma_{\SG} \sqrt{\pi}} e^{-\tfrac{1}{128 \sigma_{\SG}^2}}.
      \end{align*}
      Integrating the first term in this lower bound over the $s$ region gives,
      by a change of coordinates,
      \begin{align*}
         \int_{-0.001 + 1/\sqrt{2}}^{0.001 + 1/\sqrt{2}}
         \varphi_{\sigma_{\SG}^2}(s - 1/\sqrt{2})
         \diff s
         &=
         \int_{-0.001}^{0.001}
         \varphi_{\sigma_{\SG}^2}(s)
         \diff s
         \\
         &=
         1 - 2
         \int_0^{0.001}
         \varphi_{\sigma_{\SG}^2}(s)
         \diff s
         \\
         &\geq
         1
         -
         2
         \frac{\sigma_{\SG}}{10^{-3} \sqrt{2\pi}}
         e^{-10^{-6} / 2 \sigma_{\SG}^2},
      \end{align*}
      using also the gaussian tail estimate we applied above. 
      Thus, as soon as $\sigma_{\SG} \leq 10^{-3}$, 
      we have
      \begin{equation*}
         \int_{-0.001 + 1/\sqrt{2}}^{0.001 + 1/\sqrt{2}}
         \varphi_{\sigma_{\SG}^2}(s - 1/\sqrt{2})
         \diff s
         \geq
         \frac{1}{2},
      \end{equation*}
      and under this constraint on $\sigma_{\SG}$, we have moreover
      from our previous lower bound
      \begin{equation*}
         \int_{-0.001 + 1/\sqrt{2}}^{0.001 + 1/\sqrt{2}}
         \sC(s, t) \diff s \geq
         \frac{1}{20}.
      \end{equation*}
      Integrating over the region of $t$ adds only an additional constant
      multiple, since this expression does not depend on $t$.
      Consequently, these calculations together with
      \Cref{eq:angle-gradient-lb} imply the claimed lower bound.

      We can obtain the claimed upper bound in a similar way. Since the lower
      bound we have just established characterizes the sign of the gradient at
      all points where $0 \leq \nu \leq \pi/7$ (and similarly for negative
      $\nu$, by symmetry of the objective), it suffices to simply control the
      magnitude of the gradient.
      By
      \Cref{eq:angle-gradient-localized,%
      eq:residual-as-targetdiff,eq:residual-as-targetdiff-flippedpart}, 
      and $L^1$-$L^\infty$ control, we have
      \begin{align*}
         \abs*{
            \nabla_{\nu} \sL^{\sigma_{\SG}}(\nu, \bar{u}_{\rough})
         }
         &\leq
         4
         \left(
            \sup_{0 \leq t \leq s}\,
            \abs*{
               \varphi_{\sigma_{\SG}^2} \conv \left(
                  \bar{u}_{\rough} \bar{u}_{\rough}\adj \circ \vtau_{-\nu}
                  -
                  \bar{u}_{\rough} \bar{u}_{\rough}\adj \circ \vtau_{\nu}
               \right)(s, t)
            }
         \right)
         \int_{\set{s \geq 0, t \geq 0, s \geq t}}
         \abs*{
            \sC(s, t)
         }
         \diff s \diff t
         \\
         &\leq
         4
         \left(
            \sup_{0 \leq t \leq s}\,
            \abs*{
               \bar{u}_{\rough} \bar{u}_{\rough}\adj \circ \vtau_{-\nu}
               -
               \bar{u}_{\rough} \bar{u}_{\rough}\adj \circ \vtau_{\nu}
            }(s, t)
         \right)
         \int_{\set{0 \leq t \leq s}}
         \abs*{
            \sC(s, t)
         }
         \diff s \diff t,
      \end{align*}
      where we use Young's convolution inequality in the second line.
      By the expressions given above and the triangle inequality,
      we have
      \begin{align*}
         \int_{\set{0 \leq t \leq s}}
         \abs*{
            \sC(s, t)
         }
         \diff s \diff t
         &\leq
         \int_{\set{0 \leq t \leq s}}
         \left(
            \abs*{
               s f(s) f'(t)
            }
            + \abs*{
               t f(t) f'(s)
            }
         \right)
         \diff s \diff t
         \\
         &\leq
         2\int_{\bbR^2}
         \left(
            \abs*{
               s f(s) f'(t)
            }
         \right)
         \diff s \diff t
         \\
         &=
         2\left(
            \int_{\bbR}
            \abs*{
               s f(s)
            }
            \diff s
         \right)
         \left(
            \int_{\bbR}
            \abs{f'(t)}
            \diff t
         \right),
      \end{align*}
      where the third line uses Fubini's theorem.
      Because $f'$ is a difference of two gaussians, the integral of its
      magnitude is no larger than $2$.
      Meanwhile, we have by Fubini's theorem
      \begin{align*}
         \int_{\bbR} \abs{s f(s)} \diff s 
         &= \int_{-1/\sqrt{2}}^{1/\sqrt{2}} \int_{\bbR} \abs{s}\varphi_{\sigma_{\SG}^2}(s - x)
         \diff s \diff x
         \\
         &= \int_{-1/\sqrt{2}}^{1/\sqrt{2}} \int_{\bbR} \abs{s + x}\varphi_{\sigma_{\SG}^2}(s)
         \diff s \diff x
         \\
         &\leq
         \int_{-1/\sqrt{2}}^{1/\sqrt{2}} (\abs{x} + \sqrt{2/\pi}) \diff x
         \\
         &= \frac{1}{2} + \frac{2}{\sqrt{\pi}}.
      \end{align*}
      Thus
      \begin{equation*}
         \int_{\set{0 \leq t \leq s}}
         \abs*{
            \sC(s, t)
         }
         \diff s \diff t
         \leq 8.
      \end{equation*}
      Meanwhile, by definition (see \Cref{lem:gradient-lower-bounds-nu}),
      $\bar{u}_{\rough}$ is a $\sqrt{\pi}$-Lipschitz function of its argument,
      and is bounded by $2/\sqrt{\pi}$; this means that $\bar{u}_{\rough}
      \bar{u}_{\rough}\adj : \bbR^2 \to \bbR$ is a $2 \sqrt{2}$-Lipschitz
      function of its argument with respect to the $\ell^2$ metric on $\bbR^2$.
      Recalling moreover that $\bar{u}_{\rough}$ is compactly supported on
      $[-1, 1]$, it follows that for any $\vx = (s,t)$ at which
      $\bar{u}_{\rough} \bar{u}_{\rough}\adj \circ \vtau_{\pm \nu}$ is nonzero,
      we have
      \begin{align*}
         \abs*{
            \bar{u}_{\rough}\bar{u}_{\rough}\adj \circ \vtau_{-\nu}(\vx)
            -
            \bar{u}_{\rough}\bar{u}_{\rough}\adj \circ \vtau_{\nu}(\vx)
         }
         &\leq
         2\sqrt{2}
         \norm*{
            \vtau_{-\nu}(\vx)
            -
            \vtau_{\nu}(\vx)
         }_2
         \\
         &\leq
         2\sqrt{2}
         \norm*{
            \vR_{-\nu}
            -
            \vR_{\nu}
         }
         \norm*{\vx}_2
         \\
         &\leq
         4
         \norm*{
            \vR_{-\nu}
            -
            \vR_{\nu}
         }.
      \end{align*}
      In the last two lines, we simply pass to the operator norm of the
      difference of rotation matrices and then use that $\norm{\vx}_{\infty}
      \leq 1$.
      In dimension two, we have the representation
      \begin{equation*}
         \vR_{\nu} = (\cos \nu) \vI 
         +
         (\sin \nu) \begin{bmatrix}
            0 & -1 \\
            1 & 0
         \end{bmatrix},
      \end{equation*}
      and each of these matrices (without the prefactors) is orthogonal, hence
      has unit operator norm. Thus, the triangle inequality gives
      \begin{equation*}
         \norm*{
            \vR_{-\nu}
            -
            \vR_{\nu}
         }
         \leq 2\abs*{\sin \nu}.
      \end{equation*}
      since $\sin$ and $\cos$ are both $1$-Lipschitz.
      Combining, this shows
      \begin{equation*}
         \abs*{
            \nabla_{\nu} \sL^{\sigma_{\SG}}(\nu, \bar{u}_{\rough})
         }
         \leq
         256 \sin \nu.
      \end{equation*}

   \end{proof}

\end{lemma}

\begin{lemma}
   \label{lem:alignment-objective-sg}

   Consider the roughly-localized alignment iteration
   \Cref{eq:tilt-grad-nu-iter}. 
   Suppose
   \begin{equation*}
      \norm*{
         u_{\rough} - \bar{u}_{\rough}
      }_{L^2(\bbR)}
      \leq
      \frac{2}{\sqrt{\pi}},
   \end{equation*}
   where the nominal rough initial representation $\bar{u}_{\rough}$ is
   defined as in \Cref{lem:gradient-lower-bounds-nu}.
   For any $\sigma_{\SG} \leq \tfrac{1}{100}$ and $\alpha =
   \tfrac{1}{\sqrt{2}}$, the following holds:
   \begin{enumerate}
      \item The functions $\nabla_{\nu} \sL^{\sigma_{\SG}}(\spcdot, u_{\rough})$
         and $\nabla_{\nu} \sL^{\sigma_{\SG}}(\spcdot, \bar{u}_{\rough})$
         satisfy
         \begin{align*}
            \norm*{
               \nabla_{\nu} \sL^{\sigma_{\SG}}(\spcdot, \bar{u}_{\rough})
            }_{\Lip}
            &\leq
            \frac{4}{\pi \sigma_{\SG}^2},
            \\
            \norm*{
               \nabla_{\nu} \sL^{\sigma_{\SG}}(\spcdot, u_{\rough})
            }_{\Lip}
            &\leq
            \frac{4}{\sigma_{\SG}^2}
            \left(
               \frac{1}{\pi} 
               +
               \frac{2
                  \norm*{
                     u_{\rough} - \bar{u}_{\rough}
                  }_{L^2(\bbR)}
               }
               {
                  \sqrt{\pi}
               }
            \right),
         \end{align*}
         where $\norm{}_{\Lip}$ denotes the Lipschitz seminorm;
      \item We have the gradient estimate
         \begin{equation*}
            \abs*{
               \nabla_{\nu} \sL^{\sigma_{\SG}}(\spcdot, u_{\rough})
               -
               \nabla_{\nu} \sL^{\sigma_{\SG}}(\spcdot, \bar{u}_{\rough})
            }
            \leq
            \frac{8}{\sqrt{\pi}\sigma_{\SG}^2}
            \norm*{\bar{u}_{\rough} - u_{\rough}}_{L^2(\bbR)},
         \end{equation*}
         as well as the squared-gradient estimate
         \begin{equation*}
            \abs*{
               \left(
                  \nabla_{\nu} \sL^{\sigma_{\SG}}(\spcdot, u_{\rough})
               \right)^2
               -
               \left(
                  \nabla_{\nu} \sL^{\sigma_{\SG}}(\spcdot, \bar{u}_{\rough})
               \right)^2
            }
            \leq
            \frac{512}{\pi^{3/2}\sigma_{\SG}^4}
            \norm*{\bar{u}_{\rough} - u_{\rough}}_{L^2(\bbR)}.
         \end{equation*}
   \end{enumerate}

   \begin{proof}
      We will prove the first assertion first. 
      We recall from \Cref{lem:tilt_grad_hess} that for any $\nu \in \bbR$, $u
      \in L^2(\bbR)$,
      \begin{equation*}
         \nabla_{\nu} \sL^{\sigma_{\SG}}(\nu, u)
         =
         -\ip*{
            \varphi_{\sigma^2_{\SG}} \conv \left(
               uu\adj \circ \vtau_{\nu_{\grtr}-\nu}
            \right)
         }
         {
            \ip*{
               \nabla_{\vx}[\varphi_{\sigma_{\SG}^2} 
               \conv X_{\grtr}]
            }
            {
               \vtau_{\pi/2}
            }_{\ell^2}
         }_{L^2(\bbR^2)},
      \end{equation*}
      and using \Cref{eq:gaussian-rigid-commute} and the fact that $f \mapsto f
      \circ \vtau_{\nu}$ is a unitary transformation of $L^2(\bbR^2)$, we thus
      have
      \begin{equation*}
         \nabla_{\nu} \sL^{\sigma_{\SG}}(\nu, u)
         =
         -\ip*{
            \varphi_{\sigma^2_{\SG}} \conv uu\adj
         }
         {
            \sC^{u_{\grtr}} \circ \vtau_{\nu - \nu_{\grtr}}
         }_{L^2(\bbR^2)},
      \end{equation*}
      using notation from
      \Cref{lem:curl-field-symmetries,lem:curl-field-at-opt-estimates} as
      \begin{equation*}
         \sC^u(\vx)
         =
         \ip*{
            \nabla_{\vx}[\varphi_{\sigma^2} \ast u u\adj](\vx)
         }{
            \begin{bmatrix}
               0 & -1 \\
               1 & 0
            \end{bmatrix}
            \vx
         }_{\ell^2}.
      \end{equation*}
      It then follows that
      \begin{align*}
         \nabla_{\nu}^2 \sL^{\sigma_{\SG}}(\nu, u)
         &=
         -\ip*{
            \varphi_{\sigma^2_{\SG}} \conv uu\adj
         }
         {
            \ip*{
               \vR_{\nu - \nu_\grtr}\adj
               \nabla_{\vx} \sC^{u_\grtr} \circ \vtau_{\nu - \nu_\grtr}
            }
            {
               \vtau_{\pi/2}
            }_{\ell^2}
         }_{L^2(\bbR^2)}
         \\
         &=
         -\ip*{
            \varphi_{\sigma^2_{\SG}} \conv \left(
               uu\adj \circ \vtau_{\nu_{\grtr} - \nu}
            \right)
         }
         {
            \ip*{
               \nabla_{\vx} \sC^{u_\grtr}
            }
            {
               \vtau_{\pi/2}
            }_{\ell^2}
         }_{L^2(\bbR^2)},
      \end{align*}
      calculating as in the proof of \Cref{lem:tilt_grad_hess} for the first
      derivative.
      We can estimate the RHS with the Schwarz inequality; by Young's
      convolution inequality and the fact that $f \mapsto f \circ \vtau_{-\nu}$
      is a unitary transformation of $L^2(\bbR^2)$, we obtain
      \begin{align*}
         \norm*{
            \varphi_{\sigma^2_{\SG}} \conv \left(
               uu\adj\circ \vtau_{\nu_{\grtr} -\nu}
            \right)
         }_{L^2(\bbR^2)}
         &\leq
         \norm*{\varphi_{\sigma^2_{\SG}}}_{L^1(\bbR^2)}
         \norm*{
            uu\adj\circ \vtau_{\nu_{\grtr} -\nu}
         }_{L^2(\bbR^2)}
         \\
         &\leq
         \norm*{
            uu\adj
         }_{L^2(\bbR^2)}
         \\
         &=
         \norm{u}_{L^2(\bbR)}^2.
      \end{align*}
      Meanwhile, by the second estimate in
      \Cref{lem:curl-field-deriv-estimates}, we have
      \begin{equation*}
         \norm*{
            \ip*{
               \nabla_{\vx} \sC^{u_\grtr}
            }
            {
               \vtau_{\pi/2}
            }_{\ell^2}
         }_{L^2(\bbR^2)}
         \leq
         \left(
            \frac{3}{\pi} + 
            \frac{55}{\pi \sigma_{\SG}^2} 
            + \frac{4}{5 \pi \sigma_{\SG}^4}
         \right)^{1/2}
         \leq
         \frac{1}{\sigma_{\SG}^2},
      \end{equation*}
      where the worst-casing uses that $\sigma_{\SG}^2 \leq \tfrac{1}{100}$.
      Thus, we have
      \begin{equation*}
         \abs*{
            \nabla_{\nu}^2 \sL^{\sigma_{\SG}}(\nu, u)
         }
         \leq
         \frac{\norm{u}_{L^2(\bbR)}^2}{\sigma_{\SG}^2}.
      \end{equation*}
      For real numbers $x, y$, we have
      \begin{equation*}
         \abs{x^2 - y^2}
         \leq
         2 \max \set{\abs{x}, \abs{y}} \abs{x - y},
      \end{equation*}
      so by the triangle inequality,
      \begin{align*}
         \abs*{
            \norm*{u_{\rough}}_{L^2(\bbR)}^2
            - \norm*{\bar{u}_{\rough}}_{L^2(\bbR)}^2
         }
         &\leq
         2 \max \set*{
            \norm*{u_{\rough}}_{L^2(\bbR)},
            \norm*{\bar{u}_{\rough}}_{L^2(\bbR)}
         }
         \norm*{
            u_{\rough} - \bar{u}_{\rough}
         }_{L^2(\bbR)}
         \\
         &\leq
         2
         \left(
            \norm*{\bar{u}_{\rough}}_{L^2(\bbR)}
            +
            \norm*{
               u_{\rough} - \bar{u}_{\rough}
            }_{L^2(\bbR)}
         \right)
         \norm*{
            u_{\rough} - \bar{u}_{\rough}
         }_{L^2(\bbR)}
         \\
         &\leq
         4 
         \norm*{\bar{u}_{\rough}}_{L^2(\bbR)}
         \norm*{
            u_{\rough} - \bar{u}_{\rough}
         }_{L^2(\bbR)},
      \end{align*}
      where the last line requires that $ \norm{ u_{\rough} - \bar{u}_{\rough}
      }_{L^2(\bbR)} \leq  \norm{\bar{u}_{\rough}}_{L^2(\bbR)}$.
      Since $\norm{\bar{u}_{\rough}}_{L^2(\bbR)} = 2/\sqrt{\pi}$ by
      \Cref{lem:diamond_diagonalized},
      this, combined with our previously-derived bound, is equivalent to the
      assertion.

      For the second assertion, we have by the above inequality
      \begin{align*}
         \abs*{
            \left(
               \nabla_{\nu} \sL^{\sigma_{\SG}}(\spcdot, u_{\rough})
            \right)^2
            -
            \left(
               \nabla_{\nu} \sL^{\sigma_{\SG}}(\spcdot, \bar{u}_{\rough})
            \right)^2
         }
         \leq
         &2 \max \set*{
            \abs*{
               \nabla_{\nu} \sL^{\sigma_{\SG}}(\spcdot, u_{\rough})
            },
            \abs*{
               \nabla_{\nu} \sL^{\sigma_{\SG}}(\spcdot, \bar{u}_{\rough})
            },
         }
         \\&\quad\times
         \abs*{
            \nabla_{\nu} \sL^{\sigma_{\SG}}(\spcdot, u_{\rough})
            -
            \nabla_{\nu} \sL^{\sigma_{\SG}}(\spcdot, \bar{u}_{\rough})
         },
      \end{align*}
      so we can bound the sizes of the two factors as well as their absolute
      difference.
      Using the expression given in the proof of the previous assertion for the
      gradient, 
      we note that for any $\nu \in \bbR$
      \begin{equation*}
         \nabla_{\nu} \sL^{\sigma_{\SG}}(\nu, 0)
         =
         0,
      \end{equation*}
      so bounding the sizes of the two factors is accomplished by a bound on
      their difference.
      Now, by linearity
      and the Schwarz inequality, we have for any any $u, v \in
      L^2(\bbR)$ and any $\nu \in \bbR$
      \begin{align*}
         \abs*{
            \nabla_{\nu} \sL^{\sigma_{\SG}}(\nu, u)
            -
            \nabla_{\nu} \sL^{\sigma_{\SG}}(\nu, v)
         }
         &=
         \abs*{
            \ip*{
               \varphi_{\sigma^2_{\SG}} \conv \left(
                  \left(uu\adj - vv\adj\right) \circ \vtau_{\nu_{\grtr}-\nu}
               \right)
            }
            {
               \ip*{
                  \nabla_{\vx}[\varphi_{\sigma_{\SG}^2} 
                  \conv X_{\grtr}]
               }
               {
                  \vtau_{\pi/2}
               }_{\ell^2}
            }_{L^2(\bbR^2)}
         }
         \\
         &\leq
         \norm*{
            \varphi_{\sigma^2_{\SG}} \conv \left(
               \left(uu\adj - vv\adj\right) \circ \vtau_{\nu_{\grtr}-\nu}
            \right)
         }_{L^2(\bbR^2)}
         \norm*{
            \ip*{
               \nabla_{\vx}[\varphi_{\sigma_{\SG}^2} 
               \conv X_{\grtr}]
            }
            {
               \vtau_{\pi/2}
            }_{\ell^2}
         }_{L^2(\bbR^2)}.
      \end{align*}
      The second factor can be controlled with the second conclusion of
      \Cref{lem:curl-field-at-opt-estimates}: this gives
      \begin{equation*}
         \norm*{
            \ip*{
               \nabla_{\vx}[\varphi_{\sigma_{\SG}^2} 
               \conv X_{\grtr}]
            }
            {
               \vtau_{\pi/2}
            }_{\ell^2}
         }_{L^2(\bbR^2)}
         \leq
         \frac{1}{2\sigma_{\SG}^2}
         \left(
            1 + \sigma_{\SG}^2 
         \right)^{1/2}
         \leq
         \frac{1}{\sigma_{\SG}^2},
      \end{equation*}
      where the last bound worst-cases with our assumption on $\sigma_{\SG}$.
      For the first factor, we use Young's convolution inequality and the fact
      that $f \mapsto f \circ \vtau_{\nu_{\grtr}-\nu}$ is a unitary transformation of
      $L^2(\bbR^2)$ to obtain
      \begin{align*}
         \norm*{
            \varphi_{\sigma^2_{\SG}} \conv \left(
               \left(uu\adj - vv\adj\right) \circ \vtau_{-\nu}
            \right)
         }_{L^2(\bbR^2)}
         &\leq
         \norm*{\varphi_{\sigma^2_{\SG}}}_{L^1(\bbR^2)}
         \norm*{
            \left(uu\adj - vv\adj\right) \circ \vtau_{\nu_{\grtr}-\nu}
         }_{L^2(\bbR^2)}
         \\
         &\leq
         \norm*{
            uu\adj - vv\adj
         }_{L^2(\bbR^2)}.
      \end{align*}
      We have
      \begin{align*}
         \norm*{
            uu\adj - vv\adj
         }_{L^2(\bbR^2)}
         &=
         \abs*{
            \norm*{u}_{L^2(\bbR)}^2 -
            \norm*{v}_{L^2(\bbR)}^2
         }
         \\
         &\leq
         2\max\set*{
            \norm*{u}_{L^2(\bbR)},
            \norm*{v}_{L^2(\bbR)}
         }
         \abs*{
            \norm*{u}_{L^2(\bbR)} - 
            \norm*{v}_{L^2(\bbR)}
         }
         \\
         &\leq
         2\max\set*{
            \norm*{u}_{L^2(\bbR)},
            \norm*{v}_{L^2(\bbR)}
         }
         \norm*{u - v}_{L^2(\bbR)},
      \end{align*}
      where the two inequalities are both applications of the triangle
      inequality.
      Combining these estimates, we have shown
      \begin{align*}
         \abs*{
            \nabla_{\nu} \sL^{\sigma_{\SG}}(\nu, u)
            -
            \nabla_{\nu} \sL^{\sigma_{\SG}}(\nu, v)
         }
         \leq
         \frac{2}{\sigma_{\SG}^2}
         \max\set*{
            \norm*{u_{\rough}}_{L^2(\bbR)},
            \norm*{\bar{u}_{\rough}}_{L^2(\bbR)}
         }
         \norm*{\bar{u}_{\rough} - u_{\rough}}_{L^2(\bbR)},
      \end{align*}
      and applying this in the context of our gradient bounds, we obtain
      \begin{align*}
         \abs*{
            \left(
               \nabla_{\nu} \sL^{\sigma_{\SG}}(\spcdot, u_{\rough})
            \right)^2
            -
            \left(
               \nabla_{\nu} \sL^{\sigma_{\SG}}(\spcdot, \bar{u}_{\rough})
            \right)^2
         }
         &\leq
         \frac{8}{\sigma_{\SG}^4}
         \max\set*{
            \norm*{u_{\rough}}_{L^2(\bbR)}^3,
            \norm*{\bar{u}_{\rough}}_{L^2(\bbR)}^3
         }
         \norm*{\bar{u}_{\rough} - u_{\rough}}_{L^2(\bbR)}
         \\
         &\leq
         \frac{64}{\sigma_{\SG}^4}
         \norm*{\bar{u}_{\rough}}_{L^2(\bbR)}^3
         \norm*{\bar{u}_{\rough} - u_{\rough}}_{L^2(\bbR)},
      \end{align*}
      where the final inequality simplifies using the triangle inequality and 
      the assumption $ \norm{ u_{\rough} - \bar{u}_{\rough} }_{L^2(\bbR)} \leq
      \norm{\bar{u}_{\rough}}_{L^2(\bbR)}$, as we used before.
      This is precisely the second assertion.

   \end{proof}

\end{lemma}

\subsubsection{Technical Lemmas}

\begin{lemma}
   \label{lem:curl-field-nonnegative}

   Let $u = \Ind{[-\alpha, \alpha]}$ for some $\alpha > 0$, and for some
   smoothing level $\sigma > 0$ consider the associated curl field
   \begin{equation*}
      \sC(\vx)
      =
      \ip*{
         \nabla_{\vx}[\varphi_{\sigma^2} \ast u u\adj](\vx)
      }{
         \begin{bmatrix}
            0 & -1 \\
            1 & 0
         \end{bmatrix}
         \vx
      }_{\ell^2}.
   \end{equation*}
   If $\sigma^2 \leq \tfrac{\alpha^2}{24}$, then for any $\vx = (s, t)$ with $0
   \leq t \leq s$, one has
   \begin{equation*}
      \sC(s, t) \geq 0.
   \end{equation*}

   \begin{proof}
      The proof  uses expressions obtained in the proof
      of \Cref{lem:curl-field-at-opt-estimates}. Following
      \Cref{eq:curlfield-scalar}, if we define
      \begin{equation*}
         f(x) = \frac{1}{\sqrt{2\pi \sigma^2}} \int_{-\alpha}^{\alpha}
         e^{-\tfrac{(x - x')^2}{2\sigma^2}} \diff x', \quad x \in \bbR,
      \end{equation*}
      then we have
      \begin{equation*}
         \sC(s, t) = s f(s) f'(t) - t f(t) f'(s).
      \end{equation*}
      Moreover, note that $f > 0$, and by \Cref{eq:curlfield-gaussian-ibp} one has 
      $f'(x) = 0$ only if $x = 0$.
      To show the claim, it therefore suffices to show that
      \begin{equation*}
         \frac{f'(t)}{t f(t)} \geq \frac{f'(s)}{s f(s)}, \quad s \geq t > 0.
      \end{equation*}
      Differentiating, this monotonicity condition becomes
      \begin{equation}
         s (f'(s))^2 + f(s) f'(s) - s f''(s) f(s) \geq 0, \quad s > 0.
         \label{eq:curl-nonneg-target1}
      \end{equation}
      After a change of coordinates, we have
      \begin{equation*}
         f(s)
         = \frac{1}{\sqrt{\pi}}
         \int_{-\alpha/\sqrt{2}\sigma}^{\alpha/\sqrt{2}\sigma} e^{-(s - \sqrt{2}\sigma x)^2/2\sigma^2}
         \diff x.
      \end{equation*}
      Write $ A = \alpha / \sqrt{2}\sigma$, and define
      \begin{equation*}
         g(s) = \int_{-A}^A e^{-(s - x)^2} \diff x.
      \end{equation*}
      Noting that $\pi^{-1/2} g(s / \sqrt{2}\sigma) = f(s)$ and inspecting
      \Cref{eq:curl-nonneg-target1}, we see that it suffices to show that $g$
      satisfies the differential inequality in \Cref{eq:curl-nonneg-target1}.
      Using the fundamental theorem of calculus, we have
      \begin{align*}
         g'(s) &= e^{-(s+A)^2} - e^{-(s-A)^2} \\
         &= -2 e^{-s^2 - A^2} \sinh(2sA),
         \labelthis \label{eq:curl-nonneg-g-deriv}
      \end{align*}
      and by an additional straightforward differentiation
      \begin{equation*}
         g''(s) = 4 e^{-s^2 - A^2}( s \sinh(2sA) - A \cosh(2sA) ).
      \end{equation*}
      After substituting into \Cref{eq:curl-nonneg-target1} and cancelling some
      positive factors, the inequality to show becomes
      \begin{equation}
         2sA  \sinh(2sA) \left( \frac{A\inv e^{- A^2}}{e^{s^2} g(s)} \right)
         + 2sA \coth(2sA)
         - (1 + 2s^2)
         \geq 0.
         \label{eq:curl-nonneg-target2}
      \end{equation}
      We will prove this bound in two regimes: first, for $0 < s \leq A$, then
      for $s > A$.

      \subparagraph{Small $s$.}
      To start, we will develop a simple estimate for $g$. Notice that
      \begin{align*}
         e^{s^2} g(s) 
         &=
         \int_{-A}^A e^{-x^2} e^{2sx} \diff x \\
         &\leq
         \int_{-A}^A e^{2sx} \diff x \\
         &=
         \frac{1}{s} \sinh(2As),
      \end{align*}
      so it suffices to show
      \begin{equation*}
         2s^2 e^{-A^2}
         + 2sA \coth(2sA)
         - (1 + 2s^2)
         \geq 0.
      \end{equation*}
      For large $A$, the first term is sub-leading, and it suffices to simply
      show
      \begin{equation*}
         2sA \coth(2sA)
         - (1 + 2s^2)
         \geq 0.
      \end{equation*}
      We will show this bound on the requisite interval in two steps, since $s
      \mapsto s \coth s$ does not have a globally-convergent power series
      representation at zero. First, we have from the power series
      representation the bound $2sA \coth 2sA \geq 1 + \tfrac{(2sA)^2}{3} -
      \tfrac{(2sA)^4}{45}$ for all $s$; this bound is initially valid for
      $\abs{2sA} < \pi$, then extended to all $s$ by noticing that it is
      decreasing for $2sA \geq \pi$, whereas $2sA \coth 2sA$ is increasing for
      $s \geq 0$. With this bound, it suffices to show
      \begin{equation*}
         \frac{2s^2}{3} \left(
            2A^2 - 3 - \frac{8A^4 s^2}{15}
         \right) \geq 0,
      \end{equation*}
      which, when $A^2 \geq 6$, holds for all $0 \leq s \leq A\inv \sqrt{45/16}$. 
      Next, 
      notice that \Cref{eq:curl-nonneg-target2} can be written equivalently as
      \begin{equation}
         s\left( 
            e^{-(s-A)^2}
            - e^{-(s+A)^2}
         \right)
         + g(s) \left(
            2sA \coth(2sA)
            - (1 + 2s^2)
         \right)
         \geq 0,
         \label{eq:curl-nonneg-target3}
      \end{equation}
      where the first term is nonnegative.
      Since $\tanh(x) \leq 1$ if $x \geq 0$, it then suffices to show
      \begin{equation*}
         g(s) \left(
            2sA
            - (1 + 2s^2)
         \right)
         \geq 0.
      \end{equation*}
      The concave quadratic function $ 2sA - (1 + 2s^2)$ 
      has its two roots at $\tfrac{A}{2} \pm \tfrac{\sqrt{A^2 - 1}}{2}$;
      using the inequality $\sqrt{1 - x} \geq 1-x$ for $0 \leq x \leq 1$, it
      follows that these two roots are outside of the interval
      $[\tfrac{1}{2A}, A - \tfrac{1}{2A}]$, and hence $ 2sA - (1 + 2s^2)$
      is nonnegative on this interval. Since $\sqrt{45/16} \geq \half$, this
      establishes the inequality on $0 < s \leq A - \tfrac{1}{2A}$.

      Finally, to demonstrate the inequality on $A - \tfrac{1}{2A} \leq s \leq
      A$, we return to the sufficient expression of
      \Cref{eq:curl-nonneg-target2} given above, as
      \begin{equation*}
         s\left( 
            e^{-(s-A)^2}
            - e^{-(s+A)^2}
         \right)
         + g(s) \left(
            2sA
            - (1 + 2s^2)
         \right)
         \geq 0,
      \end{equation*}
      and note again that the concave quadratic function $ 2sA - (1 + 2s^2)$ 
      is maximized at $s =
      A/2$, hence is a decreasing function of $s$ on this interval; so it
      suffices to show
      \begin{equation*}
         s\left( 
            e^{-(s-A)^2}
            - e^{-(s+A)^2}
         \right)
         - g(s) 
         \geq 0,
      \end{equation*}
      From \Cref{eq:curl-nonneg-g-deriv}, it is clear that $g$ is a decreasing
      function of $s$, so we can show
      \begin{equation*}
         s\left( 
            e^{-(s-A)^2}
            - e^{-(s+A)^2}
         \right)
         - g(A - \tfrac{1}{2A})
         \geq 0.
      \end{equation*}
      Now, exploiting the fact that the parameter $s$ in the definition of
      $g(s)$ is similar to a ``mean'' parameter for the gaussian integrand, we
      calculate
      \begin{align*}
         g(A - \tfrac{1}{2A})
         &=
         \int_{-A}^A e^{-\left(x - (A - \tfrac{1}{2A})\right)^2} \diff x
         \\
         &=
         \int_{-A}^{A - \tfrac{1}{2A}}
         e^{-\left(x - (A - \tfrac{1}{2A})\right)^2} \diff x
         +
         \int_{A - \tfrac{1}{2A}}^{A}
         e^{-\left(x - (A - \tfrac{1}{2A})\right)^2} \diff x
         \\
         &\leq
         \frac{\sqrt{\pi}}{2}
         + \frac{1}{2A}.
      \end{align*}
      The last line above worst-cases the value of the first integral (as it is
      no larger than half of the integral over $\bbR$, by symmetry), and uses a
      $L^1$-$L^\infty$ bound to control the second. Meanwhile, using elementary
      inequalities and $A^2 \geq 12$ assumed previously, we have
      \begin{align*}
         s\left( 
            e^{-(s-A)^2}
            - e^{-(s+A)^2}
         \right)
         &\geq
         \left(A - \frac{1}{2A}\right)
         \left(
            1 - (s-A)^2 - e^{-\left( 2A - \tfrac{1}{2A} \right)^2}
         \right)
         \\
         &\geq
         \left(A - \frac{1}{2A}\right)
         \left(
            1 - \frac{1}{4A^2}
            - e^{-3A^2}
         \right),
      \end{align*}
      so it suffices to show
      \begin{equation*}
         A \left(1 - \frac{1}{2A^2}\right)
         \left(
            1 - \frac{1}{4A^2}
            - e^{-3A^2}
         \right)
         -
         \left(
            \frac{\sqrt{\pi}}{2}
            + \frac{1}{2A}
         \right)
         \geq 0,
      \end{equation*}
      which is evidently true for all $A^2 \geq 12$.
      This establishes the inequality on $0 < s \leq A$.

      \subparagraph{Large $s$.}
      For this regime, we will again proceed in steps; first for $A \leq s \leq
      A + \third$, then for $A + \third \leq s \leq 2A$, then for $s \geq 2A$.

      First, we will
      develop the bound for $A \leq s \leq A + c$, where $c>0$ is a small
      absolute constant. Proceeding as above, 
      but using now that when $s \geq A$ we have $2sA - (1 + 2s^2) = -(1 +
      2s(s-A) \leq 0$ so that we can leverage the bound 
      $g(s) = \int_{s-A}^{s+A} e^{-x^2} \diff x \leq \int_0^{s+A} e^{-x^2}
      \diff x \leq \sqrt{\pi}/2$,
      we have that it suffices to show
      \begin{equation*}
         A\left( 
            e^{-c^2}
            - e^{-4A^2}
         \right)
         - \frac{\sqrt{\pi}}{2} \left( 1 + 2c(A+c) \right) \geq 0,
      \end{equation*}
      which, after rearranging, is simply
      \begin{equation*}
         A \left( e^{-c^2} - \sqrt{\pi} c \right) - A e^{-4A^2} -
         \sqrt{\pi}\left( \half + c^2 \right) \geq 0.
      \end{equation*}
      We verify numerically that this holds for $c = \tfrac{1}{3}$ when $A^2
      \geq 12$, as we have assumed.

      Now we proceed for $s \geq A + \tfrac{1}{3}$. We will develop a sequence
      of refined upper bounds on $g(s)$. We have from a change of coordinates
      \begin{align*}
         g(s) 
         &= e^{-(s-A)^2} \int_{0}^{2A} e^{-x^2 - 2x(s-A)} \diff x.
         \labelthis \label{eq:curl-nneg-coc}
      \end{align*}
      Our previous estimate amounts to controlling the integral via $e^{-x^2}
      \leq 1$. 
      We will improve over this estimate slightly by instead
      developing a piecewise linear upper bound for the concave function $x
      \mapsto -x^2 - 2x(s-A)$. 
      Below, we will write $t = s-A$ for concision; by assumption $t \geq
      \tfrac{1}{3}$.
      For any $\veps \geq 0$, by concavity, we have for all $x \in \bbR$
      \begin{equation*}
         -x^2 - 2x(s-A)
         \leq
         -2(t + \veps)x + \veps^2.
      \end{equation*}
      Elementary algebra shows that the ``null'' upper bound for $\veps = 0$,
      that is $x \mapsto -2tx$, intersects with $x \mapsto -2(t + \veps)x +
      \veps^2$ at $x = \veps/2$. Hence, if $0 \leq \veps \leq 4A$, 
      we can estimate the integral as
      \begin{align*}
         \int_{0}^{2A} e^{-x^2 - 2x(s-A)} \diff x
         &\leq
         \int_{0}^{\veps/2} e^{-2tx} \diff x
         + e^{\veps^2} \int_{\veps/2}^{2A}e^{-2(t + \veps)x} \diff x \\
         &=
         \frac{1}{2t} \left( 1 - e^{-t\veps} \right)
         + \frac{1}{2(t+\veps)} \left(
            e^{-\veps t} - e^{\veps^2} e^{-4(t+\veps)A}
         \right)
         \\
         &\leq
         \frac{1}{2t} \left( 
            1 - \frac{\veps e^{-\veps t}}{\veps + t}
         \right).
      \end{align*}
      Choosing $\veps = 1/t$, this bound implies
      \begin{equation*}
         g(s)
         \leq
         \frac{1}{2t}
         e^{-(s-A)^2}
         \left(
            1 - \frac{e\inv}{1 + t^2}
         \right),
      \end{equation*}
      and so it suffices to show
      \begin{equation*}
         e^{-(s-A)^2}
         \left(
            s
            - \frac{1}{2(s-A)} \left(
               1 - \frac{e\inv}{1 + (s-A)^2}
            \right)
            \left(
               1 + 2s(s-A)
            \right)
         \right)
         - s e^{-(s+A)^2}
         \geq 0.
      \end{equation*}
      After some cancellation, this reads equivalently
      \begin{equation*}
         \frac{1}{2(s-A)}
         \left(
            \frac{ 1 + 2s(s-A) }{e(1+(s-A)^2)}
            - 1
         \right)
         - s e^{-4sA}
         \geq 0.
      \end{equation*}
      We will estimate the term in parenthesis. We have
      \begin{align*}
         \frac{ 1 + 2s(s-A) }{e(1+(s-A)^2)}
         - 1
         &=
         \left(
            \frac{
               -(1 - \tfrac{1}{e}) - (s-A) \left(
                  s(1 - \tfrac{2}{e}) - A
               \right)
            }{
               1 + (s-A)^2
            }
         \right).
      \end{align*}
      The numerator is a concave quadratic, which is maximized at
      $s = A \tfrac{e-1}{e-2}$; the constant is between $2$ and $3$.
      We check that when $s = A + \tfrac{1}{3}$ and $A^2 \geq 12$, the
      numerator is positive. Using $A + \tfrac{1}{3} \leq s \leq 2A$, we thus
      have that it suffices to show
      \begin{equation*}
         \left(
            \frac{
               11 + 6A - 10e
            }{
               9e(1 + A^2)
            }
         \right)
         - \frac{4A}{3} e^{-4A^2}
         \geq 0.
      \end{equation*}
      We numerically verify that this inequality holds for all $A^2 \geq 12$.

      Finally, we improve \Cref{eq:curl-nneg-coc} once more and then use $s \geq 2A$ 
      to conclude quickly.
      An improved estimate comes from \cite[Theorem 1]{Zhang2020-vu}: we apply
      this to \Cref{eq:curl-nneg-coc} to obtain
      \begin{align*}
         g(s) 
         &\leq e^{-(s-A)^2} \int_{0}^{\infty} e^{-x^2 - 2x(s-A)} \diff x
         \\
         &=
         \frac{\sqrt{\pi}}{2}  \mathrm{erfc}(s-A)
         \\
         &\leq
         \frac{e^{-(s-A)^2}}{2(s-A)}\left[
            1 - \frac{
               2 - 3 e^{-(1 + 2(s-A))} - 2(s-A)e^{-(1 + 2(s-A))}
            }
            {
               (1 + 2(s-A))^2
            }
         \right],
      \end{align*}
      where the estimate is applied in the third line, and the second line is a
      standard integral.  Following then \Cref{eq:curl-nonneg-target3} and
      using again that $\abs{\tanh} \leq 1$, it suffices to show
      \begin{equation*}
         1 - \frac{
            2 - 3 e^{-(1 + 2(s-A))} - 2(s-A)e^{-(1 + 2(s-A))}
         }
         {
            (1 + 2(s-A))^2
         }
         \leq
         \frac{
            2s(s-A)\left( 
               1 - e^{-4sA}
            \right)
         }
         {
            1 + 2s(s-A)
         }.
      \end{equation*}
      After rearranging with some algebra, it suffices to show
      \begin{equation*}
         e^{-(1 + 2(s-A))}
         \frac{
            3  + 2(s-A)
         }
         {
            (1 + 2(s-A))^2
         }
         +
         e^{-4sA}
         \frac{
            2s (s-A)
         }
         {
            1 + 2s(s-A)
         }
         \leq
         \frac{2}{
            (1 + 2(s-A))^2
         }
         -\frac{1
         }
         {
            1 + 2s(s-A)
         }.
      \end{equation*}
      By algebra,
      \begin{equation*}
         \frac{2}{
            (1 + 2(s-A))^2
         }
         -\frac{1
         }
         {
            1 + 2s(s-A)
         }
         =
         \frac{
            1 + 4sA - 4(s-A) - 4A^2
         }
         {
            (1 + 2s(s-A))
            (1 + 2(s-A))^2
         },
      \end{equation*}
      which is easily seen to be nonnegative when $s \geq A$.
      Clearing denominators, it then suffices to show
      \begin{equation*}
         ( 3  + 2(s-A))
         (1 + 2s(s-A))
         e^{-(1 + 2(s-A))}
         +
         2s (s-A) (1 + 2(s-A))^2
         e^{-4sA}
         \leq
         1 + 4sA - 4(s-A) - 4A^2.
      \end{equation*}
      We can show this holds easily by worst-casing for convenience, since the
      LHS has exponential prefactors.
      Since $s \geq 2A$, we have $2(s-A) \geq s$. We always have $s-A \leq s$,
      and since $A \geq 1$ we have $s \geq 2$,
      so it suffices to show
      \begin{equation*}
         9s^3 e^{-s}
         +
         18s^4
         e^{-4sA}
         \leq
         1 + 4sA - 4(s-A) - 4A^2.
      \end{equation*}
      Elementary calculus implies that the first term on the LHS is decreasing
      as soon as $s \geq 3$, and the second term is decreasing as soon as $s
      \geq 1/A$, both of which are implied by $s \geq 2A$ and our assumptions
      on $A$.
      Since the RHS is increasing, it suffices to check
      \begin{equation*}
         9A^3 e^{-A}
         +
         18A^4
         e^{-4A^2}
         \leq
         1 + 4A(A-1).
      \end{equation*}
      A numerical evaluation and the preceding calculus argument shows that this is
      true as soon as $A \geq 3$.

      \end{proof}
\end{lemma}

\begin{lemma}
   \label{lem:residual-field-estimates}

   Consider the residual field arising in the study of the alignment gradient: 
   for any $\nu \in \bbR$, we consider
   \begin{equation*}
      \bar{u}_{\rough} \bar{u}_{\rough}\adj \circ \vtau_{\nu}
      -
      \bar{u}_{\rough} \bar{u}_{\rough}\adj \circ \vtau_{-\nu},
   \end{equation*}
   where the (unscaled, for convenience) nominal rough initial representation
   is defined as
   \begin{equation*}
      \bar{u}_{\rough}(s) = 
      \Ind{\abs{s} \leq 1} \cos(\pi s / 2).
   \end{equation*}
   Then for any $\vx = (s, t)$ with $0 \leq t \leq s \leq 1$, the difference is
   nonnegative:
   \begin{equation*}
      \bar{u}_{\rough} \bar{u}_{\rough}\adj \circ \vtau_{\nu}(\vx)
      -
      \bar{u}_{\rough} \bar{u}_{\rough}\adj \circ \vtau_{-\nu}(\vx)
      \geq 0,
   \end{equation*}
   and moreover for any $\vx = (s, t)$ with $0 \leq t \leq s \leq 1$ and any $0
   \leq \nu \leq \pi/7$,
   it satisfies the estimate
   \begin{equation*}
      \bar{u}_{\rough} \bar{u}_{\rough}\adj \circ \vtau_{\nu}(\vx)
      -
      \bar{u}_{\rough} \bar{u}_{\rough}\adj \circ \vtau_{-\nu}(\vx)
      \geq
      \frac{7 \sin \nu}{1000}
      \Ind{-0.137 \leq t - \tfrac{1}{\sqrt{2}} \leq -0.127}
      \Ind{-0.001 \leq s - \tfrac{1}{\sqrt{2}} \leq 0.001}.
   \end{equation*}

   \begin{proof}
      We have to show
      \begin{equation*}
         \bar{u}_{\rough} \bar{u}_{\rough}\adj \circ \vtau_{-\nu}
         -
         \bar{u}_{\rough} \bar{u}_{\rough}\adj \circ \vtau_{\nu}
         \leq 0.
      \end{equation*}
      Using two trigonometric identities, we can write
      for any $\vx = (s, t)$ with $0 \leq t \leq s \leq 1$
      \begin{align*}
         &\bar{u}_{\rough} \bar{u}_{\rough}\adj \circ \vtau_{-\nu}(\vx)
         -
         \bar{u}_{\rough} \bar{u}_{\rough}\adj \circ \vtau_{\nu}(\vx)
         \\
         &\qquad=
         \sin( \tfrac{\pi}{2} (s+t) \cos \nu )
         \sin( \tfrac{\pi}{2} (s-t) \sin \nu )
         -
         \sin( \tfrac{\pi}{2} (s-t) \cos \nu )
         \sin( \tfrac{\pi}{2} (s+t) \sin \nu ).
      \end{align*}
      It is clear that this expression is identically zero when $\nu = 0$ or $s
      = t$, so assume otherwise below. To show the expression is nonpositive,
      it is equivalent to show
      \begin{equation*}
         \frac{
            \sin( \tfrac{\pi}{2} (s+t) \cos \nu )
         }{
            \sin( \tfrac{\pi}{2} (s+t) \sin \nu )
         }
         \leq
         \frac{
            \sin( \tfrac{\pi}{2} (s-t) \cos \nu )
         }{
            \sin( \tfrac{\pi}{2} (s-t) \sin \nu )
         }
      \end{equation*}
      for each $0 \leq t \leq s \leq 1$ and all $0 \leq \nu \leq \pi/4$. Given
      that under these constraints $s + t \leq 2$, it therefore suffices to
      show that
      \begin{equation*}
         x \mapsto 
         \frac{
            \sin( \tfrac{\pi}{2} x \cos \nu )
         }{
            \sin( \tfrac{\pi}{2} x \sin \nu )
         }
      \end{equation*}
      is a decreasing function of $x$ on $[0, 2]$.
      Rescaling coordinates, this is equivalent to showing that
      \begin{equation*}
         x \mapsto 
         \frac{
            \sin(x \cot \nu)
         }{
            \sin x
         }
      \end{equation*}
      is decreasing on $[0, \pi \sin \nu]$, and because $1/\cot \nu \geq \sin
      \nu$ when $0 \leq \nu \leq \pi/4$ it suffices instead to show
      decreasingness on $[0, \pi / \cot \nu]$. This is a standard calculation
      that arises in the study of the Dirichlet kernel in Fourier analysis; to
      obtain it, %
      write $A = \cot \nu$ and differentiate to obtain the sufficient
      condition
      \begin{equation*}
         Ax \cot Ax \leq x \cot x,
         \quad 0 < x < \pi/A.
      \end{equation*}
      This can be seen, for instance, from the power series expression for
      $x \cot x$, convergent for $\abs{x} < \pi$:
      \begin{equation*}
         Ax \cot Ax = 1 - 2\sum_{k=1}^\infty
         \frac{\zeta(2k)}{\pi^{2k}} A^{2k} x^{2k}
         \leq 
         1 - 2\sum_{k=1}^\infty
         \frac{\zeta(2k)}{\pi^{2k}} x^{2k}
         = x \cot x,
      \end{equation*}
      since $A \geq 1$ and all terms in the sum are nonpositive, where $\zeta$
      denotes the Riemann zeta function.
      Thus we have shown
      \begin{equation*}
         \bar{u}_{\rough} \bar{u}_{\rough}\adj \circ \vtau_{-\nu}
         -
         \bar{u}_{\rough} \bar{u}_{\rough}\adj \circ \vtau_{\nu}
         \leq 0.
      \end{equation*}

      Next, we show the quantitative bound.
      By our earlier work, we can write at any $\vx = (s, t)$ with $0 \leq t <
      s \leq 1$
      \begin{align*}
         &\frac{\sqrt{\pi}}{2}\left(
            \bar{u}_{\rough} \bar{u}_{\rough}\adj \circ \vtau_{\nu}(\vx)
            -
            \bar{u}_{\rough} \bar{u}_{\rough}\adj \circ \vtau_{-\nu}(\vx)
         \right)
         \\
         &\qquad=
         \sin( \tfrac{\pi}{2} (s+t) \sin \nu )
         \sin( \tfrac{\pi}{2} (s-t) \cos \nu )
         -
         \sin( \tfrac{\pi}{2} (s+t) \cos \nu )
         \sin( \tfrac{\pi}{2} (s-t) \sin \nu ).
         \labelthis \label{eq:gradient-lb-residual-diff}
      \end{align*}
      We will obtain a lower bound for this expression by combining
      term-by-term bounds, optimized for $t \approx s$. The difference terms
      are easiest: we can use the standard estimates
      \begin{align*}
         \sin\left( \tfrac{\pi}{2} (s - t) \cos \nu \right)  &\geq
         \tfrac{\pi}{2}(s-t) \cos \nu - \left( \tfrac{\pi}{2}(s-t) \cos
         \nu\right)^3 / 6, \\
         \sin\left( \tfrac{\pi}{2} (s - t) \sin \nu \right)
         &\leq
         \tfrac{\pi}{2} (s - t) \sin \nu,
      \end{align*}
      which follow by concavity.
      Similarly, concavity gives the estimates
      \begin{align*}
         \sin\left( \tfrac{\pi}{2} (s + t) \sin \nu \right)
         &\geq
         \sin\left( \pi s \sin \nu \right)
         +
         \tfrac{\pi}{2} \sin \nu \cos\left( \pi s \sin \nu \right) (t - s)
         - \frac{\pi^2 \sin^2 \nu}{8}(t - s)^2, \\
         \sin\left( \tfrac{\pi}{2} (s + t) \cos \nu \right)
         &\leq
         \sin\left( \pi s \cos \nu \right)
         +
         \tfrac{\pi}{2} \cos \nu \cos\left( \pi s \cos \nu \right) (t - s).
      \end{align*}
      These estimates yield a polynomial lower bound for the difference term
      when substituted into \Cref{eq:gradient-lb-residual-diff}. Since we know
      this difference is nonnegative, we can improve the bound by taking the
      maximum of it and zero, and then further simplify the bound based on its
      local behavior near $t \approx s$ to a quadratic lower bound. To this
      end, we have the unwieldy lower bound
      \begin{align*}
         \bar{u}_{\rough} \bar{u}_{\rough}\adj \circ \vtau_{\nu}(\vx)
         -
         \bar{u}_{\rough} \bar{u}_{\rough}\adj \circ \vtau_{-\nu}(\vx)
         &\geq
         \frac{\pi(s-t)}{2}\left(
            \cos(\nu) \sin(\pi s \sin\nu)
            - \sin(\nu) \sin(\pi s\cos\nu) 
         \right)
         \\
         &\quad
         - \frac{\pi^2(s-t)^2\cos \nu \sin \nu}{4} \left(
            \cos(\pi s \sin \nu) - \cos(\pi s\cos \nu)
         \right) \\
         &\quad
         - \frac{\pi^3(s-t)^3\cos \nu}{16} \left(
            \sin^2 \nu + \frac{1}{3} \cos^2 \nu \sin(\pi s \sin \nu)   
         \right) \\
         &\quad
         +
         \frac{\pi^4 (s-t)^4 \cos^3(\nu) \cos(\pi s \sin \nu) \sin \nu}{96}
         +
         \frac{\pi^5 (s-t)^5 \cos^3(\nu) \sin^2(\nu)}{384}.
      \end{align*}
      The degree four and five terms in this bound are both nonnegative, hence
      can be worst-cased out. To verify that the degree-two term dominates the
      degree-three term, we have to show for some $0 < \veps < 1$
      \begin{align*}
         (1-\veps) \sin \nu \left(
            \cos(\pi s \sin \nu) - \cos(\pi s\cos \nu)
         \right)
         &-
         \frac{\pi(s-t)}{4} \left(
            \sin^2 \nu + \frac{1}{3} \cos^2 \nu \sin(\pi s \sin \nu)   
         \right)
         \geq 0.
      \end{align*}
      We have $0 \leq s-t \leq 1$, and the LHS of the previous bound is a
      decreasing function of $s-t$. Moreover, we have $\sin(\pi s \sin \nu) /
      \sin \nu \leq \pi s$. Hence, to show this bound holds for all $0 \leq \nu
      \leq \pi/7$ and all $ 0 \leq s-t \leq 1/2$, it suffices to show for some
      $\veps$
      \begin{align*}
         (1-\veps) \left(
            \cos(\pi s \sin \nu) - \cos(\pi s\cos \nu)
         \right)
         &-
         \frac{\pi}{8} \left(
            \sin \pi/7 + \frac{\pi s}{3} 
         \right)
         \geq 0.
      \end{align*}
      We will show this by lower bounding the first term with calculus. We have
      for the second derivative of the first summand
      \begin{equation*}
         \partial_{\nu}^2[
         \cos(\pi s \sin \nu)
         ](\nu)
         =
         \sin(\pi s \sin \nu) (\pi s \sin \nu) - (\pi s \cos \nu)^2 \cos(\pi s
         \sin \nu).
      \end{equation*}
      We notice that this is an increasing function of $\nu$, because it is a
      difference of two terms which are (respectively) a product of two
      nonnegative increasing functions and a product of two nonnegative
      decreasing functions. Hence it attains its minimum value at $\nu = 0$.
      Similarly, we have
      \begin{equation*}
         \partial_{\nu}^2[
         -\cos(\pi s \cos \nu)
         ](\nu)
         =
          (\pi s \sin \nu)^2 \cos(\pi s
          - (\pi s \cos \nu )\sin(\pi s \cos \nu)
         \cos \nu),
      \end{equation*}
      which is once again a difference of a product of two nonnegative
      increasing functions and a product of two nonnegative decreasing
      functions, hence attains its minimum value at $\nu = 0$.
      It follows that the same is true of the sum, and Taylor's theorem then
      implies the lower bound
      \begin{equation*}
         \cos(\pi s \sin \nu) - \cos(\pi s\cos \nu)
         \geq
         (1 - \cos \pi s)
         -
         \frac{(\pi s)^2 + \pi s \sin \pi s}{2} \nu^2.
      \end{equation*}
      When $s \geq \half$, we have $-\cos \pi s \geq 2s - 1$.
      Worst-casing $\nu \leq \pi/7$ and choosing $\veps = 1/8$,
      it then suffices to show under these conditions
      \begin{equation*}
         \left(
            2s
            -
            \frac{(\pi s)^2 + \pi s \sin \pi s}{2} (\pi/7)^2
         \right)
         -
         \frac{\pi}{7} \left(
            \sin \pi/7 + \frac{\pi s}{3} 
         \right)
         \geq 0.
      \end{equation*}
      A numerical evaluation shows that this holds for all $\half \leq s \leq
      1$. Hence, we have the lower bound, valid for $\half \leq s \leq 1$, all
      $s-\half \leq t \leq s$, and all $0 \leq \nu \leq \pi/7$:
      \begin{align*}
         \bar{u}_{\rough} \bar{u}_{\rough}\adj \circ \vtau_{\nu}(\vx)
         -
         \bar{u}_{\rough} \bar{u}_{\rough}\adj \circ \vtau_{-\nu}(\vx)
         &\geq
         \frac{\pi(s-t)}{2}\left(
            \cos(\nu) \sin(\pi s \sin\nu)
            - \sin(\nu) \sin(\pi s\cos\nu) 
         \right)
         \\
         &\quad
         - \frac{15\pi^2(s-t)^2\cos \nu \sin \nu}{32} \left(
            \cos(\pi s \sin \nu) - \cos(\pi s\cos \nu)
         \right).
      \end{align*}
      Letting
      \begin{align*}
         A &=  
         \frac{\pi}{2}\left(
            \cos(\nu) \sin(\pi s \sin\nu)
            - \sin(\nu) \sin(\pi s\cos\nu) 
         \right),
         \\
         B &=
         \frac{15\pi^2\cos \nu \sin \nu}{32} \left(
            \cos(\pi s \sin \nu) - \cos(\pi s\cos \nu)
         \right),
      \end{align*}
      we have $B \geq 0$ since $\cos$ is decreasing and $\sin \leq \cos$ on our
      interval of interest, and
      \begin{equation*}
         \bar{u}_{\rough} \bar{u}_{\rough}\adj \circ \vtau_{\nu}(\vx)
         -
         \bar{u}_{\rough} \bar{u}_{\rough}\adj \circ \vtau_{-\nu}(\vx)
         \geq
         Ar - B r^2 = B\left(
            \left( \frac{A}{2B} \right)^2
            - \left(
               r - \frac{A}{2B}
            \right)^2
         \right),
      \end{equation*}
      where the RHS is a concave quadratic function of $r = s-t$. 
      For such a concave quadratic, the above forms make it clear that its two
      roots are at $0$ and $A/B$, and by concavity we have
      \begin{equation*}
         Ar - Br^2
         \geq
         \frac{3A^2}{16B}\Ind{\abs{r - A/2B} \leq \abs{A/4B}}.
      \end{equation*}
      We will show that this bound also applies to 
      $ \bar{u}_{\rough} \bar{u}_{\rough}\adj \circ \vtau_{\nu}(\vx) - \bar{u}_{\rough}
      \bar{u}_{\rough}\adj \circ \vtau_{-\nu}(\vx)$
      on our interval of interest, by showing that $A \geq 0$ and when $A/4B
      \leq s-t \leq 3A/4B$, $s$ and $t$ satisfy the previously assumed
      conditions uniformly in $\nu$. To see that $A \geq 0$, notice that
      \begin{equation*}
         A = \frac{\pi \sin \nu \cos \nu}{2} \left(
            \frac{\sin(\pi s \sin \nu)}{\sin \nu}
            -\frac{\sin(\pi s \cos \nu)}{\cos \nu}
         \right).
      \end{equation*}
      The function $x \mapsto \sin x / x$ is decreasing when $0 \leq x \leq
      \pi$, showing that $A \geq 0$. This means that any $s, t$ for which $s-t
      \geq A/4B$ satisfies our hypotheses. Next, note that
      \begin{equation*}
         \nu \mapsto
         \frac{\sin(\pi s \sin \nu)}{\sin \nu}
         -\frac{\sin(\pi s \cos \nu)}{\cos \nu}
      \end{equation*}
      is decreasing, as $x \mapsto \sin x / x$ is nonnegative and decreasing
      for $0 \leq x \leq \pi$ (the chain rule implies the composition is
      decreasing as a sum of decreasing functions). By the same token,
      \begin{equation*}
         \nu \mapsto
         \cos(\pi s \sin \nu) - \cos(\pi s\cos \nu)
      \end{equation*}
      is decreasing on our domain of interest. This implies
      \begin{equation*}
         \frac{
            \frac{\sin(\pi s \sin \pi/7)}{\sin \pi/7}
            -\frac{\sin(\pi s \cos \pi/7)}{\cos \pi/7}
         }
         {
            1 - \cos(\pi s)
         }
         \leq
         \frac{
            \frac{\sin(\pi s \sin \nu)}{\sin \nu}
            -\frac{\sin(\pi s \cos \nu)}{\cos \nu}
         }{
            \cos(\pi s \sin \nu) - \cos(\pi s\cos \nu)
         }
         \leq
         \frac{
            \pi s - \sin(\pi s)
         }
         {
            \cos(\pi s \sin \pi/7) - \cos(\pi s\cos \pi/7)
         }.
      \end{equation*}
      A numerical evaluation shows that both the LHS and the RHS are
      increasing. Hence, if $s \leq 0.72$, we have the bound
      \begin{equation*}
         \frac{
            \frac{\sin(\pi s \sin \nu)}{\sin \nu}
            -\frac{\sin(\pi s \cos \nu)}{\cos \nu}
         }{
            \cos(\pi s \sin \nu) - \cos(\pi s\cos \nu)
         }
         \leq
         \frac{
            0.72\pi - \sin(0.72\pi)
         }
         {
            \cos(0.72\pi \sin \pi/7) - \cos(0.72\pi \cos \pi/7)
         }
         \leq 1.483,
      \end{equation*}
      which implies
      \begin{equation*}
         \frac{3A}{4B}  
         \leq
         \frac{1.483 \cdot 4}{5\pi }
         \leq 0.378,
      \end{equation*}
      showing that any $s, t$ for which $s-t \leq 3A/4B$ satisfies our
      hypotheses. 
      Finally, as above, using $s \geq 0.7$ we can obtain the lower bound
      \begin{equation*}
         \frac{
            \frac{\sin(\pi s \sin \nu)}{\sin \nu}
            -\frac{\sin(\pi s \cos \nu)}{\cos \nu}
         }{
            \cos(\pi s \sin \nu) - \cos(\pi s\cos \nu)
         }
         \geq
         \frac{
            \frac{\sin(0.7\pi \sin \pi/7)}{\sin \pi/7}
            -\frac{\sin(0.7\pi \cos \pi/7)}{\cos \pi/7}
         }
         {
            1 - \cos(0.7\pi)
         }
         \geq
         0.544,
      \end{equation*}
      which implies
      \begin{equation*}
         \frac{A}{4B}  
         \geq
         \frac{0.544 \cdot 4}{15\pi }
         \geq 
         0.046.
      \end{equation*}
      Consequently, we have established
      \begin{equation*}
         \bar{u}_{\rough} \bar{u}_{\rough}\adj \circ \vtau_{\nu}(\vx)
         -
         \bar{u}_{\rough} \bar{u}_{\rough}\adj \circ \vtau_{-\nu}(\vx)
         \geq
         \frac{9\pi(\cos(\nu) \sin(\pi s \sin \nu) - \sin(\nu) \sin(\pi s\cos
         \nu))}{800}
         \Ind{\abs{(s-t) - A/2B} \leq \abs{A/4B}}.
      \end{equation*}
      As above, we can worst-case this bound further. Since
      \begin{align*}
         \cos(\nu) \sin(\pi s \sin \nu) - \sin(\nu) \sin(\pi s\cos \nu)
         &=
         \sin(\nu) \cos(\nu)\left(
            \frac{\sin(\pi s \sin \nu)}{\sin \nu} - \frac{\sin(\pi s\cos
            \nu)}{\cos \nu}
         \right)
         \\
         &\geq
         \sin(\nu) \cos(\pi/7)\left(
            \frac{\sin(\pi s \sin \pi/7)}{\sin \pi/7} - \frac{\sin(\pi s\cos
            \pi/7)}{\cos \pi/7}
         \right)
         \\
         &\geq
         \sin(\nu) \cos(\pi/7)\left(
            \frac{\sin(\pi/2 \sin \pi/7)}{\sin \pi/7} - \frac{\sin(\pi/2\cos
            \pi/7)}{\cos \pi/7}
         \right)
         \\
         &\geq \frac{\sin \nu}{5},
      \end{align*}
      we have
      \begin{equation*}
         \bar{u}_{\rough} \bar{u}_{\rough}\adj \circ \vtau_{\nu}(\vx)
         -
         \bar{u}_{\rough} \bar{u}_{\rough}\adj \circ \vtau_{-\nu}(\vx)
         \geq
         \frac{7 \sin \nu}{1000}
         \Ind{\abs{(s-t) - A/2B} \leq \abs{A/4B}}.
      \end{equation*}
      In addition, we have shown above
      \begin{equation*}
         0.184 \leq \frac{A}{B} \leq 0.504,
      \end{equation*}
      which implies
      \begin{equation*}
         \frac{A}{4B} \leq 0.126,
         \quad
         \frac{3A}{4B} \geq 0.138,
      \end{equation*}
      whence
      \begin{equation*}
         \bar{u}_{\rough} \bar{u}_{\rough}\adj \circ \vtau_{\nu}(\vx)
         -
         \bar{u}_{\rough} \bar{u}_{\rough}\adj \circ \vtau_{-\nu}(\vx)
         \geq
         \frac{7 \sin \nu}{1000}
         \Ind{0.126 \leq (s-t) \leq 0.138}.
      \end{equation*}
      Because we have shown that the LHS is nonnegative previously, this bound
      holds for all $t$.
      Now notice that we can write the constraint on $t$ on the RHS
      equivalently as
      \begin{equation*}
         0.126 \leq s-t \leq 0.138
         \iff
         (s - 0.132) - 0.006 \leq t \leq (s - 0.132) + 0.006.
      \end{equation*}
      Hence, if we consider a sub-interval of valid $s$, namely $s \in
      [\tfrac{1}{\sqrt{2}} - 0.001, \tfrac{1}{\sqrt{2}} + 0.001]$, 
      we have for such $s$
      \begin{equation*}
         0.126 \leq s-t \leq 0.138
         \impliedby
         (\tfrac{1}{\sqrt{2}} - 0.132) - 0.005 \leq t \leq (\tfrac{1}{\sqrt{2}} - 0.132) + 0.005.
      \end{equation*}
      In particular,
      \begin{equation*}
         \Ind{0.126 \leq (s-t) \leq 0.138}
         \Ind{0.7 \leq s \leq 0.72}
         \geq
         \Ind{-0.137 \leq t - \tfrac{1}{\sqrt{2}} \leq -0.127}
         \Ind{-0.001 \leq s - \tfrac{1}{\sqrt{2}} \leq 0.001}.
      \end{equation*}

   \end{proof}
\end{lemma}

\begin{lemma}
   \label{lem:curl-field-at-opt-estimates}

   For $\beta > 0$, let $u = \Ind{[-\beta, \beta]}$, and for some smoothing
   level $\sigma > 0$ consider the associated curl fields
   \begin{equation}
      \sC^{\beta}(\vx)
      =
      \ip*{
         \nabla_{\vx}[\varphi_{\sigma^2} \ast u u\adj](\vx)
      }{
         \begin{bmatrix}
            0 & -1 \\
            1 & 0
         \end{bmatrix}
         \vx
      }_{\ell^2}.
      \label{eq:curl-field-general-base}
   \end{equation}
   We have the following estimates: if $\sigma^2 \geq 1$ and $1/\sqrt{2} \leq
   \alpha \leq 1$, then
   \begin{equation*}
      \ip{\sC^1}{\sC^{\alpha}}_{L^2(\bbR^2)}
      \geq \frac{1}{8\pi \sigma^4},
   \end{equation*}
   and for any $\beta$ and any $\sigma^2 > 0$,
   \begin{equation*}
      \ip{\sC^\beta}{\sC^\beta}_{L^2(\bbR^2)}
      \leq
      \frac{\beta^4}{\sigma^4}
      \left(
         \sigma^2
         +
         2\beta^2
      \right).
   \end{equation*}

   \begin{proof}
      If we unravel the expression \Cref{eq:curl-field-general-base}, we have
      \begin{align*}
         \sC(s, t)
         &=
         \ip*{
            \nabla_{\vx}[\varphi_{\sigma^2} \ast uu\adj ](s, t)
         }{
            (-t, s)
         }_{\ell^2} \\
         &=
         \ip*{
            \int_{\bbR} \int_{\bbR}
            \begin{bmatrix}
               u(s') u(t') \varphi_{\sigma^2}(t - t')
               \nabla_s[\varphi_{\sigma^2}](s - s')
               \\
               u(s') u(t') \varphi_{\sigma^2}(s - s')
               \nabla_t[\varphi_{\sigma^2}](t - t')
            \end{bmatrix}
            \diff s \diff t
         }{
            (-t, s)
         }_{\ell^2} \\
         &=
         \One\adj
         \begin{bmatrix}
            -t 
            \left(
               \int_{\bbR} u(s') \nabla_s \varphi_{\sigma^2}(s - s') \diff s'
            \right)
            \left(
               \int_{\bbR} u(t') \varphi_{\sigma^2}(t - t') \diff t'
            \right) \\
            s 
            \left(
               \int_{\bbR} u(s') \varphi_{\sigma^2}(s - s') \diff s'
            \right)
            \left(
               \int_{\bbR} u(t') \nabla_t \varphi_{\sigma^2}(t - t') \diff t'
            \right)
         \end{bmatrix}.
      \end{align*}
      Defining
      \begin{align*}
         f_1(x) &= \int_{\bbR} u(x') \nabla \varphi_{\sigma^2}(x - x') \diff
         x'; \\
         f_2(x) &= -x \int_{\bbR} u(x') \varphi_{\sigma^2}(x - x') \diff x',
      \end{align*}
      the above implies
      \begin{equation}
         \sC(s, t) = f_1(s) f_2(t) - f_1(t) f_2(s).
         \label{eq:curlfield-scalar}
      \end{equation}
      Now notice that, by the fundamental theorem of calculus,
      \begin{align*}
         f_1(x) 
         &= \int_{\bbR} u(x - x') \nabla \varphi_{\sigma^2}(x') \diff x'
         \\
         &= \int_{x - \beta}^{x + \beta} \nabla \varphi_{\sigma^2}(x') \diff x'
         \\
         &= \varphi_{\sigma^2}(x + \beta) - \varphi_{\sigma^2}(x - \beta).
         \labelthis \label{eq:curlfield-gaussian-ibp}
      \end{align*}
      For the estimates we need, we introduce more general notation: for any
      $\gamma > 0$ (following
      \Cref{eq:curlfield-gaussian-ibp}), let
      \begin{align*}
         f_1^{\gamma}(x) &= 
         \varphi_{\sigma^2}(x + \gamma) - \varphi_{\sigma^2}(x - \gamma);
         \\
         f_2^{\gamma}(x) &= -x \int_{\bbR} \Ind{[-\gamma, \gamma]}(x') \varphi_{\sigma^2}(x - x') \diff x'.
      \end{align*}
      Our task is to estimate $\Xi$, defined as
      \begin{equation*}
         \Xi(\alpha, \beta)
         =
         \ip*{
            f_1^{\alpha} (f_2^{\alpha})\adj
            -
            f_2^{\alpha} (f_1^{\alpha})\adj
         }{
            f_1^{\beta} (f_2^{\beta})\adj
            -
            f_2^{\beta} (f_1^{\beta})\adj
         }_{L^2(\bbR^2)},
      \end{equation*}
      since by the analysis above we have
      $\ip{\sC^\alpha}{\sC^\beta}_{L^2(\bbR^2)} = \Xi(\alpha,\beta)$.
      Distributing in the inner product, we have
      \begin{equation}
         \Xi(\alpha, \beta)
         =
         2\left(
            \ip{f_1^\alpha}{f_1^{\beta}} \ip{f_2^\alpha}{f_2^{\beta}}
            - \ip{f_1^\alpha}{f_2^{\beta}} \ip{f_2^\alpha}{f_1^{\beta}}
         \right).
         \label{eq:nu-second-deriv-to-est}
      \end{equation}
      We first estimate the cross terms $\ip{f_2^\alpha}{f_1^\beta}$.
      We have by
      \Cref{eq:curlfield-gaussian-ibp}
      \begin{align*}
         \ip{f_2^\alpha}{f_1^\beta}
         &=
         \int_{\bbR}
         x (\Ind{[-\alpha,\alpha]} \ast \varphi_{\sigma^2})(x) \left(
            \varphi_{\sigma^2}(x - \beta) 
            - \varphi_{\sigma^2}(x + \beta) 
         \right)
         \diff x.
      \end{align*}
      As above, we can integrate this out once again. We have, following the
      argument in \Cref{eq:curlfield-gaussian-ibp}
      \begin{align*}
         \int_{\bbR} x \Ind{[-\alpha,\alpha]}(x- x') \varphi_{\sigma^2}(x - \beta) \diff x
         &=
         \int_{\bbR} (x - \beta) \Ind{[-\alpha,\alpha]}(x- x') \varphi_{\sigma^2}(x - \beta) \diff x
         + \beta \int_{\bbR} \Ind{[-\alpha,\alpha]}(x - x') \varphi_{\sigma^2}(x - \beta) \diff x
         \\
         &=
         -\sigma^2 \int_{\bbR} \Ind{[-\alpha,\alpha]}(x- x') \nabla \varphi_{\sigma^2}(x - \beta) \diff x
         + \beta \int_{\bbR} \Ind{[-\alpha,\alpha]}(x - x') \varphi_{\sigma^2}(x - \beta) \diff x \\
         &=
         \sigma^2 \left(
            \varphi_{\sigma^2}(x' - \alpha - \beta) - \varphi_{\sigma^2}(x' +
            \alpha - \beta)
         \right)
         + \beta \int_{\bbR} \Ind{[-\alpha,\alpha]}(x - x') \varphi_{\sigma^2}(x - \beta) \diff x.
      \end{align*}
      Reasoning symmetrically, we get
      \begin{align*}
         \int_{\bbR} x \Ind{[-\alpha,\alpha]}(x- x')\left(
            \varphi_{\sigma^2}(x - \beta) 
            - \varphi_{\sigma^2}(x + \beta) 
         \right)\diff x
         &=
         \sigma^2\bigl(
            \varphi_{\sigma^2}(x' - \alpha - \beta) 
            - \varphi_{\sigma^2}(x' + \alpha - \beta) \\
            &\qquad\quad- \varphi_{\sigma^2}(x' - \alpha + \beta)
            + \varphi_{\sigma^2}(x' + \alpha + \beta)
         \bigr) \\
         &\qquad + \beta \int_{\bbR} \Ind{[-\alpha,\alpha]}(x - x') \left(
            \varphi_{\sigma^2}(x - \beta) 
            + \varphi_{\sigma^2}(x + \beta) 
         \right)
         \diff x.
      \end{align*}
      To obtain $\ip{f_2^\alpha}{f_1^\beta}$ from this last expression, we
      integrate against $\varphi_{\sigma^2}(x')$. Integrating this function
      against the first term on the RHS of the previous expression yields a
      convolution between gaussians, which is another gaussian:
      \begin{align*}
         &\sigma^2
         \int_{\bbR} \varphi_{\sigma^2}(x')
         \bigl(
            \varphi_{\sigma^2}(x' - \alpha - \beta) 
            - \varphi_{\sigma^2}(x' + \alpha - \beta)
            - \varphi_{\sigma^2}(x' - \alpha + \beta)
            + \varphi_{\sigma^2}(x' + \alpha + \beta)
         \bigr) \diff x' \\
         &\qquad\qquad=
         2\sigma^2\left(
            \varphi_{2\sigma^2}(\alpha + \beta)
            -
            \varphi_{2\sigma^2}(\alpha - \beta)
         \right),
         \labelthis \label{eq:tilt-2ndderiv-crossterm-1}
      \end{align*}
      where we used even symmetry of the gaussian. Integrating against the
      second term can be similarly manipulated to give
      \begin{align*}
         &\beta \int_{\bbR} \int_{\bbR}
         \Ind{[-\alpha,\alpha]}(x - x') \left(
            \varphi_{\sigma^2}(x - \beta) 
            + \varphi_{\sigma^2}(x + \beta) 
         \right) \varphi_{\sigma^2}(x')
         \diff x \\
         &\qquad=
         \beta \int_{\bbR} \int_{\bbR}
         \Ind{[-\alpha,\alpha]}(x) \left(
            \varphi_{\sigma^2}(x + x' - \beta) 
            + \varphi_{\sigma^2}(x + x' + \beta) 
         \right) \varphi_{\sigma^2}(x')
         \diff x \\
         &\qquad=
         \beta \int_{\bbR} \int_{\bbR}
         \Ind{[-\alpha,\alpha]}(x) \left(
            \varphi_{\sigma^2}(\beta - x - x')
            + \varphi_{\sigma^2}(-\beta - x - x')
         \right) \varphi_{\sigma^2}(x')
         \diff x \\
         &\qquad=
         \beta \int_{\bbR}
         \Ind{[-\alpha,\alpha]}(x) \left(
            \varphi_{2\sigma^2}(\beta - x)
            + \varphi_{2\sigma^2}(-\beta - x)
         \right)
         \diff x \\
         &= \beta \left(
            \Ind{[-\alpha,\alpha]} \ast \varphi_{2\sigma^2}(\beta)
            +\Ind{[-\alpha,\alpha]} \ast \varphi_{2\sigma^2}(-\beta)
         \right).
      \end{align*}
      Thus
      \begin{equation*}
         \ip{f_2^\alpha}{f_1^\beta}
         =
         2\sigma^2\left(
            \varphi_{2\sigma^2}(\alpha + \beta)
            -
            \varphi_{2\sigma^2}(\alpha - \beta)
         \right)
         +
         2\beta %
         \Ind{[-\alpha,\alpha]} \ast \varphi_{2\sigma^2}(\beta)
         .
      \end{equation*}
      The remaining calculations proceed along similar lines. We have by
      symmetry
      \begin{align*}
         \ip{f_1^\alpha}{f_1^\beta}
         &=
         2\ip{\varphi_{\sigma^2}(\spcdot + \alpha)}{\varphi_{\sigma^2}(\spcdot +
         \beta)}_{L^2}
         -
         2\ip{\varphi_{\sigma^2}(\spcdot + \alpha)}{\varphi_{\sigma^2}(\spcdot -
         \beta)}_{L^2}
         \\
         &=
         2\left(
            \varphi_{2\sigma^2}(\beta - \alpha)
            -
            \varphi_{2\sigma^2}(\beta + \alpha)
         \right),
         \labelthis \label{eq:f1f1-expr}
      \end{align*}
      because the integrals are gaussian convolutions. Notice that this implies
      \begin{equation}
         \ip{f_2^\alpha}{f_1^\beta}
         =
         2\beta
         (\Ind{[-\alpha,\alpha]} \ast \varphi_{2\sigma^2})
         (\beta)
         -
         \sigma^2 \ip{f_1^\alpha}{f_1^\beta}.
         \label{eq:f1f2-expr}
      \end{equation}
      For the remaining integral, we start with
      \begin{align*}
         \ip{f_2^\alpha}{f_2^\beta}
         &=
         \ip*{
            (\spcdot)\int_{\bbR} \Ind{[-\alpha, \alpha]}(x')
            \varphi_{\sigma^2}(\spcdot - x') \diff x'
         }{
            (\spcdot)\int_{\bbR} \Ind{[-\beta, \beta]}(x')
            \varphi_{\sigma^2}(\spcdot - x') \diff x'
         },
      \end{align*}
      which motivates us to consider
      \begin{align*}
         \int_{\bbR}x^2 \varphi_{\sigma^2}(x - x') \varphi_{\sigma^2}(x - x'') \diff x
         &=
         \varphi_{2\sigma^2}(x' - x'')
         \int_{\bbR}x^2 \varphi_{\sigma^2/2}\left(x - \tfrac{x' + x''}{2}\right) \diff x
         \\
         &=
         \varphi_{2\sigma^2}(x' - x'')
         \int_{\bbR}\left(x + \tfrac{x' + x''}{2}\right)^2
         \varphi_{\sigma^2/2}(x) \diff x \\
         &=
         \varphi_{2\sigma^2}(x' - x'')
         \left(
            \frac{\sigma^2}{2}
            +
            \frac{(x'+x'')^2}{4}
         \right),
      \end{align*}
      where the first line follows by completing the square. In particular,
      this shows that
      \begin{equation}
         \ip{f_2^\alpha}{f_2^\beta}
         =
         \int_{\bbR}\int_{\bbR}
         \Ind{[-\alpha, \alpha]}(x') \Ind{[-\beta, \beta]}(x'')
         \varphi_{2\sigma^2}(x' - x'')
         \left(
            \frac{\sigma^2}{2}
            +
            \frac{(x'+x'')^2}{4}
         \right)
         \diff x' \diff x''.
         \label{eq:f2f2-expr}
      \end{equation}

      We turn to using these calculations to obtain the remaining estimates.
      From \Cref{eq:f2f2-expr}, we have (because all terms in the integral are
      nonnegative)
      \begin{align*}
         \ip{f_2^\alpha}{f_2^\beta}
         &\geq
         \frac{\sigma^2}{2}\int_{\bbR}\int_{\bbR}
         \Ind{[-\alpha, \alpha]}(x') \Ind{[-\beta, \beta]}(x'')
         \varphi_{2\sigma^2}(x' - x'')
         \diff x' \diff x'' \\
         &=
         \frac{\varphi_{2\sigma^2}(0) \sigma^2}{2}\int_{\bbR}\int_{\bbR}
         \Ind{[-\alpha, \alpha]}(x') \Ind{[-\beta, \beta]}(x'')
         \\
         &\qquad-
         \frac{\sigma^2}{2}\int_{\bbR}\int_{\bbR}
         \Ind{[-\alpha, \alpha]}(x') \Ind{[-\beta, \beta]}(x'')
         \left(
            \varphi_{2\sigma^2}(0)
            -
            \varphi_{2\sigma^2}(x' - x'')
         \right)
         \diff x' \diff x'' \\
         &\geq
         2\alpha\beta \sigma^2\varphi_{2\sigma^2}(0) 
         -
         \frac{\varphi_{2\sigma^2}(0)}{8}\int_{\bbR}\int_{\bbR}
         \Ind{[-\alpha, \alpha]}(x') \Ind{[-\beta, \beta]}(x'') (x' - x'')^2
         \diff x' \diff x'' \\
         &=
         2\alpha\beta \sigma^2\varphi_{2\sigma^2}(0) 
         -
         \frac{\varphi_{2\sigma^2}(0)}{8}
         \int_{-\alpha}^{\alpha} \int_{-\beta}^{\beta}
          (x' - x'')^2
         \diff x' \diff x'' \\
         &=
         2\alpha\beta \sigma^2\varphi_{2\sigma^2}(0) 
         -
         \frac{\varphi_{2\sigma^2}(0)}{6}
         \left(
            \alpha^3 \beta + \beta^3 \alpha
         \right),
         \labelthis \label{eq:f2f2-lb}
      \end{align*}
      where the second line applies the triangle inequality, and the third uses
      the inequality $1 - e^{-x} \leq x$.
      From \Cref{eq:f1f2-expr,eq:f1f1-expr}, we require upper and lower bounds
      on \Cref{eq:f1f1-expr}. We have
      \begin{align*}
         \ip{f_1^\alpha}{f_1^\beta}
         &=
         2\varphi_{2\sigma^2}(0) 
         e^{-\frac{1}{4\sigma^2}(\beta - \alpha)^2}
         \left(
            1 - 
            e^{-\frac{1}{4\sigma^2}((\beta+\alpha)^2 - (\alpha-\beta)^2)}
         \right).
      \end{align*}
      Upper bounds from here are straightforward, using that $e^{-x} \leq 1$
      for $x \geq 0$ and $1 - e^{-x} \leq x$. We get
      \begin{equation}
         \ip{f_1^\alpha}{f_1^\beta}
         \leq
         \frac{2\alpha \beta \varphi_{2\sigma^2}(0) }{\sigma^2} .
         \label{eq:f1f1-ub}
      \end{equation}
      Lower bounds can be obtained similarly: by the mean value theorem, there
      is a $\xi \in (1/(4\sigma^2))[(\beta-\alpha)^2, (\beta + \alpha)^2]$ such
      that
      \begin{align*}
         \left(
            e^{-\frac{(\beta - \alpha)^2}{4\sigma^2}}
            - e^{-\frac{(\alpha + \beta)^2}{4\sigma^2}}
         \right)
         &=e^{-\xi}
         \left(
            \frac{(\alpha + \beta)^2}{4\sigma^2}
            -
            \frac{(\beta - \alpha)^2}{4\sigma^2}
         \right)
         \\
         &=e^{-\xi}
         \frac{\alpha\beta}{\sigma^2}.
      \end{align*}
      Using the lower bound on $\xi$ and the fact that $e^{-x} \geq 1-x$
      gives the lower bound
      \begin{equation}
         \ip{f_1^\alpha}{f_1^\beta}
         \geq
         \frac{2\alpha \beta \varphi_{2\sigma^2}(0) }{\sigma^2}
         \left(
            1 - \frac{(\beta - \alpha)^2}{4\sigma^2}
         \right).
         \label{eq:f1f1-lb}
      \end{equation}
      It remains to estimate the remaining term in \Cref{eq:f1f2-expr}. We
      write
      \begin{align*}
         2\beta
         (\Ind{[-\alpha,\alpha]} \ast \varphi_{2\sigma^2})
         (\beta)
         &=
         4\alpha \beta
         \int_{\bbR}
         \frac{1}{2\alpha} \Ind{[-\alpha, \alpha]}(x')
         \varphi_{2\sigma^2}(\beta - x')
         \diff x' \\
         &\leq
         4\alpha \beta \varphi_{2\sigma^2}(0)
         \int_{\bbR}
         \frac{1}{2\alpha} \Ind{[-\alpha, \alpha]}(x')
         \left(
            1 - \frac{(\beta - x')^2}{4\sigma^2}
            + \frac{(\beta - x')^4}{32\sigma^4}
         \right)
         \diff x' \\
         &=
         4\alpha \beta \varphi_{2\sigma^2}(0)
         \left(
            1 
            - \frac{1}{4\sigma^2}\left( \frac{\alpha^2}{3} + \beta^2 \right)
            + \frac{1}{32\sigma^4}\left( \frac{\alpha^4}{5} + 2\alpha^2\beta^2
            + \beta^4 \right)
         \right)
         \labelthis \label{eq:f1f2-lastterm-est}
      \end{align*}
      using again $e^{-x} \leq 1 - x + \half x^2$ in the second line.
      Plugging \Cref{eq:f1f1-ub,eq:f1f1-lb,eq:f1f2-lastterm-est,eq:f2f2-lb}
      into \Cref{eq:nu-second-deriv-to-est}, we have the estimate
      \begin{align*}
         \frac{1}{2}\Xi(\alpha, \beta)
         \geq
         &4\alpha^2 \beta^2 \left(\varphi_{2\sigma^2}(0)\right)^2
         \Biggl(
            \left(
               1 - \frac{(\beta - \alpha)^2}{4\sigma^2}
            \right)
            \left(
               1
               -
               \frac{1}{12\sigma^2}
               \left(
                  \alpha^2 + \beta^2
               \right)
            \right)
            \\
            &\qquad-\Biggl[
               \left(
                  1
                  -
                  \frac{ 1 }{2\sigma^2}
                  \left(
                     \frac{\alpha^2}{3} + \beta^2
                  \right)
                  + \frac{1}{16\sigma^4}\left( \frac{\alpha^4}{5} + 2\alpha^2\beta^2
                  + \beta^4 \right)
               \right)
               \\
               &\qquad\hphantom{-}\times
               \left(
                  1
                  -
                  \frac{ 1 }{2\sigma^2}
                  \left(
                     \frac{\beta^2}{3} + \alpha^2
                  \right)
                  + \frac{1}{16\sigma^4}\left( \frac{\beta^4}{5} + 2\alpha^2\beta^2
                  + \alpha^4 \right)
               \right)
            \Biggr]
         \Biggr),
      \end{align*}
      where plugging in in this manner is justified by the fact that both
      factors in the product to the right of the minus sign are positive as
      long as $\sigma^2 \geq \alpha^2/6 + \beta^2/2$. Specializing to our
      setting of interest where $\alpha \leq 1$ and $\beta = 1$ and
      collecting terms makes this bound become (after simplifying constants
      numerically)
      \begin{equation*}
         \frac{1}{2}\Xi(\alpha, 1)
         \geq
         \frac{1}{2\pi \sigma^4}
         \left(
            {2/3 + \alpha/2}
            -
            \frac{0.845}{\sigma^2}
            -
            \frac{0.04}{\sigma^6}
         \right),
      \end{equation*}
      and the requirement is $\sigma^2 \geq 2/3$. Choosing $\sigma \geq 1$ and
      $\alpha \geq 1/\sqrt{2}$,
      the term in parentheses is no smaller than $1/8$, which gives the lower
      bound
      \begin{equation*}
         \Xi(\alpha, 1)
         \geq
         \frac{1}{8\pi \sigma^4}.
      \end{equation*}

      The remaining upper bounds can be obtained easily from our work above.
      Notice that
      \begin{align*}
         \frac{1}{2} \Xi(\alpha, \alpha)
         &=
         \norm{f_1^\alpha}_{L^2}^2 
         \norm{f_2^\alpha}_{L^2}^2 
         -
         \ip{f_1^\alpha}{f_2^\alpha}^2
         \\
         &\leq
         \norm{f_1^\alpha}_{L^2}^2 
         \norm{f_2^\alpha}_{L^2}^2 .
      \end{align*}
      \Cref{eq:f1f1-ub} gives a suitable upper bound on the first term; we only
      need to develop an upper bound on the second term. From
      \Cref{eq:f2f2-expr}, we proceed as
      \begin{align*}
         \ip{f_2^\alpha}{f_2^\alpha}
         &=
         \int_{\bbR}\int_{\bbR}
         \Ind{[-\alpha, \alpha]}(x') \Ind{[-\alpha, \alpha]}(x'')
         \varphi_{2\sigma^2}(x' - x'')
         \left(
            \frac{\sigma^2}{2}
            +
            \frac{(x'+x'')^2}{4}
         \right)
         \diff x' \diff x''
         \\
         &\leq
         \int_{\set{s^2 + t^2 \leq 2\alpha^2}}
         \varphi_{2\sigma^2}(s - t)
         \left(
            \frac{\sigma^2}{2}
            +
            \frac{(s+t)^2}{4}
         \right)
         \diff x' \diff x'' \\
         &\leq
         \int_{\set{s^2 + t^2 \leq 2\alpha^2}}
         \varphi_{2\sigma^2}(s - t)
         \left(
            \frac{\sigma^2}{2}
            +
            \alpha^2
         \right)
         \diff x' \diff x'' \\
         &\leq
         \left(
            \frac{\sigma^2}{2}
            +
            \alpha^2
         \right)
         \int_{\set{s^2 + t^2 \leq 2\alpha^2}}
         \varphi_{2\sigma^2}(\sqrt{2}s)
         \diff x' \diff x'' \\
         &\leq 
         \varphi_{2\sigma^2}(0)
         \left(
            \frac{\sigma^2}{2}
            +
            \alpha^2
         \right)
         2\pi \alpha^2,
      \end{align*}
      where we pass to an enclosing circular domain in the second line by the
      fact that the integrand is nonnegative, use Cauchy-Schwarz in the third
      line and replace $(s + t)^2 \leq 2s^2 + 2t^2$ by its maximum over the
      domain of integration, apply an orthogonal change of coordinates in the
      fourth line, and use H\"{o}lder's inequality for the fifth line.
      Thus, invoking also \Cref{eq:f1f1-ub}, we have
      \begin{equation*}
         \Xi(\alpha, \alpha)
         \leq
         \frac{\alpha^4}{\sigma^4}
         \left(
            \sigma^2
            +
            2\alpha^2
         \right).
      \end{equation*}

   \end{proof}
\end{lemma}

\begin{lemma}
   \label{lem:curl-field-deriv-estimates}

   Let $u = \Ind{[-\alpha, \alpha]}$ for some $\alpha > 0$, and for some
   smoothing level $\sigma > 0$ consider the associated curl field
   \begin{equation*}
      \sC(\vx)
      =
      \ip*{
         \nabla_{\vx}[\varphi_{\sigma^2} \ast uu\adj ](\vx)
      }{
         \begin{bmatrix}
            0 & -1 \\
            1 & 0
         \end{bmatrix}
         \vx
      }_{\ell^2}.
   \end{equation*}
   Writing $\vx = (s, t)$, and defining
   \begin{equation*}
      f(s) = \int_{-\alpha}^{\alpha} \varphi_{\sigma^2}(s - x) \diff x,
   \end{equation*}
   as in the proof of \Cref{lem:curl-field-at-opt-estimates},
   we have the explicit expression
   \begin{equation}
      \nabla_{\vx} \sC(\vx)
      =
      \begin{bmatrix}
         s f'(s) f'(t) + f(s) f'(t) - t f(t) f''(s) \\
         s f(s) f''(t) - f(t) f'(s) - t f'(s) f'(t)
      \end{bmatrix},
      \label{eq:curl-field-gradient}
   \end{equation}
   and the `iterated curl field' satisfies the estimate
   \begin{equation*}
      \int_{\bbR^2}
      \left(
         \ip*{
            \nabla_{\vx} \sC(s, t)
         }{
            \begin{bmatrix}
               -t \\
               s
            \end{bmatrix}
         }_{\ell^2}
      \right)^2
      \diff s \diff t
      \leq
      \frac{28 \alpha^4}{\pi \sigma^2}
      + \frac{3\alpha^6(20\sigma^2 + 4\alpha^2)}{10\pi \sigma^6 },
   \end{equation*}
   and if $\alpha^2 \leq 1$, it also satisfies the estimate (which is better
   when $\sigma$ is small)
   \begin{equation*}
      \int_{\bbR^2}
      \left(
         \ip*{
            \nabla_{\vx} \sC(s, t)
         }{
            \begin{bmatrix}
               -t \\
               s
            \end{bmatrix}
         }_{\ell^2}
      \right)^2
      \diff s \diff t
      \leq
      \frac{3}{\pi} + 
      \frac{55}{\pi \sigma^2} 
      + \frac{4}{5 \pi \sigma^4}.
   \end{equation*}
   Above, we use $\varphi_{\sigma^2}$ interchangeably for a one-dimensional
   gaussian function and a two-dimensional gaussian function, with the meaning
   clear from the dimensionality of its argument.

   \begin{proof}
      Following the proof of \Cref{lem:curl-field-at-opt-estimates}, we have
      \begin{equation*}
         \sC(s, t) = s f(s) f'(t) - t f(t) f'(s),
      \end{equation*}
      where 
      by the fundamental theorem of calculus,
      \begin{equation*}
         f'(x) 
         = \varphi_{\sigma^2}(x + \alpha) - \varphi_{\sigma^2}(x - \alpha).
      \end{equation*}
      It is straightforward to calculate \Cref{eq:curl-field-gradient} from
      this expression. We have
      \begin{equation*}
         \ip*{
            \nabla_{\vx} \sC(s, t)
         }{
            \begin{bmatrix}
               -t \\
               s
            \end{bmatrix}
         }_{\ell^2}
         =
         f(t) \left(
            t^2 f''(s) - s f'(s)
         \right)
         + f(s) \left(
            s^2 f''(t) - t f'(t)
         \right)
         - 2st f'(s) f'(t).
      \end{equation*}
      We square and integrate this expression in order to take care of the
      permutation symmetry. Using computer algebra software, one obtains
      \begin{align*}
         \int_{\bbR^2}
         \left(
            \ip*{
               \nabla_{\vx} \sC(s, t)
            }{
               \begin{bmatrix}
                  -t \\
                  s
               \end{bmatrix}
            }_{\ell^2}
         \right)^2
         \diff s \diff t
         &=
         2 \norm{p_x f'}_{L^2}^2 \norm{f}_{L^2}^2
         + 8 \norm{p_x f'}_{L^2}^2 \int_{\bbR} p_x f f'
         + 2 \norm{f''}_{L^2}^2 \norm{p_{x^2} f}_{L^2}^2 \\
         &\hphantom{=}
         - 4 \left( \int_{\bbR} f f'' \right) \left( \int_{\bbR} p_{x^3} f f' \right)
         - 8 \left( \int_{\bbR} p_x f' f'' \right)\left( \int_{\bbR} p_{x^3} f
         f' \right) \\
         &\hphantom{=}
         - 4 \norm{p_x f}_{L^2}^2 \int_{\bbR} p_x f' f''
         + 2 \left( \int_{\bbR} p_x f f' \right)^2
         + 4 \norm{p_x f'}_{L^2}^4
         + 2 \left( \int_{\bbR} p_{x^2} f f'' \right)^2.
      \end{align*}
      In this expression, if $x \mapsto g(x)$ is a polynomial in $x$ we write $p_{g(x)}$
      to denote the function $x \mapsto g(x)$. We can simplify further using
      integration by parts. It is clear that $f$ vanishes at infinity faster
      than any polynomial, and the expression for $f'$ as a difference of
      gaussians shows this is also true of every derivative of $f$. Thus, we
      find straightforwardly
      \begin{align*}
         \int_{\bbR} p_{x^3} f f'
         &= -\frac{3}{2} \norm{p_x f}_{L^2}^2, \\
         \int_{\bbR} p_{x} f' f''
         &= -\frac{1}{2} \norm{f'}_{L^2}^2, \\
         \int_{\bbR} f f''
         &= -\norm{f'}_{L^2}^2, \\
         \int_{\bbR} p_{x} f f'
         &= -\frac{1}{2}\norm{f}_{L^2}^2, \\
         \int_{\bbR} p_{x^2} f f''
         &= \norm{f}_{L^2}^2 - \norm{p_x f'}_{L^2}^2.
      \end{align*}
      Applying these identities, we simplify the previous expression to
      \begin{align*}
         \int_{\bbR^2}
         \left(
            \ip*{
               \nabla_{\vx} \sC(s, t)
            }{
               \begin{bmatrix}
                  -t \\
                  s
               \end{bmatrix}
            }_{\ell^2}
         \right)^2
         \diff s \diff t
         &=
         6\norm{p_{x} f'}_{L^2}^2 \left(
            \norm{p_{x}f'}_{L^2}^2 - \norm{f}_{L^2}^2
         \right)
         + \frac{5}{2} \norm{f}_{L^2}^4 \\
         &\hphantom{=}
         + 2 \norm{f''}_{L^2}^2 \norm{p_{x^2} f}_{L^2}^2
         - 10 \norm{f'}_{L^2}^2 \norm{p_x f}_{L^2}^2 \\
         &\leq
         6\norm{p_{x} f'}_{L^2}^4
         + \frac{5}{2} \norm{f}_{L^2}^4
         + 2 \norm{f''}_{L^2}^2 \norm{p_{x^2} f}_{L^2}^2
      \end{align*}
      We can estimate the integrals involving $f$ using Jensen's inequality.
      In particular, notice that
      \begin{align*}
         \left( f(s) \right)^2
         &= 
         \left(
            \int_{-\alpha}^\alpha \varphi_{\sigma^2}(s - x) \diff x
         \right)^2
         \\
         &=
         (2\alpha)^2
         \left(
            \frac{1}{2\alpha}
            \int_{-\alpha}^\alpha \varphi_{\sigma^2}(s - x) \diff x
         \right)^2 \\
         &\leq
         2\alpha
         \int_{-\alpha}^\alpha \varphi_{\sigma^2}(s - x)^2 \diff x,
      \end{align*}
      by Jensen's inequality for the convex function $x \mapsto x^2$.
      Since
      \begin{equation*}
         \varphi_{\sigma^2}(s - x)^2 
         =
         \frac{1}{2\sqrt{\pi \sigma^2}} \varphi_{\sigma^2/2}(s-x),
      \end{equation*}
      we obtain
      \begin{equation*}
         \left( f(s) \right)^2
         \leq
         \frac{\alpha}{\sigma\sqrt{\pi}}
         \int_{-\alpha}^\alpha \varphi_{\sigma^2/2}(s - x) \diff x,
      \end{equation*}
      which is a scaled version of $f$ with the variance of the gaussian
      smoothing halved. From here, it follows by Fubini's theorem and
      standard (non-centered) gaussian moment calculations
      \begin{align*}
         \norm{f}_{L^2}^2 
         &\leq
         \frac{2\alpha^2}{\sigma\sqrt{\pi}}; \\
         \norm{p_{x^2} f}_{L^2}^2
         &\leq
         \frac{\alpha}{\sigma \sqrt{\pi}} \int_{-\alpha}^\alpha 
         \int_{\bbR} s^4 \varphi_{\sigma^2 / 2}(s - x) \diff s \diff x \\
         &=
         \frac{\alpha}{4\sigma \sqrt{\pi}} \int_{-\alpha}^\alpha 
         \left(
            3\sigma^4 + 12 \sigma^2 x^2 + 4x^4
         \right) \diff x \\
         &=
         \frac{\alpha^2
            \left(
               15\sigma^4 + 20 \sigma^2 \alpha^2 + 4\alpha^4
            \right)
         }{10\sigma \sqrt{\pi}}.
      \end{align*}
      The remaining terms are gaussian integrals, and can be calculated easily.
      We evaluate
      \begin{align*}
         \norm{p_{x} f'}_{L^2}^2
         &=
         \frac{1}{2\sigma \sqrt{\pi}} \left(
            2\alpha^2 + \sigma^2\left( 
               1 - e^{-\tfrac{\alpha^2}{\sigma^2}}
            \right)
         \right); \\
         \norm{f''}_{L^2}^2
         &=
         \frac{1}{2\sigma^5 \sqrt{\pi}} \left(
            2\alpha^2 e^{-\tfrac{\alpha^2}{\sigma^2}} 
            + \sigma^2\left( 
               1 - e^{-\tfrac{\alpha^2}{\sigma^2}}
            \right)
         \right).
      \end{align*}
      We can simplify
      these expressions further: applying the inequality $1 - x
      \leq e^{-x}$ %
      gives 
      \begin{align*}
         \norm{p_{x} f'}_{L^2}^2
         &\leq
         \frac{3\alpha^2}{2\sigma \sqrt{\pi}}; \\
         \norm{f''}_{L^2}^2
         &\leq
         \frac{3\alpha^2}{2\sigma^5 \sqrt{\pi}} .
      \end{align*}
      Combining, we thus get
      \begin{align*}
         \int_{\bbR^2}
         \left(
            \ip*{
               \nabla_{\vx} \sC(s, t)
            }{
               \begin{bmatrix}
                  -t \\
                  s
               \end{bmatrix}
            }_{\ell^2}
         \right)^2
         \diff s \diff t
         &\leq
         \frac{28 \alpha^4}{\pi \sigma^2}
         + \frac{3\alpha^6(20\sigma^2 + 4\alpha^2)}{10\pi \sigma^6 }.
      \end{align*}
      We can obtain improved estimates when $\sigma$ is small:
      writing
      \begin{align*}
         \norm{f''}_{L^2}^2
         &=
         \frac{1}{2\sigma^3 \sqrt{\pi}} \left(
            \frac{2\alpha^2}{\sigma^2} e^{-\tfrac{\alpha^2}{\sigma^2}} 
            + 1 - e^{-\tfrac{\alpha^2}{\sigma^2}}
         \right),
      \end{align*}
      evidently
      \begin{equation*}
         \frac{2\alpha^2}{\sigma^2} e^{-\tfrac{\alpha^2}{\sigma^2}} 
         + 1 - e^{-\tfrac{\alpha^2}{\sigma^2}}
         \leq
         1 + \frac{2}{e},
      \end{equation*}
      and the RHS is no larger than $2$; hence
      \begin{equation*}
         \norm{f''}_{L^2}^2
         \leq
         \frac{1}{\sigma^3 \sqrt{\pi}}.
      \end{equation*}
      Combining in this case gives the estimate (together with $\alpha^2 \leq
      1$)
      \begin{equation*}
         \int_{\bbR^2}
         \left(
            \ip*{
               \nabla_{\vx} \sC(s, t)
            }{
               \begin{bmatrix}
                  -t \\
                  s
               \end{bmatrix}
            }_{\ell^2}
         \right)^2
         \diff s \diff t
         \leq
         \frac{3}{\pi} + 
         \frac{55}{\pi \sigma^2} 
         + \frac{4}{5 \pi \sigma^4}.
      \end{equation*}

   \end{proof}
\end{lemma}

\subsection{Auxiliary Results}
\label{sec:proofs_aux}

\begin{lemma}
   \label{lem:lipschitz_grid_1d}
   Let $f : [-1, +1] \to \bbR$ be a $L$-Lipschitz function, and let $\pi_1(G)$
   be the projection of the rectangular grid $G$ onto its first coordinate. One
   has
   \begin{equation*}
      \abs*{
         \frac{2}{n}\sum_{i \in \pi_1(G)} f(i)
         -
         \int_{[-1, 1]} f(t) \diff t
      }
      \leq \frac{2L}{n}.
   \end{equation*}

   \begin{proof}
      Define
      \begin{equation*}
         \delta_i = -1 + i \frac{2}{n-1}, \quad i = 0, 1, \dots, n-1,
      \end{equation*}
      so that
      \begin{equation*}
         \int_{[-1, 1]} f(t) \diff t
         = \int_{\delta_0}^{\delta_{0} + \tfrac{2}{n}} f(t) \diff t
         + \int_{\delta_{n-1} - \tfrac{2}{n}}^{\delta_{n-1}} f(t) \diff t
         + \sum_{i=1}^{n-2} \int_{\delta_i - \tfrac{1}{n}}^{\delta_{i} +
         \tfrac{1}{n}} f(t) \diff t.
      \end{equation*}
      This is a `midpoint' estimate of the integral, given the boundary. Since
      \begin{equation*}
         \sum_{i \in \pi_1(G)} f(i)
         =
         \sum_{i=0}^{n-1} f(\delta_i),
      \end{equation*}
      we obtain from the triangle inequality and the Lipschitz property of $f$
      \begin{align*}
         \abs*{
            \frac{2}{n}\sum_{i \in \pi_1(G)} f(i)
            -
            \int_{-1}^{1} f(t) \diff t
         }
         &\leq
         \int_{\delta_0}^{\delta_{0}+\tfrac{2}{n}} 
         \abs*{
            f(\delta_0) - f(t)
         }
         \diff t
         +
         \int_{\delta_{n-1}-\tfrac{2}{n}}^{\delta_{n-1}}
         \abs*{
            f(\delta_{n-1}) - f(t)
         }
         \diff t
         + \sum_{i=1}^{n-2} 
         \int_{\delta_i-\tfrac{1}{n}}^{\delta_{i}+\tfrac{1}{n}} 
         \abs*{
            f(\delta_i) - f(t)
         }
         \diff t \\
         &\leq
         L 
         \left(
            \int_{\delta_0}^{\delta_{0}+\tfrac{2}{n}} 
            (t - \delta_0) \diff t
            +
            \int_{\delta_{n-1}-\tfrac{2}{n}}^{\delta_{n-1}}
            (\delta_{n-1} - t) \diff t
            + \sum_{i=1}^{n-2}
            \int_{\delta_i-\tfrac{1}{n}}^{\delta_{i}+\tfrac{1}{n}} \abs{t - \delta_i}
            \diff t
         \right)  \\
         &= L\left(
            \frac{4}{n^2}
            + \sum_{i=1}^{n-2} \frac{1}{n^2}
         \right)\\
         &\leq \frac{2L}{n},
      \end{align*}
      where the last estimate holds if $n \geq 2$.
      
   \end{proof}
\end{lemma}

\begin{lemma}
   \label{lem:matrix-am-gm}

   Let $\vU, \vV \in \bbR^{m \times n}$, and let $\vD \in \bbR^{n \times n}$ be
   a diagonal matrix.  Let $\nnorm{}$ be any unitarily invariant matrix norm.
   Then one has
   \begin{equation*}
      \nnorm{\vU \vD \vV\adj}
      \leq
      \frac{1}{2} \left(
         \nnorm{\abs{\vD}^{1/2} \vU\adj \vU\abs{\vD}^{1/2}}
         + \nnorm{\abs{\vD}^{1/2} \vV\adj \vV\abs{\vD}^{1/2}}
      \right),
   \end{equation*}
   where $\abs{\vA} =(\vA\adj\vA)^{1/2} $ denotes the positive part 
   of a matrix, and the matrix norms in this expression are to be interpreted
   in terms of the `dilation norm' of the larger size matrix
   norm.\footnote{That is, if $n > m$, $\nnorm{}$ is the matrix norm on $n
      \times n$ matrices, and if $\vA \in \bbR^{m \times m}$
      \begin{equation*}
         \nnorm{\vA} = \nnorm{
            \begin{bmatrix}
               \vA & \Zero \\
               \Zero & \Zero
            \end{bmatrix}
         },
      \end{equation*}
      and likewise if $m > n$.  Compare \cite[Exercise IV.2.15]{Bhatia1997-ly}.
   }

   \begin{proof}
      We apply a slight modification of a matrix arithmetic-geometric mean
      inequality. There exists a diagonal matrix $\vS \in \bbR^{n \times n}$
      with diagonal entries either $1$ or $\sqrt{-1}$ such that $\vS\adj \vD
      \vS\adj =
      \abs{\vD}$. Then $\vS\adj\vS = \vI$, so $\vS$ is unitary, and by
      \cite[Corollary IX.4.4]{Bhatia1997-ly},
      \begin{align*}
         \nnorm{\vU \vD \vV\adj}
         =\nnorm{\vU \vS \abs{\vD} \vS \vV\adj} 
         &= \nnorm{(\vU \vS \abs{\vD}^{1/2}) (\vV \vS\adj \abs{\vD}^{1/2})\adj}
         \\
         &\leq
         \frac{1}{2} \left(
            \nnorm{
               \abs{\vD}^{1/2} \vS\adj \vU\adj \vU \vS \abs{\vD}^{1/2}
               + \abs{\vD}^{1/2} \vS \vV\adj \vV \vS\adj \abs{\vD}^{1/2}
            }
         \right).
      \end{align*}
      Now apply the triangle inequality and use the fact that diagonal matrices
      commute and that $\nnorm{}$ is unitarily invariant to establish the
      claim.
   \end{proof}
\end{lemma}

\begin{lemma}
   \label{lem:l2-sampling}

   For $\sigma^2 > 0$, let $\varphi_{\sigma^2}(t) = 1/\sqrt{2\pi\sigma^2}
   \exp(-\tfrac{1}{2 \sigma^2}t^2)$ denote the one-dimensional standard
   gaussian, and let $m_{\sigma^2} = \varphi_{\sigma^2}^{\otimes 2}$.
   Let $f, g \in L^1(\bbR^2) \cap L^2(\bbR^2)$, 
   and let $\bar{G}$ denote the infinite extension of the image sampling grid
   $G$ defined in \Cref{eq:image_ndc}:
   \begin{equation*}
      \bar{G} = \set*{ \left(1 + \frac{2k}{n-1}, 1 + \frac{2l}{n-1}\right)
      \given (k, l) \in \bbZ^2}
   \end{equation*}
   (notice that $G \subset \bar{G}$). Let $\ell^2(\bar{G})$ denote the space of
   square-summable sequences defined on $\bar{G}$.
   Then it holds
   \begin{equation*}
      \abs*{ 
         \left(\frac{n-1}{2}\right)^2
         \ip{m_{\sigma^2} \ast f}{m_{\sigma^2} \ast g}_{L^2(\bbR^2)}
         - 
         \ip{m_{\sigma^2} \ast f}{m_{\sigma^2} \ast g}_{\ell^2(\bar{G})}
      } 
      \leq 
      \frac{\norm{f}_{L^1}\norm{g}_{L^1}}{(2\pi \sigma^2)^2}
      \left(
         1 + \frac{(n-1)\sigma}{\sqrt{2}}
      \right).
   \end{equation*}
   \begin{proof}
      We will rely on machinery from the theory of tempered distributions
      throughout the proof, following notation and results contained in
      \cite[Ch.\ I, \S 3]{Stein1971-ed}. Let $\cS \subset L^2(\bbR^2)$ denote
      the class of real-valued Schwartz functions (a dense subset of $L^2(\bbR^2)$). 
      For concision, write $a = m_{\sigma^2}\ast f$ and $b = m_{\sigma^2} \ast
      g$.
      Then because $f, g$ are in $L^1$, $a, b$ are in $\cS$. 
      Let $\delta_{\vx}$ denote the ``Dirac distribution'' at $\vx \in
      \bbR^2$, the tempered distribution defined by $\delta_{\vx}(h) =
      h(\vx)$ for every $h \in \cS$. 
      Let $\Delta_n$ denote the ``Dirac comb'' for the grid $\bar{G}$, the
      tempered distribution defined by
      \begin{equation*}
         \Delta_n = \sum_{(i, j) \in \bar{G}} \delta_{(i, j)}.
      \end{equation*}
      Notice that when $n$ is odd, we have $\bar{G} = (2/(n-1)) \bbZ^2$, and when
      $n$ is even we have $\bar{G} = 1/(n-1) + (2/(n-1)) \bbZ^2$.
      Then from the definition of the product, convolution, and Fourier
      transform of tempered distributions, we have
      \begin{align*}
         \ip{a}{b}_{\ell^2(\bar{G})}
         &= \Delta_n (ab) \\
         &= (\Delta_n a)(b) \\
         &= (\hat{\Delta}_n \ast \hat{a})\hat{\hphantom{\enspace}}(b)
         \labelthis \label{eq:tempered-convolution} \\
         &= (\hat{\Delta}_n \ast \hat{a})(\hat{b}) \\
         &= \hat{\Delta}_n(\tilde{\hat{a}} \ast \hat{b}),
         \labelthis \label{eq:tempered-parseval}
      \end{align*}
      where $\ast$ additionally denotes convolution of a tempered distribution
      with a Schwartz function, %
      for a Schwartz function or
      a tempered distribution $\hat{\psi}$ denotes its Fourier transform, and
      for a Schwartz function $\tilde{g}$ denotes its reversal $\tilde{g}(\vx)
      = g(-\vx)$. Above, \Cref{eq:tempered-convolution} applies the convolution
      formula for tempered distributions (c.f.\ \cite[Proof of Ch.\ I, Thm.\
      3.18]{Stein1971-ed}), and the remaining manipulations are unraveling
      definitions. The tempered distribution $\hat{\varphi}$ is defined by the
      relation $\hat{\varphi}(h) = \varphi(\hat{h})$ for all $h \in \cS$; so we
      have for the Dirac comb and for any $h \in \cS$
      \begin{align*}
         \hat{\Delta}_n(h) 
         = 
         \sum_{(i, j) \in \bar{G}} \delta_{(i, j)} (\hat{h})
         &=
         \sum_{(k, l) \in \bbZ^2} \int_{\bbR^2} h(\vx) e^{-\iu 2\pi\left(
               \ip*{ \tfrac{2\vx}{n-1}}{(k, l)}
               + \Ind{\text{n even}}\ip*{ \tfrac{\vx}{n-1}}{(1, 1)}
         \right)}\diff \vx \\
         &=
         \left(\frac{n-1}{2}\right)^2
         \sum_{(k, l) \in \bbZ^2} \int_{\bbR^2} 
         h\left(\frac{n-1}{2}\vx\right) 
         e^{-\iu \pi \ip{\vx}{  \Ind{\text{n even}}(1, 1)}}
         e^{-\iu 2\pi \ip{\vx}{(k, l)}}\diff \vx \\
         &=
         \left(\frac{n-1}{2}\right)^2
         \sum_{(k, l) \in \bbZ^2} 
         \bigl(
            h_{\tfrac{n-1}{2}}
            \cdot
            e^{-\iu \pi \ip{\spcdot}{(1, 1)\Ind{\text{n even}}}}
         \bigr)\hat{\hphantom{\enspace}}
         (k,l),
      \end{align*}
      where in the final line $h_{(n-1)/2}$ denotes the dilation of $h$ (as in
      the previous line). Now, because $h \in \cS$, it holds that
      $\bar{h}(\vx)
      =
      h_{\tfrac{n-1}{2}}(\vx) e^{-\iu \pi \ip{\vx}{(1, 1)}}$
      satisfies $\bar{h} \in \cS$, because the complex exponential function is
      infinitely differentiable with uniformly bounded derivatives on $\bbR^2$.
      We can thus apply the Poisson
      summation formula \cite[Ch.\ VII, Cor.\ 2.6]{Stein1971-ed} to obtain from
      the previous
      \begin{align*}
         \hat{\Delta}_n(h) 
         &= 
         \left(\frac{n-1}{2}\right)^2
         \sum_{(k, l) \in \bbZ^2} 
         \bigl(
            h_{\tfrac{n-1}{2}}
            \cdot
            e^{-\iu \pi \ip{\spcdot}{(1, 1)\Ind{\text{n even}}}}
         \bigr)(k, l) \\
         &=
         \left(\frac{n-1}{2}\right)^2
         \sum_{(k, l) \in \bbZ^2} 
         e^{-\iu \pi k \Ind{\text{n even}}}
         e^{-\iu \pi l \Ind{\text{n even}}}
         h_{\tfrac{n-1}{2}} (k, l).
      \end{align*}
      This shows that $\hat{\Delta}_n$ is equal to a modulated Dirac comb on a
      rescaled grid. 
      Continuing from \Cref{eq:tempered-parseval}, we therefore have
      \begin{align*}
         \ip{a}{b}_{\ell^2(\bar{G})}
         &=
         \left(\frac{n-1}{2} \right)^2 \left[
            \sum_{(k, l) \in \bbZ^2} 
            e^{-\iu \pi k \Ind{\text{n even}}}
            e^{-\iu \pi l \Ind{\text{n even}}}
            \int_{\bbR^2} \hat{a}(\vxi) \hat{b}(\vxi +
            \tfrac{n-1}{2}(k, l)) \diff \vxi
         \right] \\
         &=
         \left(\frac{n-1}{2} \right)^2 \left[
            \int_{\bbR^2} \hat{a}(\vxi) \hat{b}(\vxi) \diff \vxi
            +
            \sum_{\substack{(k, l) \in \bbZ^2 \\ (k,l)\neq \Zero}}
            e^{-\iu \pi k \Ind{\text{n even}}}
            e^{-\iu \pi l \Ind{\text{n even}}}
            \int_{\bbR^2} \hat{a}(\vxi) \hat{b}(\vxi +
            \tfrac{n-1}{2}(k, l)) \diff \vxi
         \right],
      \end{align*}
      where we applied a change of variables to simplify the
      convolution integrals to cross-correlations. Now, by Parseval's theorem
      on Schwartz functions, we have
      \begin{align*}
         \ip{a}{b}_{\ell^2(\bar{G})}
         =
         \left(\frac{n-1}{2} \right)^2
         \ip{a}{b}_{L^2(\bbR^2)}
         +
         \left(\frac{n-1}{2} \right)^2
         \sum_{\substack{(k, l) \in \bbZ^2 \\ (k,l)\neq \Zero}}
         e^{-\iu \pi k \Ind{\text{n even}}}
         e^{-\iu \pi l \Ind{\text{n even}}}
         \int_{\bbR^2} \hat{a}(\vxi) \hat{b}(\vxi +
         \tfrac{n-1}{2}(k, l)) \diff \vxi,
         \labelthis \label{eq:sampling-target-residual}
      \end{align*}
      so our task is to bound the residual in the previous expression.
      We have $\hat{a} = (\varphi_{\sigma^2}^{\otimes
      2})\hat{\hphantom{\enspace}} \hat{f}$ by the convolution formula for $L^2$
      functions (and similarly for $\hat{b}$), and the Fourier transform of a gaussian is another gaussian,
      suitably scaled (\cite[Theorem 1.13]{Stein1971-ed}):
      \begin{equation*}
         (\varphi_{\sigma^2}^{\otimes 2})\hat{\hphantom{\enspace}}(\vxi)
         = e^{-2\pi^2 \sigma^2 \norm{\vxi}_2^2}
         = \frac{1}{2\pi \sigma^2} \varphi_{1/{(2\pi\sigma)^2}}^{\otimes 2}.
      \end{equation*}
      Because $f, g \in L^1(\bbR^2)$, we have $\norm{\hat{f}}_{L^\infty} \leq
      \norm{f}_{L^1}$ and $\norm{\hat{g}}_{L^\infty} \leq
      \norm{g}_{L^1}$. For the residual term in
      \Cref{eq:sampling-target-residual}, we thus have the estimate
      \begin{align*}
         &\left(\frac{n-1}{2} \right)^2
         \abs*{
            \sum_{\substack{(k, l) \in \bbZ^2 \\ (k,l)\neq \Zero}}
            e^{-\iu \pi k \Ind{\text{n even}}}
            e^{-\iu \pi l \Ind{\text{n even}}}
            \int_{\bbR^2} \hat{g}(\vxi) \hat{g}(\vxi +
            \tfrac{n-1}{2}(k, l)) \diff \vxi
         } \\
         &\qquad\leq
         \norm{f}_{L^1} \norm{g}_{L^1}
         \left(\frac{n-1}{4\pi \sigma^2}\right)^2
         \sum_{\substack{(k, l) \in \bbZ^2 \\ (k,l)\neq \Zero}}
         \int_{\bbR^2} \varphi_{1/(2\pi\sigma)^2}^{\otimes 2}(\vxi) 
         \varphi_{1/(2\pi\sigma)^2}^{\otimes 2}(\vxi + \tfrac{n-1}{2}(k, l))
         \diff \vxi,
         \labelthis \label{eq:samplingl2-est-tobound}
      \end{align*}
      where we applied the triangle inequality.
      The integral in the previous
      expression is a convolution integral; as is well-known, the convolution
      of two gaussians is another gaussian, with mean equal to the sum of the
      means of the factors and variance equal to the sum of the variances.
      In particular, we have (using reflection symmetry of the gaussian
      function)
      \begin{equation*}
         \int_{\bbR^2} \varphi_{1/(2\pi\sigma)^2}^{\otimes 2}(\vxi) 
         \varphi_{1/(2\pi\sigma)^2}^{\otimes 2}(\vxi + \tfrac{n-1}{2}(k, l))
         \diff \vxi
         =
         \varphi_{2/(2\pi\sigma)^2}^{\otimes 2}\left(\tfrac{n-1}{2}(k,
         l)\right).
      \end{equation*}
      Because the gaussian function factors across components of its argument,
      we have
      \begin{equation*}
         \sum_{\substack{(k, l) \in \bbZ^2 \\ (k,l)\neq \Zero}}
         \varphi_{2/(2\pi\sigma)^2}^{\otimes 2}\left(\tfrac{n-1}{2}(k,
         l)\right)
         =
         \left(\sum_{k \in \bbZ}
            \varphi_{2/(2\pi\sigma)^2}\left(\tfrac{n-1}{2}k\right)
         \right)^2
         -
         \left(\varphi_{2/(2\pi\sigma)^2}(0)\right)^2.
      \end{equation*}
      Let $\gamma^2 = 2/(2\pi \sigma^2)$. We have
      \begin{align*}
         \varphi_{2/(2\pi\sigma)^2}\left(\tfrac{n-1}{2}k\right)
         &=
         \frac{1}{\sqrt{2\pi \gamma^2}} e^{-\tfrac{1}{2\gamma^2} ((n-1)/2)^2
         k^2} \\
         &=
         \frac{2}{n-1} \frac{1}{\sqrt{2\pi \bar{\gamma}^2}}
         e^{-\tfrac{1}{2\bar{\gamma}^2} k^2} \\
         &= \frac{2}{n-1} \varphi_{\bar{\gamma}^2}(k),
      \end{align*}
      where we have defined $\bar{\gamma}^2 = \gamma^2 / ((n-1)/2)^2$.
      Estimating the sum with the integral test estimate gives
      \begin{align*}
         \sum_{k \in \bbZ} \varphi_{\bar{\gamma}^2}(k)
         &\leq
         2 \left(
            \varphi_{\bar{\gamma}^2}(0)
            + \int_0^\infty \varphi_{\bar{\gamma}^2}(\xi)\diff\xi
         \right)
         -
         \varphi_{\bar{\gamma}^2}(0) \\
         &= 1 + \varphi_{\bar{\gamma}^2}(0),
      \end{align*}
      so in particular
      \begin{align*}
         \left(\sum_{k \in \bbZ}
            \varphi_{2/(2\pi\sigma)^2}\left(\tfrac{n-1}{2}k\right)
         \right)^2
         -
         \left(\varphi_{2/(2\pi\sigma)^2}(0)\right)^2
         &\leq
         \left( \frac{2}{n-1} + \varphi_{\gamma^2}(0) \right)^2
         - 
         \left(\varphi_{\gamma^2}(0)\right)^2 \\
         &=
         \left(\frac{2}{n-1}\right)^2
         + \frac{4\varphi_{\gamma^2}(0)}{n-1}.
      \end{align*}
      With this, \Cref{eq:samplingl2-est-tobound} can be bounded as
      \begin{align*}
         \left(\frac{n-1}{2} \right)^2
         \abs*{
            \sum_{\substack{(k, l) \in \bbZ^2 \\ (k,l)\neq \Zero}}
            e^{-\iu \pi k \Ind{\text{n even}}}
            e^{-\iu \pi l \Ind{\text{n even}}}
            \int_{\bbR^2} \hat{g}(\vxi) \hat{g}(\vxi +
            \tfrac{n-1}{2}(k, l)) \diff \vxi
         }
         &\leq
         \norm{f}_{L^1} \norm{g}_{L^1}
         \left(\frac{1}{2\pi \sigma^2}\right)^2
         \left(
            1 + \frac{(n-1)\sigma}{\sqrt{2}}
         \right),
      \end{align*}
      which implies the claim.

   \end{proof}
\end{lemma}

\begin{lemma}
   \label{lem:mollified-orthogonality}
   Let $f, g \in L^1(\bbR) \cap L^2(\bbR)$,
   and for $\sigma^2 > 0$, let $\varphi_{\sigma^2}(t) = 1/\sqrt{2\pi\sigma^2}
   \exp(-\tfrac{1}{2 \sigma^2}t^2)$ denote the one-dimensional standard
   gaussian. 
   Then one has
   \begin{equation*}
      \abs*{\ip{f}{g}_{L^2} -
      \ip{\varphi_{\sigma^2} \ast f}{\varphi_{\sigma^2} \ast g}_{L^2}}
      \leq 
      \sigma^2
      \norm{f'}_{L^2(\bbR^2)} \norm{g'}_{L^2(\bbR^2)}.
   \end{equation*}

   \begin{proof}
      One calculates with Plancherel's theorem and the convolution theorem for
      the Fourier transform
      \begin{align*}
         \abs*{\ip{f}{g}_{L^2} -
         \ip{\varphi_{\sigma^2} \ast f}{\varphi_{\sigma^2} \ast g}_{L^2}}
         &=
         \abs*{
            \ip{\hat{\varphi}_{\sigma^2}\hat{f}}{\hat{\varphi}_{\sigma^2}\hat{g}}
            -
            \ip{\hat{f}}{\hat{g}}
         } \\
         &= \abs*{
            \ip{(\hat{\varphi}_{\sigma^2})^2 \hat{f}}{\hat g}
            - \ip{\hat{f}}{\hat{g}}
         } \\
         &= \abs*{
            \ip{((\hat{\varphi}_{\sigma^2})^2 - 1) \hat{f}}{\hat g}
         } \\
         &= \abs*{
            \ip{\sqrt{1 - (\hat{\varphi}_{\sigma^2})^2} \hat{f}}
            {\sqrt{1 - (\hat{\varphi}_{\sigma^2})^2} \hat{g}}
         },
      \end{align*}
      where we use that the Fourier transform of a gaussian is another
      gaussian (and in particular, is positive and bounded by $1$):
      \begin{equation*}
         \hat{\varphi}_{\sigma^2}(\xi) = e^{-2\pi^2 \sigma^2 \xi^2}.
      \end{equation*}
      We thus have, by the triangle inequality,
      \begin{align*}
         \abs*{
            \ip{\sqrt{1 - (\hat{\varphi}_{\sigma^2})^2} \hat{f}}
            {\sqrt{1 - (\hat{\varphi}_{\sigma^2})^2} \hat{g}}
         }
         &\leq
         \int_{\bbR^2}
         \abs{\hat{f}\hat{g}}(\xi) (1 - e^{-4\pi^2 \sigma^2 \xi^2})\diff \xi
         \\
         &\leq
         \sigma^2
         \int_{\bbR^2}
      \abs{\iu 2\pi \xi \hat{f}(\xi)}
      \abs{\iu 2\pi \xi \hat{g}(\xi)}\diff \xi
      \end{align*}
      using $1-e^{-x} \leq x$ in the second line. By \cite[Theorem \S I,
      1.8]{Stein1971-ed}, we have $\iu 2\pi \xi \hat{f}(\xi) =
      (f')\hat{\hphantom{\,}}(\xi)$, whence by the Schwarz inequality and 
      Parseval's theorem
      \begin{equation*}
         \sigma^2
         \int_{\bbR^2}
         \abs{\iu 2\pi \xi \hat{f}(\xi)}
         \abs{\iu 2\pi \xi \hat{g}(\xi)}\diff \xi
         \leq
         \sigma^2
         \norm{f'}_{L^2(\bbR^2)} \norm{g'}_{L^2(\bbR^2)}.
      \end{equation*}

   \end{proof}
\end{lemma}

\begin{lemma}
   \label{lem:r1approx-to-power}
   For $X \in L^2(\bbR^2)$, consider the rank-one factorization objective
   \begin{equation*}
      \min_{u \in L^2(\bbR)}\, \frac{1}{2}
      \norm*{ X - uu\adj }_{L^2(\bbR^2)}^2.
   \end{equation*}
   Suppose that there exists a nonzero $v \in L^2(\bbR)$ such that $v\adj
   \sT_{X} v \geq 0$, where
   $\sT_{X} : L^2(\bbR) \to L^2(\bbR)$ denotes the integral operator
   $u \mapsto \int_{\bbR} X(\spcdot, t) u(t) \diff t$ associated to $X$.
   Then this optimization problem is equivalent to the constrained problem
   \begin{equation*}
      \max_{\norm{u}_{L^2(\bbR)} = 1}\,
      u\adj \left( \sT_{X} + \sT_{X}\adj \right) u;
   \end{equation*}
   precisely, if $u$ is an optimal solution to the second problem, then
   $( \half u\adj (\sT_{X} + \sT_{X}\adj) u) uu\adj$ is an optimal solution to the first problem.
   Here, $\sT_{X}\adj$ is the adjoint of $\sT_{X}$.
   \begin{proof}
      The problem
      \begin{equation*}
         \min_{u \in L^2(\bbR)}\, \frac{1}{2}
         \norm*{ X - uu\adj }_{L^2(\bbR^2)}^2.
      \end{equation*}
      is equivalent to the problem
      \begin{equation*}
         \min_{\norm{u}_{L^2(\bbR)} = 1,\, c \geq 0}\,
         \frac{1}{2} \norm*{ X - c uu\adj }_{L^2(\bbR^2)}^2.
      \end{equation*}
      Expanding the square, the objective in this latter problem satisfies
      \begin{equation*}
         \norm*{ X - c uu\adj }_{L^2(\bbR^2)}^2
         =
         \norm{X}_{L^2(\bbR^2)}^2
         - 2c \ip*{X}{uu\adj}_{L^2(\bbR^2)}
         + c^2,
      \end{equation*}
      since $u$ is constrained to be unit norm.
      By elementary calculus, the minimization over $c$ in this problem can be
      calculated in closed form; we find that the optimal $c$ is equal to
      $\ip{X}{uu\adj}_{L^2(\bbR^2)} = u\adj \sT_{X} u = \half u\adj(\sT_{X} +
      \sT_{X}\adj) u$, where $\sT_{X}\adj u = \int_{\bbR} X(s, \spcdot) u(s)
      \diff s$ is the adjoint of $\sT_{X}$.
      Hence, the original problem is equivalent to the problem
      \begin{equation*}
         \min_{\norm{u}_{L^2(\bbR)} = 1, u\adj(\sT_{X} + \sT_{X}\adj) u \geq 0}\,
         \frac{1}{2} \norm*{ X - (\half u\adj (\sT_{X} + \sT_{X}\adj) u) uu\adj
         }_{L^2(\bbR^2)}^2.
      \end{equation*}
      Expanding the square as before, this objective satisfies
      \begin{equation*}
         \frac{1}{2} \norm*{ X - (\half u\adj (\sT_{X} + \sT_{X}\adj) u) uu\adj
         }_{L^2(\bbR^2)}^2
         =
         \norm{X}_{L^2(\bbR^2)}^2
         - (\half u\adj (\sT_{X} + \sT_{X}\adj) u)^2
      \end{equation*}
      at any point where
      $u\adj(\sT_{X} + \sT_{X}\adj) u \geq 0$; otherwise, the objective equals
      $\norm{X}_{L^2(\bbR^2)}^2$.
      Now, if for every nonzero $u$ we have $u\adj(\sT_{X} + \sT_{X}\adj) u <
      0$, then evidently the only optimal solution to the problem is $u=0$.
      If for some nonzero $u$ we have $u\adj(\sT_{X} + \sT_{X}\adj) u \geq 0$,
      then the previous expression shows that the problem is equivalent to
      \begin{equation*}
         \max_{\norm{u}_{L^2(\bbR)} = 1}\,
         u\adj (\sT_{X} + \sT_{X}\adj) u,
      \end{equation*}
      which is feasible.

   \end{proof}

\end{lemma}

\subsection{Background on Image Resampling}
\label{sec:app_resampling}

We give a precise definition of the vector field representation underlying
\Cref{eq:2d_rotation} in the discrete setting
\Cref{eq:tilt-objective-discrete}.
For the template image $\vX_{\grtr} \in \bbR^{m \times n}$,
$\vX_{\grtr} \circ \vtau_\nu$ denotes image resampling:
\begin{equation}
   \vX_{\grtr} \circ \vtau_{\nu}
   =
   \sum_{(k, l) \in G} (\vX_{\grtr})_{kl} \phi\left(
      \tfrac{n-1}{2} \left(
         \vtau_{\nu}^0 - k \One \One\adj
      \right)
   \right)
   \odot
   \phi \left(
      \tfrac{n-1}{2}
      \left(
         \vtau_{\nu}^1 - l \One \One\adj
      \right)
   \right).
   \label{eq:resampling}
\end{equation}
Here, $\phi : \bbR \to \bbR$ is the interpolation kernel; it is applied
elementwise, and is independent of the image content and resolution. Typical choices
for this kernel in practice are the bilinear interpolation kernel (which is
continuous, but not continuously differentiable; we adopt it in our
experiments) and the cubic convolution interpolation kernel \cite{Keys1981-jx}
(which is continuously differentiable, with an absolutely continuous
derivative). Both of these kernels are compactly supported, which allows
\Cref{eq:resampling} to be computed with cost proportional to the image size.
The transformation field $\vtau_{\nu} \in \bbR^{m \times n \times 2}$ is
defined as
\begin{align}
   \label{eq:motion_field_param_0}
   \vtau_{\nu}^0 &=
   \cos \nu \, \left( \frac{2}{n-1} \vn - \One\right) \One\adj 
   + \sin \nu \, \One \left(\frac{2}{n-1}\vn - \One \right)\adj 
   \\
   \vtau_{\nu}^1 &=
   -\sin \nu \, \left( \frac{2}{n-1} \vn - \One\right) \One\adj 
   + \cos \nu \, \One \left(\frac{2}{n-1}\vn - \One \right)\adj 
   \label{eq:motion_field_param_1}
\end{align}
where 
$\vn = [0, 1, \dots, n-1]$ (c.f.\ \cite[\S A.1]{Buchanan2022-qp} and
\Cref{eq:image_ndc}). Note that this definition ensures that the resampled
image $\vX_{\grtr} \circ \vtau_{\nu}$ corresponds to a rotation of the image content
by an angle of $\nu$ (with the usual ``counterclockwise'' positive
orientation): in particular,
\begin{equation*}
   (\vtau_\nu)_{ij}
   =
   \begin{bmatrix}
      \cos \nu & -\sin \nu \\
      \sin \nu & \cos \nu
   \end{bmatrix}\adj
   \begin{bmatrix}
      i \\ j
   \end{bmatrix}
\end{equation*}
for $(i, j) \in G$ defined in \Cref{eq:image_ndc}.

\end{document}